\documentclass[10pt,journal,compsoc]{IEEEtran}

\usepackage{amsmath}
\usepackage{amssymb}
\usepackage{amsfonts}
\usepackage[cmintegrals]{newtxmath}
\usepackage{graphicx}
\usepackage{epsfig}
\usepackage{epstopdf}
\usepackage{booktabs}
\usepackage{multirow}
\usepackage{array}
\usepackage[table]{xcolor}     
\usepackage{pifont}
\usepackage{bbding}            
\usepackage{enumitem}
\usepackage{textcomp}
\usepackage{stfloats}
\usepackage{url}
\usepackage{ragged2e}          % \justifying
\usepackage{caption}
\usepackage[caption=false,font=normalsize,labelfont=sf,textfont=sf]{subfig}
\usepackage[british,english,american]{babel}

\ifCLASSOPTIONcompsoc
  \usepackage[nocompress]{cite}
\else
  \usepackage{cite}
\fi

\usepackage[pagebackref,breaklinks,colorlinks=true,citecolor=blue]{hyperref}

\def\etal{{\emph{et al}}}

\newlist{todolist}{itemize}{2}
\setlist[todolist]{label=$\square$}

\newcolumntype{I}{!{\vrule width 3pt}}

\newlength\savedwidth

\newlength\savewidth
\newcommand\shline{\noalign{\global\savewidth\arrayrulewidth
  \global\arrayrulewidth 1pt}\hline
  \noalign{\global\arrayrulewidth\savewidth}}

\newcommand{\tablestyle}[2]{\setlength{\tabcolsep}{#1}%
  \renewcommand{\arraystretch}{#2}\centering\footnotesize}
\renewcommand{\paragraph}[1]{\vspace{1.25mm}\noindent\textbf{#1}}

\begin{document}
\title{Ultra-High-Definition Restoration Transformers with Correlation Matching Transformation}

\author{Cong Wang,
        Liyan Wang,
        Jinshan Pan, 
        Wei Wang,
        Wenqi Ren,
        Jun Liu,
        Xiaochun Cao
        % <-this % stops a space
\IEEEcompsocitemizethanks{

\IEEEcompsocthanksitem This work was supported in part by the National Natural Science Foundation of China (Grant No.62306343), the Guangdong Basic and Applied Basic Research Foundation (Grant No. 2025A1515011322).

\IEEEcompsocthanksitem Cong Wang, Wei Wang, Wenqi Ren, and Xiaocun Cao are with the School of Cyber Science and Technology, Shenzhen campus of Sun Yat-Sen University, Shenzhen, China (E-mail: supercong94@gmail.com, \{wangwei29, renwq3, caoxiaochun\}@mail.sysu.edu.cn).

\IEEEcompsocthanksitem Cong Wang is also with the Department of Radiology and Biomedical Imaging, University of California, San Francisco, CA, USA.

\IEEEcompsocthanksitem Liyan Wang is with the School of Mathematical Sciences, Dalian University of Technology, Dalian, China (E-mail: wangliyan@mail.dlut.edu.cn).

\IEEEcompsocthanksitem Jinshan Pan is with the School of Computer Science and Engineering, Nanjing University of Science and Technology, Nanjing, China (E-mail: sdluran@gmail.com).

\IEEEcompsocthanksitem Jun Liu is with the School of Computing and Communications, Lancaster University (E-mail: j.liu81@lancaster.ac.uk).

\IEEEcompsocthanksitem Wei Wang is the corresponding author.
% note need leading \protect in front of \\ to get a newline within \thanks as
% \\ is fragile and will error, could use \hfil\break instead.
% E-mail: see http://www.michaelshell.org/contact.html
}% <-this % stops an unwanted space
%\thanks{Manuscript received April 19, 2005; revised August 26, 2015.}
}

% The paper headers
\markboth{IEEE Transactions on Pattern Analysis and Machine Intelligence}%
{Shell \MakeLowercase{\textit{et al.}}: Bare Demo of IEEEtran.cls for Computer Society Journals}

\IEEEtitleabstractindextext{%
\begin{abstract}
\justifying
We propose UHDformer++, a general Transformer-based framework to solve numerous Ultra-High-Definition (UHD) image restoration tasks. 
UHDformer++ operates across $4$ coordinated learning spaces: 1) a high-resolution space (HR) for multi-level feature extraction, 2) a low-resolution space (LR) for learning compact, representative features, 3) a super-resolution space (SR) for upsampling low-resolution features from SR, and 4) a low-high fusion and reconstruction space (LHFR) for final image restoration.
Specifically, HR extracts multi-scale high-resolution features and fuses them with low-resolution cues to produce residual images, while LR distills complementary representations from HR to improve restoration quality. 
To supply LHFR with richer features, SR super-resolves LR outputs before fusion. 
We further introduce two modules to bridge the high- and low-resolution spaces. 
The Feature-Refined Correlation Matching Transformation (FR-CMT) module selects the top $C/r~(C~\text{denotes the number of channels;~}r\geq1~\text{controls the squeezing level})$, from the fusion between max- and mean-pooled high-resolution features to replace less informative channels in the low-resolution Transformer. 
The Adaptive Channel Modulator (ACM) adaptively recalibrates multi-scale high-resolution features, ensuring that only task-relevant information propagates to LR.
Extensive experiments demonstrate that UHDformer++ reduces model parameters by at least 86\% compared with recent state-of-the-art methods while achieving substantial performance gains across $5$ UHD restoration tasks, including low-light image enhancement, dehazing, deblurring, deraining, and desnowing. Code will be released at \url{https://github.com/supersupercong/uhdformerplus}.
\end{abstract}

}

% make the title area
\maketitle
% To allow for easy dual compilation without having to reenter the
% abstract/keywords data, the \IEEEtitleabstractindextext text will
% not be used in maketitle, but will appear (i.e., to be "transported")
% here as \IEEEdisplaynontitleabstractindextext when the compsoc 
% or transmag modes are not selected <OR> if conference mode is selected 
% - because all conference papers position the abstract like regular
% papers do.
\IEEEdisplaynontitleabstractindextext
% \IEEEdisplaynontitleabstractindextext has no effect when using
% compsoc or transmag under a non-conference mode.

% For peer review papers, you can put extra information on the cover
% page as needed:
% \ifCLASSOPTIONpeerreview
% \begin{center} \bfseries EDICS Category: 3-BBND \end{center}
% \fi
%
% For peerreview papers, this IEEEtran command inserts a page break and
% creates the second title. It will be ignored for other modes.
\IEEEpeerreviewmaketitle

% \IEEEraisesectionheading{\section{Introduction}\label{sec:introduction}}
% Computer Society journal (but not conference!) papers do something unusual
% with the very first section heading (almost always called "Introduction").
% They place it ABOVE the main text! IEEEtran.cls does not automatically do
% this for you, but you can achieve this effect with the provided
% \IEEEraisesectionheading{} command. Note the need to keep any \label that
% is to refer to the section immediately after \section in the above as
% \IEEEraisesectionheading puts \section within a raised box.

% The very first letter is a 2 line initial drop letter followed
% by the rest of the first word in caps (small caps for compsoc).
% 
% form to use if the first word consists of a single letter:
% \IEEEPARstart{A}{demo} file is ....
% 
% form to use if you need the single drop letter followed by
% normal text (unknown if ever used by the IEEE):
% \IEEEPARstart{A}{}demo file is ....
% 
% Some journals put the first two words in caps:
% \IEEEPARstart{T}{his demo} file is ....
% 
% Here we have the typical use of a "T" for an initial drop letter
% and "HIS" in caps to complete the first word.
% \IEEEPARstart{T}{his} demo file is intended to serve as a ``starter 

\section{Introduction}\label{sec:introduction}

%Most computer vision applications are built on clear input signal.
%However, images captured in low-light environments usually suffer from varies of causes, such as low-light environment, limited ability of photography equipment, and inappropriate configurations for the equipment.
%These causes will lead to poor visibility that will affect further vision analysis and processing.
%Hence, restoring a normal result from a given low-light image becomes a significant task.
% \IEEEPARstart{T}{aking} high-quality images in low-illumination \section{Introduction} Ultra High-Definition (UHD, i.e., 12 megapixels or 4K) videos have become a trend during the last several years. 
%
\IEEEPARstart{I}{n} recent years, the rapid development of advanced imaging sensors and displays has greatly contributed to the progress of Ultra-High-Definition (UHD, i.e., 12 megapixels or 4K) imaging~\cite{uhd_video_deblurring}. 
Although UHD imaging provides extensive applications and improves picture quality significantly, the increased pixel count poses efficiency challenges for existing image processing algorithms. 
Especially, UHD images captured under low-light, hazy, or high-speed movement conditions often suffer from undesirable degradation, resulting in visually low quality and hindering high-level vision tasks. 
To solve these problems, this paper focuses on presenting a unified framework that effectively addresses the challenging problem of UHD image restoration.

\begin{figure*}[!t]
\centering
\begin{center}
\begin{tabular}{ccccccccc}
\includegraphics[width=0.2\linewidth]{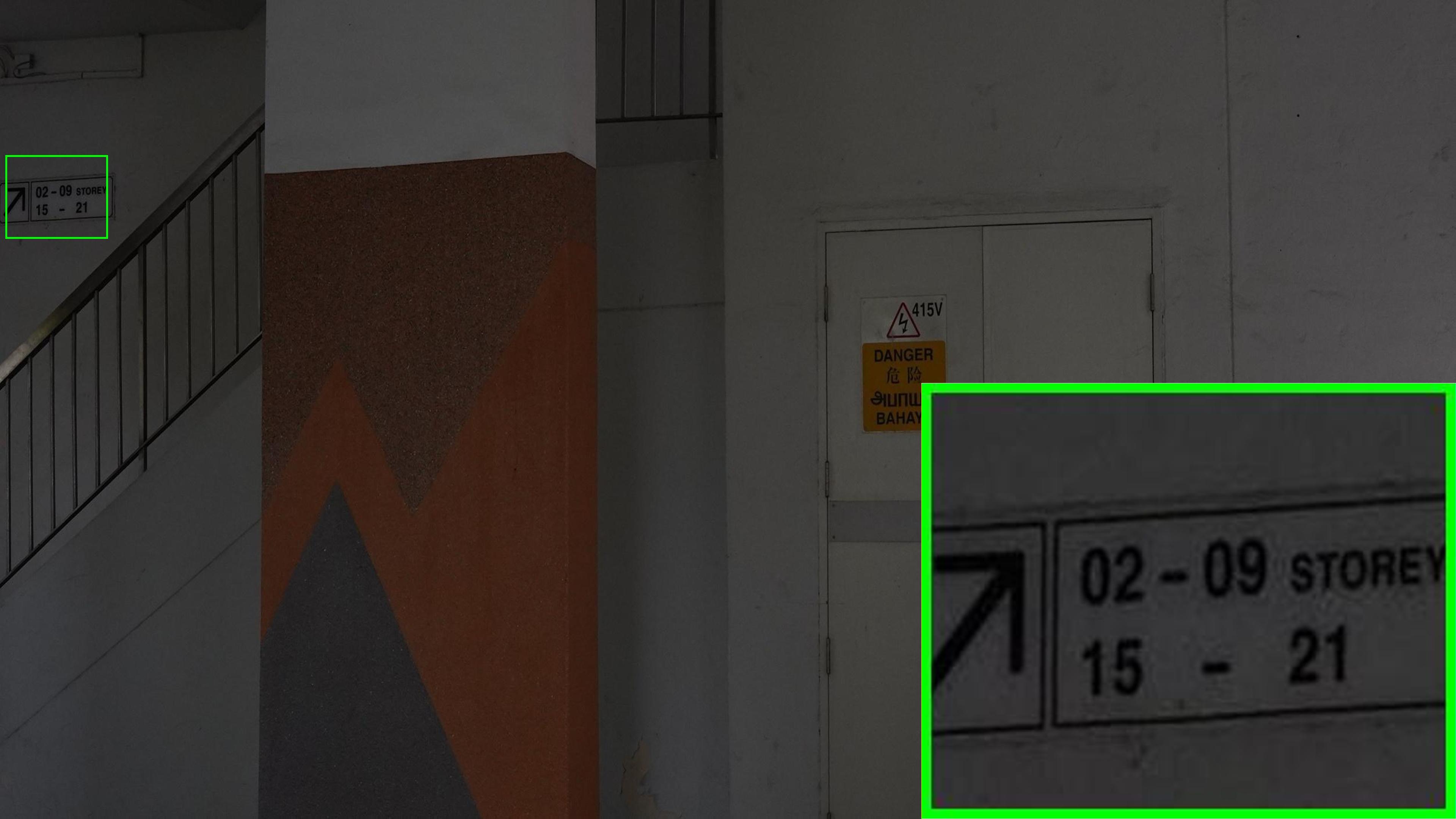} &\hspace{-4.75mm}
\includegraphics[width=0.2\linewidth]{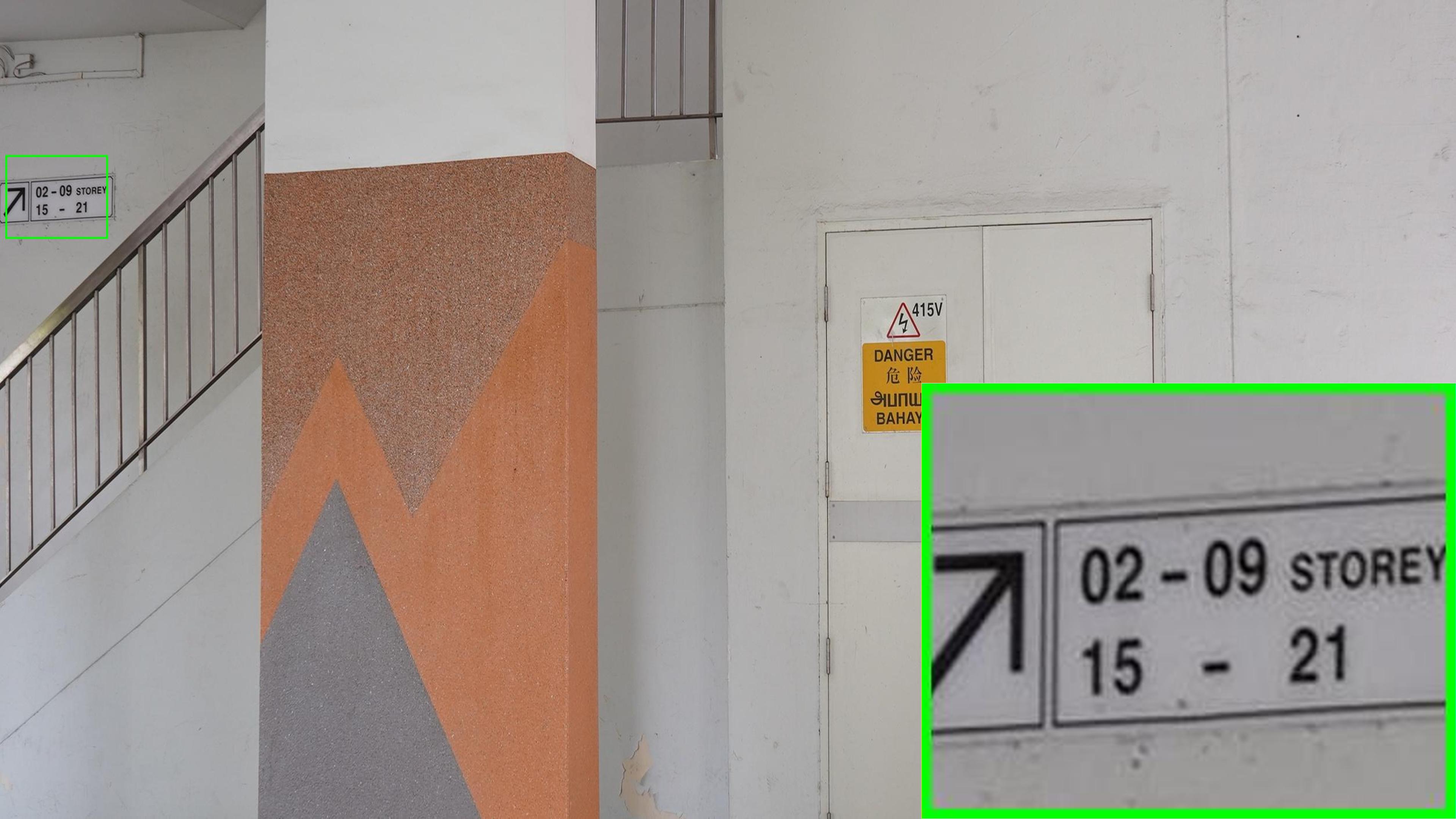} &\hspace{-4.75mm}
\includegraphics[width=0.2\linewidth]{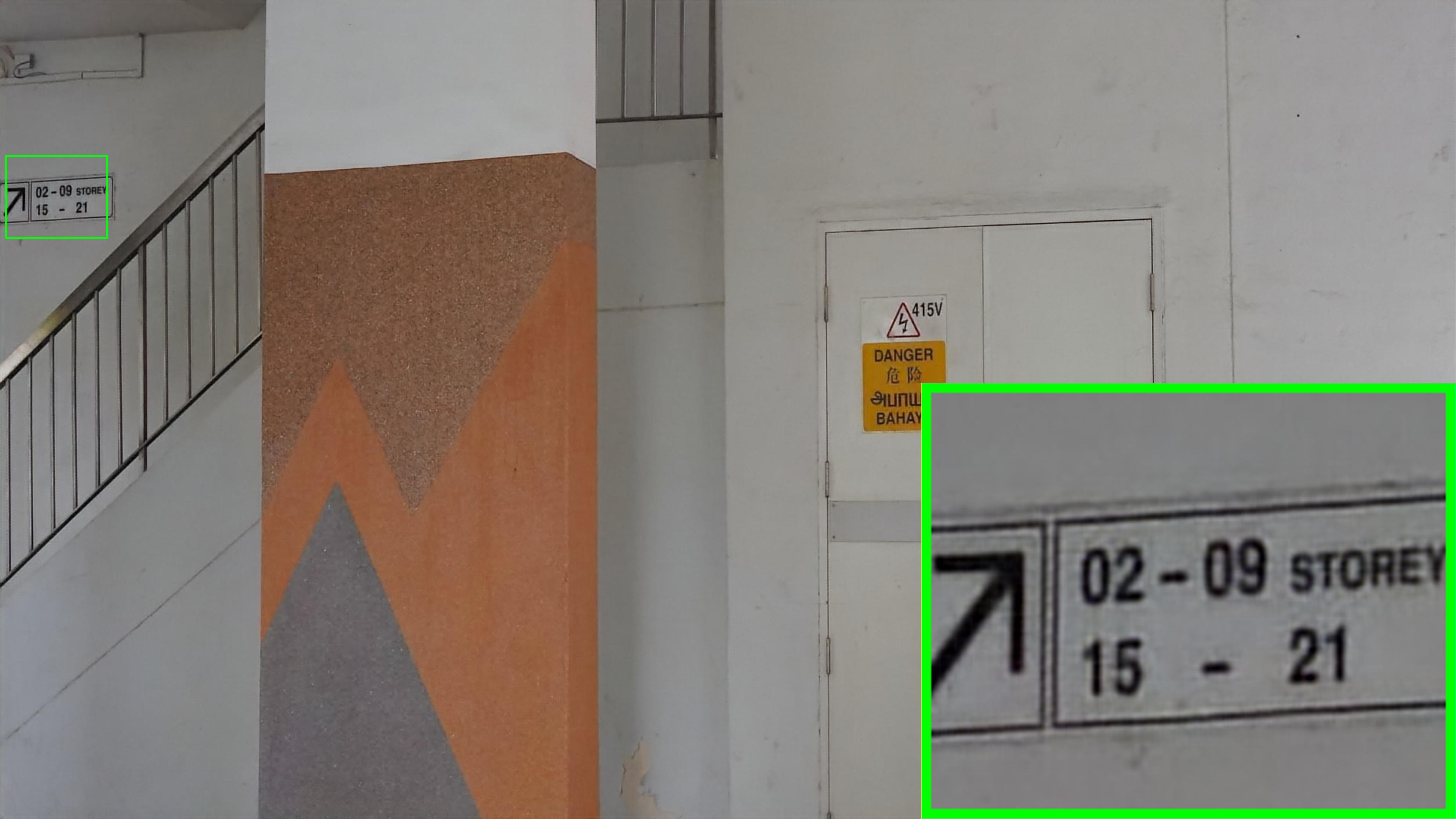} &\hspace{-4.75mm}
\includegraphics[width=0.2\linewidth]{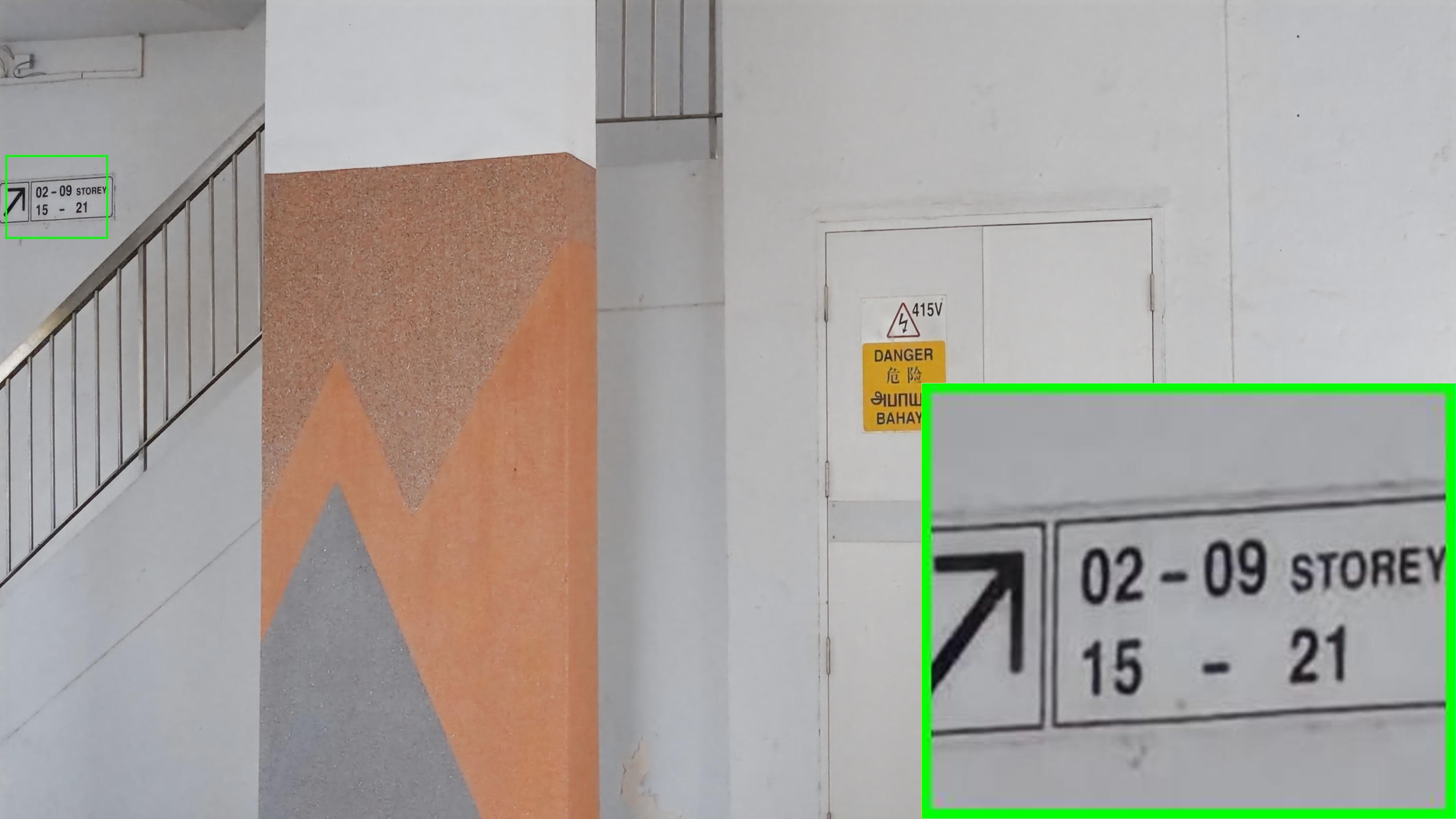} 
\\
Low-light&\hspace{-4.75mm}  GT&\hspace{-4.75mm}  UHDFour~\cite{Li2023ICLR_uhdfour}&\hspace{-4.75mm}   \textbf{UHDformer++}
\\
\includegraphics[width=0.2\linewidth]{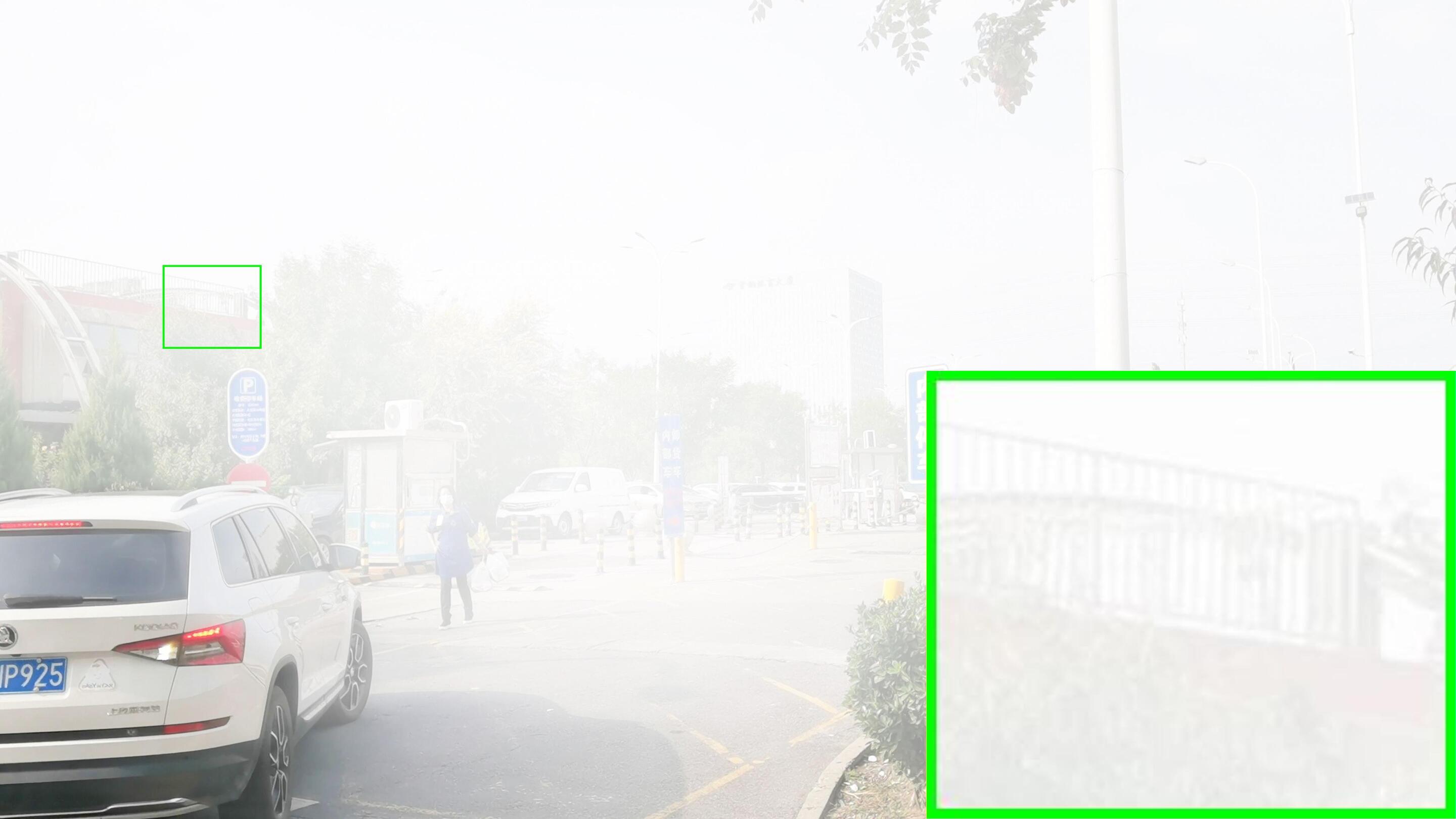} &\hspace{-4.75mm}
\includegraphics[width=0.2\linewidth]{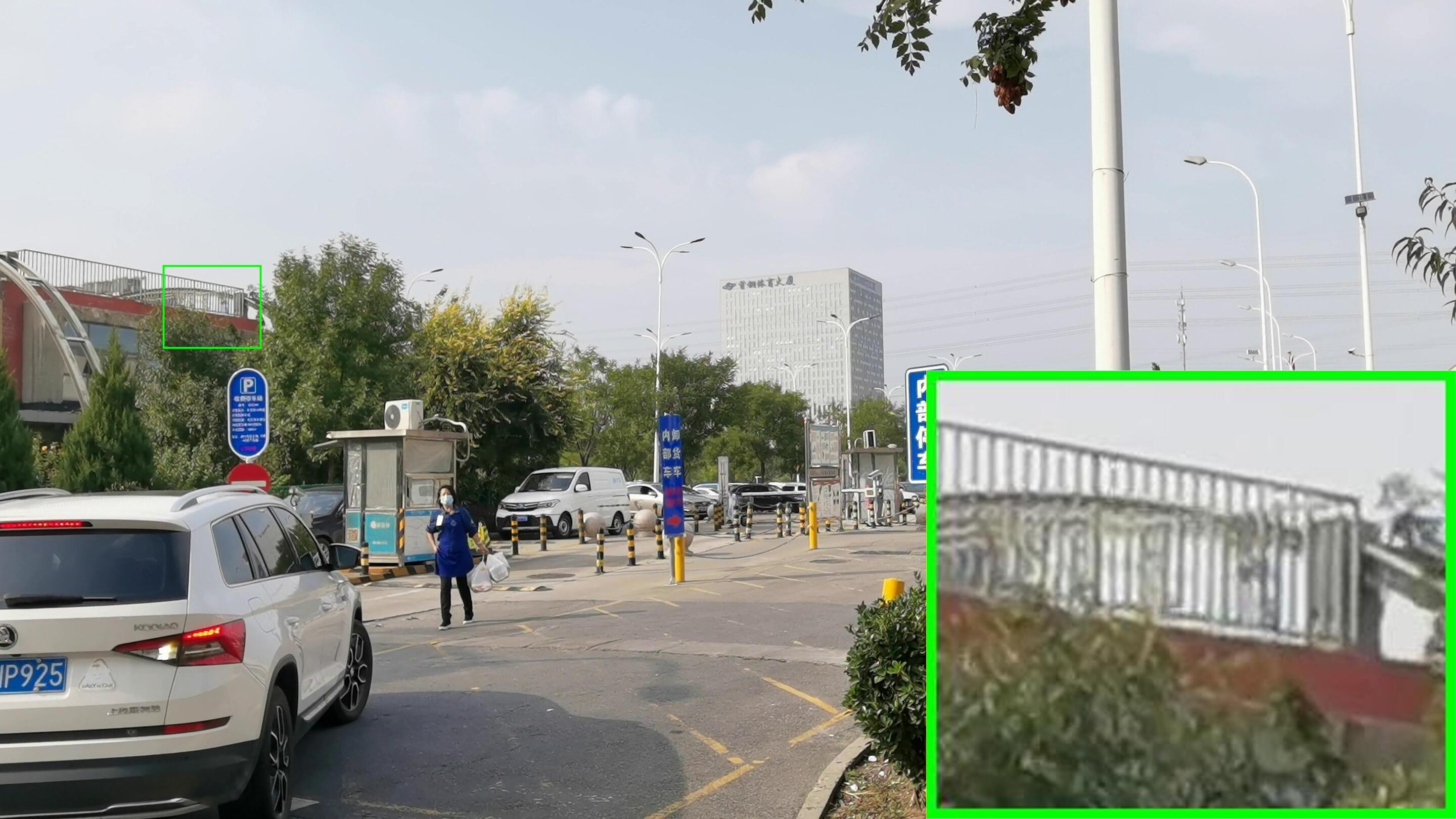} &\hspace{-4.75mm}
\includegraphics[width=0.2\linewidth]{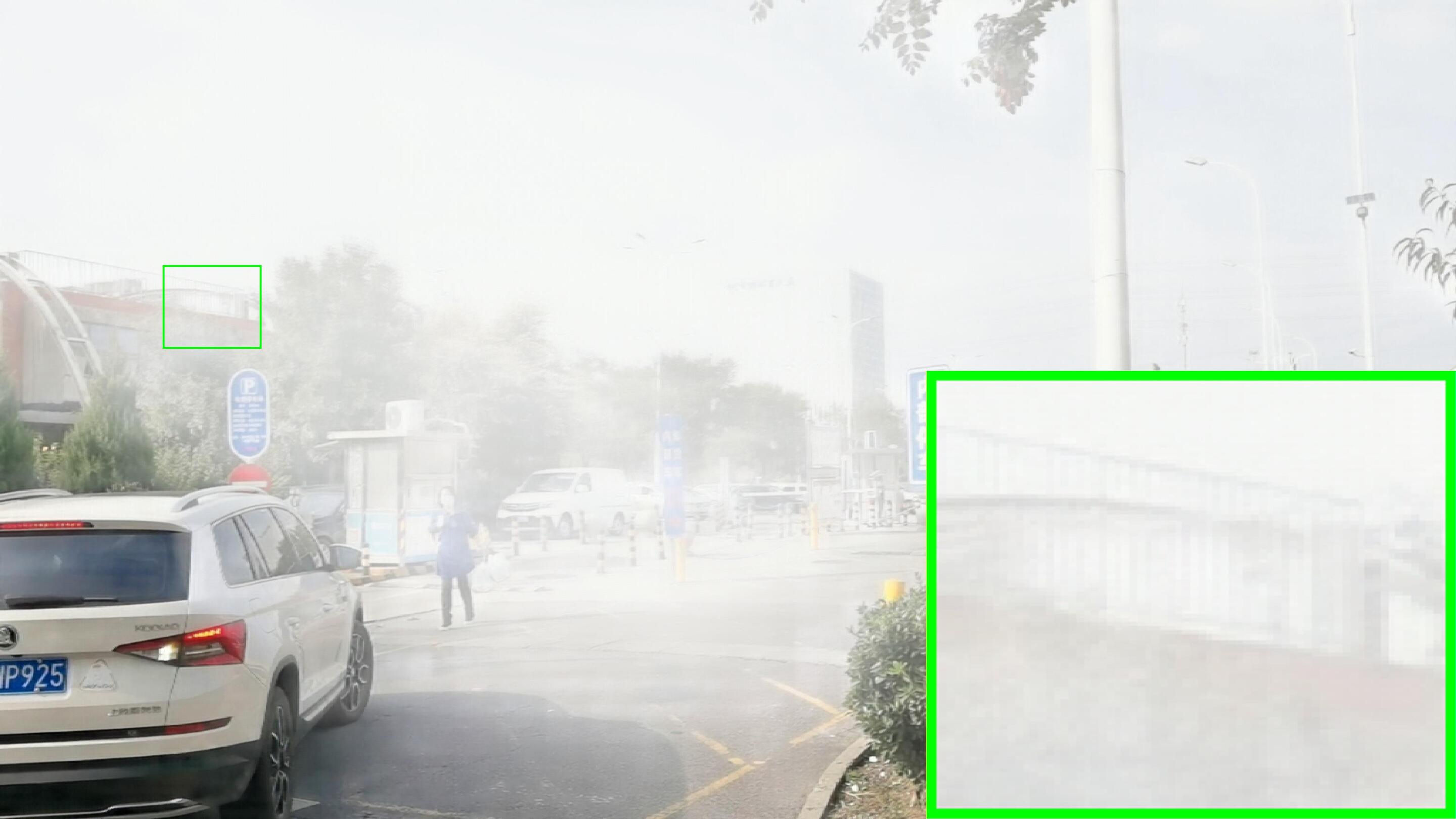} &\hspace{-4.75mm}
\includegraphics[width=0.2\linewidth]{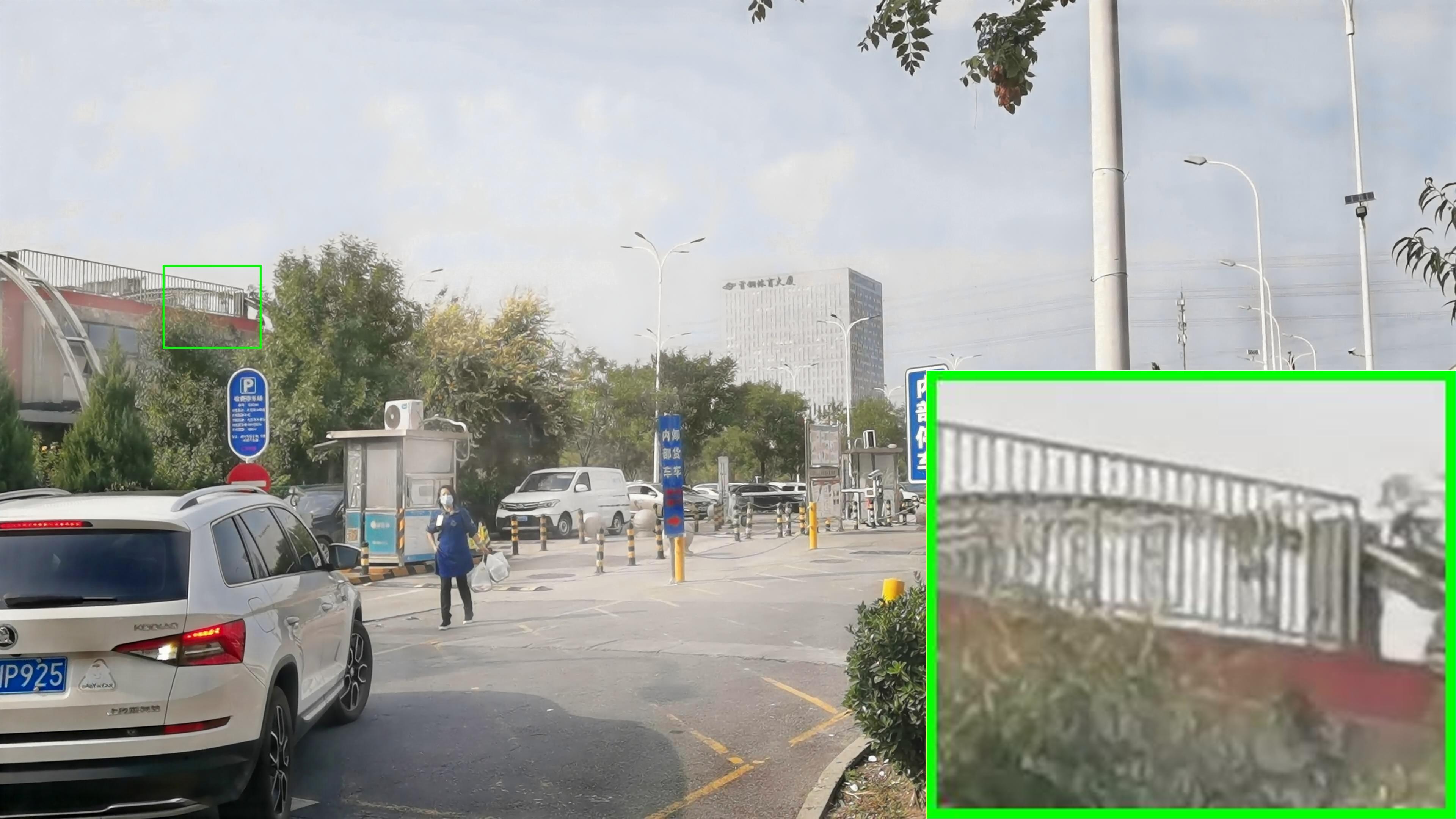}
\\
 Haze&\hspace{-4.75mm}  GT&\hspace{-4.75mm}  DehazeFormer~\cite{DehazeFormer}&\hspace{-4.75mm}  \textbf{UHDformer++}
\\
\includegraphics[width=0.2\linewidth]{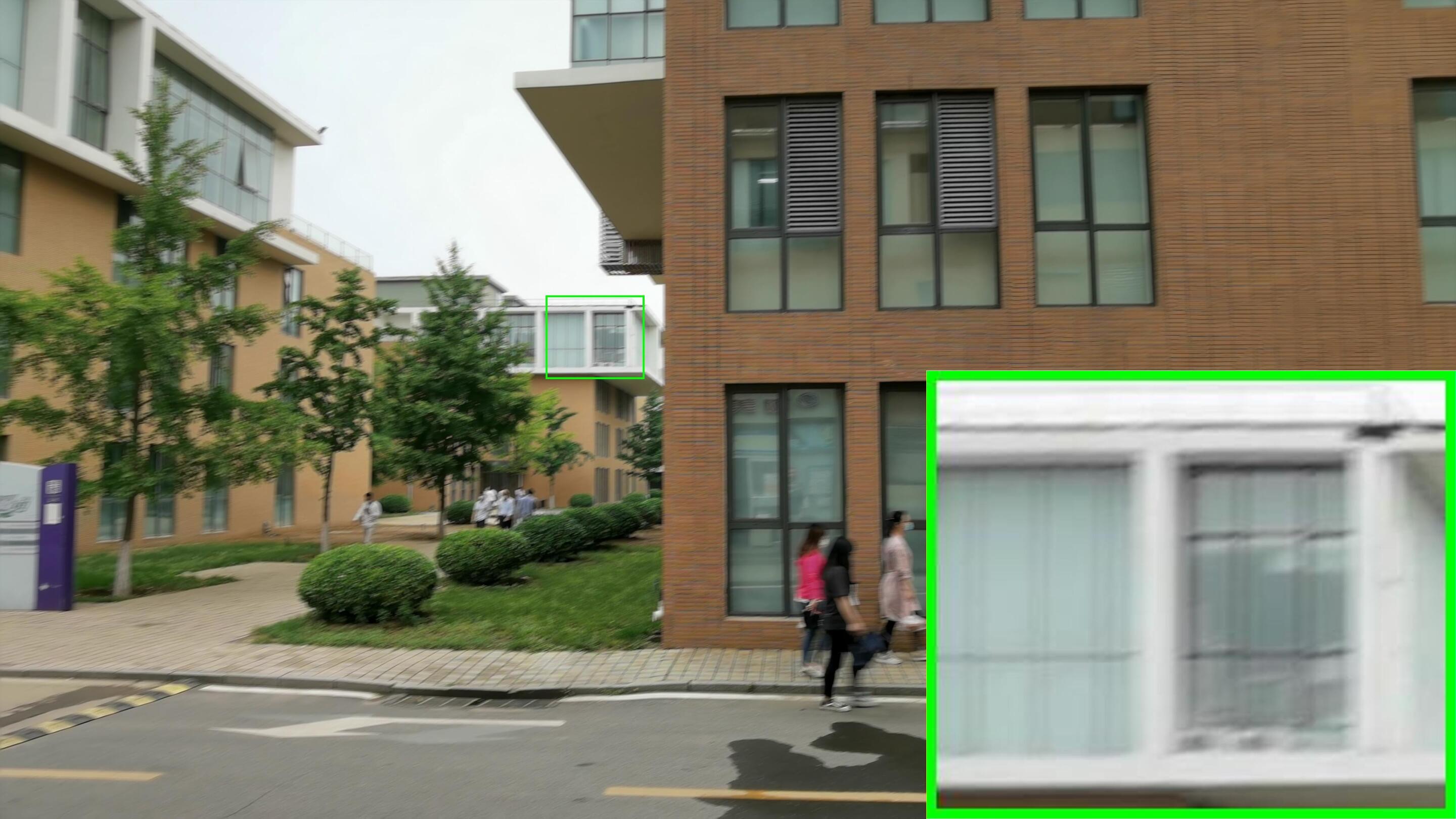} &\hspace{-4.75mm}
\includegraphics[width=0.2\linewidth]{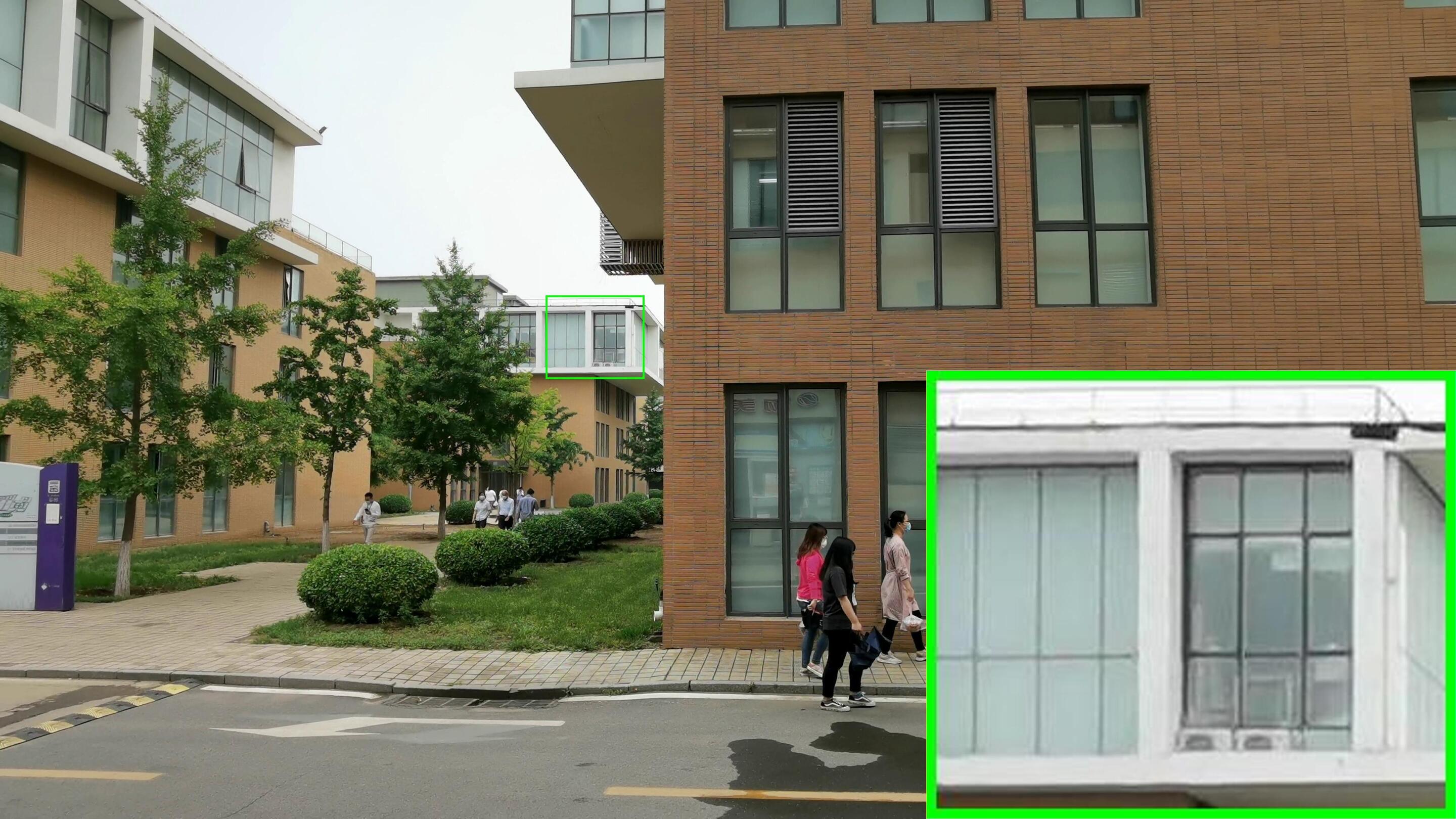} &\hspace{-4.75mm}
\includegraphics[width=0.2\linewidth]{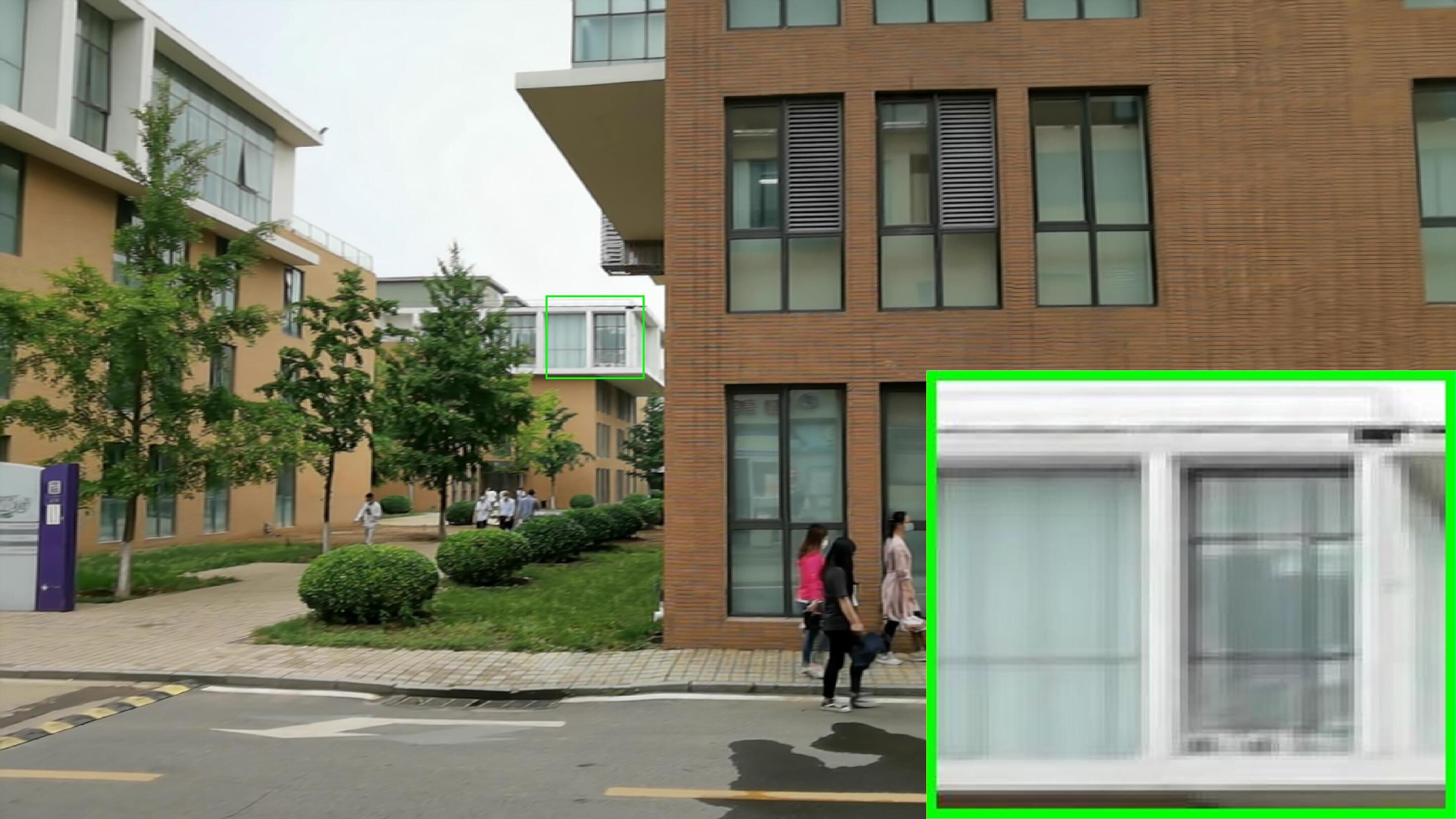} &\hspace{-4.75mm}
\includegraphics[width=0.2\linewidth]{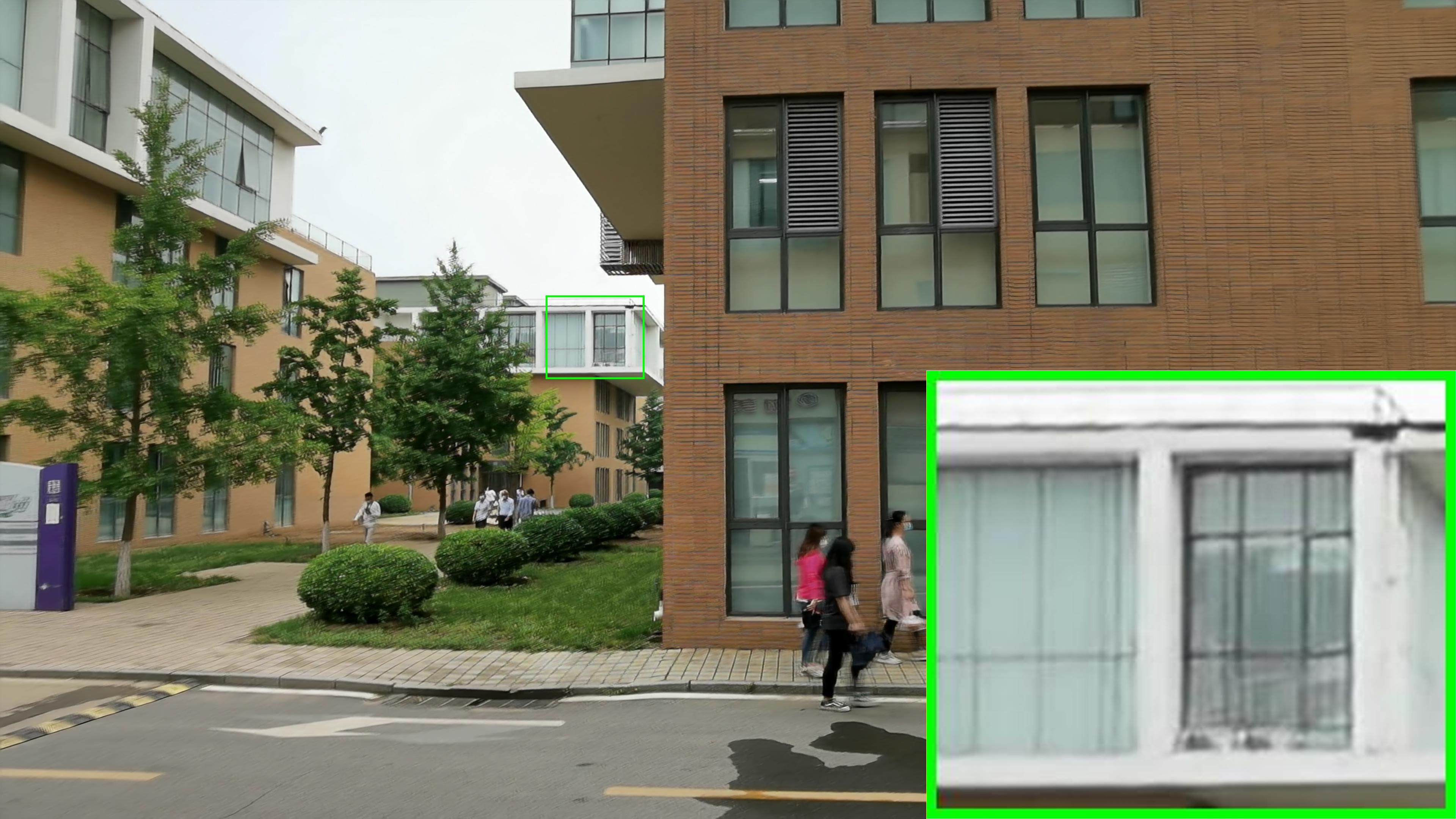}
\\
Blur&\hspace{-4.75mm}  GT&\hspace{-4.75mm}  FFTformer~\cite{Kong_2023_CVPR_fftformer}&\hspace{-4.75mm}   \textbf{UHDformer++}
\end{tabular}
\vspace{-3mm}
\caption{
\textbf{Challenging Ultra-High-Definition image restoration examples.}
\textbf{Top}: Low-light image enhancement;
\textbf{Middle}: Image dehazing;
\textbf{Bottom}: Image deblurring.
Compared with recent state-of-the-art approaches UHDFour~\cite{Li2023ICLR_uhdfour} for low-light image enhancement, DehazeFormer~\cite{DehazeFormer} for image dehazing, and FFTformer~\cite{Kong_2023_CVPR_fftformer} for image deblurring,
our proposed \textbf{UHDformer++} is able to recover clear results with finer structures and more natural colors while consuming only $0.3962$M parameters with about $50 \times$ fewer parameters than state-of-the-art methods (see Tabs.~\ref{tab:Low-light image enhancement.}-\ref{tab:Image deblurring.}) due to the simple and effective designs introduced in this paper.
}
\label{fig:Low-light image enhancement on UHD-LL}
\end{center}
\vspace{-4mm}
\end{figure*}

With the great success of Convolutional Neural Networks (CNNs)~\cite{AlexNet,res,dense_network_cvpr17} and Vision Transformers (ViT)~\cite{vision_transformer,khan2021transformers,liu2021swin} on various computer vision tasks, learning-based methods have achieved impressive performance on general image restoration~\cite{tao2018scale,dehazing_AOD,ren2019progressive,Zamir_2021_CVPR_mprnet,liang2021swinir,wang2021uformer,guo2022dehamer,Tsai2022Stripformer,d4_dehze,liu2018desnownet,chen2022msp,PromptRestorer,wang2024progressive,mm24_perceplie,selfpromer,wang2025intra,cui_pami24_fs,cui_pami24_irnext,cui_tip25,cui_ijcv26}. 
These general image restoration methods are usually designed without considering the high-resolution image size, which makes them incapable of handling UHD image sizes or effectively recovering high-quality images due to computational constraints, e.g., inference on a single 3090 GPU. 
This limits the potential applications of UHD imaging systems.

Recently, with the demand for handling UHD degraded images, several approaches have been developed for UHD image restoration~\cite{uhd_video_deblurring,Zheng_uhd_CVPR21,aaai24wang_UHDformer,wang2025deep,wang2025ultra}.
For example, Zheng~\etal.~\cite{Zheng_uhd_CVPR21} propose a multi-guided bilateral upsampling model for UHD image dehazing.
To solve the UHD video deblurring problem, Deng~\etal.~\cite{uhd_video_deblurring} develop a separable-patch architecture by collaborating with a multi-scale integration scheme.
%
% In contrast, \citet{Li2023ICLR_uhdfour} incorporate Fourier transform into low-light image enhancement by leveraging the amplitude and phase within a cascaded network.
Different from the above methods, which handle restoration in the spatial domain, Li~\etal.~\cite{Li2023ICLR_uhdfour} incorporate the Fourier transform into low-light image enhancement by leveraging the amplitude and phase within a cascaded network.
However, these methods usually fail to explore the valuable content for low-resolution space from high-resolution ones, which could contain useful information that significantly affects restoration quality. 
Furthermore, existing methods often rely on large-capacity models to achieve optimal performance. 
For instance, methods of UHDFour~\cite{Li2023ICLR_uhdfour} and UHD~\cite{Zheng_uhd_CVPR21} consume $17.7$M and $34.5$M trainable parameters, respectively, making them unfriendly for deployment on small-capacity devices.
To address the aforementioned limitations, we propose UHDformer++, a simple yet effective unified Transformer for UHD image restoration.
UHDformer++ is composed of $4$ complementary spaces: 1) a high-resolution space (HR) for multi-level feature extraction and residual prediction, 2) a low-resolution space (LR) for learning compact representations, 3) a super-resolution space (SR) for feature upsampling from LR, and 4) a low-high fusion and reconstruction space (LHFR) for final image restoration.
Specifically, HR extracts multi-scale high-resolution features and fuses them with low-resolution cues to generate residual images. 
In parallel, LR leverages knowledge from HR to learn more discriminative low-resolution features, thereby facilitating restoration. 
To bridge the resolution gap, SR super-resolves LR features and feeds them to LHFR, which integrates multi-resolution cues to reconstruct the final output.

A key challenge is transferring informative features from high- to low-resolution space while suppressing redundancy. 
To this end, we introduce $2$ modules. First, the Feature-Refined Correlation Matching Transformation (FR-CMT) module computes channel-wise correlations between high-resolution features pooled by max and mean operations. 
It then selects the top $C/r~(C~\text{denotes the number of channels;~}r\geq1~\text{controls the squeezing level})$, to replace the query features of in Correlation Matching Transformation Attention (CMTA) and the feed-forward features Correlation Matching Transformation Feed-forward Network (CMTFN) in Correlation Matching Transformation Transformer Blocks (CMT-TB) at LR. 
This suppresses redundant activations and enhances feature compactness. 
Second, the Adaptive Channel Modulator (ACM) adaptively recalibrates multi-level features from HR before they are transmitted to LR, ensuring that only task-relevant information propagates across scales.

\begin{figure*}[!t]
% \color{blue}
\begin{center}
\begin{tabular}{cccccccccc}
\hspace{-2mm}\includegraphics[width=1.05\linewidth]{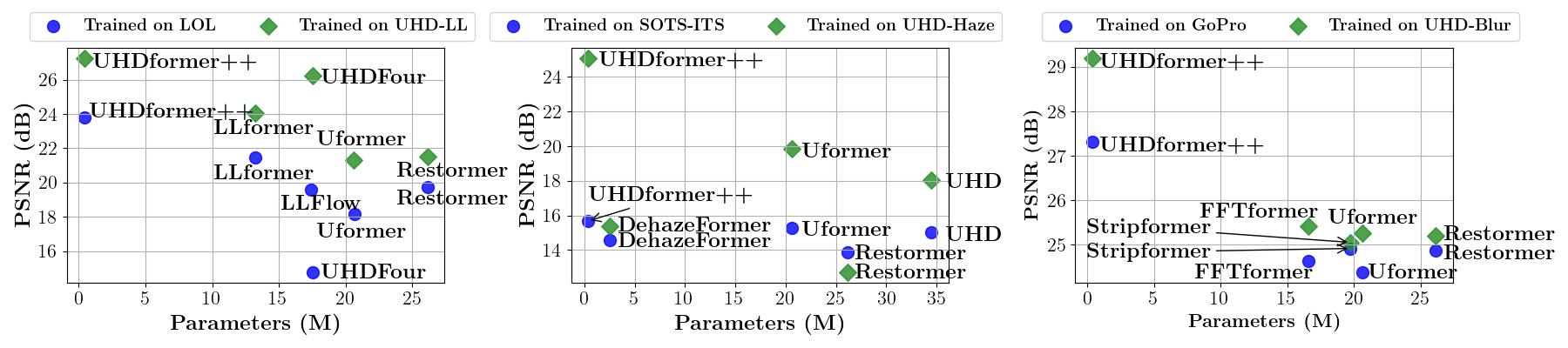}
\\
\hspace{-16mm}(a) Low-Light Image Enhancement (Tab.~\ref{tab:Low-light image enhancement.})  ~~~~~~~~~~~  (b)  Image Dehazing (Tab.~\ref{tab:Image dehazing.})~~~~~~~~~~~~~~~~~~~   (c) Image Deblurring (Tab.~\ref{tab:Image deblurring.})
\\
\end{tabular}
\vspace{-3mm}
\caption{\textbf{Parameters-performance trade-off.}
Compared with recent state-of-the-art methods UHDFour~\cite{Li2023ICLR_uhdfour} for low-light image enhancement, DehazeFormer-B~\cite{DehazeFormer} for image dehazing, and FFTformer~\cite{Kong_2023_CVPR_fftformer} for image deblurring, our UHDformer++ respectively reduces at least $98\%$, $86\%$, and $97\%$ model sizes while obtaining state-of-the-art UHD image restoration performance under different training sets, significantly improving the trade-off on these $3$ UHD image restoration tasks.
}
\label{fig: Parameters-Performance trade-off.}
\end{center}
\vspace{-4mm}
\end{figure*}

With these designs, UHDformer++ achieves state-of-the-art performance while substantially reducing model size. As shown in Fig.~\ref{fig: Parameters-Performance trade-off.}, it attains the best parameter-performance trade-off on three UHD restoration tasks: low-light image enhancement, dehazing, deblurring, image deraining, and image desnowing.

We summarize our main contributions to this paper as follows:
\begin{itemize}
% \begin{compactitem}
     \item We propose UHDformer++, to the best of our knowledge, the first general Transformer framework for UHD image restoration. It establishes a principled transformation from high-resolution to low-resolution spaces via correlation matching in both attention and feed-forward networks, enabling efficient multi-scale feature interaction.
     \item We design FR-CMT to identify and transfer the most informative channels from high-resolution to low-resolution features. By replacing low-resolution query features with top- $C/r$ correlated channels from pooled high-resolution features, FR-CMT suppresses redundancy and enhances representation compactness for improved restoration.
    \item  We introduce ACM to adaptively recalibrate multi-level high-resolution features before cross-scale transmission. ACM ensures that more task-relevant information is propagated to the low-resolution space, mitigating feature mismatch and noise accumulation.
    \item Extensive experiments demonstrate that UHDformer++ reduces model parameters by at least 86\% relative to recent state-of-the-art methods while achieving superior results on {\color{blue} $5$} UHD restoration tasks, including low-light enhancement, dehazing, deblurring, deraining, and desnowing.
    % 
% \end{compactitem}
\end{itemize}

This work substantially extends our preliminary conference version~\cite{aaai24wang_UHDformer} in both methodology and analysis. 
The key differences are:
First, we introduce a Feature-Refined Correlation Matching Transformation (FR-CMT) module that replaces the original correlation matching scheme. FR-CMT selects more representative features from the fusion between max-pooled and mean-pooled high-resolution features, followed by a newly designed gated feature refinement module, yielding more discriminative cross-scale correspondences.
Second, in the Correlation Matching Transformation Transformer Block (CMT-TB), we redesign Correlation Matching Transformation Feed-forward Networks (CMTFN) in Transformers. The new design increases modeling capacity while maintaining efficiency. Moreover, we replace the layer normalization with more powerful and effective Dynamic Tanh (DyT) operations.
Third, we incorporate a dedicated super-resolution space (SR) to upsample low-resolution features before final fusion and reconstruction. Ablations show that SRS contributes about 0.8dB PSNR gains by providing LHFR with spatially aligned features.
Fourth, we provide visualizations of the high-to-low correlation matching process, revealing that transformed high-resolution features preserve clearer structural patterns than before matching. This analysis validates the core motivation behind FR-CMT.
Fifth, we conduct extensive ablation studies to quantify the contribution of each component and present failure case analysis to delineate the operational scope of UHDformer++.
Finally, we evaluate on general image restoration benchmarks beyond UHD tasks, including standard low-light enhancement datasets, where UHDformer++ remains competitive with task-specific state-of-the-art methods. We also discuss current limitations that our framework is less effective on conventional motion deblurring benchmarks that require explicit kernel estimation, suggesting future work on blur-aware modules.

\section{Related Work}
In this section, we review image restoration techniques and Ultra-High-Definition restoration approaches.
% \\
\subsection{Image Restoration}\label{sec: Image Restoration}
CNN-based architectures~\cite{DnCNN,RCAN,zhang2020rdn,dualcnn_cvpr18,Zamir_2021_CVPR_mprnet,DCSFN,wang_aaai22,wang_tcsvt22,mm20_wang_jdnet,wang_icme2020,wang_tcsvt22,selfpromer} and Transformer-based models~\cite{yang2020learning, liang2021swinir,wang2021uformer,PromptRestorer} have been shown to outperform conventional restoration approaches~\cite{he2010single,timofte2013anchored,michaeli2013nonparametric}. 
These learning-based methods usually adopt U-Net architectures~\cite{cho2021rethinking_mimo,Zamir_2021_CVPR_mprnet,wang2021uformer,Zamir2021Restormer}, which have been demonstrated to be effective because of hierarchical multi-scale representation and effective learning between shallow and deeper layers by skip connection~\cite{zhang2019residual,liu2019dual}. 
Zhang~\etal.~\cite{RCAN} suggest the very deep residual channel attention network for single image super-resolution, where the authors are the first to build over 100 convolutional layers for image restoration.
Pan~\etal.~\cite{dualcnn_cvpr18} propose the dual convolution neural network to solve low-level vision tasks.
They formulate the image restoration problem by learning details and structures that are inspired by a physics-based formulation.
Zamir~\etal.~\cite{Zamir_2021_CVPR_mprnet} propose the multi-stage progressive image restoration network. 
They gradually recover the clear images within multi-patch progressive learning perspectives.
Liang~\etal.~\cite{liang2021swinir} adopt the Transformers to solve the image restoration problem, where they suggest a residual Swin Transformer block to learn the hierarchy features within shifted windows.
Zamir~\etal.~\cite{Zamir2021Restormer} propose an efficient Transformer to solve high-resolution image restoration tasks.
Different from previous works~\cite{liang2021swinir,chen2021IPT,wang2021uformer} that build the attention mechanism from the spatial perspectives, they propose to apply SA across feature dimension rather than the spatial dimension, which can effectively avoid exponential computational cost with respect to image resolution.
We refer the readers to recent excellent literature reviews on image restoration~\cite{anwar2019deep,li2019single,tian2020deep}, which summarize the main designs in deep image restoration models.

Although these Transformer-based methods have achieved promising performance for general image restoration, they usually cannot handle UHD images, which limits further potential applications on UHD imaging devices.

% \\
\subsection{Ultra-High-Definition Restoration}\label{sec: Ultra-High-Definition Restoration}
% \noindent \textbf{Ultra-High-Definition Restoration.}
%
With the demand for processing UHD-degraded images on imaging systems, a few methods have been proposed to recover clear UHD images.
Kim~\etal.~\cite{SR-ITM} propose the learning bilateral regularizer for UHD image super-resolution.
Motivated by bilateral learning, Zheng~\etal.~\cite{Zheng_uhd_CVPR21} suggest multi-guided bilateral learning for UHD image dehazing, where they adopt the bilateral learning to downsample the features to reduce the feature resolution so that the UHD images can be processed in a resource-constrained device. 
Deng~\etal.~\cite{uhd_video_deblurring} develop multi-scale networks with a separable-patch integration scheme to achieve a large receptive field for UHD video deblurring, which collaborates with a multi-scale integration scheme to achieve a large receptive field without adding the number of generic convolutional layers and kernels.
Wang~\etal.~\cite{LLformer} propose a Transformer with axis-based multi-head self-attention and cross-layer attention fusion for low-light image enhancement, where they split the images into multiple patches at the inference stage and merge the processed patches into one image.
However, such an inference strategy is not flexible and consumes more time.
Inspired by unique characteristics observed in the Fourier domain, Li~\etal.~\cite{Li2023ICLR_uhdfour} propose to embed the Fourier transform into a cascaded network by implementing amplitude and phase enhancement.
Please refer to the recent survey paper~\cite{wang2025deep} for a comprehensive review of model design in the UHD image restoration field.

%
%%%%%%%%%%%%%%%%%%%%%%%%%%%%%%%
\begin{figure*}[!t]
% \vspace{-12mm}
% \footnotesize
\centering
\begin{center}
\begin{tabular}{c}
\includegraphics[width=0.8\linewidth]{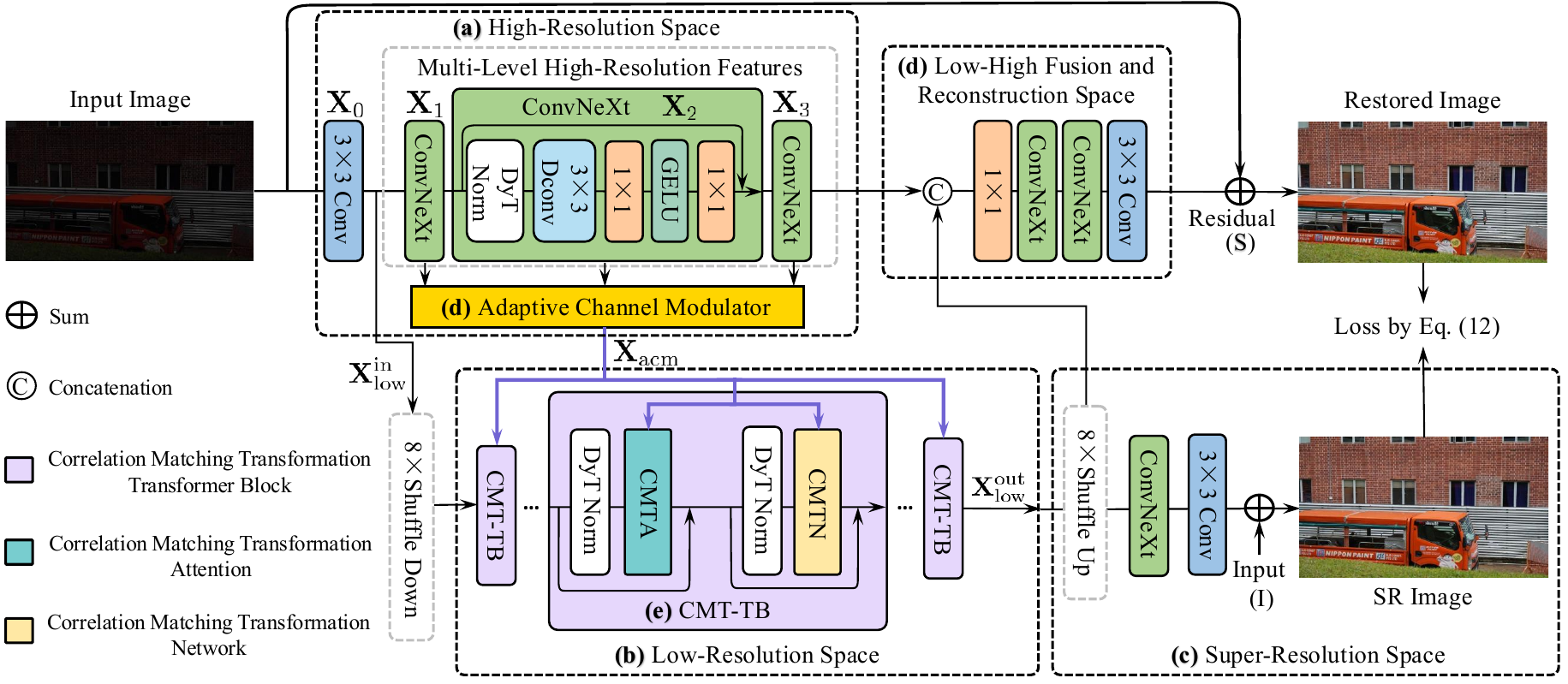} 
\end{tabular}
\vspace{-3mm}
\caption{
\textbf{Overall framework of the proposed UHDformer++}. It mainly contains $4$ spaces: (a) learning in high-resolution space (HR), (b) learning in low-resolution space (LR), (c) super-resolution space (SR), and (d) Low-high fusion and reconstruction space (LHFR).
The HR explores multi-level high-resolution features and fuses low-high features and reconstructs the residual images, while the LR learns the low-resolution features via the Feature-Refined Correlation Matching Transformation Transformer Block, as shown in Fig.~\ref{fig:Visual effect on Maximal Correlation Matching Module}.
Before transforming high-resolution features to low-resolution space, we use an Adaptive Channel Modulator to adaptively modulate multi-level high-resolution features to provide more representative content to low-resolution space.
Finally, we super-resolve the low-resolution features from LR by SR, which is further fused and reconstructed for the final reconstruction via LHFR.
}
\label{fig:Overall framework of our UHDformer}
\end{center}
\vspace{-4mm}
\end{figure*}

However, all these methods neglect to explore useful content in high-resolution space for the contributions of low-resolution ones.
In this paper, we propose to build the transformation from high- and low-resolution spaces to provide more representative features to conduct effective Transformer computation in the low-resolution space.

\section{Proposed UHDformer++}
Our objective is to transfer informative features from the high-resolution space to the low-resolution space within Transformer blocks, thereby enriching low-resolution representations for UHD image restoration. 
As high-resolution features contain fine-grained details, while low-resolution features provide context, an effective cross-scale transformation is essential for balancing detail and efficiency.
To this end, we design two complementary modules: the Feature-Refined Correlation Matching Transformation (FR-CMT) and the Adaptive Channel Modulator (ACM).
FR-CMT establishes channel-wise correspondences between high- and low-resolution features in both attention and feed-forward networks. 
It replaces low-resolution query and feed-forward features with the most correlated high-resolution channels, which suppresses redundant activations and yields a more compact representation in the low-resolution space.
ACM operates on the high-resolution path. 
It adaptively recalibrates multi-level high-resolution features before they are transmitted to the low-resolution space, ensuring that only task-relevant information is propagated and mitigating feature mismatch across scales.
Together, FR-CMT and ACM form a bidirectional refinement mechanism that ACM selects what to send from high-resolution, while FR-CMT decides what to keep in low-resolution.
\subsection{Overall Pipeline}\label{sec:Overall Pipeline}
Fig.~\ref{fig:Overall framework of our UHDformer} illustrates the overall architecture of UHDformer++, which operates across $4$ coordinated spaces: (a) High-Resolution space (HR), (b) Low-Resolution space (LR), (c) Super-Resolution space (SR), and (d) Low-High Fusion and Reconstruction space (LHFR).
HR extracts multi-level high-resolution features and fuses them with low-resolution cues to predict residual images. 
In parallel, LR learns compact low-resolution representations by establishing correlations with HR, thereby enriching its feature space with high-resolution priors.
To supply LHFR with spatially aligned features for final reconstruction, SR upsamples the outputs of LRS to the high-resolution scale. 
LHFR then integrates the refined features from HR and SR to generate the restored UHD image.
% \\
\subsubsection{High-Resolution Space}
% {\flushleft \bf Learning in High-Resolution Space.}
Given a UHD input image $\mathbf{I}$~$\in$~$\mathbb{R}^{H\times W \times 3}$, we first applies a $3$$\times$$3$ convolution to obtain low-level embeddings $\mathbf{X}_0$~$\in$~$\mathbb{R}^{H\times W \times C}$; where $H\times W$ denotes the spatial dimension and $C$ is the number of channels. 
Next, the shallow features $\mathbf{X}_{0}$ are hierarchically encoded into multi-level features $\{\mathbf{X}_1, \mathbf{X}_2, \mathbf{X}_3\}$~$\in$~$\mathbb{R}^{H \times W \times C}$ via $3$ ConvNeXt blocks~\cite{liu2022convnet_convnext}.
Then, the multi-level features are modulated via an Adaptive Channel Modulator, and the adjusted feature $\mathbf{X}_{\text{acm}}$~$\in$~$\mathbb{R}^{H\times W \times 3C}$ is sent to a low-resolution space to participate in conducting RF-CMT in Transformers, which will be performed for use in LiLR.
Finally, a low-high fusion and reconstruction layer containing a concatenation operation and two ConvNeXt blocks, and a $3 \times 3$ convolution receives $\mathbf{X}_3$ and shuffled-up features learned from low-resolution space and generates residual image $\mathbf{S}$~$\in$~$\mathbb{R}^{H\times W \times 3}$ to which UHD input image is added to obtain the restored image: $\mathbf{\hat{H}} = \mathbf{I} + \mathbf{S}$. 
Fig.~\ref{fig:Overall framework of our UHDformer}(a) shows the learning in high-resolution space.
\begin{figure*}[!t]
\centering
\begin{center}
\begin{tabular}{c}
\hspace{-2mm}\includegraphics[width=0.8\linewidth]{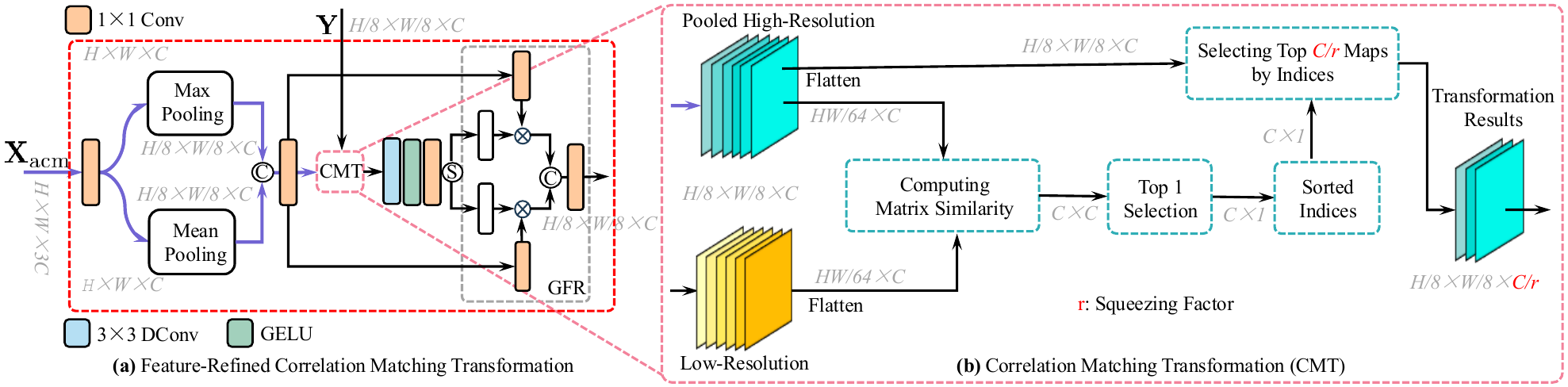} 
\end{tabular}
\vspace{-3mm}
\caption{
\textbf{(a) Feature-Refined Correlation Matching Transformation (FR-CMT) and (b) Correlation Matching Transformation (CMT)}.
FR-CMT contains a fusion of max-pooling and mean-pooling operations, followed by CMT, convolutions, and a Gated Feature Refinement (GFR) module.
CMT selects the top $C/r~(C~\text{denotes the number of channels;~}r\geq1~\text{is the squeezing factor which controls the squeezing level})$ channels from the pooling high-resolution features to replace low-resolution features.
GFR is used to refine the features for better representation.
}
\label{fig:Maximal Correlation Selection Module}
\end{center}
\vspace{-4mm}
\end{figure*}
%%%%%%%%%%%%%%%%%%%%%%%%%%%%%%%%%%%%%%%%%%%%%%%%%%%%%%%%%%%%%%%%%%%%%
% \\
% \noindent \textbf{Learning in Low-Resolution Space.} 
%
\subsubsection{Low-Resolution Space}
% {\flushleft \bf Learning in Low-Resolution Space.}
The low-resolution space, as shown in Fig.~\ref{fig:Overall framework of our UHDformer}(b), first receives the shuffled-down features $\mathbf{X}_{\text{low}}^{\text{in}}$~$\in$~$\mathbb{R}^{\frac{H}{8} \times \frac{W}{8} \times C}$ from $\mathbf{X}_{0}$ of the high-resolution space.
Then, $\mathbf{X}_{\text{low}}^{\text{in}}$ is sent to several Correlation Matching Transformation Transformer Blocks (CMT-TB), shown in Fig.~\ref{fig:Maximal Correlation Selection Module}, to learn the low-resolution features.
Meanwhile, each CMT-TB also matches the correlation between the modulated feature $\mathbf{X}_{\text{acm}}$ in ACM and attention/forward networks in CMT-TBs to generate more representative features to facilitate better restoration.
Finally, the learned feature $\mathbf{X}_{\text{low}}^{\text{out}}$~$\in$~$\mathbb{R}^{\frac{H}{8} \times \frac{W}{8} \times C}$ in the low-resolution space is sent to high-resolution space to participate in reconstructing the final recovery images. 

\subsubsection{Super-Resolution Space}
The SR contains an 8$\times$ shuffle up operation, a ConvNeXt, and a 3$\times$ convolution to super-resolve the low-resolution features, as shown in Fig.~\ref{fig:Overall framework of our UHDformer}(c).
The simple operation is experimentally demonstrated to have a positive effect on high restoration quality.

\subsubsection{Low-High Fusion and Reconstruction Space}
We use the LHFR to reconstruct the final image, as shown in Fig.~\ref{fig:Overall framework of our UHDformer}(d).
The LHFR receives the concatenated features from shuffled-up features from SR and features from LiHR, followed by a 1$\times$1 convolution, two ConvNeXt, and a 3$\times$3 convolution, to reconstruct the final image.

\subsection{Correlation Matching Transformation Transformer}
\begin{figure}[!t]
\centering
\begin{center}
\begin{tabular}{c}
\hspace{-2mm}\includegraphics[width=0.65\linewidth]{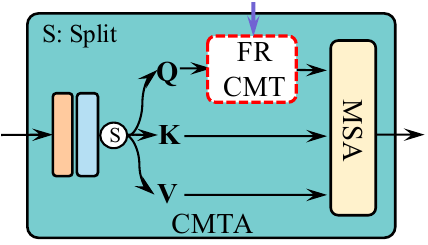} 
\end{tabular}
\vspace{-3mm}
\caption{
\textbf{Correlation Matching Transformation Attention}.
}
\label{fig:CMTA}
\end{center}
\vspace{-4mm}
\end{figure}

\begin{figure}[!t]
\centering
\begin{center}
\begin{tabular}{c}
\hspace{-2mm}\includegraphics[width=0.75\linewidth]{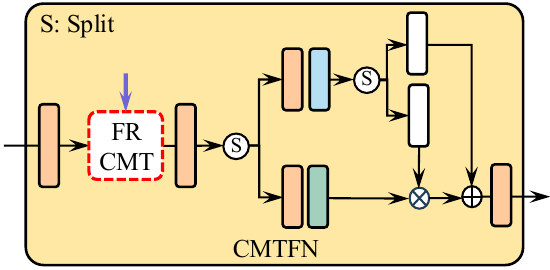} 
\end{tabular}
\vspace{-3mm}
\caption{
\textbf{Correlation Matching Transformation Feed-Forward Network.}
}
\label{fig:CMTN}
\end{center}
\vspace{-4mm}
\end{figure}

% \label{sec:Exact Correlation Matching Transformer Block}
%
To better improve the representation of the low-resolution features, we propose to build the transformation from high-resolution features to low-resolution ones to replace the low-resolution features by squeezing useless features of the high-resolution space.
To that end, we propose the Correlation Matching Transformation Transformer Block (CMT-TB), as shown in Fig.~\ref{fig:Overall framework of our UHDformer}(b), to provide low-resolution space with more representative features learned from multi-level high-resolution features to replace the low-resolution ones via the Feature-Refined Correlation Matching module shown in Fig.~\ref{fig:Maximal Correlation Selection Module}.
Each CMT-TB contains the Correlation Matching Transformation Attention (CMTA) as shown in Fig~\ref{fig:CMTA} and Correlation Matching Transformation Feed-Forward Network (CMTFN) as shown in Fig~\ref{fig:CMTN} to respectively explore the correlation matching transformation in both attention and feed-forward networks:
\begin{subequations} 
    \begin{align}
&\textrm{$\textbf{X}$}^{'} = \textit{CMTA}\Big(\textit{DyT}(\textrm{$\textbf{X}$}_{\text{low}}^{l-1}), \textrm{$\textbf{X}$}_{\text{acm}} \Big) + \textrm{$\textbf{X}$}_{\text{low}}^{l-1},
\\
&\textrm{$\textbf{X}$}_{\text{low}}^{l} = \textit{CMTFN}\Big(\textit{DyT}(\textrm{$\textbf{X}$}^{'}), \textrm{$\textbf{X}$}_{\text{acm}}\Big) + \textrm{$\textbf{X}$}^{'},~~l = 1, 2, \cdots, L.
\end{align}
\label{eq: ECM-TB}
\end{subequations}
where $\textrm{$\textbf{X}$}_{\text{low}}^{l}$ means the output features of $l^{\text{th}}$ CMT-TB;
Especially, $\textrm{$\textbf{X}$}_{\text{low}}^{0}$ is the $\textrm{$\textbf{X}$}_{\text{low}}^{\text{in}}$ and $\textrm{$\textbf{X}$}_{\text{low}}^{L}$ is the $\textrm{$\textbf{X}$}_{\text{low}}^{\text{out}}$;
$\textit{CMTA}(\cdot,\cdot)$ and $\textit{CMTN}(\cdot,\cdot)$ are respectively defined in Eq.~\eqref{eq:ECMAttention} and Eq.~\eqref{eq:ECMFFN};
$\textit{DyT}(\cdot)$ means the operation of Dynamic Tanh~\cite{zhu2025dyt}.
% \\
% \noindent \textbf{Correlation Matching Transformation Attention.} 
%
%%%%%%%%%%%%%%%%%%%%%%%%%%%%%%
\subsubsection{Correlation Matching Transformation Attention}
% \label{sec:Correlation Matching Transformation Attention}
The Query typically describes its relationship with others in the attention of Transformers~\cite{query}.
Hence, one more powerful query vector may significantly influence the results.
In this paper, we empower the query with more representative content by matching it with adjusted multi-level high-resolution features $\textrm{$\textbf{X}$}_{\text{acm}}$ for better restoration.
To that end, we propose the Correlation Matching Transformation Attention (CMTA), as shown in Fig.~\ref{fig:CMTA}, to improve the query representation to better conduct attention.
Specifically, the CMTA first generates the \emph{query} (\textbf{Q}), \emph{key} (\textbf{K}), and \emph{value} (\textbf{V}) projections from the normalized low-resolution features $\textrm{$\textbf{X}$}_{\text{low}}^{l}$ via $1$$\times$$1$ point-wise convolution $W_{p}$ and $3$$\times$$3$ depth-wise convolution $W_{d}$, and then conducts the DualCMT between \textbf{Q} and ACM features $\textrm{$\textbf{X}$}_{\text{acm}}$.
Next, the CMTA conducts the attention~\cite{Zamir2021Restormer}:
\begin{equation}
\begin{array}{ll}
\textit{CMTA}\left(\textit{DyT}\left(\textrm{$\textbf{T}_{1}$}\right),  \textrm{$\textbf{T}_{2}$}\right) =  \mathcal{A}\Big(\textit{FR-CMT}(\textbf{Q}, \textrm{$\textbf{T}_{2}$}), \textbf{K}, \textbf{V}\Big),
\end{array}
\label{eq:ECMAttention}
\end{equation}
where
\begin{equation}
\begin{array}{ll}
\textbf{Q}, \textbf{K}, \textbf{V} = \textit{Split}\Big(W_{d}W_{p}(\textit{DyT}\left(\textrm{$\textbf{T}_{1}$}\right)\Big),
\end{array}
\label{eq:ECMAttention111}
\end{equation}
where $\textrm{$\textbf{T}_{1}$}$ is the $\textrm{$\textbf{X}$}_{\text{low}}^{l}$ in Eq.~\ref{eq: ECM-TB} and $\textrm{$\textbf{T}_{2}$}$ is the $\textrm{$\textbf{X}$}_{\text{acm}}$ in Eq.~\ref{eq: ECM-TB};
$\textit{FR-CMT}$ means the Feature-Refined Correlation Matching Transformation module;
$\mathcal{A}\left(\hat{\textbf{Q}}, \hat{\textbf{K}}, \hat{\textbf{V}}\right) = \hat{\textbf{V}}\cdot \textrm{Softmax$\left( \hat{\textbf{K}} \cdot \hat{\textbf{Q}}/\alpha \right)$}$; 
Here, $\alpha$ is a learnable scaling parameter to control the magnitude of the dot product of $\hat{\mathbf{K}}$ and $\hat{\mathbf{Q}}$ before applying the softmax function;
$\textit{Split}$ denotes the split operation;
Similar to the MSA in \cite{Zamir2021Restormer}, we divide the number of channels into `heads' and learn separate attention maps.

\subsubsection{Correlation Matching Transformation Feed-Forward Network}
% \label{sec:Correlation Matching Transformation Forward Network}
%%%%%%%%%%%%%%%%%%%%%%%%%%%%%%
Similarly to CMTA, we note that improving the feature representation for feed-forward networks would better restore images.
Hence, we propose the Correlation Matching Transformation Feed-Forward Network (CMTFN), which is used to learn more useful features from the high-resolution features of FR-CMT for better recovery.
The first generates the features from attention features with normalization and $1$$\times$$1$ convolution and $3$$\times$$3$ depth-wise convolution, and then conducts FR-CMT between the generated features and features of ACM:
\begin{equation}
\begin{array}{ll}
\textit{CMTFN}\Big(\textit{DyT}\left(\textrm{$\textbf{T}_{1}$}\right),  \textrm{$\textbf{T}_{2}$}\Big) = \Phi\Big(\textit{FR-CMT}\big(W_{p}\left(\textit{DyT}\left(\textrm{$\textbf{T}_{1}$}\right)\right),  \textrm{$\textbf{T}_{2}$} \big)\Big),
\end{array}
\label{eq:ECMFFN}
\end{equation}
where $\textrm{$\textbf{T}_{1}$}$ is the $\textrm{$\textbf{X}$}^{'}$ in Eq.~\eqref{eq: ECM-TB} and $\textrm{$\textbf{T}_{2}$}$ is the $\textrm{$\textbf{X}$}_{\text{acm}}$ in Eq.~\eqref{eq: ECM-TB}, while $\Phi(\cdot)$ denotes the operation:
\begin{equation}
\begin{array}{ll}
\Phi(\textrm{$\textbf{T}_{3}$}) = \varphi\left(W_{p}\left(W_{p}\textrm{$\textbf{T}_{4}$}\right)\right) * \textrm{$\textbf{T}_{5}$} + \textrm{$\textbf{T}_{6}$}),
\end{array}
\label{eq:ECMFFN}
\end{equation}
where
\begin{subequations} 
    \begin{align}
    & \textrm{$\textbf{T}_{4}$}, \textrm{$\widetilde{\textbf{T}}_{4}$} = \textit{Split}\left(W_{p}(\textrm{$\textbf{T}_{3}$})\right),
     \\
     & \textrm{$\textbf{T}_{5}$}, \textrm{$\textbf{T}_{6}$} = \textit{Split}\left(W_{d}W_{p}(\textrm{$\widetilde{\textbf{T}}_{4}$})\right).
    \end{align}
    \label{eq:t456}
\end{subequations}
Here $\varphi$ is the GELU activation.
% \\
%%%%%%%%%%%%%%%%%%%%%%%%%%%%%%
% \noindent \textbf{Dual-Branch Correlation Matching Module.} 
\subsection{Feature-Refined Correlation Matching Transformation}
% \label{sec:Dual-path Correlation Matching Transformation}
%
The Feature-Refined Correlation Matching Transformation (FR-CMT), as shown in Fig.~\ref{fig:Maximal Correlation Selection Module}, aims to match the features between high-resolution features and low-resolution ones to provide the low-resolution space with more representative features for better restoration.
Given the learned features $\mathbf{X}_{\text{acm}}$~$\in$~$\mathbb{R}^{H \times W \times 3C}$ from ACM, we first exploit $1$$\times$$1$ convolution to generate channel-reduced features \textrm{$\textbf{Y}$}~$\in$~$\mathbb{R}^{H \times W \times C}$, and then utilize pooling fusion including max-pooling and mean-pooling to reduce the spatial dimension to $\textrm{$\textbf{Y}$}_{\text{max}}$ and $\textrm{$\textbf{Y}$}_{\text{mean}}$~$\in$~$\mathbb{R}^{\frac{H}{8} \times \frac{W}{8} \times C}$.
Then, the $\textrm{$\textbf{Y}$}_{\text{max}}$ and $\textrm{$\textbf{Y}$}_{\text{mean}}$ are matched with the features $\textrm{$\textbf{Y}$}$~$\in$~$\mathbb{R}^{\frac{H}{8} \times \frac{W}{8} \times C}$ in the Transformer of low-resolution space via Correlation Matching Transformation $\mathcal{M}(\cdot,\cdot)$ which is computed as the process in Eq.~\eqref{eq:exact-matching}:
\begin{subequations} 
    \begin{align}
    & \hat{\textrm{$\textbf{Y}$}} =W_{p}\left(\mathbf{X}_{\text{acm}}\right),
     \\
     & \textrm{$\textbf{Y}$}_{\text{max}} = \textit{Max-Pool}\left(\hat{\textrm{$\textbf{Y}$}}\right); \textrm{$\textbf{Y}$}_{\text{mean}} = \textit{Mean-Pool}\left(\hat{\textrm{$\textbf{Y}$}}\right),
    \\
     & \textrm{$\textbf{Y}^{\text{selected}}$} = \mathcal{M}\left(\textrm{$\textbf{Y}$}_{\text{fusion}}, \textrm{$\textbf{Y}$}\right)
    \end{align}
    \label{eq:dual-ecm}
\end{subequations}
where $\textrm{$\textbf{Y}$}_{\text{fusion}}=W_{p}(\textit{Concat}[\textbf{Y}_{\text{max}}, \textbf{Y}_{\text{mean}}])$ means the fusion between max-pooling and mean-pooling features; $\textit{Concat}$ means the concatenation operations.

The Correlation Matching Transformation first computes the matrix similarity ($\textit{MatSim}(\cdot, \cdot)$) at the channel dimension between given two tensors $\mathbf{R}_{1}$~$\in$~$\mathbb{R}^{\frac{H}{8} \times \frac{W}{8} \times C}$ and $\mathbf{R}_{2}$~$\in$~$\mathbb{R}^{\frac{H}{8} \times \frac{W}{8} \times C}$ after flattening to $\mathbf{\tilde{R}}_{1}$~$\in$~$\mathbb{R}^{\frac{HW}{64} \times C}$ and $\mathbf{\tilde{R}}_{2}$~$\in$~$\mathbb{R}^{\frac{HW}{64} \times C}$ to generate similarity matrix $\mathbf{M}$~$\in$~$\mathbb{R}^{C \times C}$.
Then, we select the Top-1 vector $\mathbf{D}$~$\in$~$\mathbb{R}^{C \times 1}$ for each tensor in $\mathbf{M}$ and sort the values of $\mathbf{D}$ to produce the sorted indices $\mathbf{S}$~$\in$~$\mathbb{R}^{C \times 1}$.
Finally, we select $\textit{Select}_{C/r}(\cdot|\cdot)$ the Top $C/r$ ($r\geq1$ means the squeezing factor which controls the squeezing level) features $\mathbf{Y}^{\text{selected}}$~$\in$~$\mathbb{R}^{\frac{H}{8} \times \frac{W}{8} \times \frac{C}{r}}$ from feature maps $\mathbf{R}_{1}$ by the sorted indices $\mathbf{S}$:
\begin{subequations} 
    \begin{align}
    &\mathbf{M} = \textit{MatSim}\left(\mathbf{\tilde{R}}_{1}, \mathbf{\tilde{R}}_{2}\right),
    \\
    &\mathbf{D}=\textit{Top}_{1}(\mathbf{M}); \mathbf{S} = \textit{Sort}\left(\mathbf{D}\right),
    \\
    &\mathbf{Y}^{\text{selected}} = \textit{Select}_{C/r}\left(\mathbf{R}_{1}|\mathbf{S}\right).
    \end{align}
    \label{eq:exact-matching}
\end{subequations}

Finally, with the selected features $\textrm{$\textbf{Y}$}^{\text{selected}}$~$\in$~$\mathbb{R}^{\frac{H}{8} \times \frac{W}{8} \times \frac{C}{r}}$, we then use several operations on it and then use Gated Feature Refinement $\textit{GFR}(\cdot)$ to refine the features:
\begin{subequations} 
    \begin{align}
    &\textrm{$\textbf{Y}$}_{1}^{\text{selected}}, \textrm{$\textbf{Y}$}_{2}^{\text{selected}} = \textit{Split}\Big(W_{p}\varphi W_{d}\textrm{$\textbf{Y}$}^{\text{selected}}\Big), 
    \\
    &\textit{GFR}\Big(\textrm{$\textbf{Y}$}_{1}^{\text{selected}}, \textrm{$\textbf{Y}$}_{2}^{\text{selected}}\Big) = W_{p}\Big(\textit{Concat}[\textrm{$\textbf{Z}$}_{1}, \textrm{$\textbf{Z}$}_{2}]\Big).
    \end{align}
    \label{eq:gfr}
\end{subequations}
where $\textrm{$\textbf{Z}$}_{1} = \textrm{$\textbf{Y}$}_{1}^{\text{selected}}\odot \textrm{$\textbf{Y}$}_{\text{fusion}}$ and $\textrm{$\textbf{Z}$}_{2} = \textrm{$\textbf{Y}$}_{2}^{\text{selected}}\odot \textrm{$\textbf{Y}$}_{\text{fusion}}$.

\subsection{Adaptive Channel Modulator}
The Adaptive Channel Modulator (ACM), as illustrated in Fig.~\ref{fig:Overall framework of our UHDformer}(d), aims to adaptively modulate high-resolution features to better balance the importance of channel-wise features, enabling to provide more representative features for low-resolution space.
Given the multi-level high-resolution tensors $\{\mathbf{X}_1, \mathbf{X}_2, \mathbf{X}_3\}$~$\in$~$\mathbb{R}^{H \times W \times C}$, we first concatenate $\textit{concat}$ them to generate a wider tensor $\mathbf{X}_{\text{concat}}$~$\in$~$\mathbb{R}^{H \times W \times 3C}$.
Then, we use $1$$\times$$1$ convolution and $3$$\times$$3$ depth-wise convolution to expand the concatenated features $\mathbf{X}_{\text{concat}}$ to wider features and split the features into two tensors: $\mathbf{Z}_1$~$\in$~$\mathbb{R}^{H \times W \times 3C}$ and $\mathbf{Z}_2$~$\in$~$\mathbb{R}^{H \times W \times 3C}$.
Next, we conduct the softmax $\mathcal{S}_{\text{channel}}$ at channel dimension on $\mathbf{Z}_1$ to obtain channel-wise weights and finally conduct element-wise addition and element-wise multiplication among $\mathbf{X}_{\text{concat}}$, $\mathbf{Z}_1$, and $\mathbf{Z}_2$:
\begin{figure}[!t]
\centering
\begin{center}
\begin{tabular}{c}
\hspace{-2mm}\includegraphics[width=0.75\linewidth]{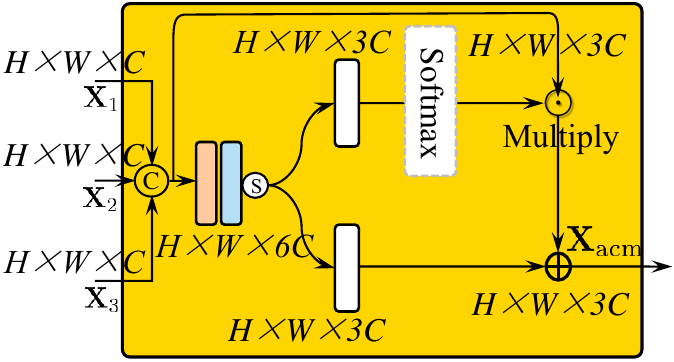} 
\end{tabular}
\vspace{-3mm}
\caption{
\textbf{Adaptive Channel Modulator}.
}
\label{fig:acm}
\end{center}
\vspace{-4mm}
\end{figure}
\begin{equation} 
   \begin{array}{ll}
    \textit{ACM}\big(\mathbf{X}_1, \mathbf{X}_2, \mathbf{X}_3\big) = \mathbf{X}_{\text{concat}} {\odot} \mathcal{S}_{\text{channel}}\left(\mathbf{Z}_1\right) +\mathbf{Z}_2,
    \end{array}
    \label{eq:Adaptive Channel Modulator}
\end{equation}
where
\begin{subequations} 
    \begin{align}
    &\mathbf{X}_{\text{concat}} = \textit{Concat}\big[\mathbf{X}_1, \mathbf{X}_2, \mathbf{X}_3\big],
    \\
    &\mathbf{Z}_1, \mathbf{Z}_2 = \textit{Split}\Big(W_{d}W_{p}\big(\mathbf{X}_{\text{concat}}\big) \Big).
    \end{align}
    \label{eq:Adaptive Channel Modulator}
\end{subequations}

\subsection{Learning Strategy} 
% \label{sec:Learning Strategy}
%%%%%%%%%%%%%%%%%%%%%%%%%%%%%%%%%%%%%%%%%
As we use both LHFR and SR to reconstruct the high-resolution images, the loss of our UHDformer++ contains two parts, including $\mathcal{L}_{LHFR}$ and $\mathcal{L}_{SR}$:
\begin{equation}
\begin{aligned}
\mathcal{L}_{\mathrm{UHDformer++}}
=&\underbrace{\|\mathbf{\hat{H}}-\mathbf{H}\|_1
+\lambda_{freq}\|\mathcal{F}(\mathbf{\hat{H}})-\mathcal{F}(\mathbf{H})\|_1}_{\mathcal{L}_{\mathrm{LHFR}}} \\
&+\underbrace{\lambda_{re\_sr}\|\mathbf{\hat{H}}-\mathbf{H}\|_1
+\lambda_{freq\_sr}\|\mathcal{F}(\mathbf{\hat{H}})-\mathcal{F}(\mathbf{H})\|_1}_{\mathcal{L}_{\mathrm{SR}}},
\end{aligned}
\label{eq:loss_uhdformerpp}
\end{equation}
where $\mathcal{F}$ denotes the Fast Fourier transform.
For each loss, it constrains the restored results $\mathbf{\hat{H}}$ with Ground Truth (GT) images $\mathbf{H}$, including image reconstruction loss for pixel recovery and frequency loss for detail enhancement~\cite{cho2021rethinking_mimo}.
In our experiments, the loss weights are empirically set to 
$\lambda_{freq}=0.05$,
$\lambda_{re\_sr}=0.5$, 
and $\lambda_{freq\_sr}=1$. 
The analysis of these loss weights is presented in Tab.~\ref{tab:Effect on weight of loss function}.
%%%%%%%%%%%%%%%%%%%%%%%%%%%%%%%%%%%%%

%
\section{Experiments}
In this section, we conduct extensive experiments to demonstrate the effectiveness of our proposed UHDformer++.
We evaluate our proposed UHDformer++ on benchmarks for $5$ UHD image restoration tasks: \textbf{(a)} low-light image enhancement, \textbf{(b)} image dehazing, \textbf{(c)} image deblurring, \textbf{(d)} image deraining, and \textbf{(e)} image desnowing. 
We train separate models for different UHD tasks. 

\begin{table}[!t]
\caption{\textbf{Datasets Statistics}.
}
\tablestyle{5pt}{1.1}
\label{tab: Datasets Statistics.} 
\begin{tabular}{l|cccccc}
\shline
 \textbf{Dataset} &\textbf{Training samples} & \textbf{Testing samples} & \textbf{Resolution}
\\
\shline
UHD-LL &2,000 &150 & 3,840 $\times$ 2,160 \\
UHD-Haze&2,290 &231 & 3,840 $\times$ 2,160 \\
UHD-Blur&1,964 &300 & 3,840 $\times$ 2,160 \\
UHD-Rain&3,000 &200 & 3,840 $\times$ 2,160 \\
UHD-Snow&3,000 &200 &3,840 $\times$ 2,160 \\
\shline
\end{tabular}
\end{table}
\subsection{Datasets} 
For the UHD low-light image enhancement experiment, we adopt UHD-LL~\cite{Li2023ICLR_uhdfour} as the evaluation benchmark. 
For UHD image dehazing and deblurring, we curate new benchmarks by re-sampling from the datasets of~\cite{Zheng_uhd_CVPR21} and~\cite{Deng_2021_ICCV_uhd_denlurring}, designated as UHD-Haze and UHD-Blur, respectively. 
For UHD image deraining and desnowing, we adopt the datasets collected by Wang et al.~\cite{wang2025ultra} to assess model performance. 
The statistics of all five datasets are summarized in Tab.~\ref{tab: Datasets Statistics.}.
In addition to training on UHD images, we further investigate the generalization capability of our model through cross-dataset evaluation, wherein models are trained on commonly used general image restoration datasets and subsequently tested on UHD images for low-light enhancement, dehazing, and deblurring. 
Specifically, we employ the well-established LOL~\cite{retinexnet_wei_bmvc18}, SOTS-ITS~\cite{RESIDE_dehazingbenchmarking_tip2019}, and GoPro~\cite{gopro2017} datasets for low-light enhancement, dehazing, and deblurring, respectively, using their training splits to optimize the deep models. 
For deraining and desnowing, following the protocol of UHDDIP~\cite{wang2025ultra}, models are trained on UHD images and evaluated directly on the UHD test set.
%
%%%%%%%%%%%%%%%%%%%%%%%%%%%%%%%%%%%%%%
\begin{table*}[!t]
\caption{\textbf{Low-light image enhancement}. 
% \
%
The best and the second results are marked in \textbf{bold} and \underline{underline}, respectively.
Full Size means that if the method can directly infer a full UHD image on one 3090 GPU.
}
\vspace{-2mm}
\label{tab:Low-light image enhancement.} 
\tablestyle{2pt}{1}
\centering
\begin{tabular}{l|ccc|ccc|ccc|c}
\shline
 \multicolumn{1}{c|}{\multirow{3}{*}{\textbf{Method}}} & \multicolumn{6}{c|}{\textbf{Metrics}}&\multicolumn{3}{c|}{\multirow{2}{*}{\textbf{Computational Complexity}}}& \multirow{3}{*}{\textbf{Full Size}} 
 \\
  \cline{2-7}
&\multicolumn{3}{c|}{\textbf{Full Reference}}&\multicolumn{3}{c|}{\textbf{No-Reference}}&\multicolumn{3}{c|}{}
 \\
 \cline{2-10}
&\textbf{PSNR (dB)~$\uparrow$} 
&\textbf{SSIM~$\uparrow$}
&\textbf{LPIPS~$\downarrow$}
&\textbf{PI~$\downarrow$} 
&\textbf{NIQE~$\downarrow$} 
&\textbf{MIUSIQ~$\uparrow$}
&\textbf{Params. (M)~$\downarrow$} 
&\textbf{FLOPs (G)~$\downarrow$}
&\textbf{Time (s)~$\downarrow$}
\\
\shline
 \multicolumn{10}{c}{\textbf{Training Set on LOL}}\\
\shline
SwinIR~\cite{liang2021swinir}&17.900 &0.7379&0.5217&6.5902&8.5836&28.3047&11.5 &11992.17& 1.41&\XSolidBrush  \\
Restormer~\cite{Zamir2021Restormer}&19.728 & 0.7703 &0.4566&6.5099& 8.3787&30.2802&26.1 &2255.85& 1.27&\XSolidBrush\\
Uformer~\cite{wang2021uformer}&18.168 & 0.7201 &0.5593&7.1781&9.8494&29.5157& 20.6 &657.45& 0.43&\XSolidBrush  \\
LLFlow~\cite{wang2021llflow}&19.596 &0.7333 &0.4606&\underline{4.3283}&\underline{4.3790}&\underline{35.2491}& 17.4  &450.52& 0.11& \CheckmarkBold\\
LLformer~\cite{LLformer}&21.440 & \underline{0.7763}&0.4528&6.6185&8.5009&28.9955& 13.2  &221.64&1.10& \XSolidBrush\\
UHDFour~\cite{Li2023ICLR_uhdfour}& 14.771 & 0.3760  &0.7608&\textbf{4.0232}&\textbf{4.3363}&31.7022&  17.5 &75.63&\textbf{0.02}&\CheckmarkBold \\
UHDformer~\cite{aaai24wang_UHDformer}&\underline{22.615} & 0.7754  &\underline{0.4241}&4.3995&4.4026&33.8086& \textbf{0.3393} &\underline{51.63}& 0.16& \CheckmarkBold\\
\textbf{UHDformer++ (Ours)}&\textbf{23.796}&\textbf{0.7795}&\textbf{0.4232}&4.3740&4.3909&\textbf{36.3919}&\underline{0.3962}&\textbf{49.65}&\underline{0.12}& \CheckmarkBold\\
\shline
 \multicolumn{10}{c}{\textbf{Training Set on UHD-LL}}\\
\shline
SwinIR~\cite{liang2021swinir}&21.165 &0.8450 &0.3995&7.2767&8.3811 &21.6384& 11.5   &11992.17& 1.41&\XSolidBrush  \\
Restormer~\cite{Zamir2021Restormer}&21.536 & 0.8437   &0.3608&7.0071&8.2193&23.2947& 26.1 &2255.85& 1.27&\XSolidBrush\\
Uformer~\cite{wang2021uformer}&21.303  & 0.8233   &0.4013&6.9021&8.8050&21.8972& 20.6  &657.45& 0.43&\XSolidBrush  \\
LLformer~\cite{LLformer}&24.065 & 0.8580 &0.3516&7.4717&8.2977&21.1605& 
 13.2  &221.64&1.10&\XSolidBrush\\
UHDFour~\cite{Li2023ICLR_uhdfour}&26.226 & 0.9000  &\textbf{0.2194}&\textbf{6.2463}&\textbf{6.0459}&28.5186& 17.5 &75.63&\textbf{0.02}& \CheckmarkBold \\
UHDformer~\cite{aaai24wang_UHDformer}&\underline{27.113} & \underline{0.9271}&\underline{0.2240}&6.5761&7.0614&\underline{35.8419}& \textbf{0.3393}&\underline{51.63}& \underline{0.16}& \CheckmarkBold\\
\textbf{UHDformer++ (Ours)}&\textbf{27.256}&\textbf{0.9308}&0.2304&\underline{6.5199}&\underline{6.9849}&\textbf{37.1814}&\underline{0.3962}&\textbf{49.65}&\textbf{0.12}& \CheckmarkBold\\
\shline
\end{tabular}
% \vspace{-2mm}
\end{table*}

%%%%%%%%%%%%%%%%%%%%%%%
\begin{figure*}[!t]
% \footnotesize
\centering
\begin{center}
\begin{tabular}{ccccccccc}
\includegraphics[width=0.2\linewidth]{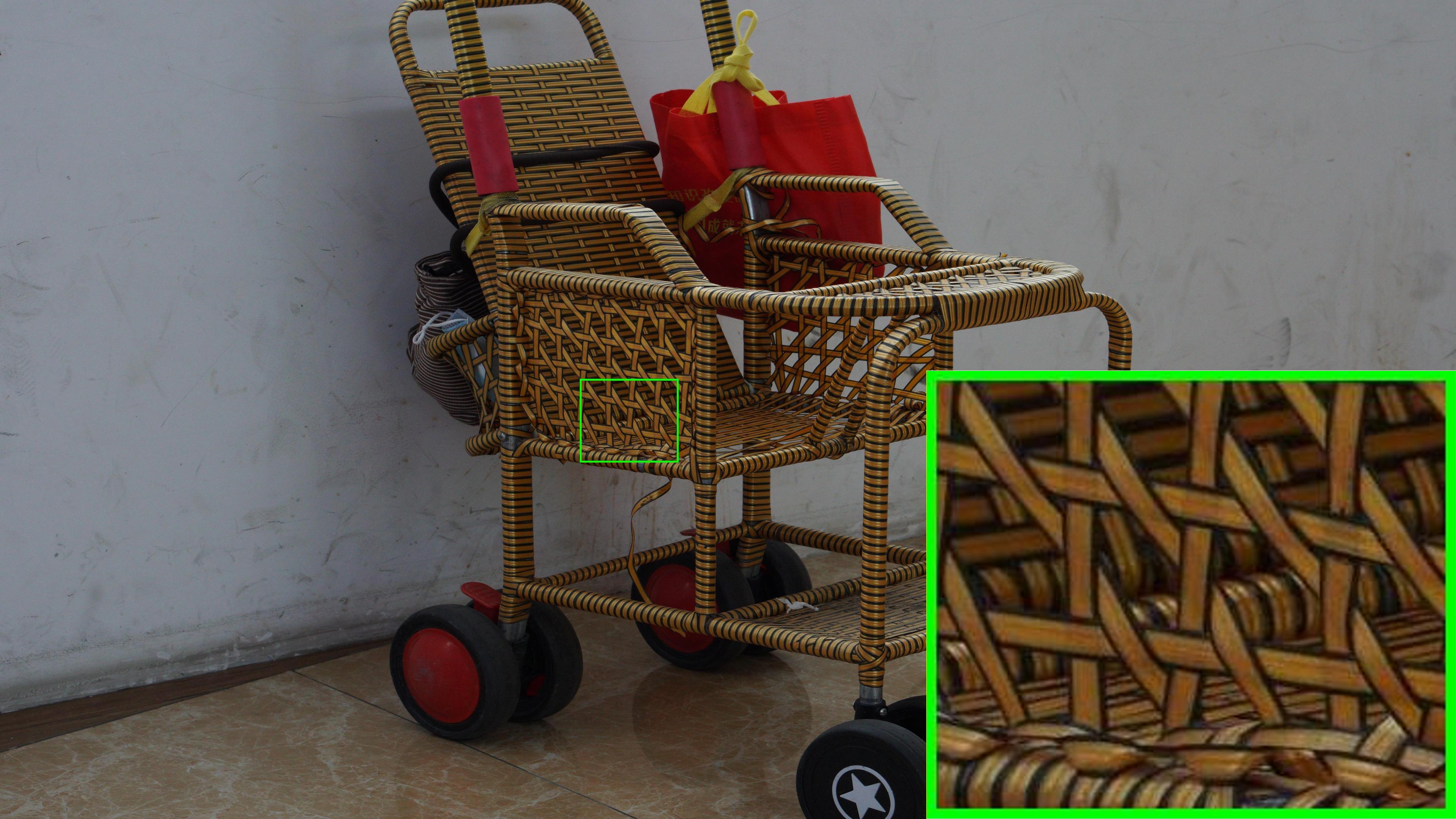} &\hspace{-4.75mm}
\includegraphics[width=0.2\linewidth]{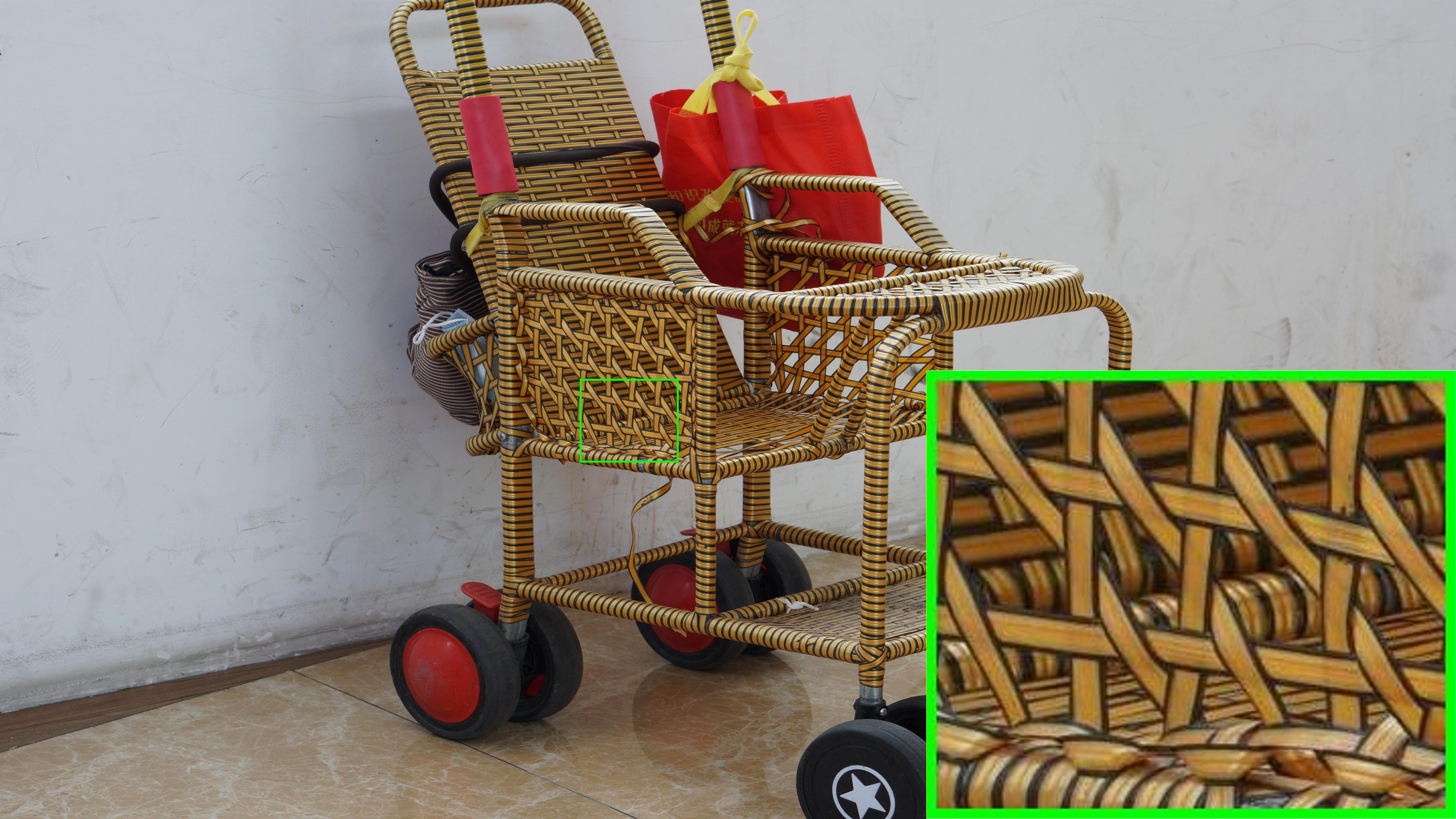} &\hspace{-4.75mm}
\includegraphics[width=0.2\linewidth]{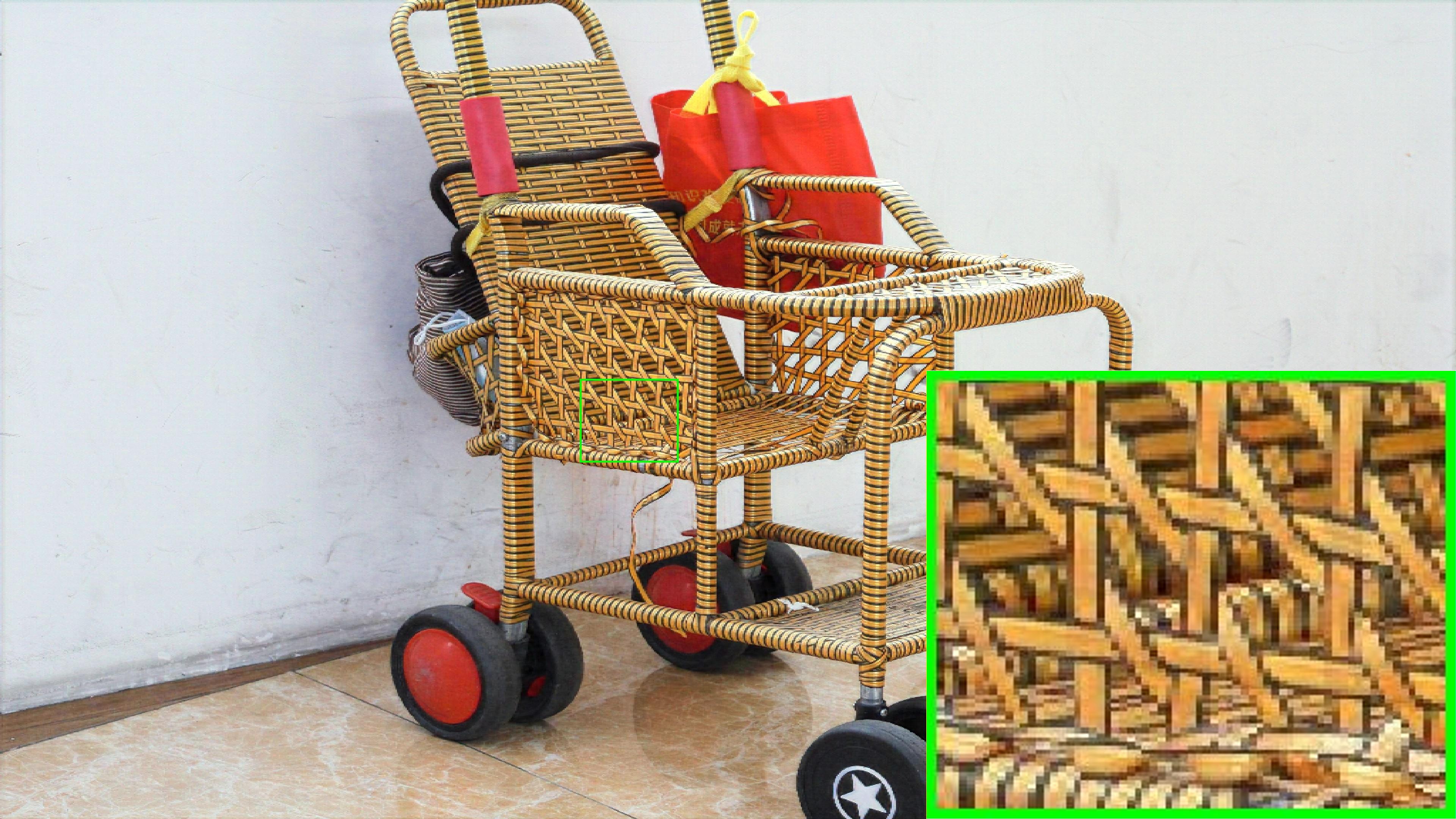} 
&\hspace{-4.75mm}
\includegraphics[width=0.2\linewidth]{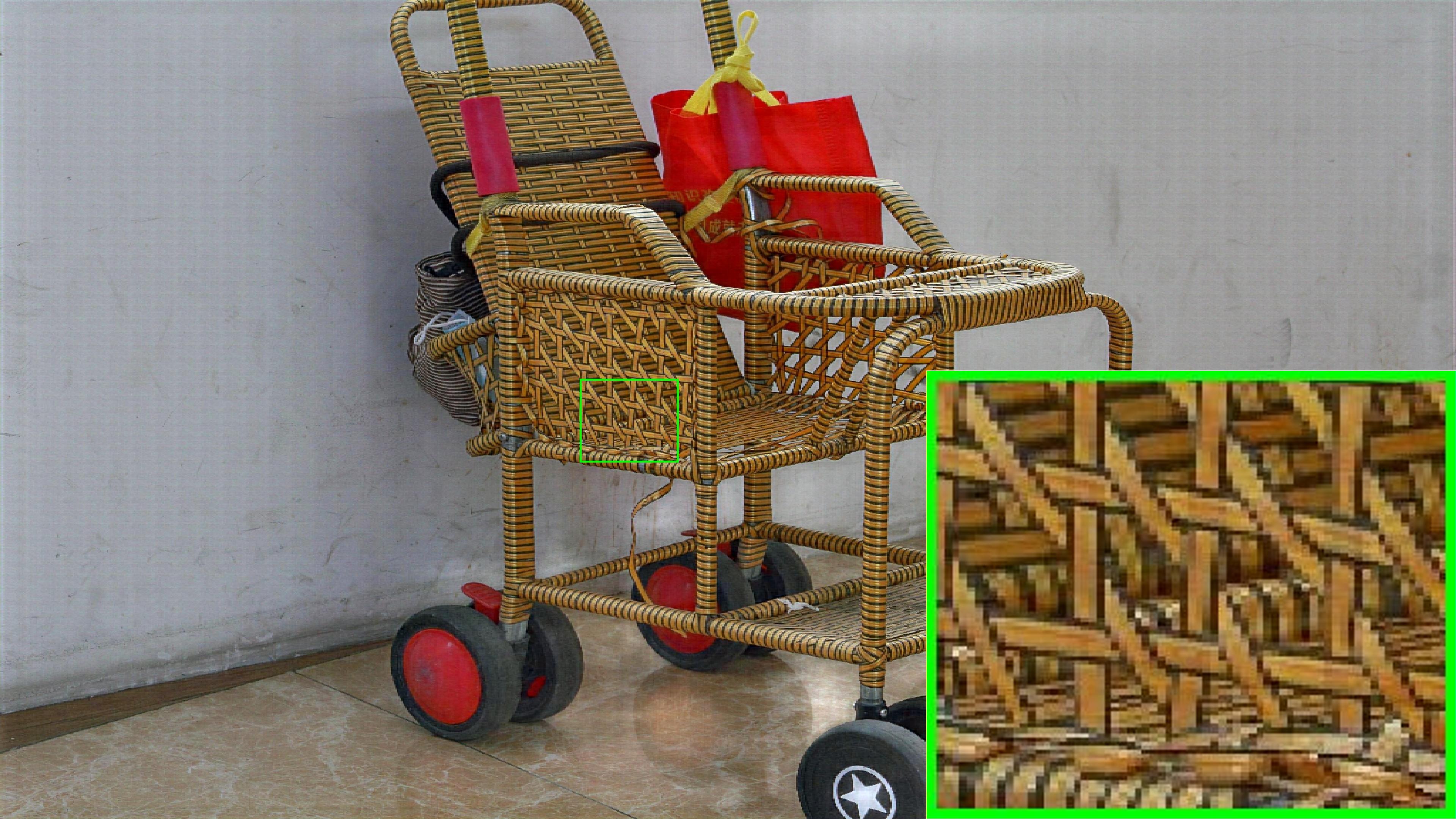} 
\\
 (a) Input&\hspace{-4.75mm} (b) GT&\hspace{-4.75mm} (c) Restormer~\cite{Zamir2021Restormer} &\hspace{-4.75mm} (d) Uformer~\cite{wang2021uformer}
\\
\includegraphics[width=0.2\linewidth]{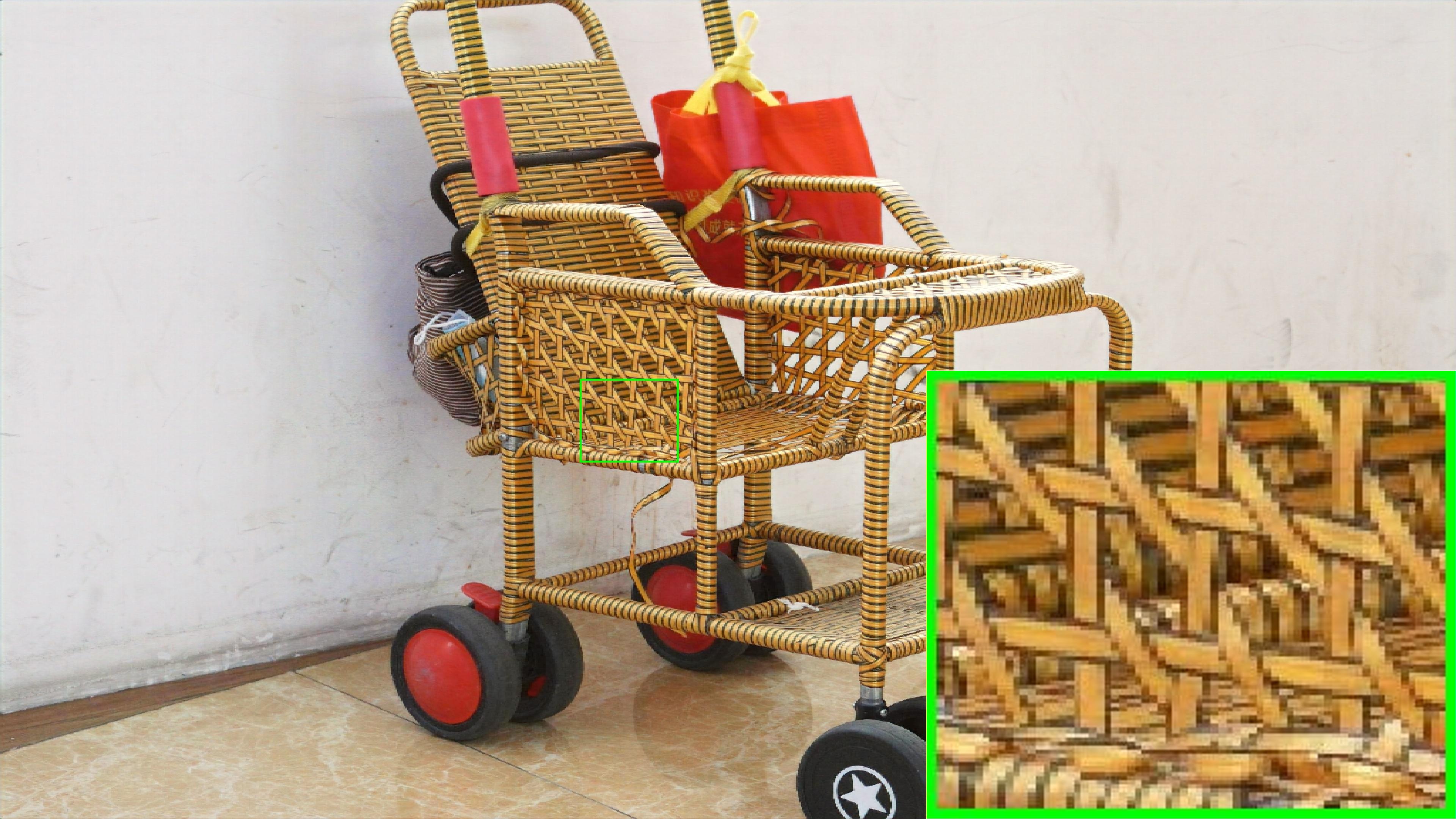} &\hspace{-4.75mm}
\includegraphics[width=0.2\linewidth]{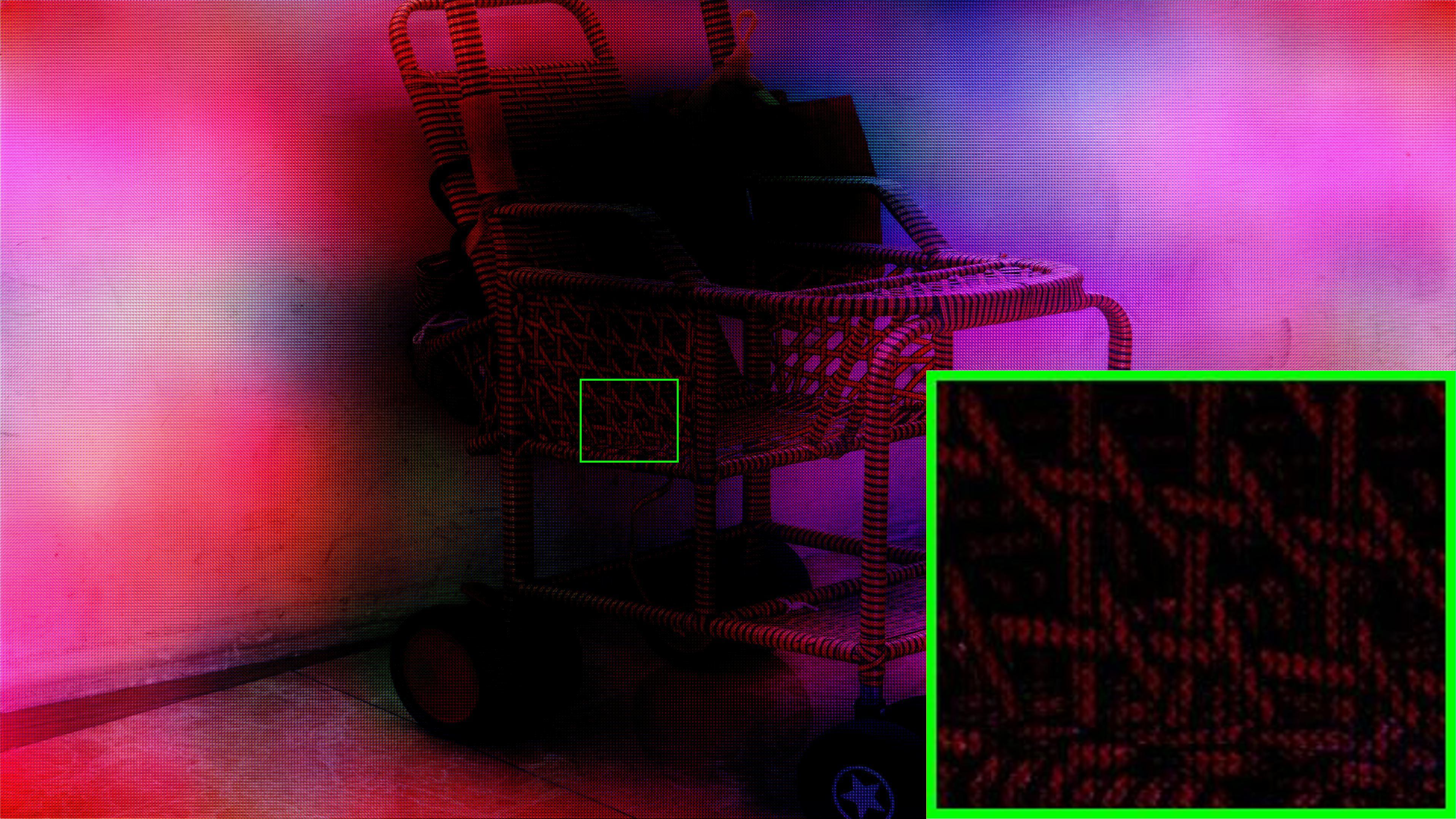} &\hspace{-4.75mm}
\includegraphics[width=0.2\linewidth]{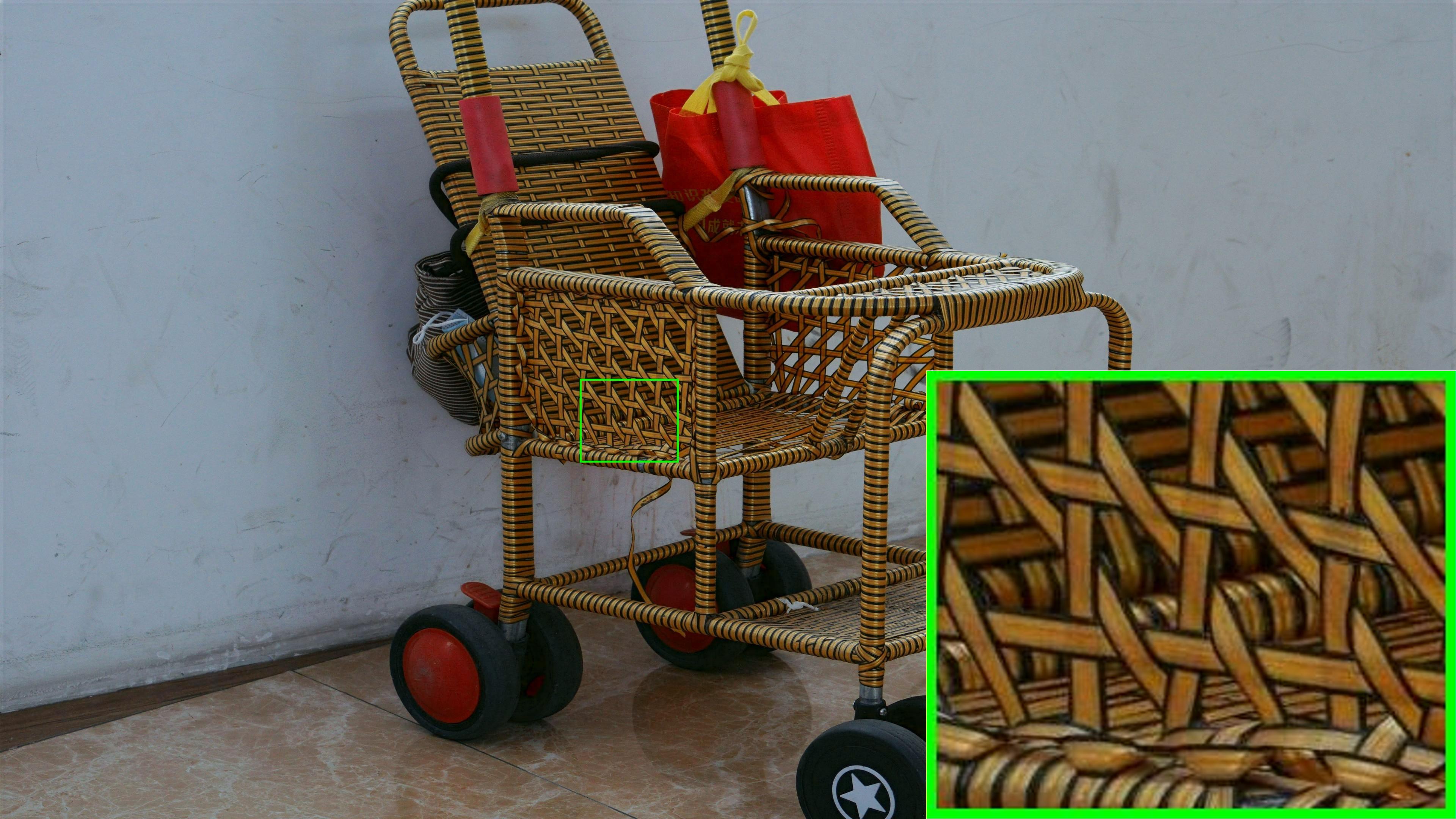}&\hspace{-4.75mm}
\includegraphics[width=0.2\linewidth]{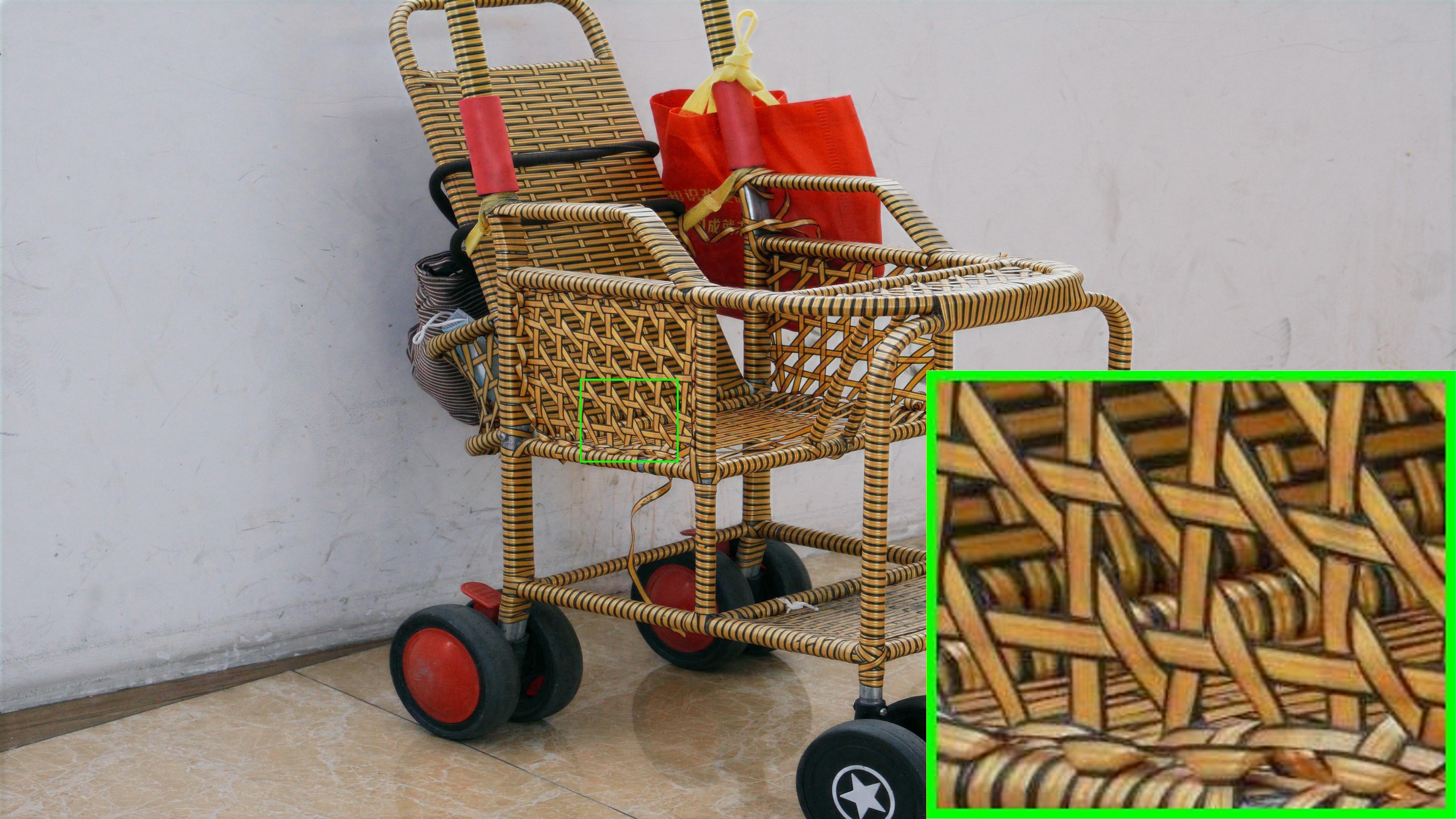}
\\
(e) LLformer~\cite{LLformer}&\hspace{-4.75mm} (f) UHDFour~\cite{Li2023ICLR_uhdfour}&\hspace{-4.75mm} (g) UHDformer~\cite{aaai24wang_UHDformer} &\hspace{-4.75mm}  (h) \textbf{UHDformer++}
\end{tabular}
\vspace{-2mm}
\caption{\textbf{Low-light image enhancement on UHD-LL \textit{trained on the LOL dataset}~\cite{retinexnet_wei_bmvc18}.}
}
\label{fig:Low-light image enhancement on UHD-LL}
\end{center}
\vspace{-4mm}
\end{figure*}
%%%%%%%%%%%%%%%%%%%%%%%%%%%%%%%%%%%%%%%%%%%%%%%%%%%%%%%%%%%%
\begin{figure*}[!t]
% \footnotesize
\centering
\begin{center}
\begin{tabular}{ccccccccc}
\includegraphics[width=0.2\linewidth]{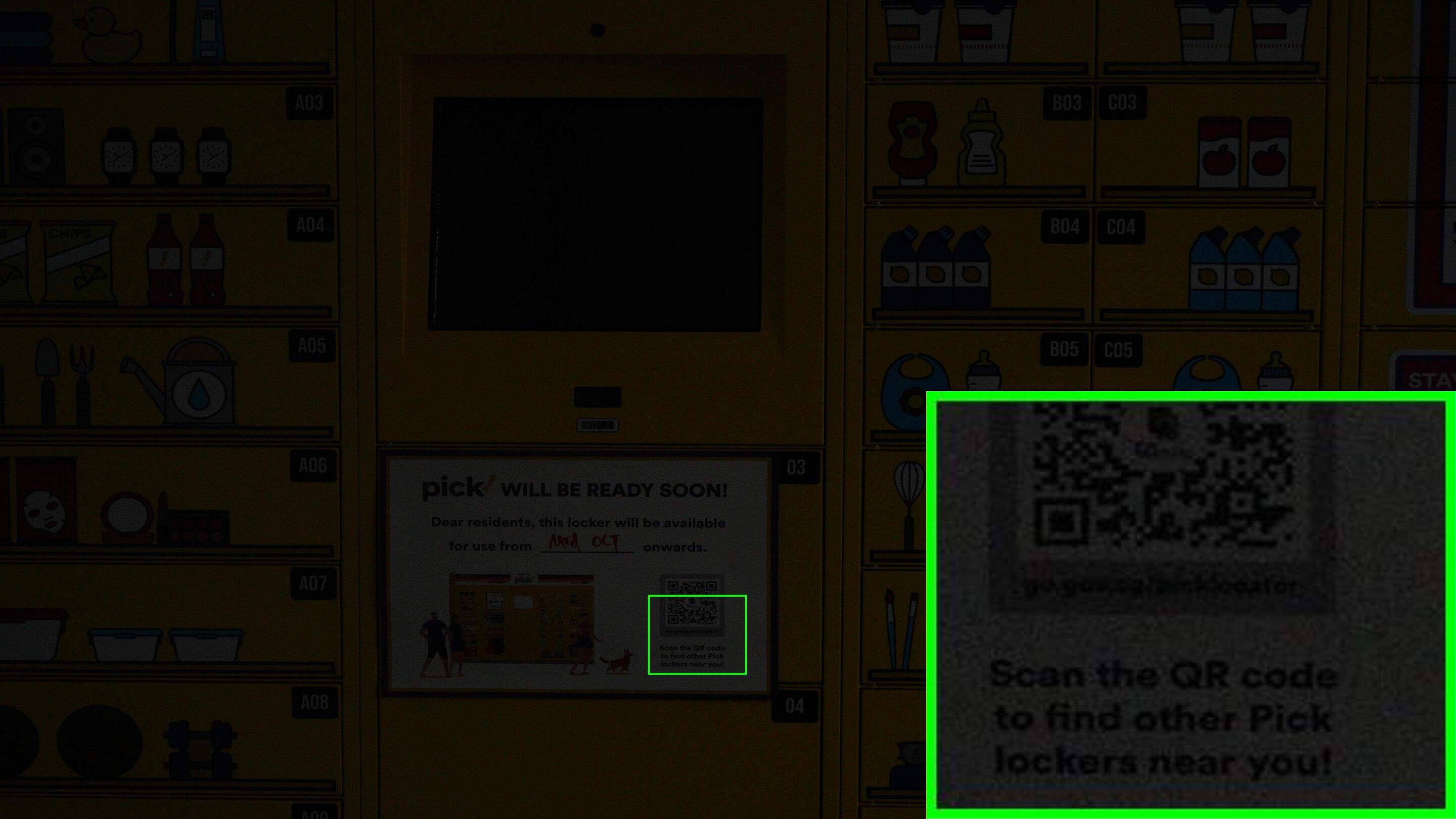} &\hspace{-4.75mm}
\includegraphics[width=0.2\linewidth]{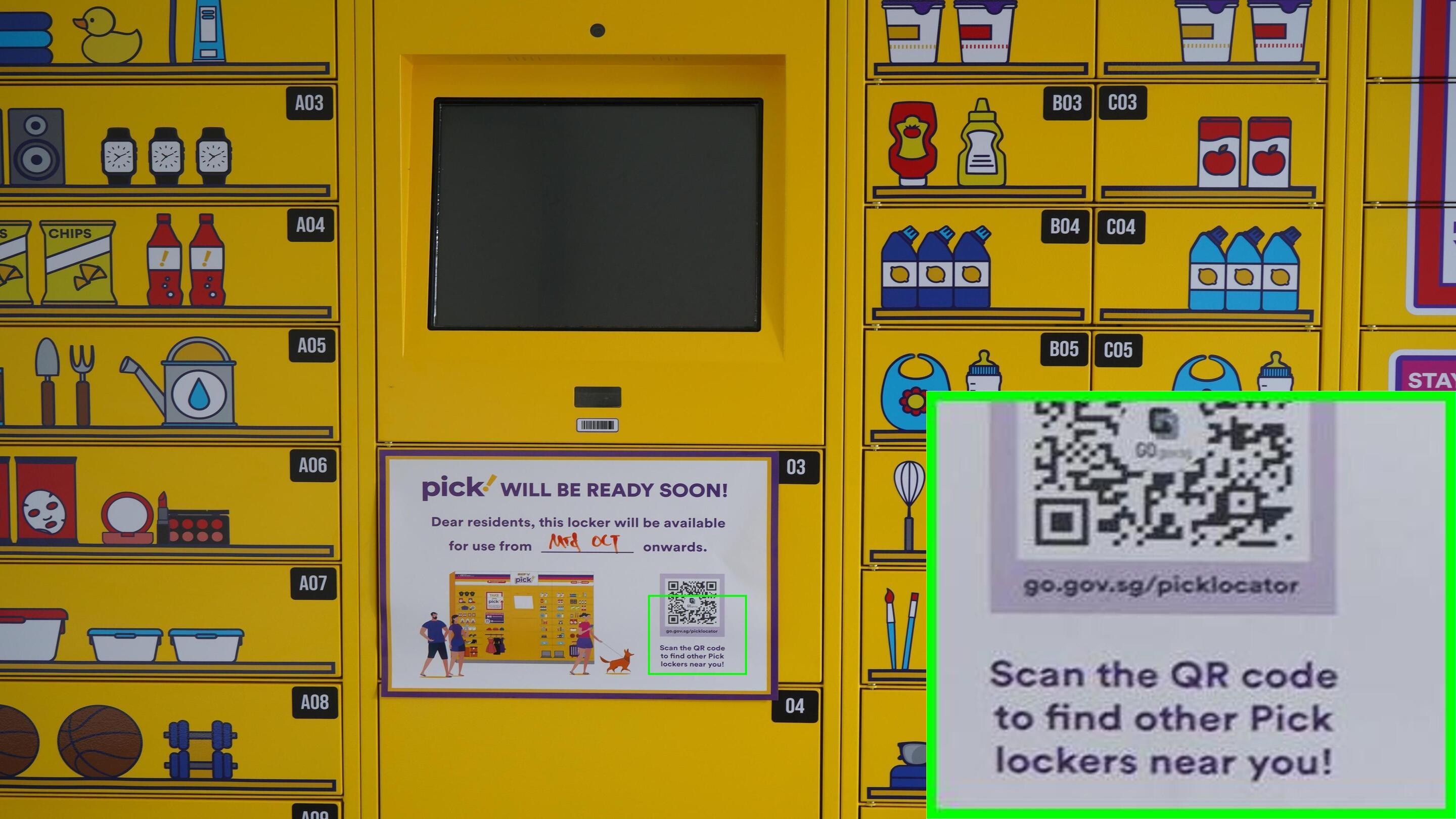} &\hspace{-4.75mm}
\includegraphics[width=0.2\linewidth]{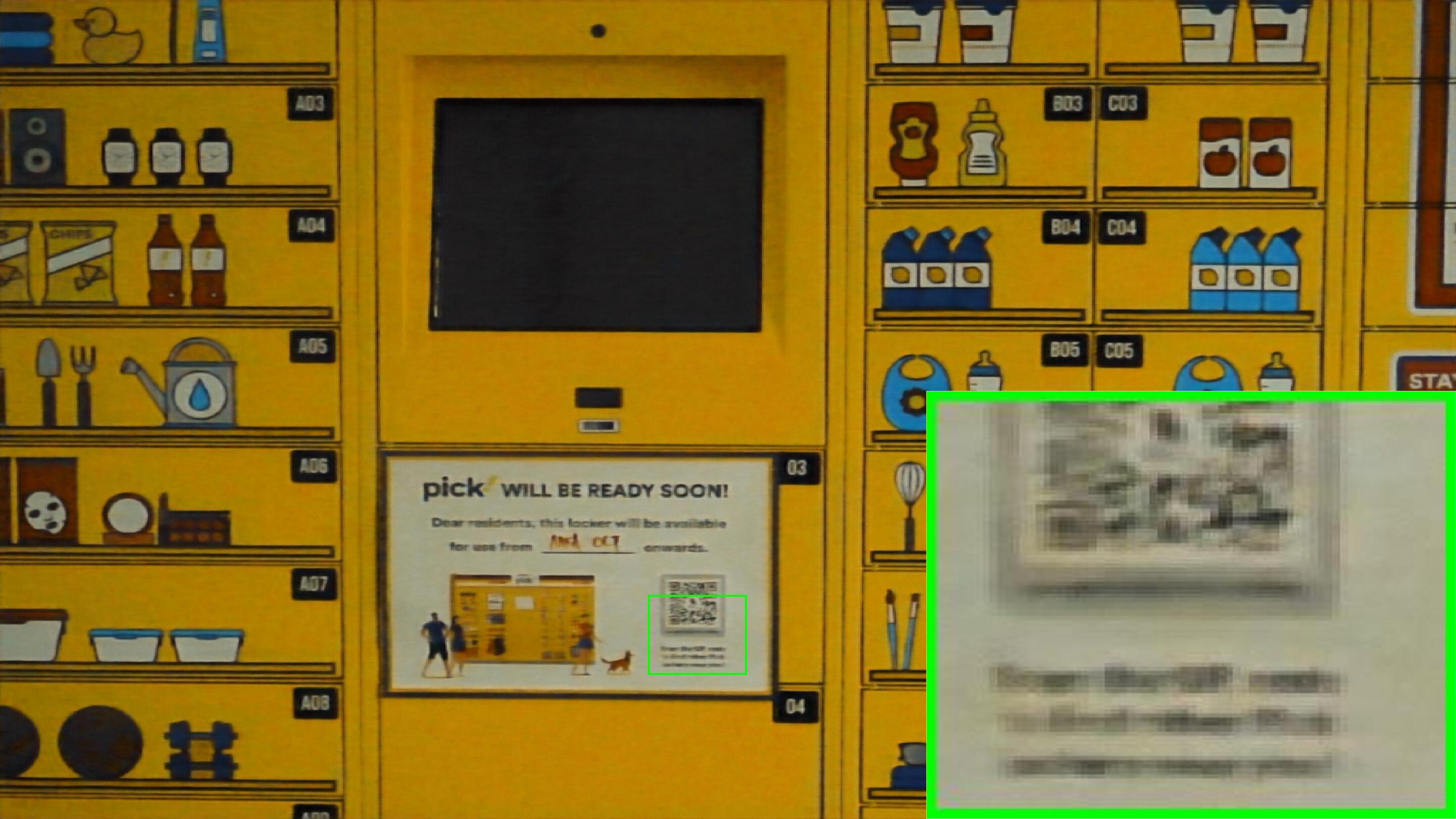} 
&\hspace{-4.75mm}
\includegraphics[width=0.2\linewidth]{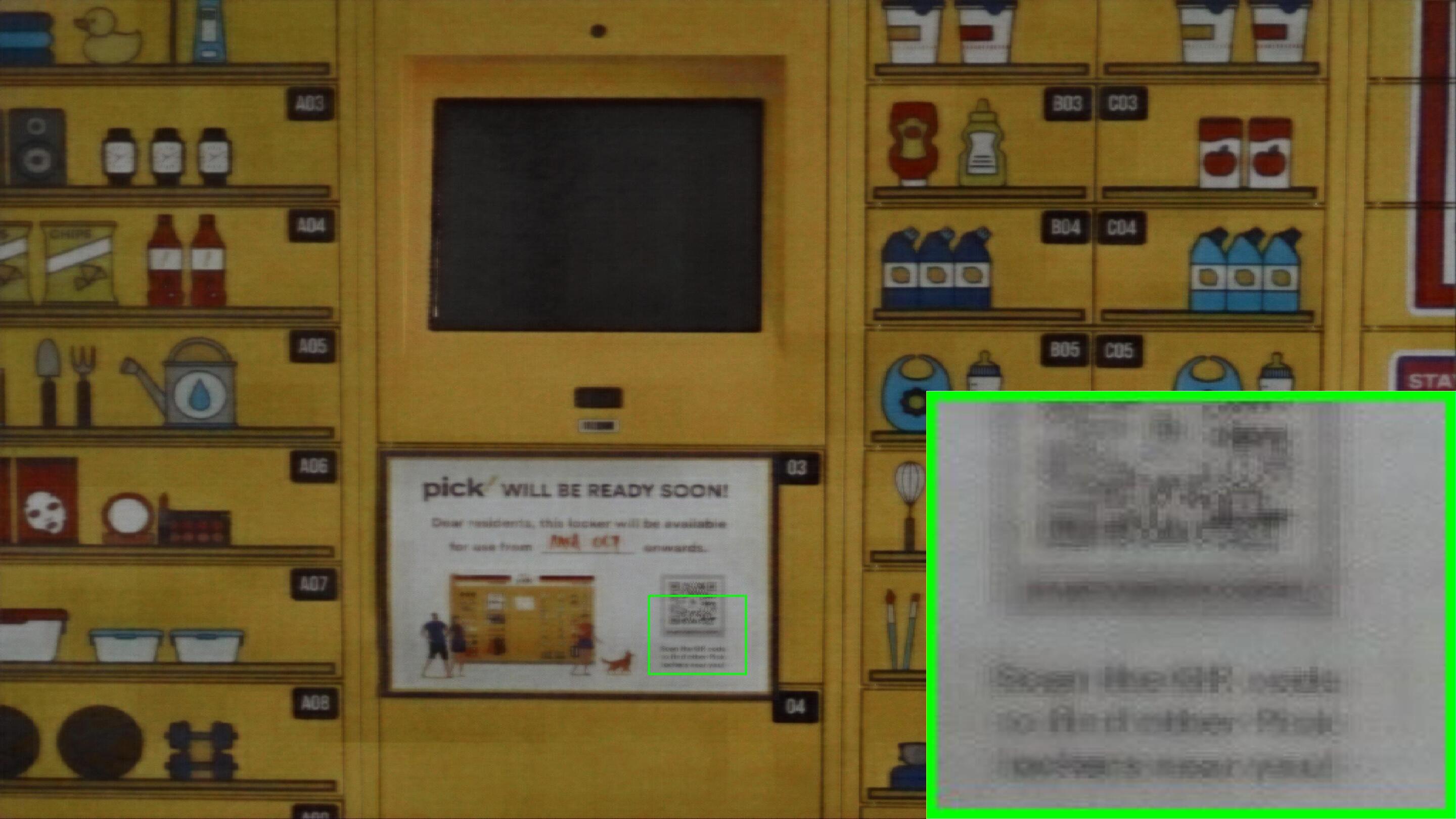} 
\\
(a) Input&\hspace{-4.75mm} (b) GT&\hspace{-4.75mm} (c) Restormer~\cite{Zamir2021Restormer} &\hspace{-4.75mm}(d) Uformer~\cite{wang2021uformer}
\\
\includegraphics[width=0.2\linewidth]{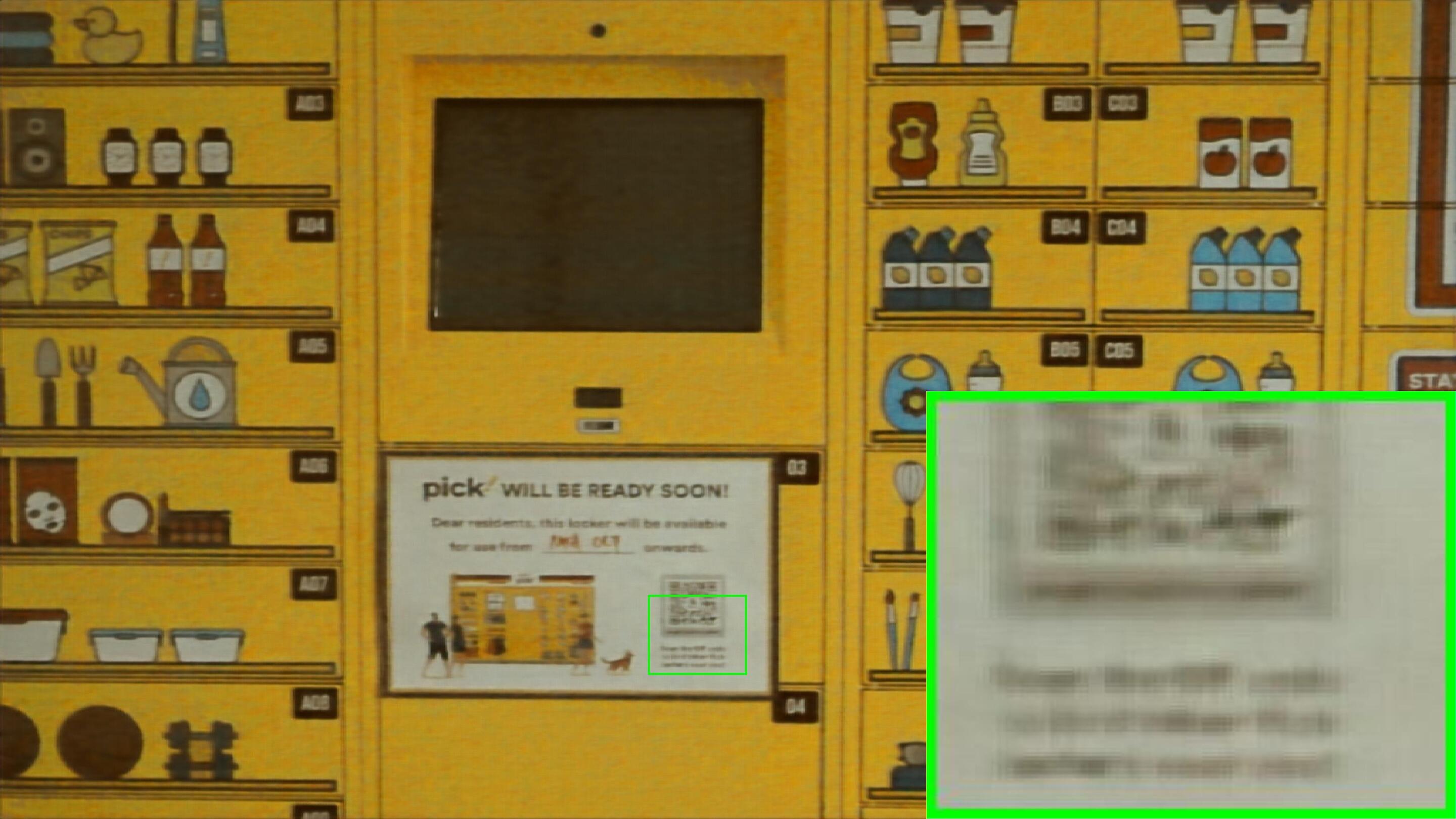} &\hspace{-4.75mm}
\includegraphics[width=0.2\linewidth]{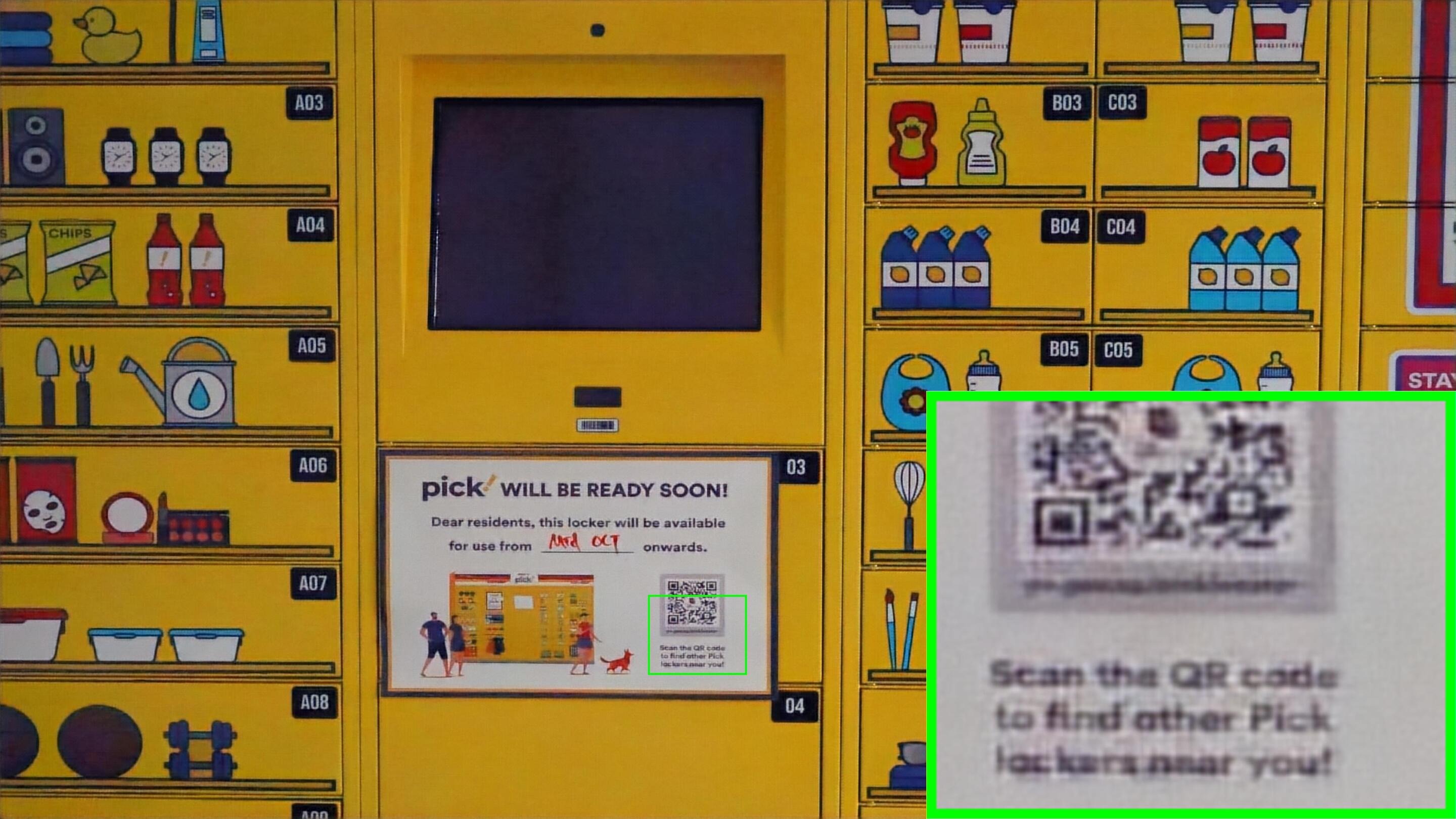} &\hspace{-4.75mm}
\includegraphics[width=0.2\linewidth]{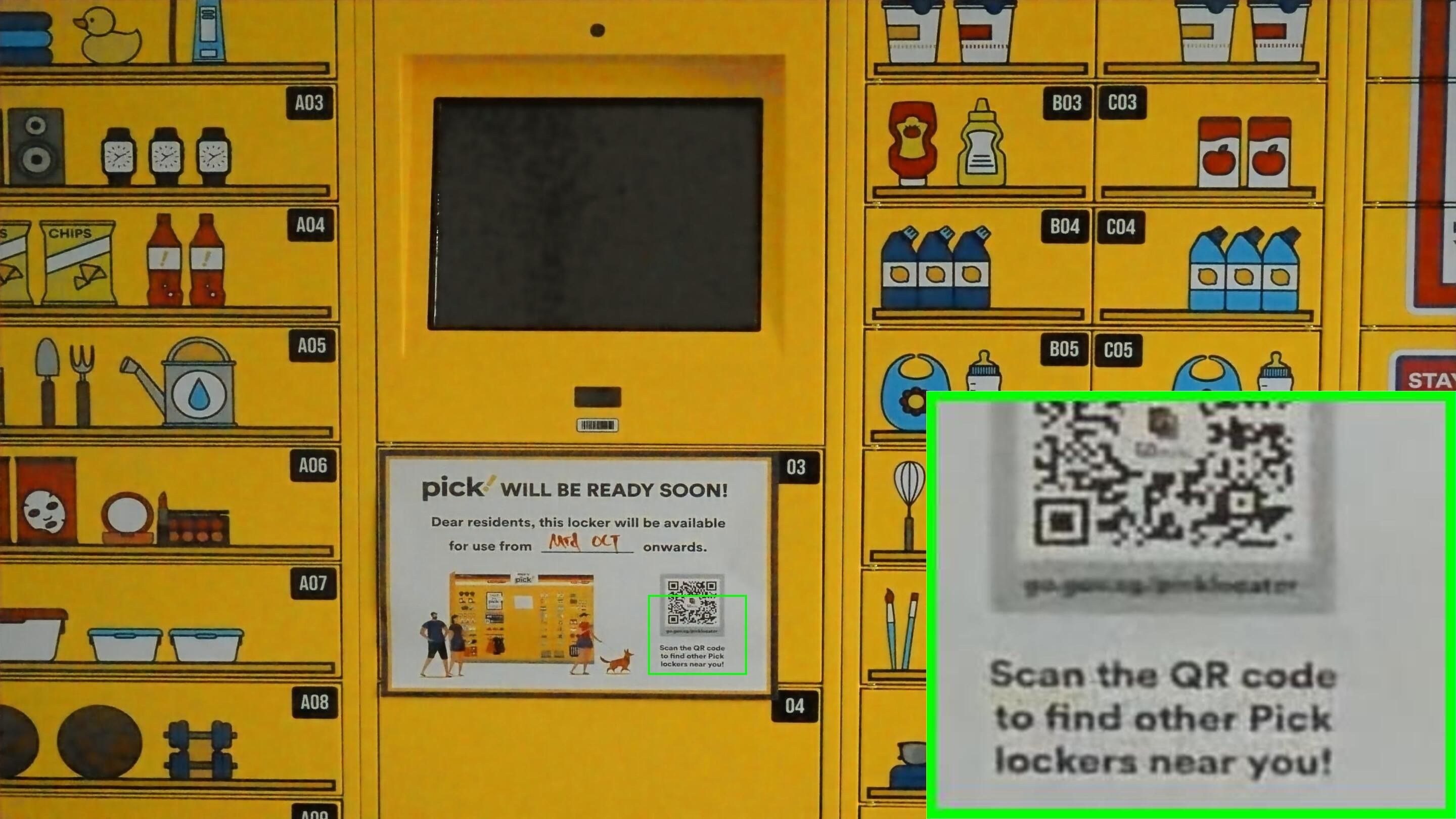}
&\hspace{-4.75mm}
\includegraphics[width=0.2\linewidth]{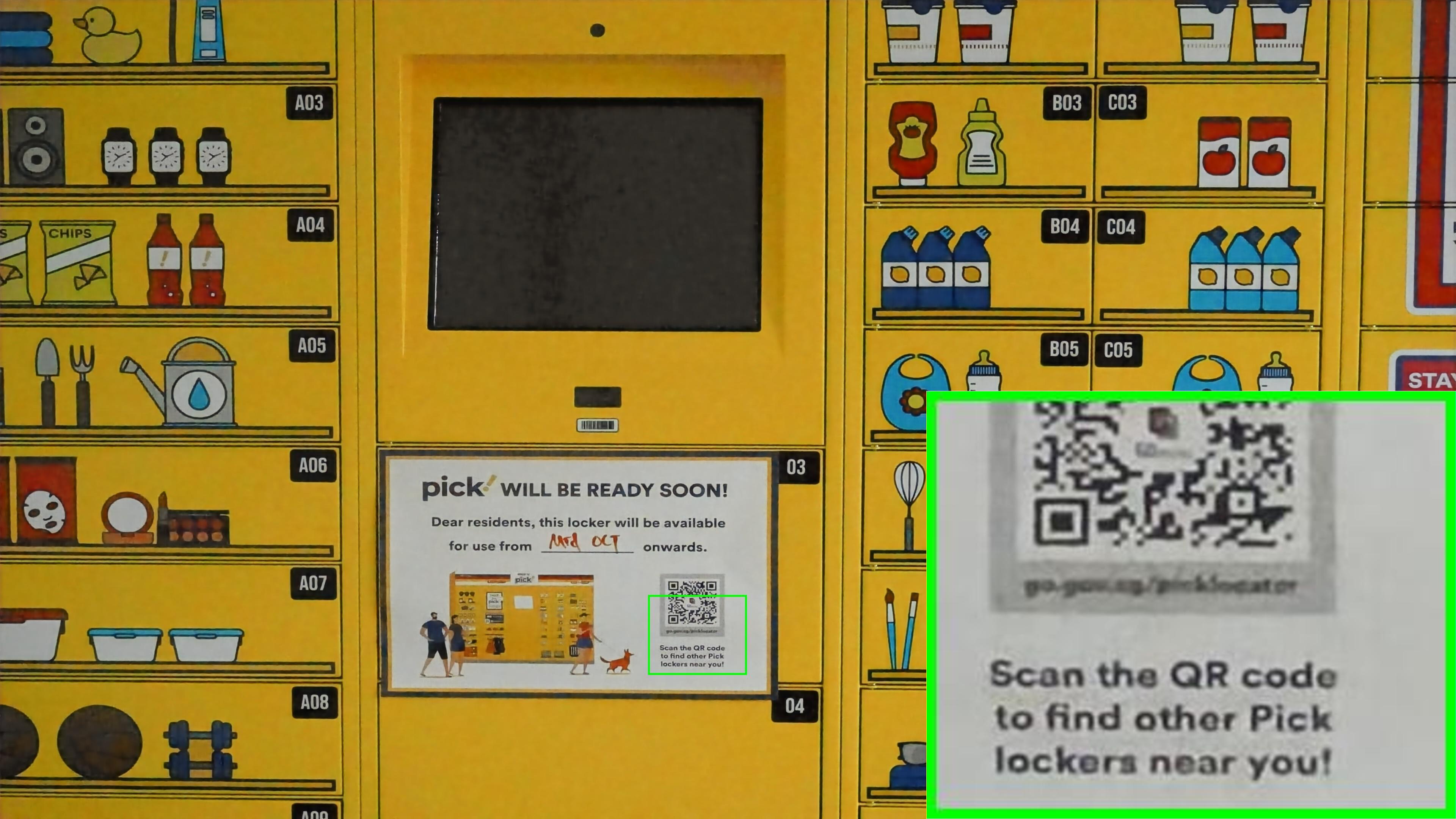}
\\
(e)  LLformer~\cite{LLformer} &\hspace{-4.75mm} (f) UHDFour~\cite{Li2023ICLR_uhdfour}&\hspace{-4.75mm} (g) UHDformer~\cite{aaai24wang_UHDformer} &\hspace{-4.75mm} (h) \textbf{UHDformer++}
\end{tabular}
\vspace{-2mm}
\caption{\textbf{Low-light image enhancement on UHD-LL \textit{trained on the UHD-LL dataset}.}
% \textbf{UHDformer++} is able to generate results with finer structures.
}
\label{fig:Low-light image enhancement on UHD-LL}
\end{center}
\vspace{-3mm}
\end{figure*}

%%%%%%%%%%%%%%%%%%%%%%%%%%%%%%%%%%%
\begin{table*}[!t]
\caption{\textbf{Image dehazing}. 
% \textbf{UHDformer} family with the fewest parameters significantly advances state-of-the-art methods.
}
\vspace{-2mm}
\label{tab:Image dehazing.} 
\tablestyle{2.25pt}{1}
\centering
\begin{tabular}{l|ccc|ccc|ccc|c}
\shline
 \multicolumn{1}{c|}{\multirow{3}{*}{\textbf{Method}}} & \multicolumn{6}{c|}{\textbf{Metrics}}&\multicolumn{3}{c|}{\multirow{2}{*}{\textbf{Computational Complexity}}}& \multirow{3}{*}{\textbf{Full Size}} 
 \\
  \cline{2-7}
&\multicolumn{3}{c|}{\textbf{Full Reference}}&\multicolumn{3}{c|}{\textbf{No-Reference}}&\multicolumn{3}{c|}{}
 \\
 \cline{2-10}
&\textbf{PSNR (dB)~$\uparrow$} 
&\textbf{SSIM~$\uparrow$}
&\textbf{LPIPS~$\downarrow$}
&\textbf{PI~$\downarrow$} 
&\textbf{NIQE~$\downarrow$} 
&\textbf{MIUSIQ~$\uparrow$}
&\textbf{Params. (M)~$\downarrow$} 
&\textbf{FLOPs (G)~$\downarrow$}
&\textbf{Time (s)~$\downarrow$}
\\
\shline
 \multicolumn{10}{c}{\textbf{Training Set on SOTS-ITS}}\\
\shline
GridNet~\cite{grid_dehaze_liu}&14.783 &0.8466&0.2518&\underline{4.4077}&\textbf{4.0874}&35.6738&0.96 &343.92 &0.10 & \CheckmarkBold   \\
MSBDN~\cite{msbdn_cvpr20_dong}&15.043 &\underline{0.8570}&0.2709&4.5582&4.4640&35.5302&31.4&664.87&0.09& \CheckmarkBold    \\
%SwinIR~\cite{liang2021swinir}&17.900 &0.7379&&&&&11.5M (-97\%) &&   \\
Restormer~\cite{Zamir2021Restormer}&13.875 & 0.6405 &0.4477&5.2136&6.1129&35.4127& 26.1 &2255.85& 1.27&\XSolidBrush \\
Uformer~\cite{wang2021uformer}&15.264 & 0.6724 &0.4288&5.1385&6.1000&35.4127& 20.6 &657.45& 0.43&\XSolidBrush  \\
DehazeFormer-B~\cite{DehazeFormer} &  14.551 &  0.6710  &0.4100&5.2742&6.1963&\textbf{36.3794}& 2.5 &375.40&0.43  &\XSolidBrush\\
UHD~\cite{Zheng_uhd_CVPR21}&11.708&0.6569&0.5948&5.5382&5.2163&24.6878&34.5 &113.46&\textbf{0.03}& \CheckmarkBold  \\
UHDformer~\cite{aaai24wang_UHDformer}&\underline{15.325}&0.8560 &\textbf{0.2294}&4.5272&4.4382&34.2755& \textbf{0.3393}&\underline{51.63}& 0.16& \CheckmarkBold\\
%UHDPromer&16.927 &0.8666&0.2034&&&&0.7430M&32.56&0.07\\
\textbf{UHDformer++ (Ours)}&\textbf{15.675}&\textbf{0.8591}&\underline{0.2406}&\textbf{4.3740}&\underline{4.3361}&\underline{35.8625}&\underline{0.3962}&\textbf{49.65}&\underline{0.12}& \CheckmarkBold\\
\shline
 \multicolumn{10}{c}{\textbf{Training Set on UHD-Haze}}\\
\shline
UHD~\cite{Zheng_uhd_CVPR21}&17.628 & 0.8791 &0.3593&5.4557&\textbf{4.9918}& 27.9019& 34.5 &113.46&\textbf{0.03}& \CheckmarkBold  \\
%Dehamer~\cite{liang2021swinir}&17.900 &0.7379&&&&& 132.4M (-97\%) &&  \\
Restormer~\cite{Zamir2021Restormer}& 12.718 &  0.6930 &0.4560&6.1032&6.7613&29.2544& 26.1 &2255.85& 1.27&\XSolidBrush \\
Uformer~\cite{wang2021uformer}& 19.828 &  0.7374 &0.4220&5.8680&6.6796&26.6380&20.6 &657.45& 0.43&\XSolidBrush  \\
DehazeFormer-B~\cite{DehazeFormer}&15.372  & 0.7245  &0.3998&6.5477&7.1774&29.5049&2.5 &375.40&0.43&\XSolidBrush  \\
UHDformer~\cite{aaai24wang_UHDformer}& \underline{22.586}& \underline{0.9427}&\underline{0.1188}&\underline{5.2078}&5.3911&31.7302& \textbf{0.3393} &\underline{51.63}& 0.16& \CheckmarkBold\\
%UHDPromer&22.725 &0.9432 &0.1134&&&&0.7430M&32.56&0.07\\
\textbf{UHDformer++ (Ours)}&\textbf{25.089}&\textbf{0.9557}&\textbf{0.1043}&\textbf{5.1713}&\underline{5.3662}&\textbf{33.2766}&\underline{0.3962}&\textbf{49.65}&\underline{0.12}& \CheckmarkBold\\
\shline
\end{tabular}
% \vspace{-2mm}
\end{table*}
%%%%%%%%%%%%%%%%%%%%%%%%%%%%%%%
\begin{figure*}[!t]
\centering
\begin{center}
\begin{tabular}{ccccccccc}
\includegraphics[width=0.2\linewidth]{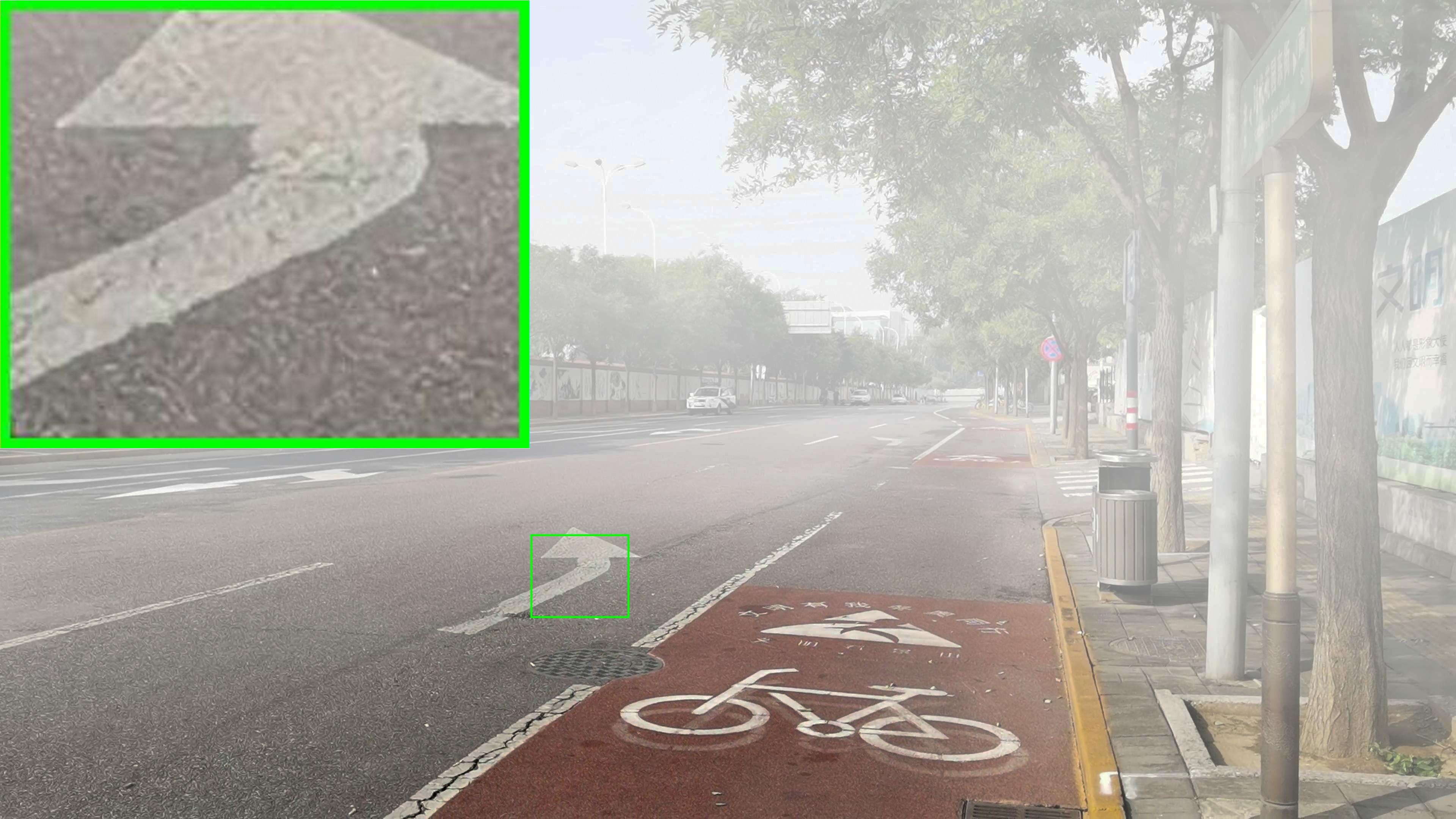} &\hspace{-4.75mm}
\includegraphics[width=0.2\linewidth]{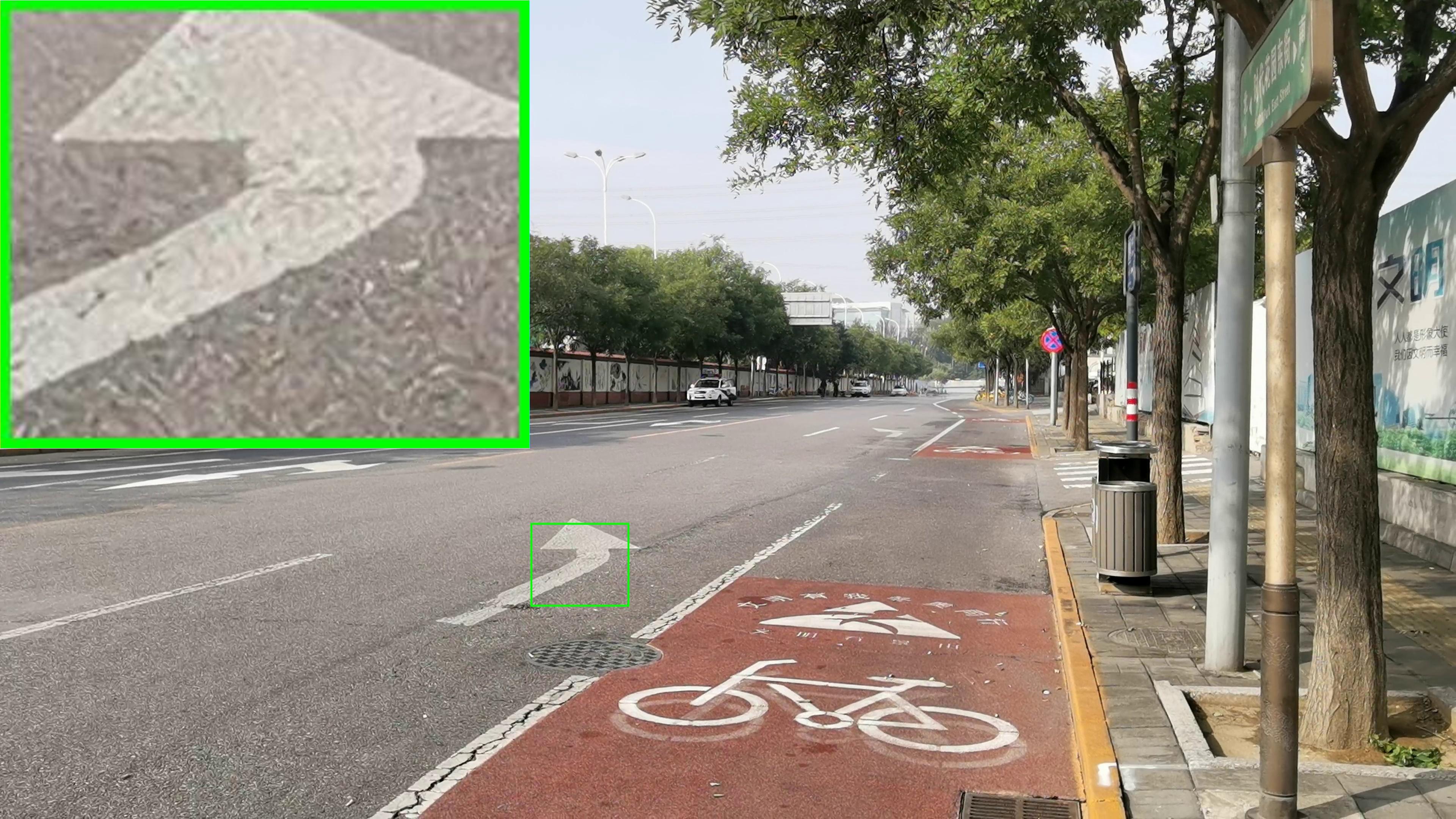} &\hspace{-4.75mm}
\includegraphics[width=0.2\linewidth]{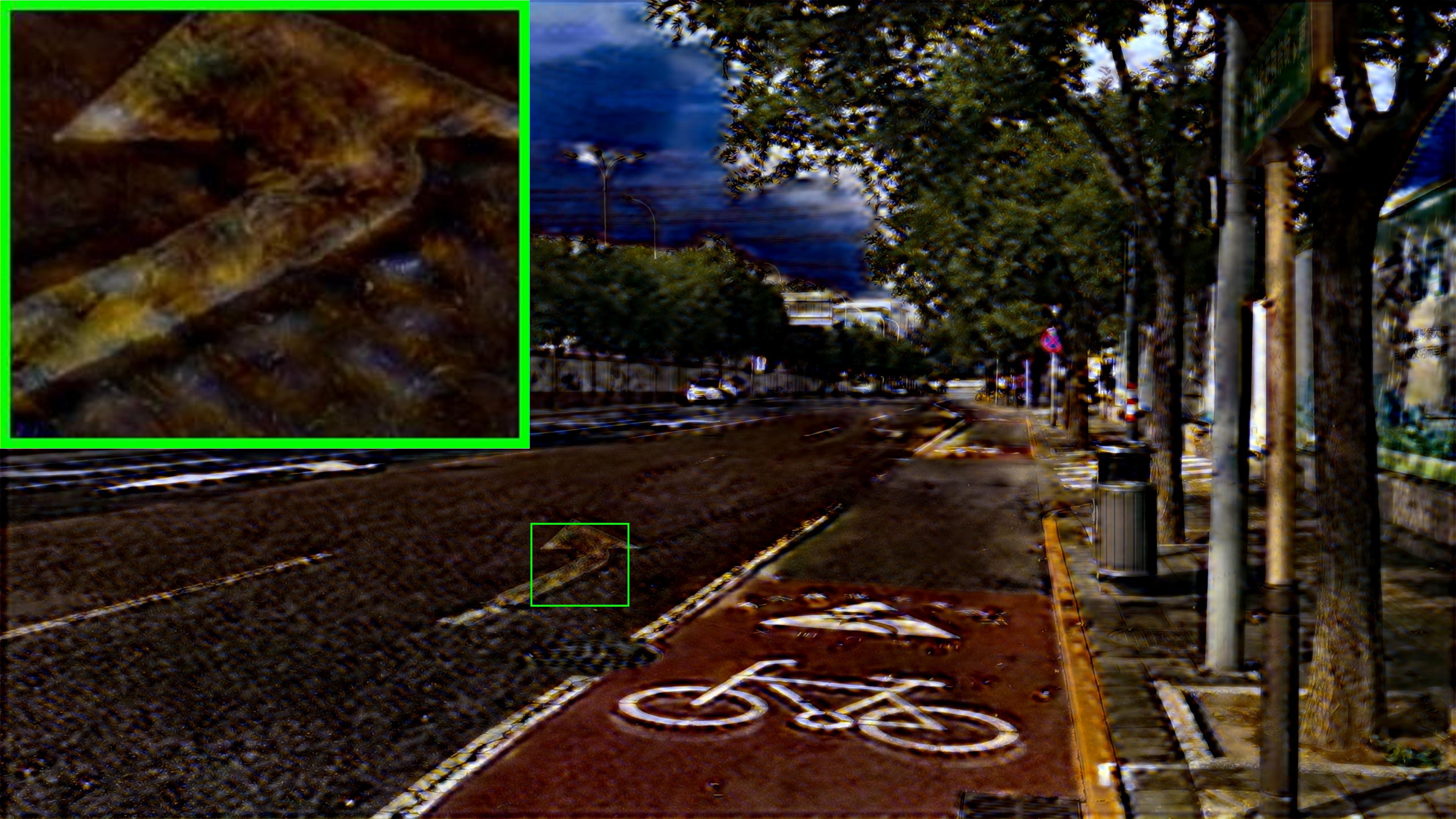} &\hspace{-4.75mm}
\includegraphics[width=0.2\linewidth]{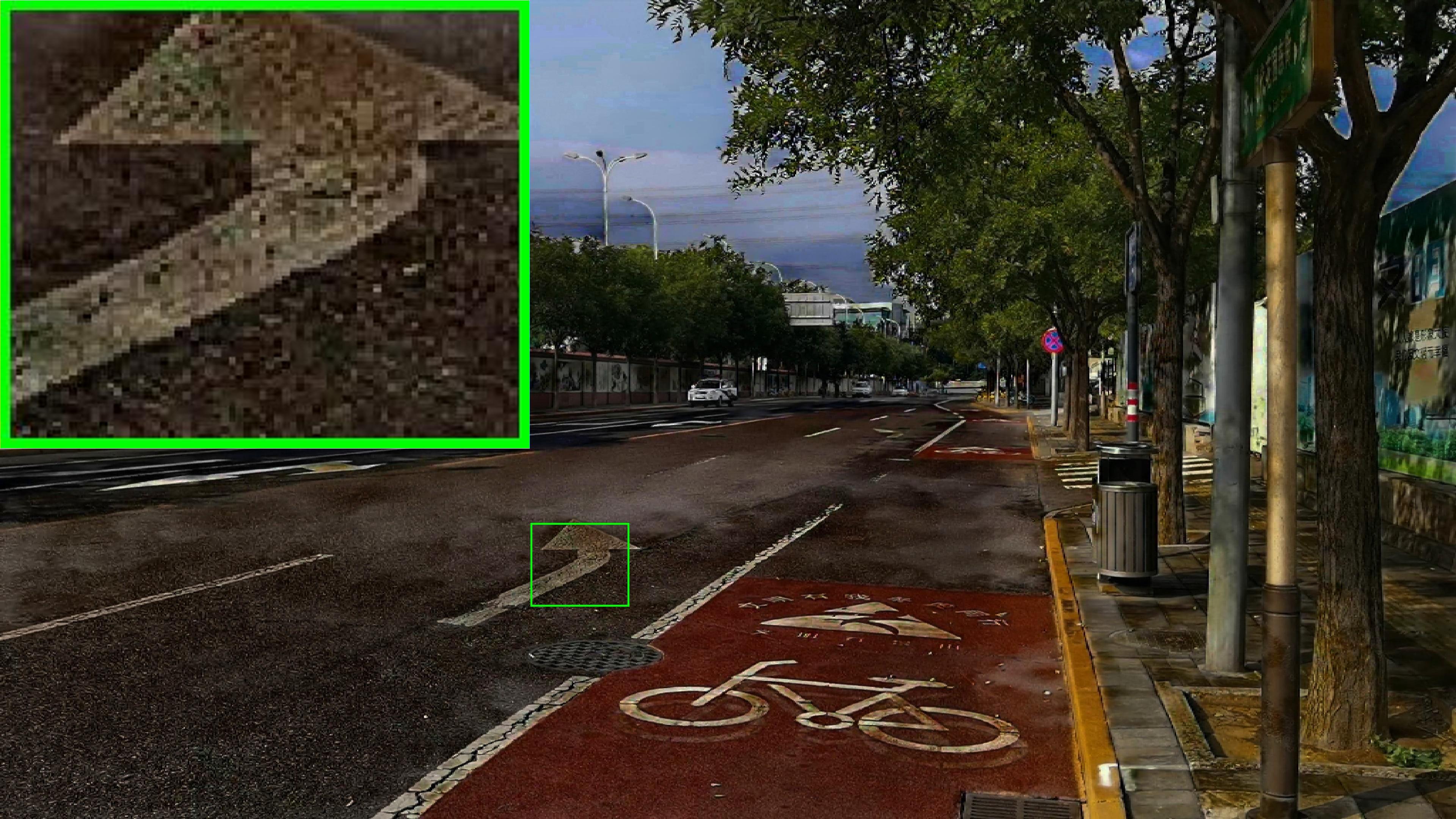}  
\\
 (a) Input&\hspace{-4.75mm} (b) GT&\hspace{-4.75mm} (c) UHD~\cite{Zheng_uhd_CVPR21} &\hspace{-4.75mm}(d)   Restormer~\cite{Zamir2021Restormer}
\\

\includegraphics[width=0.2\linewidth]{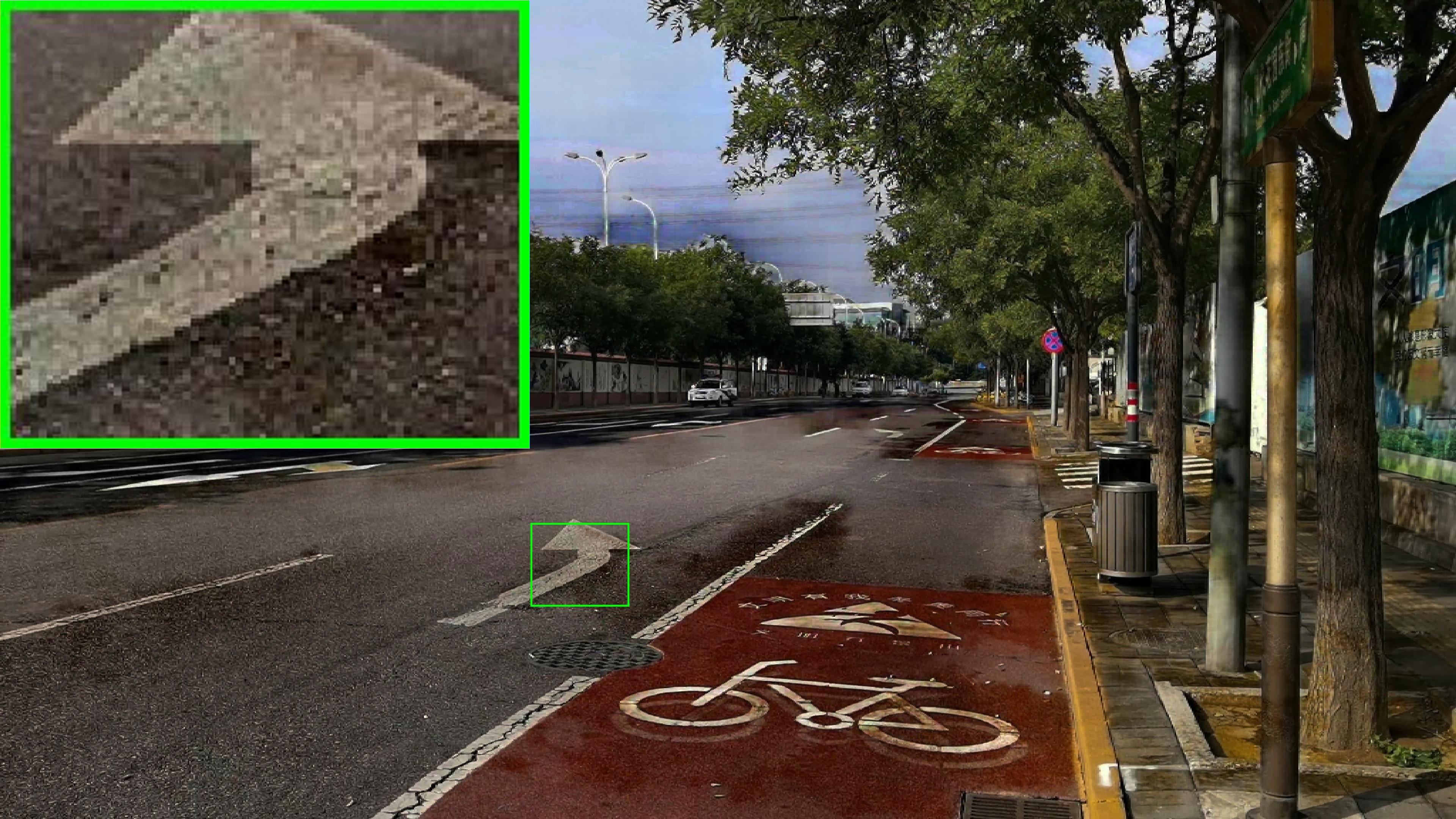} &\hspace{-4.75mm}
\includegraphics[width=0.2\linewidth]{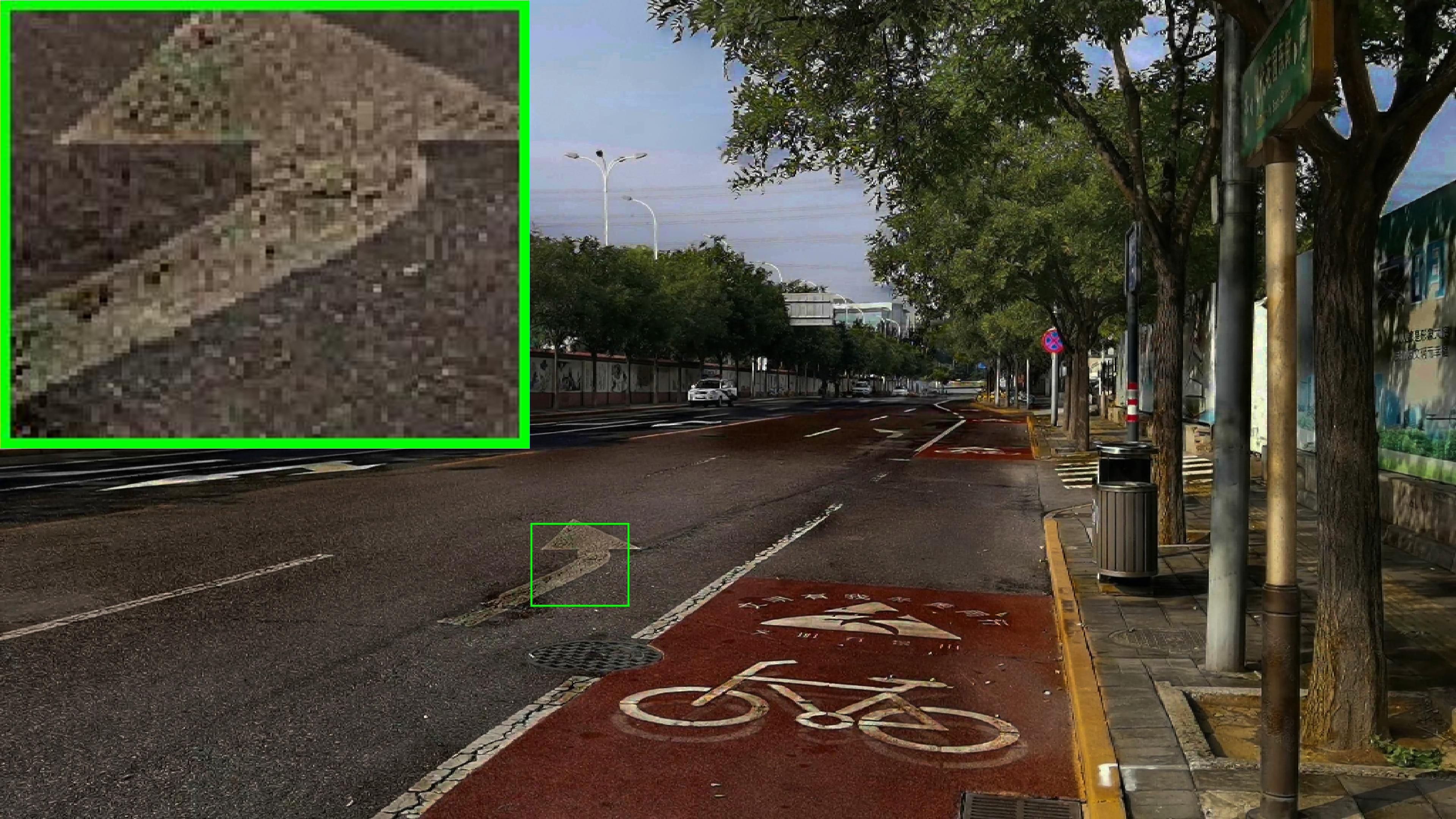} &\hspace{-4.75mm}
\includegraphics[width=0.2\linewidth]{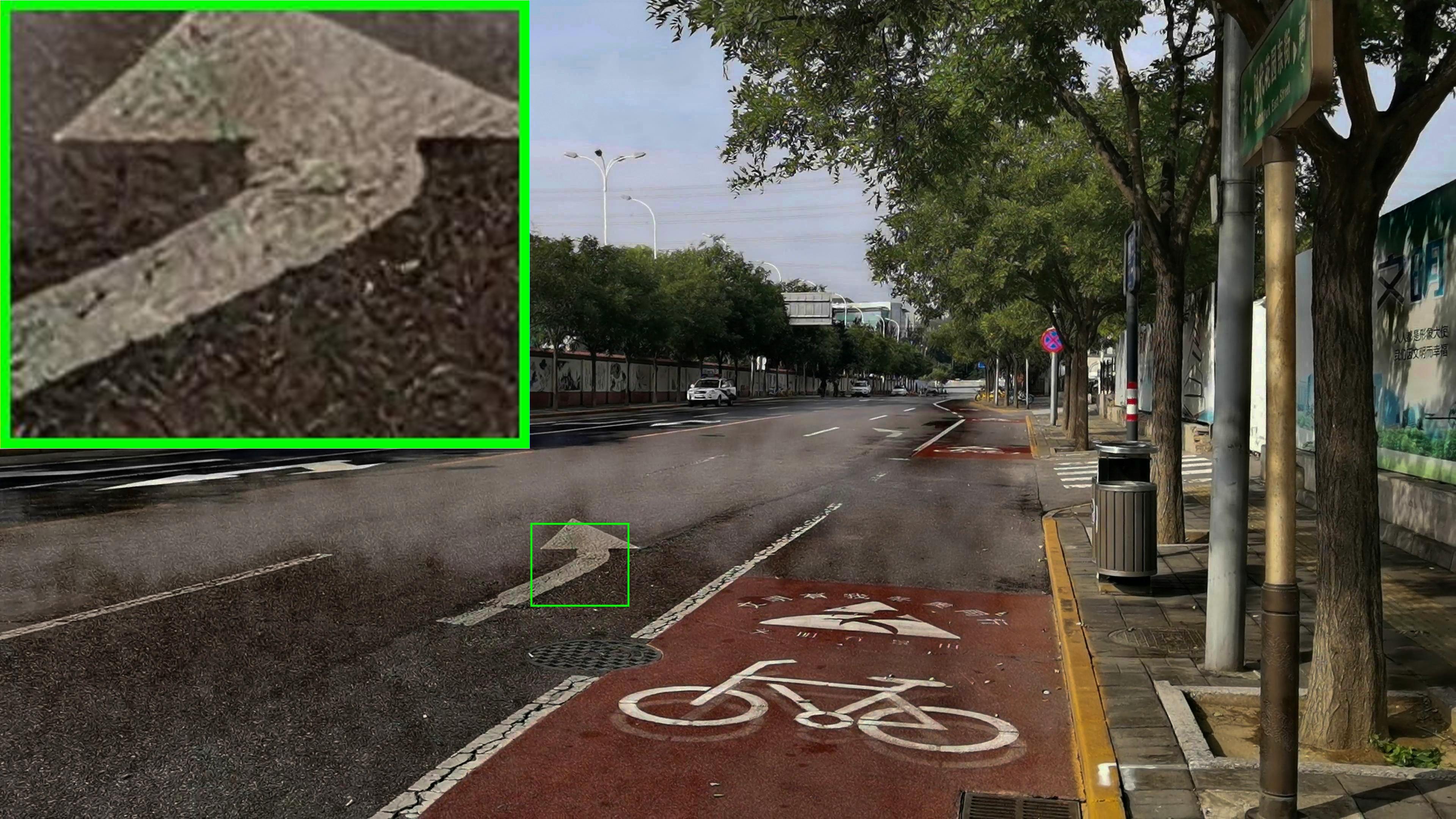}
&\hspace{-4.75mm}
\includegraphics[width=0.2\linewidth]{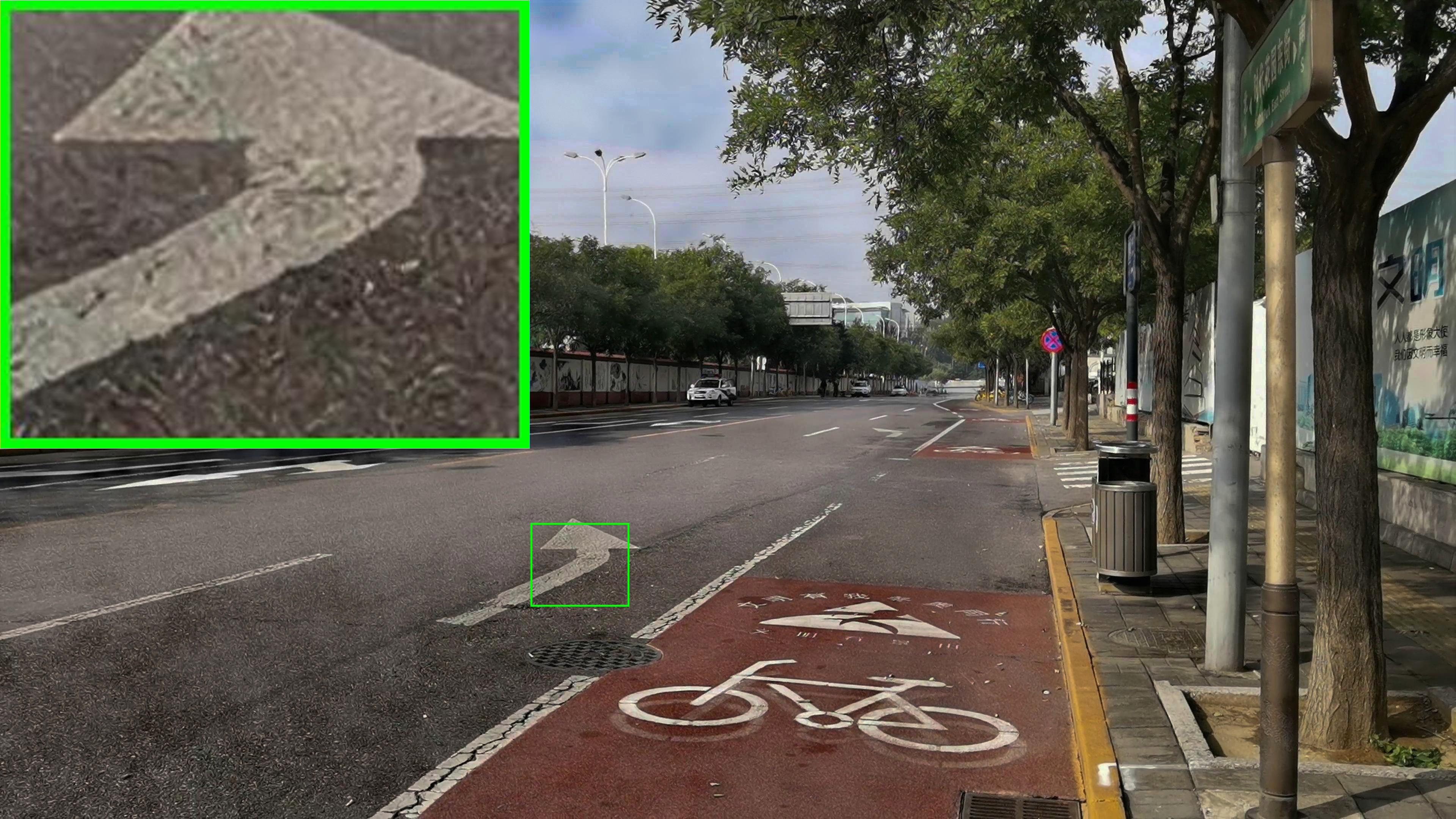}
\\
(e) Uformer~\cite{wang2021uformer}&\hspace{-4.75mm} (f) DehazeFormer~\cite{DehazeFormer} &\hspace{-4.75mm}(g) UHDformer~\cite{aaai24wang_UHDformer} &\hspace{-4.75mm} (h) \textbf{UHDformer++}
\end{tabular}
\vspace{-2mm}
\caption{\textbf{Image dehazing on UHD-Haze \textit{trained on the SOTS-ITS dataset}~\cite{RESIDE_dehazingbenchmarking_tip2019}.}
% \textbf{UHDformer++} is able to generate much clearer results.
}
\label{fig:Image dehazing on UHD-Haze.}
\end{center}
\vspace{-3mm}
\end{figure*}
%
%%%%%%%%%%%%%%%%%%%%%%%%%%%%%%%%%%%%%%%%%%%%%%%%%%%%%%%%%%%%%%%%%%%%%%%%%%%%
\begin{figure*}[!t]
\centering
\begin{center}
\begin{tabular}{ccccccccc}
\includegraphics[width=0.2\linewidth]{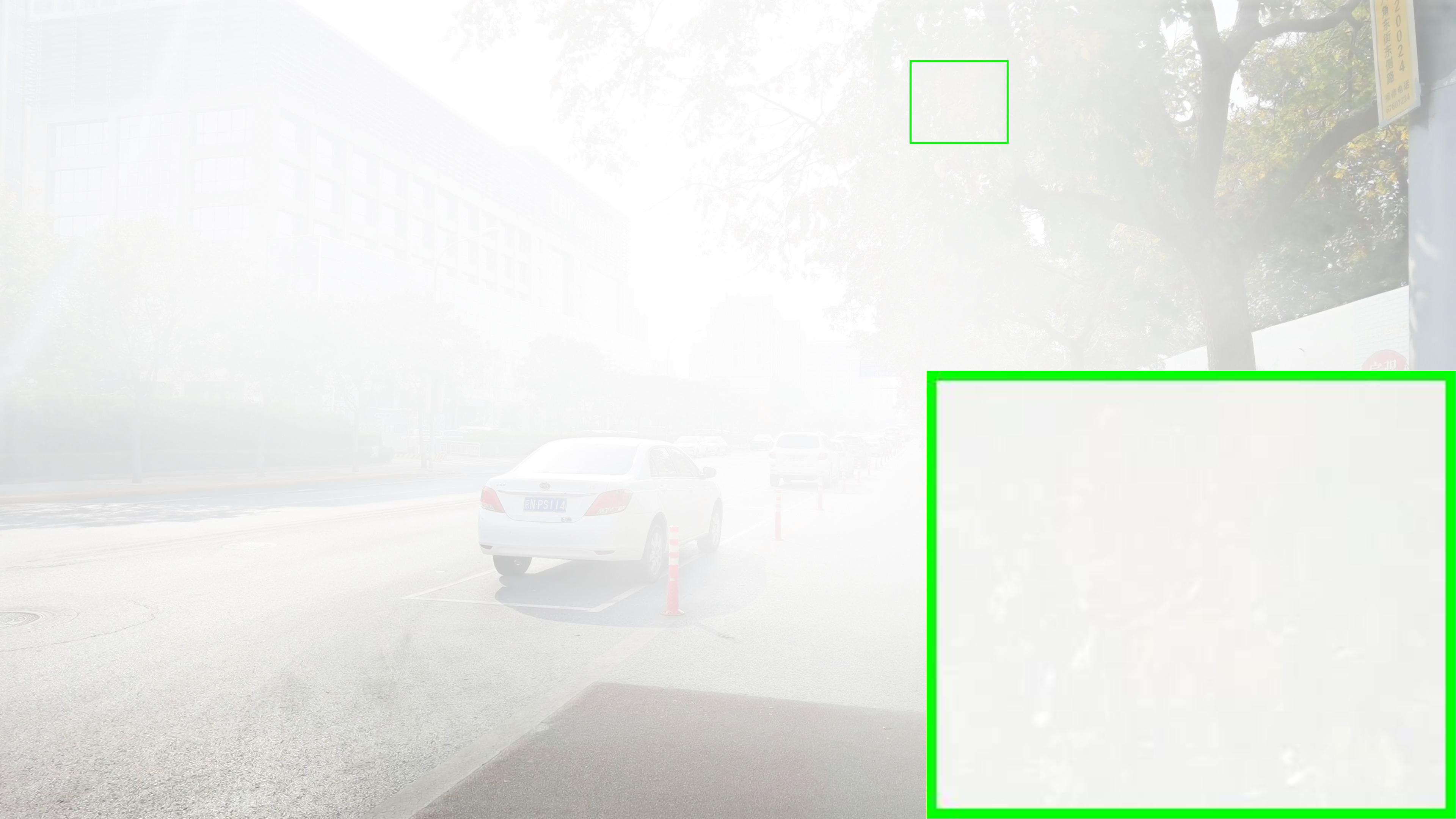} &\hspace{-4.75mm}
\includegraphics[width=0.2\linewidth]{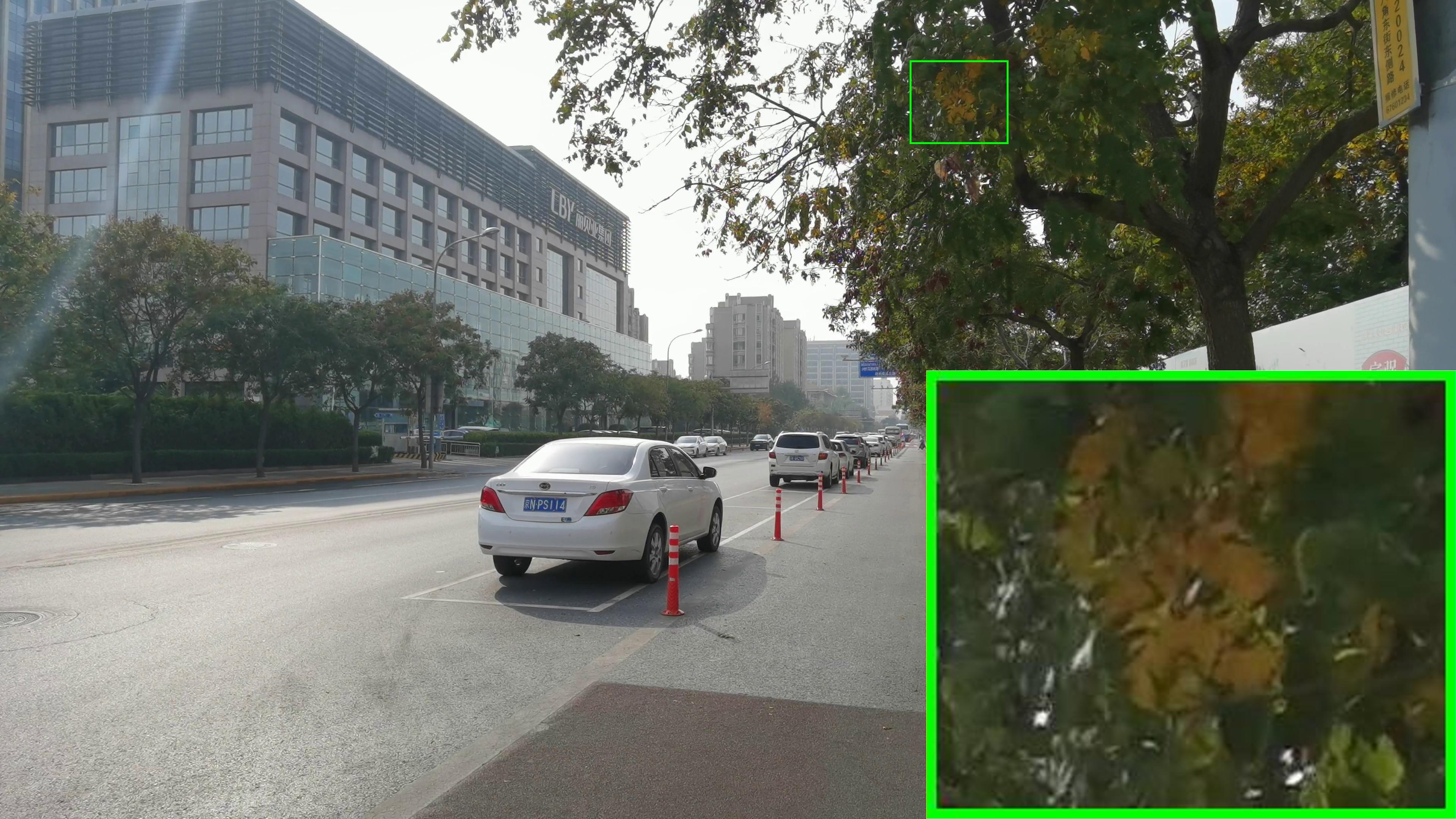} &\hspace{-4.75mm}
\includegraphics[width=0.2\linewidth]{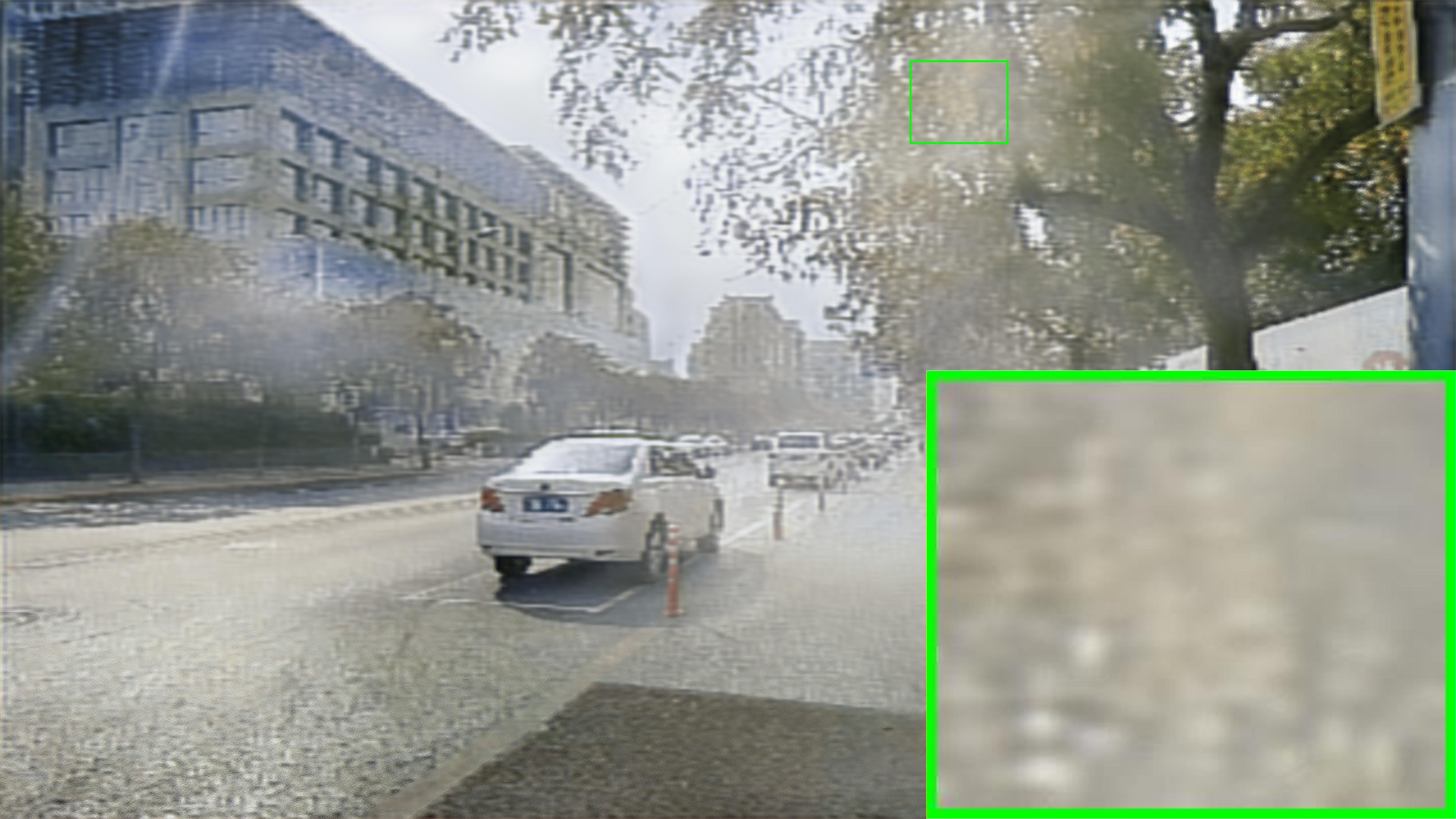} &\hspace{-4.75mm}
\includegraphics[width=0.2\linewidth]{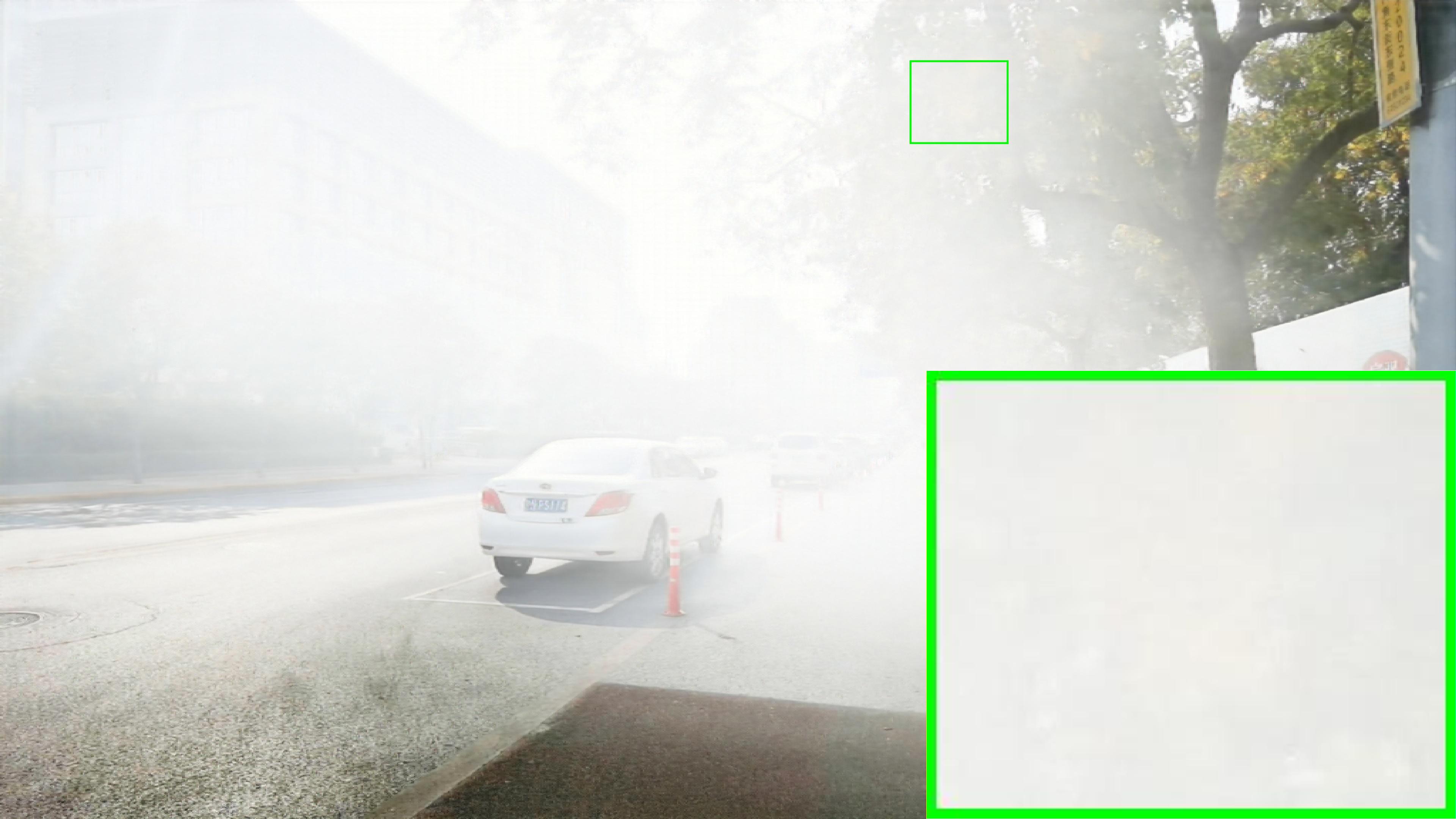} 
\\
 (a) Input&\hspace{-4.75mm} (b) GT&\hspace{-4.75mm} (c) UHD~\cite{Zheng_uhd_CVPR21} &\hspace{-4.75mm}(d)  Restormer~\cite{Zamir2021Restormer}
\\
\includegraphics[width=0.2\linewidth]{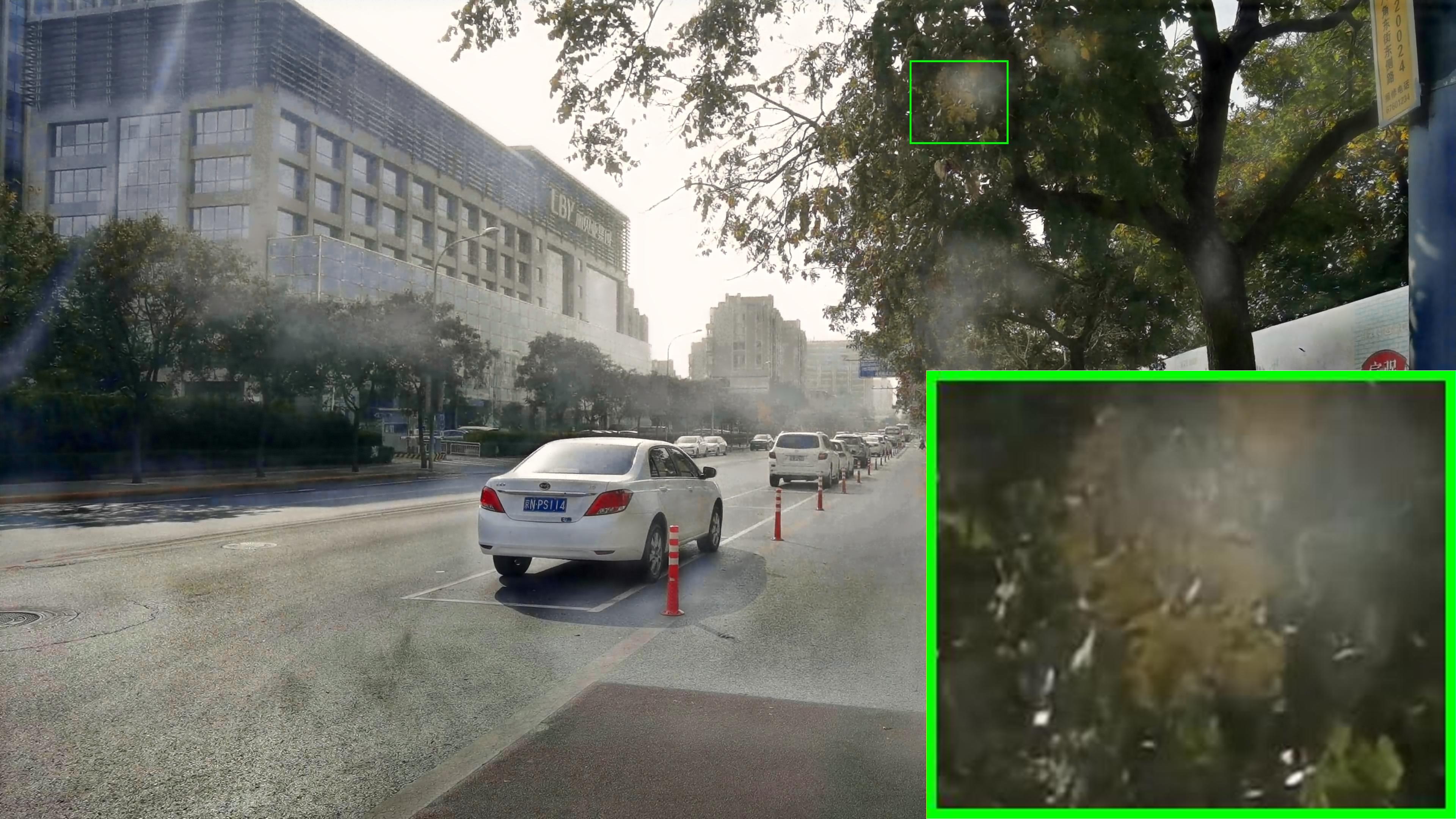} &\hspace{-4.75mm}
\includegraphics[width=0.2\linewidth]{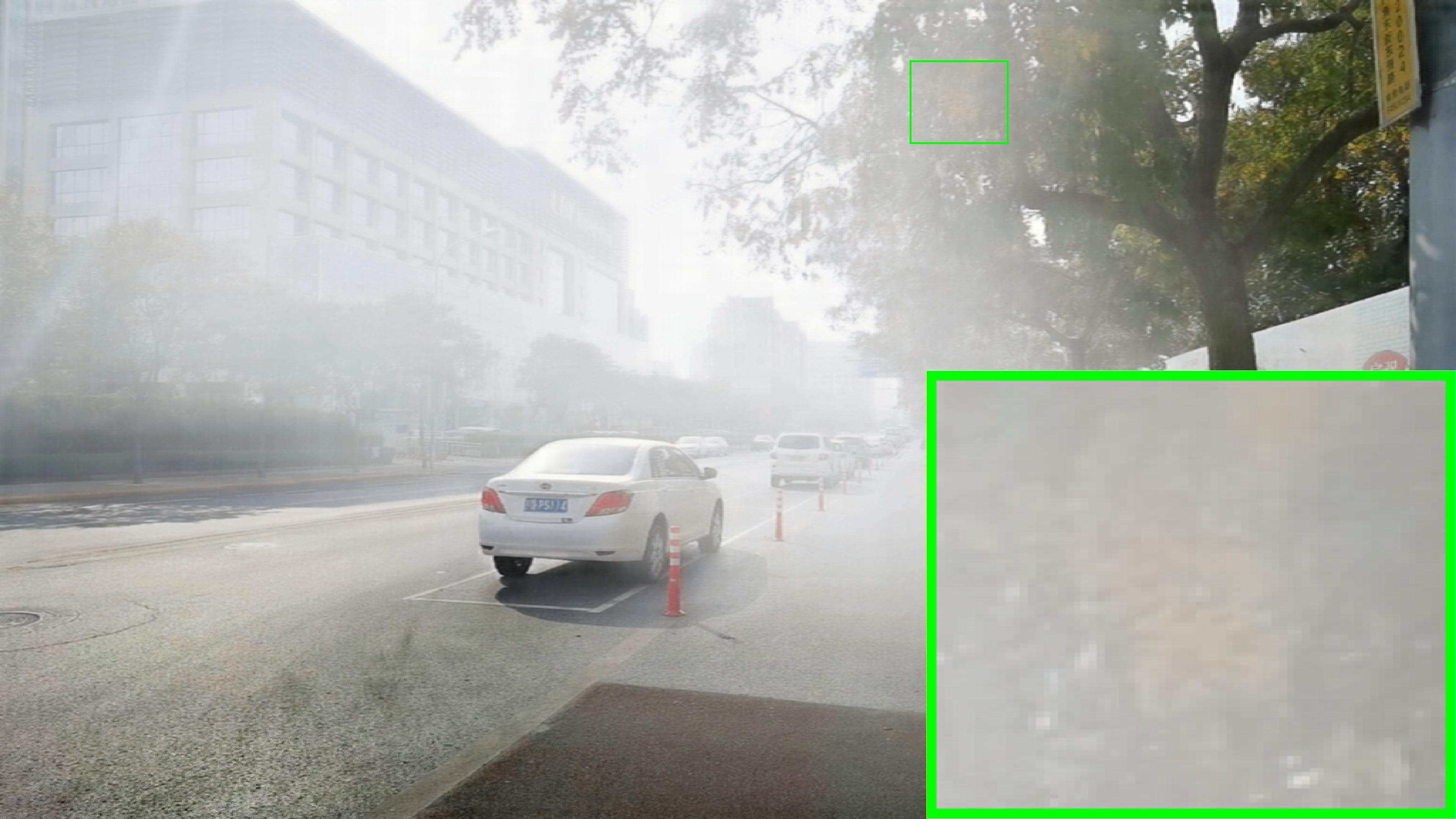} &\hspace{-4.75mm}
\includegraphics[width=0.2\linewidth]{Rect_40_400000112-uhdhaze-uhdformer_.jpg}
&\hspace{-4.75mm}\includegraphics[width=0.2\linewidth]{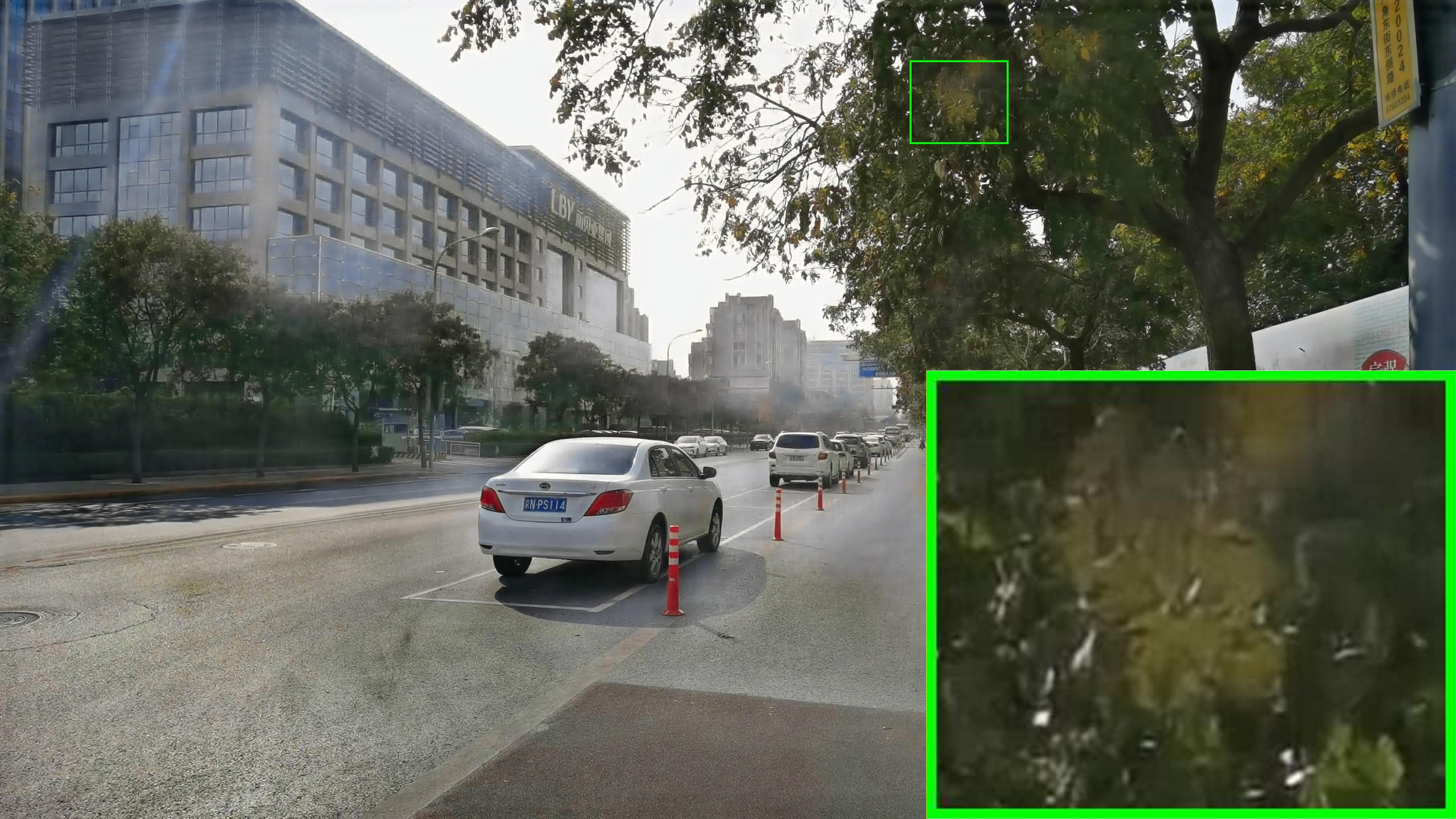}
\\
%(e) DeHamer~\cite{guo2022dehamer}&\hspace{-4.75mm} 
(e) Uformer~\cite{wang2021uformer}&\hspace{-4.75mm} (f) DehazeFormer~\cite{DehazeFormer} &\hspace{-4.75mm}(g) UHDformer~\cite{aaai24wang_UHDformer} &\hspace{-4.75mm}(h) \textbf{UHDformer++}
\end{tabular}
\vspace{-2mm}
\caption{\textbf{Image dehazing on UHD-Haze \textit{trained on the UHD-Haze dataset}.}
% \textbf{UHDformer++} is able to generate much cleaner results with fewer haze residuals.
}
\label{fig:Image dehazing on UHD-Haze.}
\end{center}
\vspace{-3mm}
\end{figure*}
\subsection{Implementation Details}
In the low-resolution space, the number of CMT-TBs (i.e., $L$ in Eq.~\eqref{eq: ECM-TB}) is set to $16$, with residual connections~\cite{res} applied at every $4$ CMT-TBs. 
The number of attention heads is set to $8$, and the feature channel dimension $C$ is set to 16. The matching factor $r$ is set to $4$; an ablation analysis of this parameter is provided in Sec.~\ref{sec: Effect on Dual-path Correlation Matching Transformation}.
We train models using the AdamW optimizer~\cite{adamw} with the initial learning rate $5e^{-4}$ gradually reduced to $1e^{-7}$ with the cosine annealing~\cite{loshchilov2016sgdr}.
The patch size is set as $512$$\times$$512$.
We randomly perform horizontal and vertical flips as data augmentation.
We train our model until converged.

% \\
% \noindent \textbf{Evaluation.} 
%

\subsection{Evaluation} 
To evaluate model performance, we employ both full-reference and no-reference image quality metrics. 
For full-reference evaluation, we adopt the Peak Signal-to-Noise Ratio (PSNR)~\cite{PSNR_thu}, Structural Similarity Index Measure (SSIM)~\cite{SSIM_wang}, and Learned Perceptual Image Patch Similarity (LPIPS)~\cite{LPIPS}. 
For no-reference evaluation, we utilize the Perceptual Index (PI)~\cite{pi}, Naturalness Image Quality Evaluator (NIQE)~\cite{niqe}, and Multi-Scale Image Quality Transformer (MUSIQ)~\cite{ke2021musiq}.
Beyond perceptual quality metrics, we also report computational complexity in terms of model parameters (Params.), Floating-Point Operations (FLOPs), and inference time to assess efficiency, all of which are measured on images of resolution 1024$\times$1024.
Since certain baseline methods are unable to directly process full-resolution UHD images, an adaptation strategy must be applied during evaluation. 
As demonstrated in UHDFour~\cite{Li2023ICLR_uhdfour}, resizing the input to the maximum resolution a model can handle yields superior results compared to splitting the input into four non-overlapping patches and subsequently stitching the outputs. 
Accordingly, we apply the resizing strategy to those methods incapable of processing UHD images at full resolution, and explicitly indicate each model's full-resolution processing capability in Tabs.~\ref{tab:Low-light image enhancement.}-\ref{tab:Image desnowing.}.

\subsection{Main Results}
We provide qualitative and quantitative comparisons with state-of-the-art approaches on $5$ UHD image restoration tasks, including low-light image enhancement, image dehazing, image deblurring, image deraining, and image desnowing.
\subsubsection{Low-Light Image Enhancement Results} 

We evaluate UHD low-light image enhancement on the UHD-LL benchmark under two training settings: models trained on the general-purpose LOL dataset and models trained on UHD-LL itself, with all methods tested on the UHD-LL test set. 
Tab.~\ref{tab:Low-light image enhancement.} summarizes the results, demonstrating that our UHDformer++ achieves state-of-the-art performance under both training configurations.
Specifically, UHDformer++ consistently surpasses UHDformer~\cite{aaai24wang_UHDformer} across both training strategies while maintaining a comparable model size, reduced FLOPs, and faster inference time. 
Compared to UHDFour~\cite{Li2023ICLR_uhdfour}, UHDformer++ reduces the number of trainable parameters by at least 98\% while consistently outperforming it across different training settings in terms of widely-used PSNR and SSIM. 
Notably, UHDFour, trained on LOL~\cite{retinexnet_wei_bmvc18}, fails to generalize well to UHD images, suggesting limited robustness under cross-dataset evaluation. 
In contrast, UHDformer++ maintains strong enhancement performance in this setting, demonstrating its superior generalization capability.
Furthermore, UHDformer++ outperforms methods that natively support full-resolution UHD image processing without resizing, including~\cite{zhao_lie} and URetinex~\cite{Wu_2022_CVPR}, further validating the effectiveness of our approach. 
Visual comparisons on UHD-LL are presented in Fig.~\ref{fig:Low-light image enhancement on UHD-LL}, where UHDformer++ produces results with noticeably more natural and faithful color rendition.

\begin{table*}[!t]
\caption{\textbf{Image deblurring}. 
}
% \vspace{-3mm}
\label{tab:Image deblurring.} 
\tablestyle{2pt}{1}
\centering
\begin{tabular}{l|ccc|ccc|ccc|c}
\shline
 \multicolumn{1}{c|}{\multirow{3}{*}{\textbf{Method}}} & \multicolumn{6}{c|}{\textbf{Metrics}}&\multicolumn{3}{c|}{\multirow{2}{*}{\textbf{Computational Complexity}}}& \multirow{3}{*}{\textbf{Full Size}} 
 \\
  \cline{2-7}
&\multicolumn{3}{c|}{\textbf{Full Reference}}&\multicolumn{3}{c|}{\textbf{No-Reference}}&\multicolumn{3}{c|}{}
 \\
 \cline{2-10}
&\textbf{PSNR (dB)~$\uparrow$} 
&\textbf{SSIM~$\uparrow$}
&\textbf{LPIPS~$\downarrow$}
&\textbf{PI~$\downarrow$} 
&\textbf{NIQE~$\downarrow$} 
&\textbf{MIUSIQ~$\uparrow$}
&\textbf{Params. (M)~$\downarrow$} 
&\textbf{FLOPs (G)~$\downarrow$}
&\textbf{Time (s)~$\downarrow$}
\\
\shline
 \multicolumn{10}{c}{\textbf{Training Set on GoPro}}\\
\shline
DMPHN~\cite{dmphn2019}&26.490 &0.7985&\textbf{0.2696}& 6.1592&6.0393& 26.1362&21.7 &1251.76 & 0.15& \CheckmarkBold \\
% MTRNN&ECCV'20&24.706 &0.7415&2.6M & \CheckmarkBold  \\
MIMO-Unet++~\cite{cho2021rethinking_mimo}&24.290 &0.7354&0.3723&6.0395&6.1247&25.3302&16.1 & 2470.52& 0.31  & \CheckmarkBold\\
MPRNet~\cite{Zamir_2021_CVPR_mprnet}&24.571&0.7426&0.3597&6.9655&7.9493&27.6944&20.1 & 27317.71& 1.13 & \CheckmarkBold\\
Restormer~\cite{Zamir2021Restormer}&24.872 & 0.7484&0.3550&6.9825&7.9199&27.1289& 26.1 & 2255.85&1.27&\XSolidBrush\\
Uformer~\cite{wang2021uformer}&24.382 & 0.7209 &0.3901& 7.3765&8.9162&27.3757& 20.6 & 657.45&0.43&\XSolidBrush \\
Stripformer~\cite{Tsai2022Stripformer}&24.915 &0.7463&0.3299& \textbf{5.8254}&\underline{5.8983}&\textbf{28.8320}&19.7 & 2728.08& 0.81&\XSolidBrush \\
FFTformer~\cite{Kong_2023_CVPR_fftformer}&24.625 & 0.7396 &0.3428&\underline{5.8407}&\textbf{5.8325}&\underline{27.2571}& 16.6 &2107.94&  1.99&\XSolidBrush\\
% LMAR~\cite{lmar24}&23.040	&0.7894&0.3513&&&&1.314M&208.34&0.03\\
UHDformer~\cite{aaai24wang_UHDformer}&\textbf{27.436} & \textbf{0.8231}&0.2774& 6.4616& 6.4781&25.3380& \textbf{0.3393} & 51.63&0.16& \CheckmarkBold\\
\textbf{UHDformer++ (Ours)}&\underline{27.318}&\underline{0.8209}&\underline{0.2706}&6.3320&6.2669& 25.8895&\underline{0.3962}&\textbf{49.65}&\textbf{0.12}& \CheckmarkBold\\
\shline
 \multicolumn{10}{c}{\textbf{Training Set on UHD-Blur}}\\
\shline
% MTRNN&ECCV'20&24.991 & 0.7424 & 2.6M  & \CheckmarkBold \\
Restormer~\cite{Zamir2021Restormer}&25.210 & 0.7522 &0.3695&6.9449&7.9269&25.4992& 26.1 & 2255.85&1.27&\XSolidBrush\\
Uformer~\cite{wang2021uformer}&25.267 & 0.7515 &0.3851&7.0959&7.9875&23.8844& 20.6 & 657.45&0.43 &\XSolidBrush\\
Stripformer~\cite{Tsai2022Stripformer}&25.052 &0.7501&0.3740&7.3679&8.2872&25.1366&19.7 & 2728.08& 0.81 &\XSolidBrush\\
FFTformer~\cite{Kong_2023_CVPR_fftformer}&25.409 & 0.7571 &0.3708&7.0416&\textbf{5.8325}&\underline{27.2380}& 16.6M &2107.94&  1.99&\XSolidBrush\\
% LMAR~\cite{lmar24}&26.472	&0.7989&0.2470&&&&1.314M&208.34&0.03\\
UHDformer~\cite{aaai24wang_UHDformer}&\underline{28.821}& \underline{0.8440}&\underline{0.2350}&\textbf{6.0820}&6.4782&25.3655& \textbf{0.3393} & \underline{51.63}&\underline{0.16}& \CheckmarkBold\\
\textbf{UHDformer++ (Ours)}&\textbf{29.195}&\textbf{0.8500}&\textbf{0.2310}& \underline{6.1483}& \underline{6.2241}&\textbf{27.339}8&\underline{0.3962}&\textbf{49.65}&\textbf{0.12}& \CheckmarkBold\\
\shline
\end{tabular}
% \vspace{-3mm}
\end{table*}

%%%%%%%%%%%%%%%%%%%%%%%%%%%%%%%%%%%%%%%%%%%%%%%%%%%%%%%%%%%%%%%%%%%%%%%%%%%%
\begin{figure*}[!t]
\centering
\begin{center}
\begin{tabular}{ccccccccc}
\includegraphics[width=0.2\linewidth]{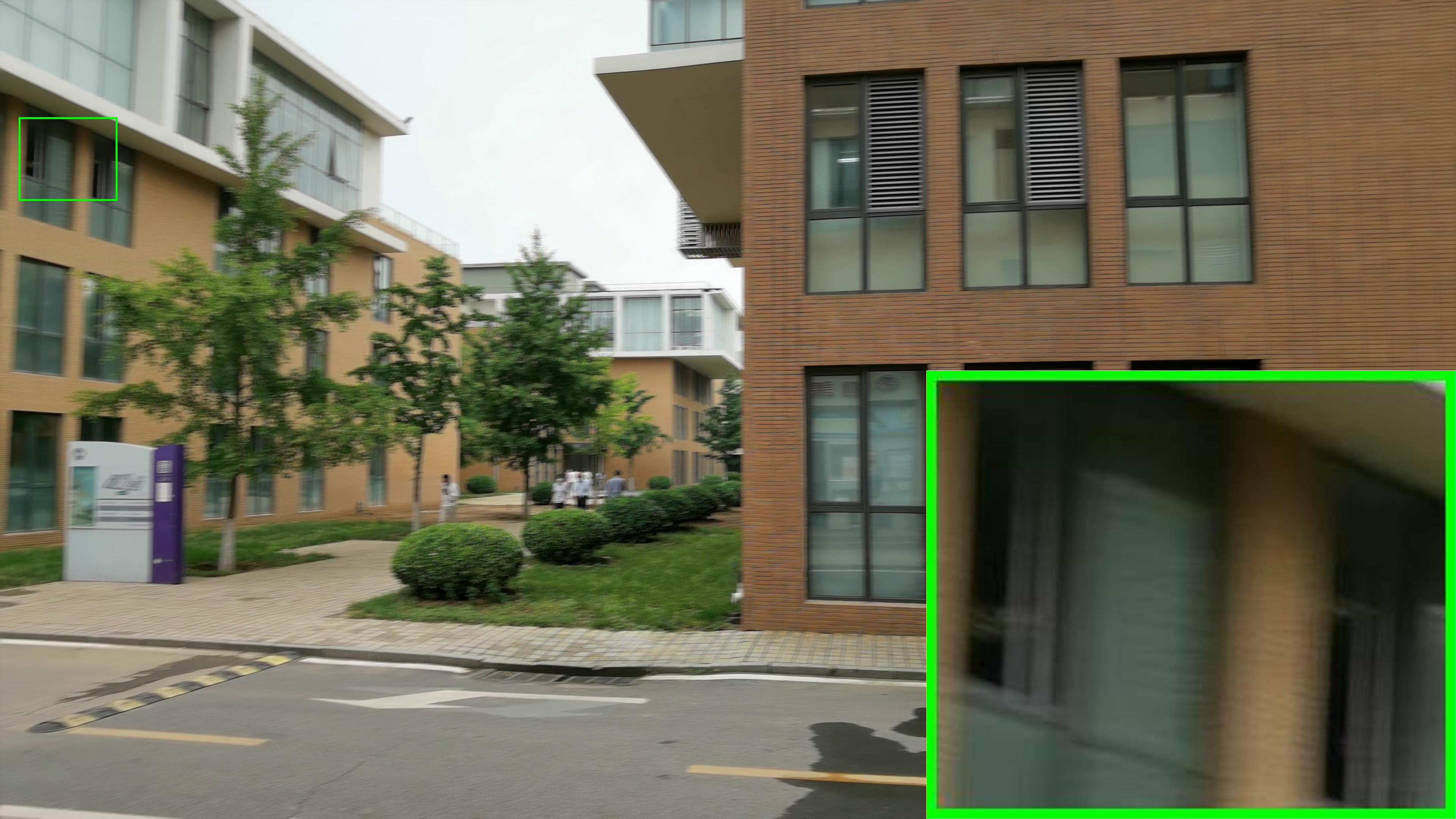} &\hspace{-4.75mm}
\includegraphics[width=0.2\linewidth]{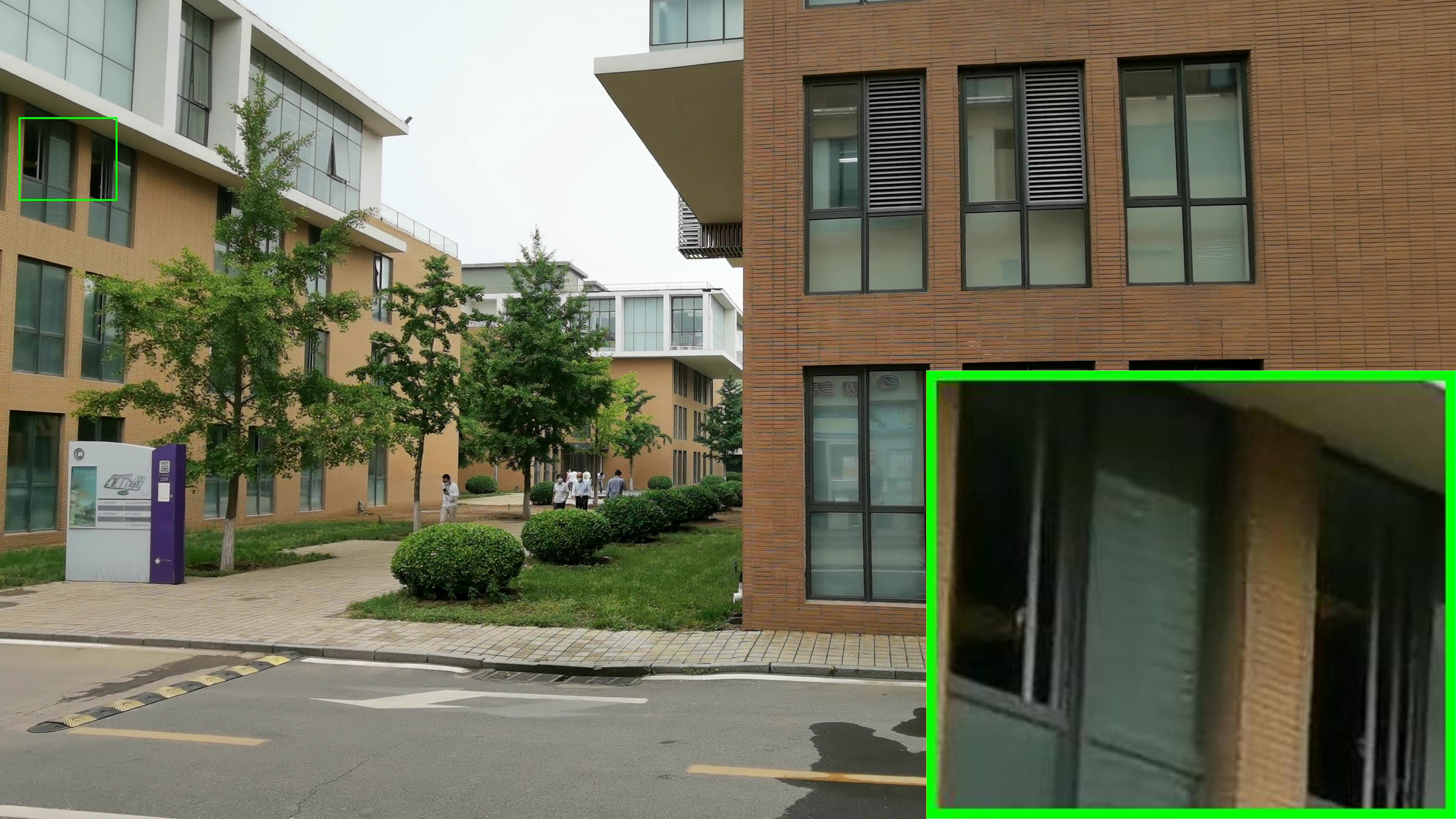} &\hspace{-4.75mm}
\includegraphics[width=0.2\linewidth]{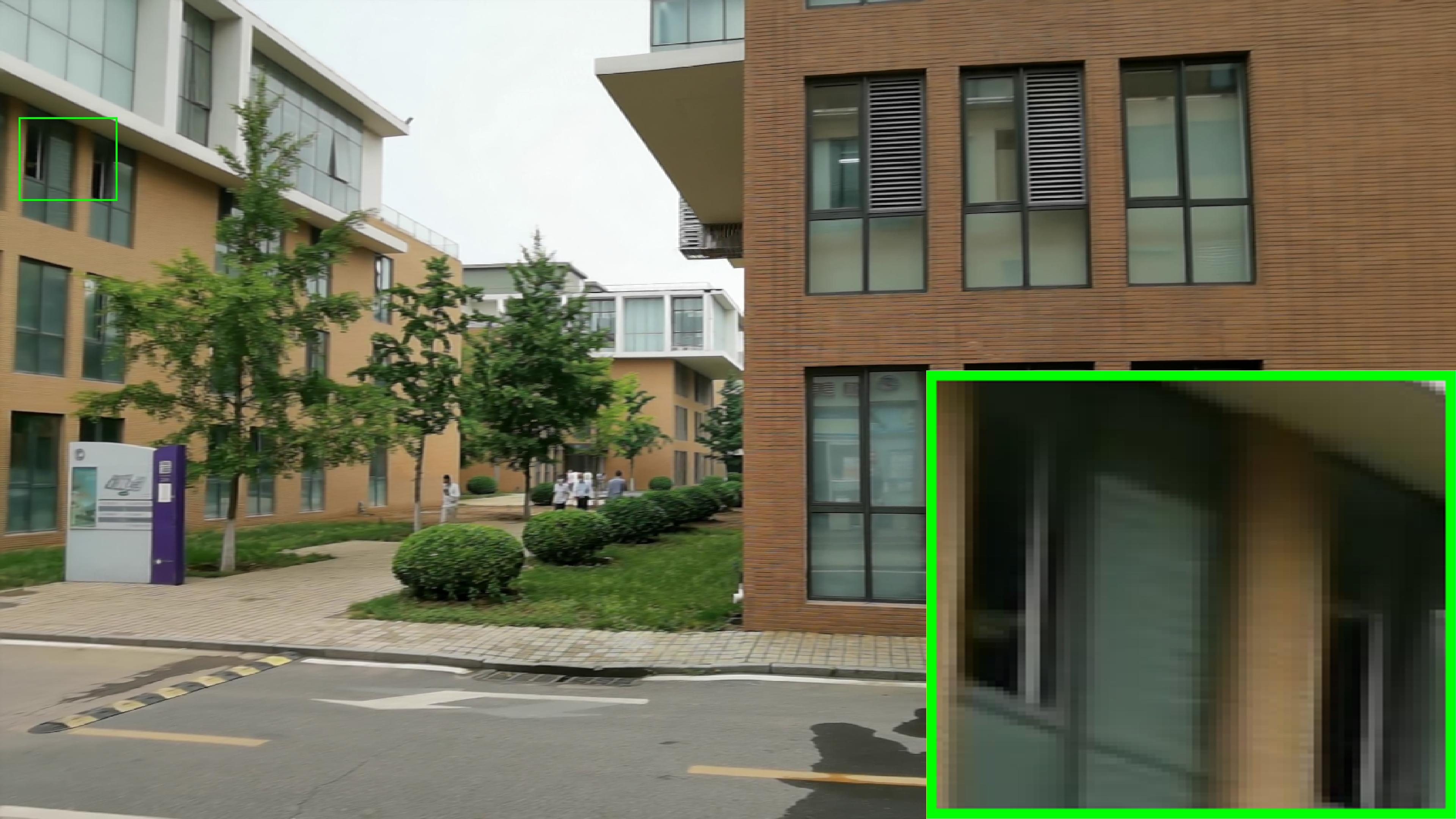}
&\hspace{-4.75mm}
\includegraphics[width=0.2\linewidth]{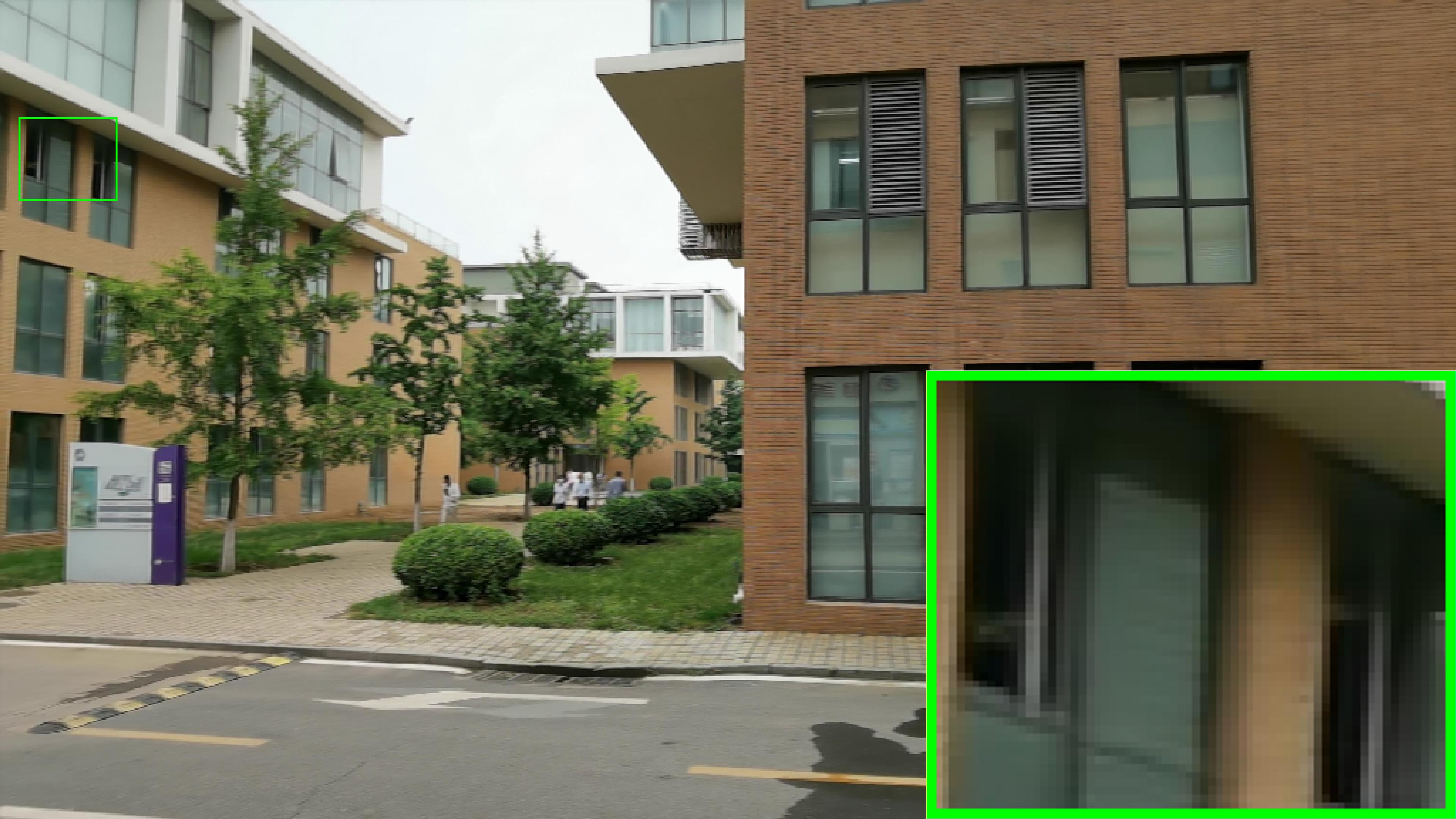} 
\\
 (a) Input&\hspace{-4.75mm} (b) GT&\hspace{-4.75mm}(c) Restormer~\cite{Zamir2021Restormer}   &\hspace{-4.75mm} (d) Uformer~\cite{wang2021uformer} 
\\

\includegraphics[width=0.2\linewidth]{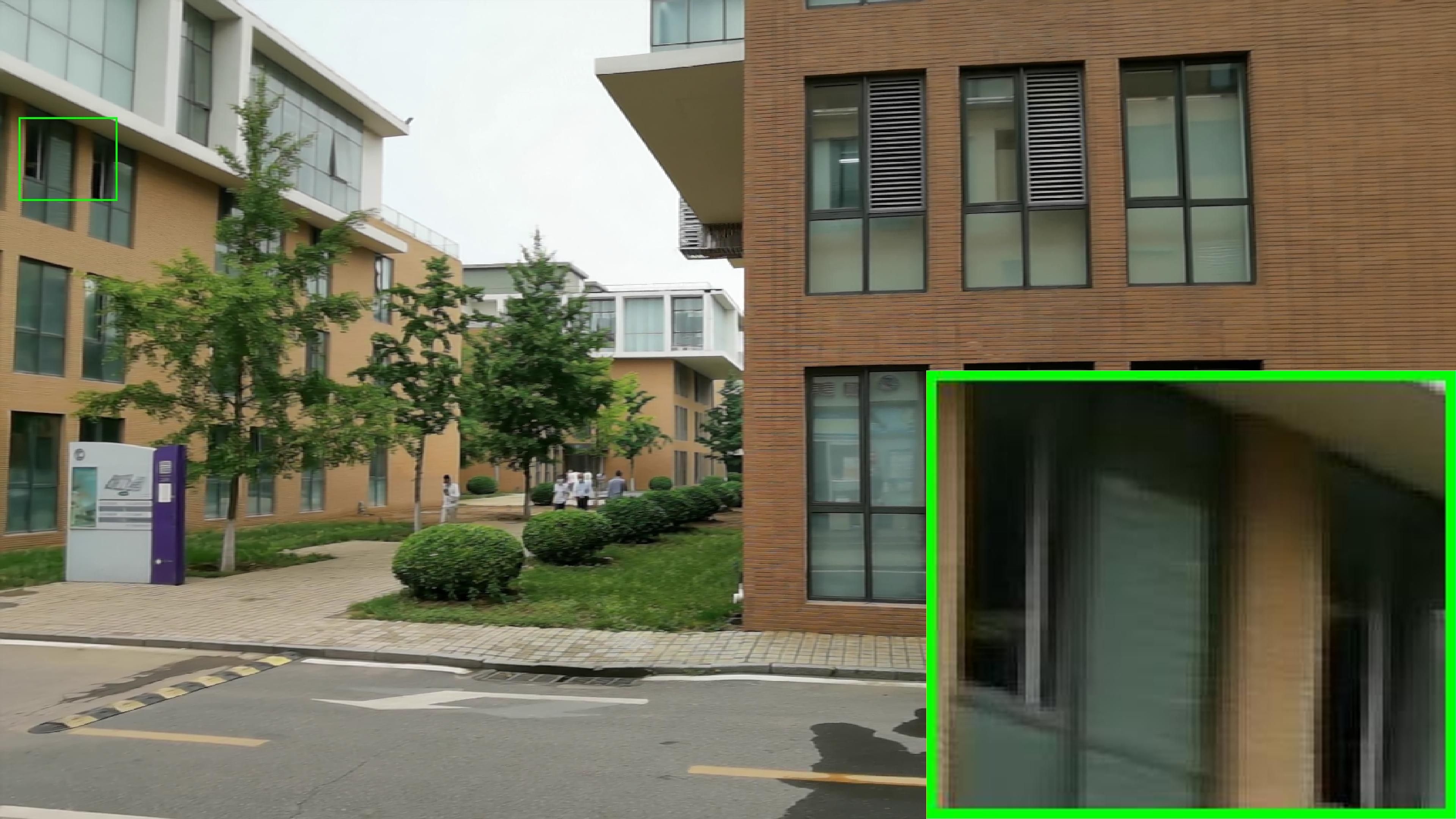} &\hspace{-4.75mm}
\includegraphics[width=0.2\linewidth]{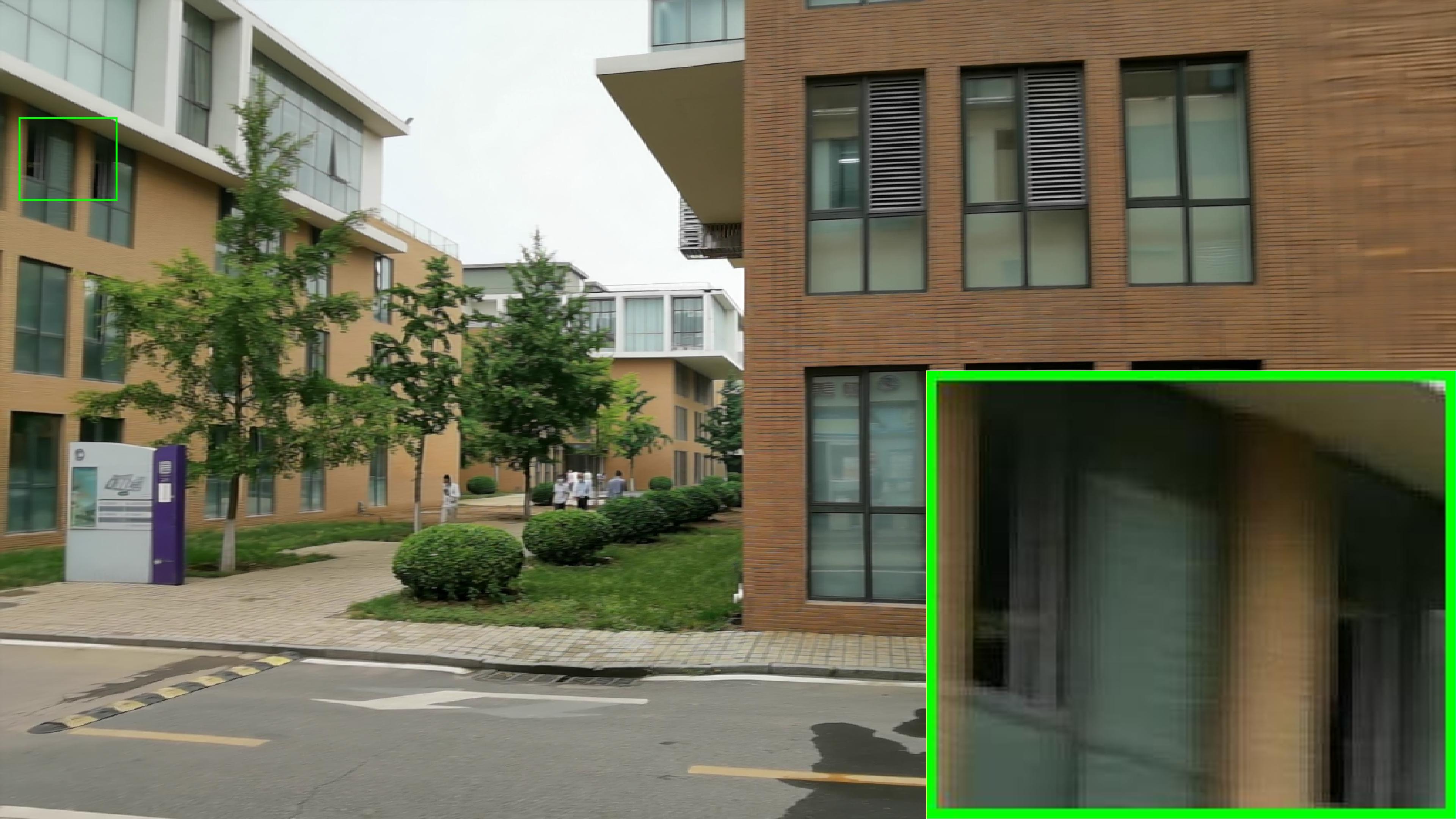} &\hspace{-4.75mm}
\includegraphics[width=0.2\linewidth]{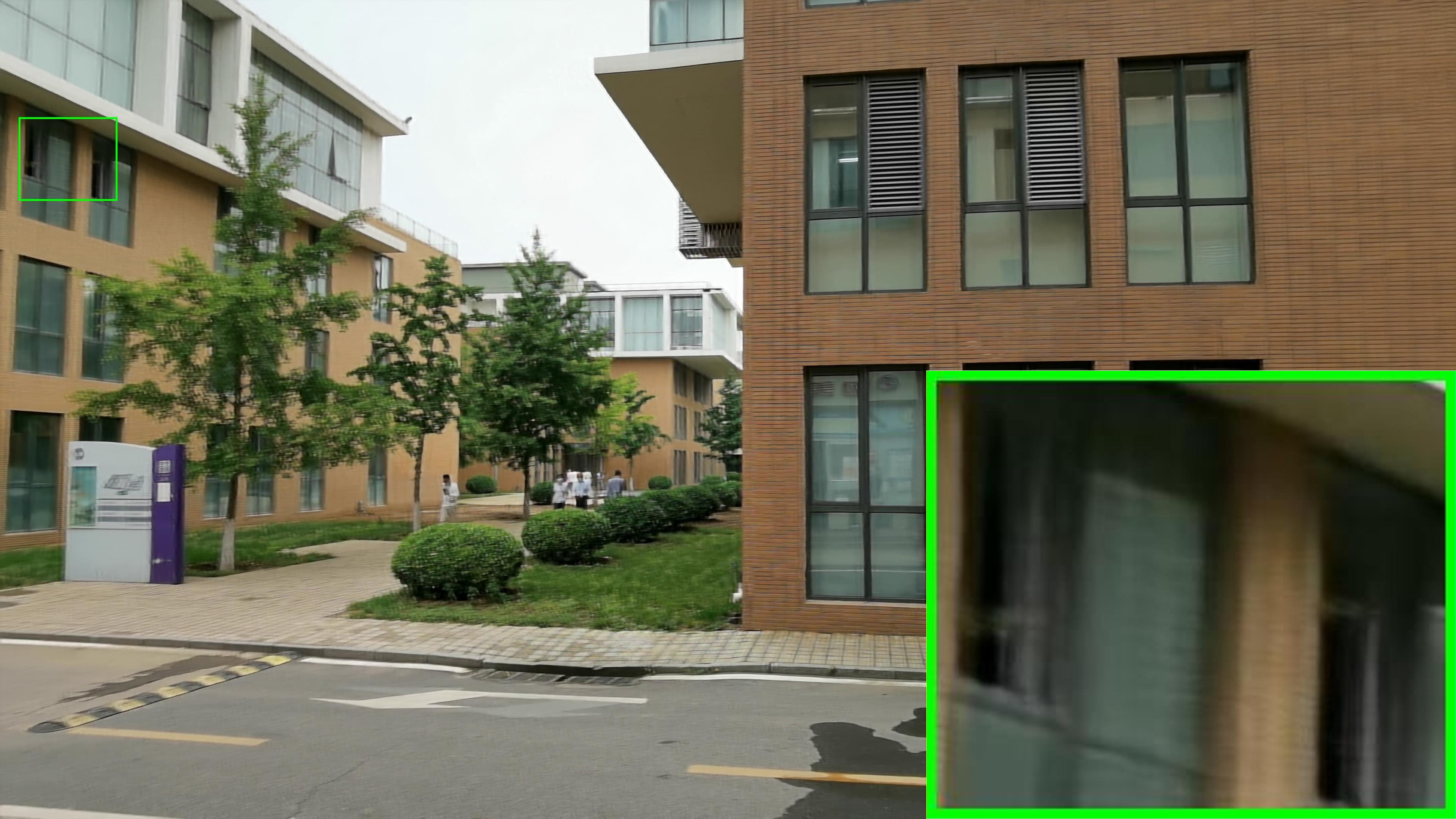}
&\hspace{-4.75mm}
\includegraphics[width=0.2\linewidth]{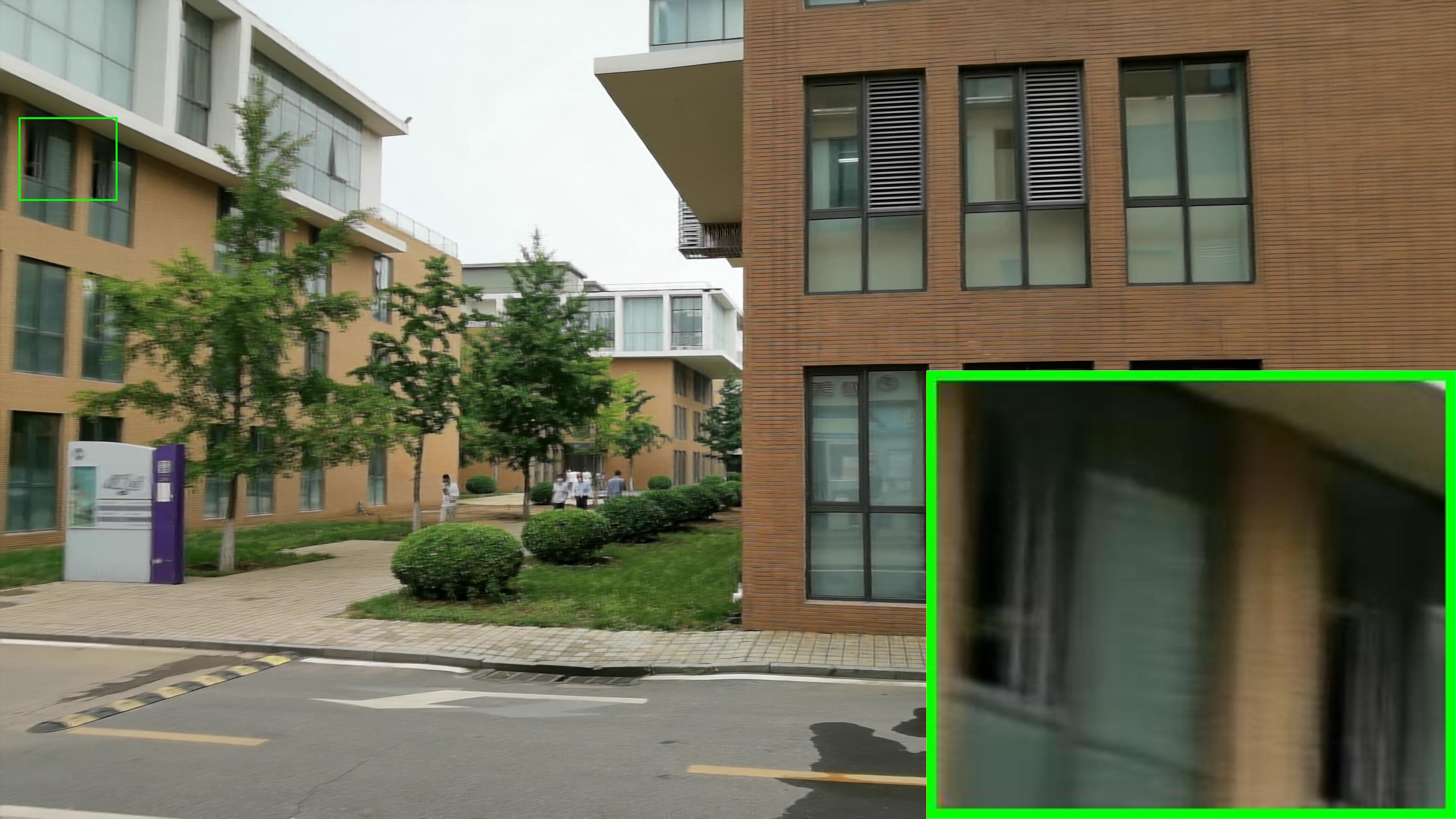}
\\
(e) Stripformer~\cite{Tsai2022Stripformer} &\hspace{-4.75mm} (f) FFTformer~\cite{Kong_2023_CVPR_fftformer} &\hspace{-4.75mm} (g) UHDformer~\cite{aaai24wang_UHDformer}
&\hspace{-4.75mm} (h) \textbf{UHDformer++}
\end{tabular}
\vspace{-2mm}
\caption{\textbf{Image deblurring on UHD-Blur \textit{trained on the GoPro dataset}~\cite{gopro2017}.}
}
\label{fig:Image deblurring on UHD-Blur.}
\end{center}
\vspace{-3mm}
\end{figure*}

\begin{figure*}[!t]
\centering
\begin{center}
\begin{tabular}{ccccccccc}
\includegraphics[width=0.2\linewidth]{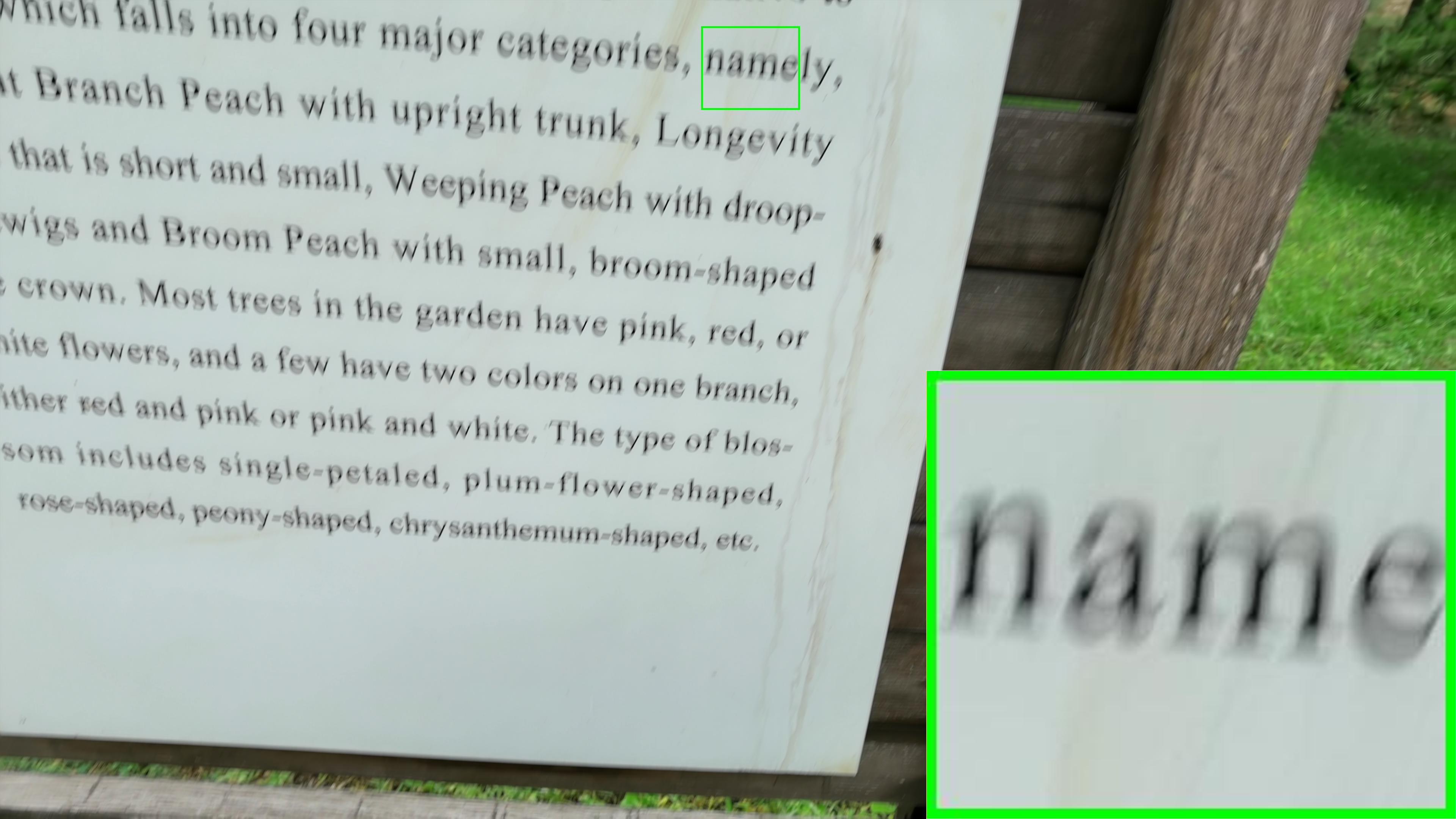} &\hspace{-4.75mm}
\includegraphics[width=0.2\linewidth]{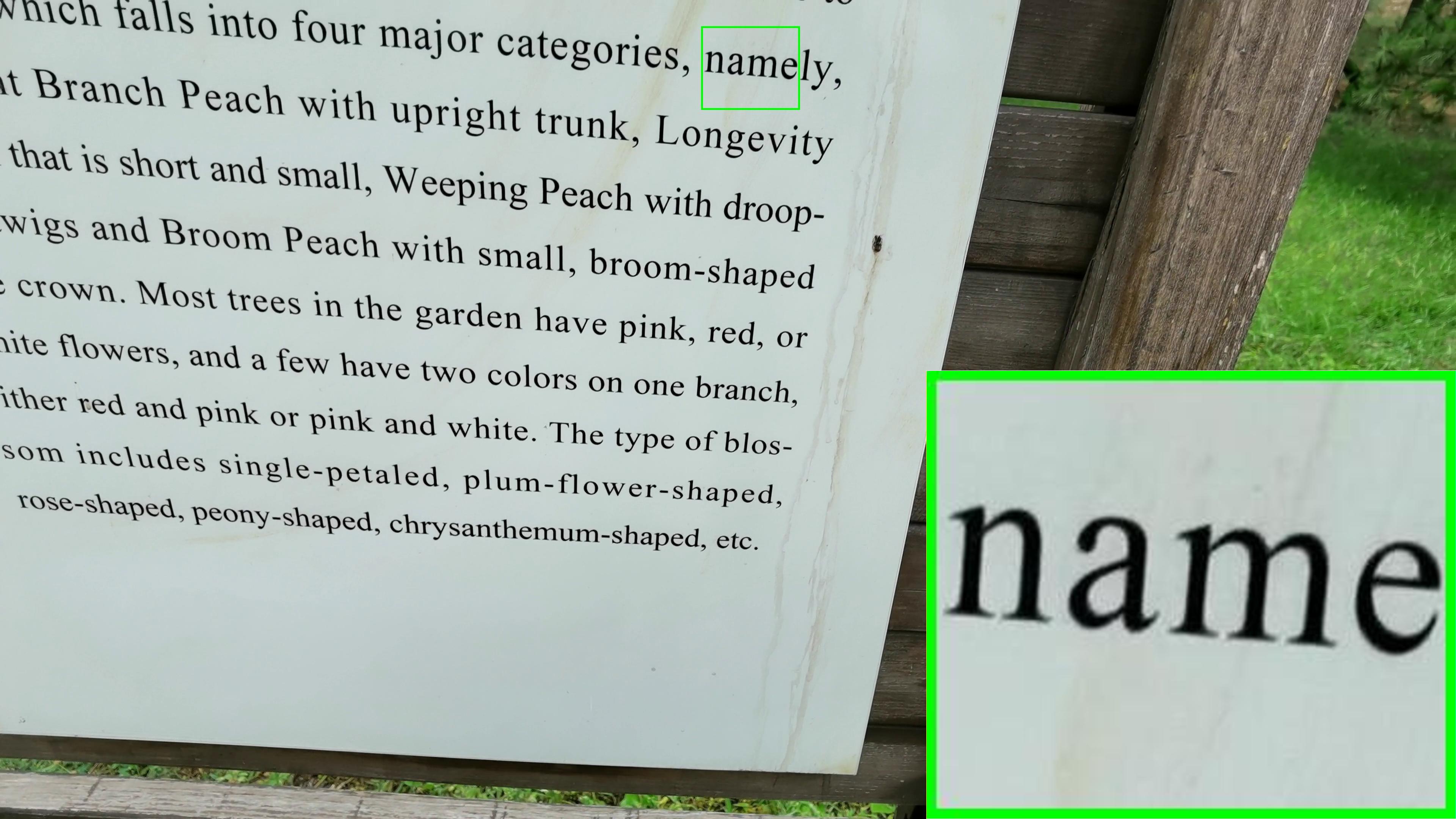} &\hspace{-4.75mm}
\includegraphics[width=0.2\linewidth]{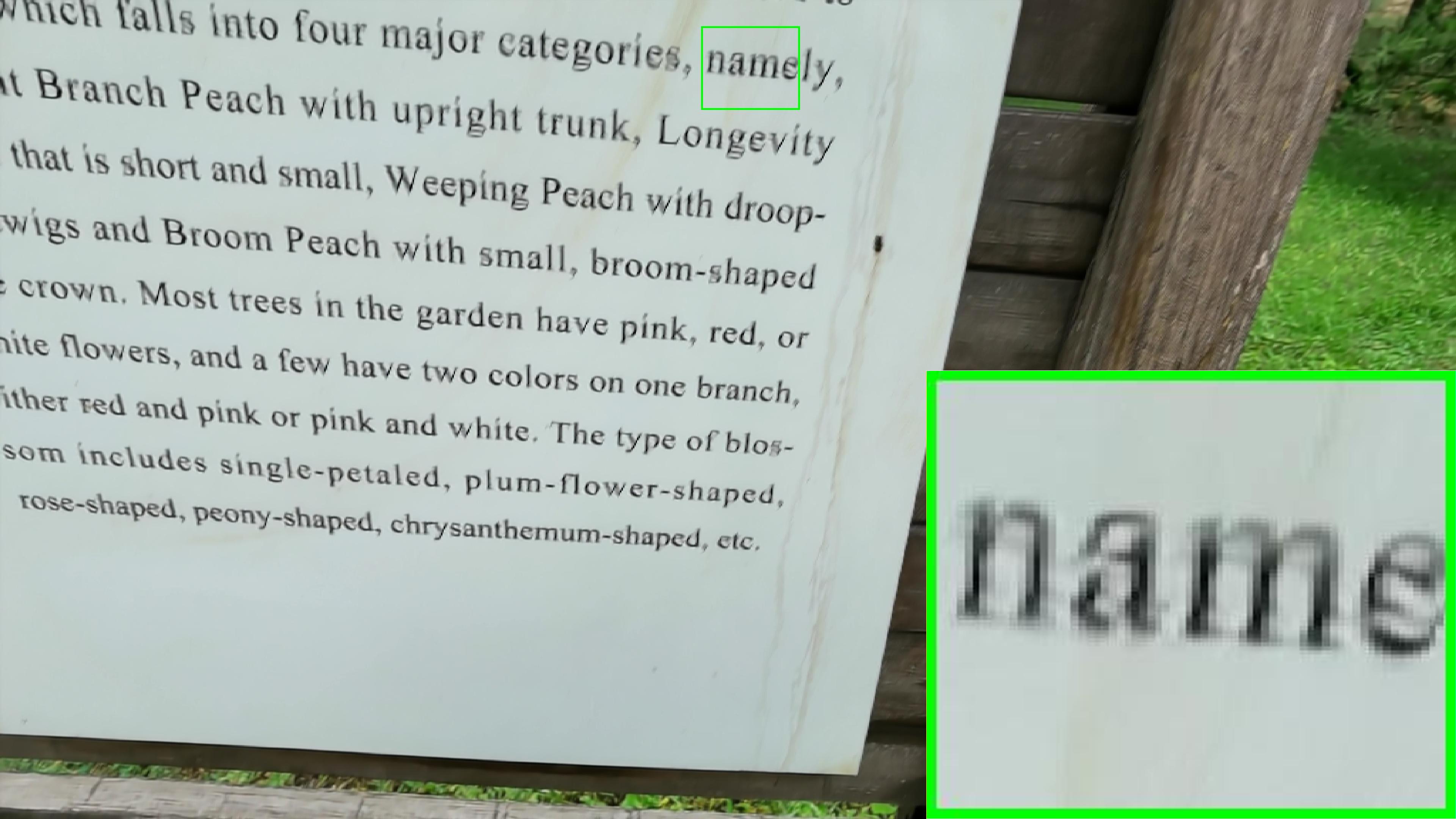} &\hspace{-4.75mm}
\includegraphics[width=0.2\linewidth]{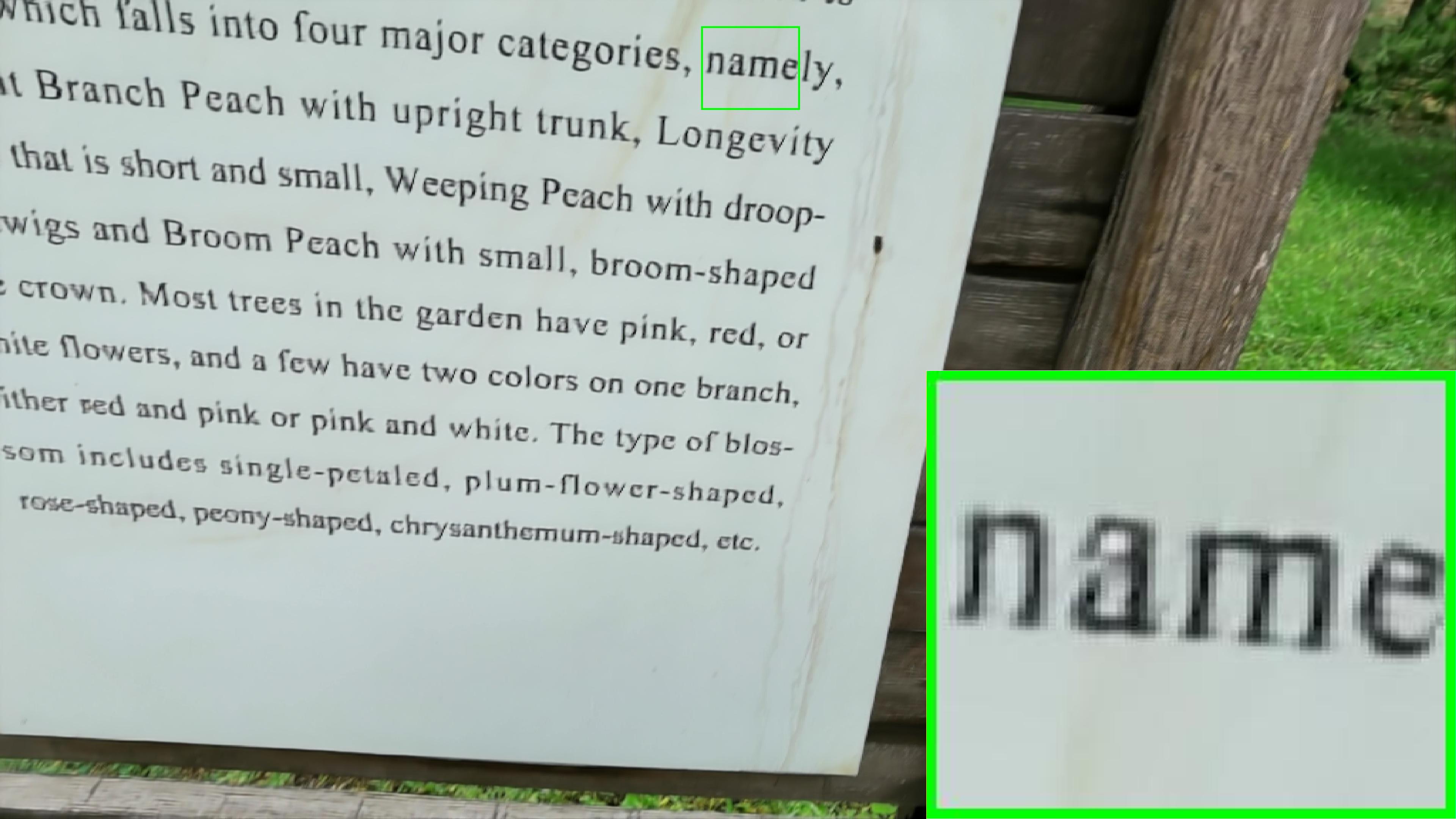} 

\\
 (a) Input&\hspace{-4.75mm} (b) GT&\hspace{-4.75mm}(c) Restormer~\cite{Zamir2021Restormer}  &\hspace{-4.75mm} (d) Uformer~\cite{wang2021uformer} 
\\

\includegraphics[width=0.2\linewidth]{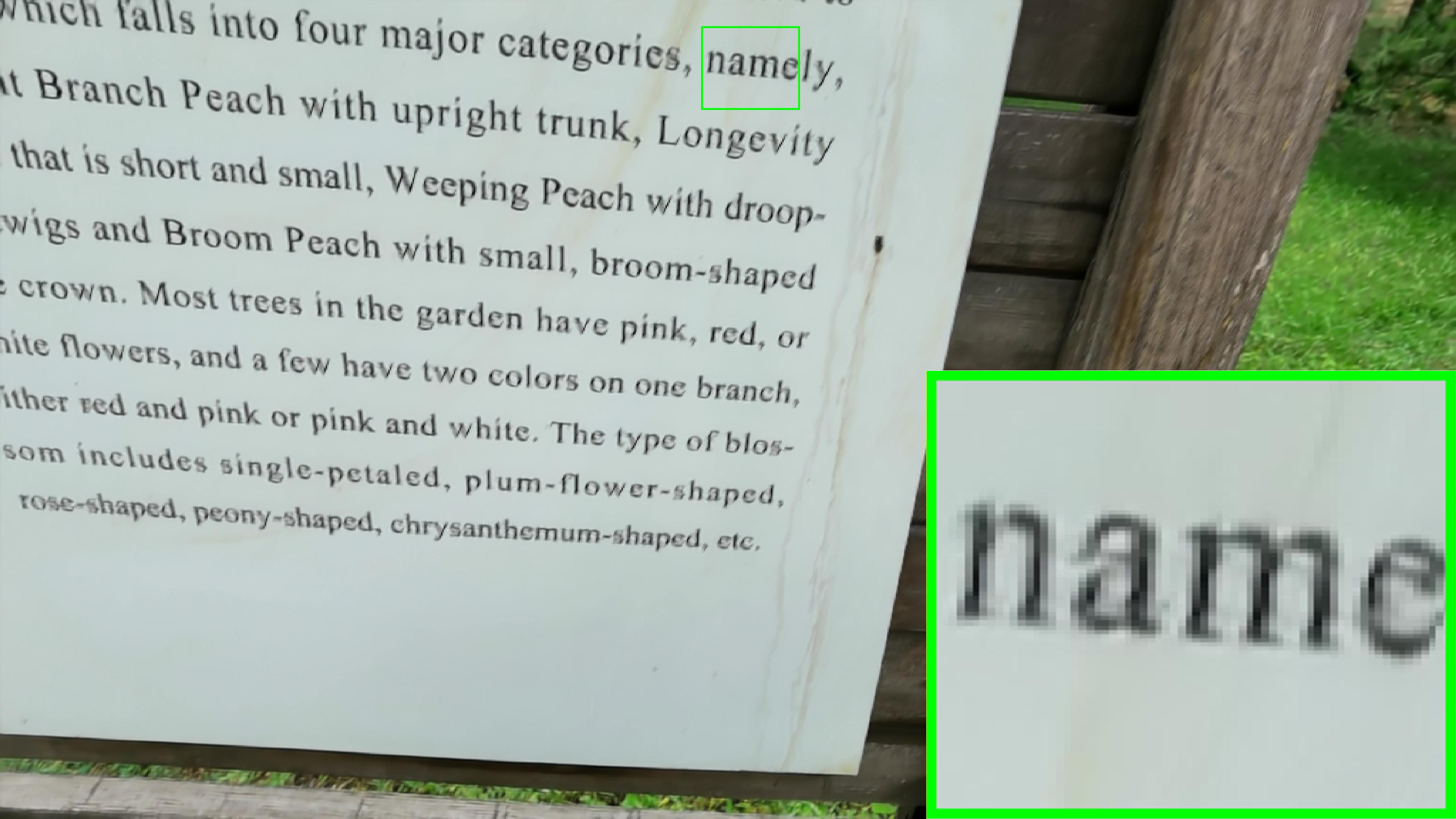} &\hspace{-4.75mm}
\includegraphics[width=0.2\linewidth]{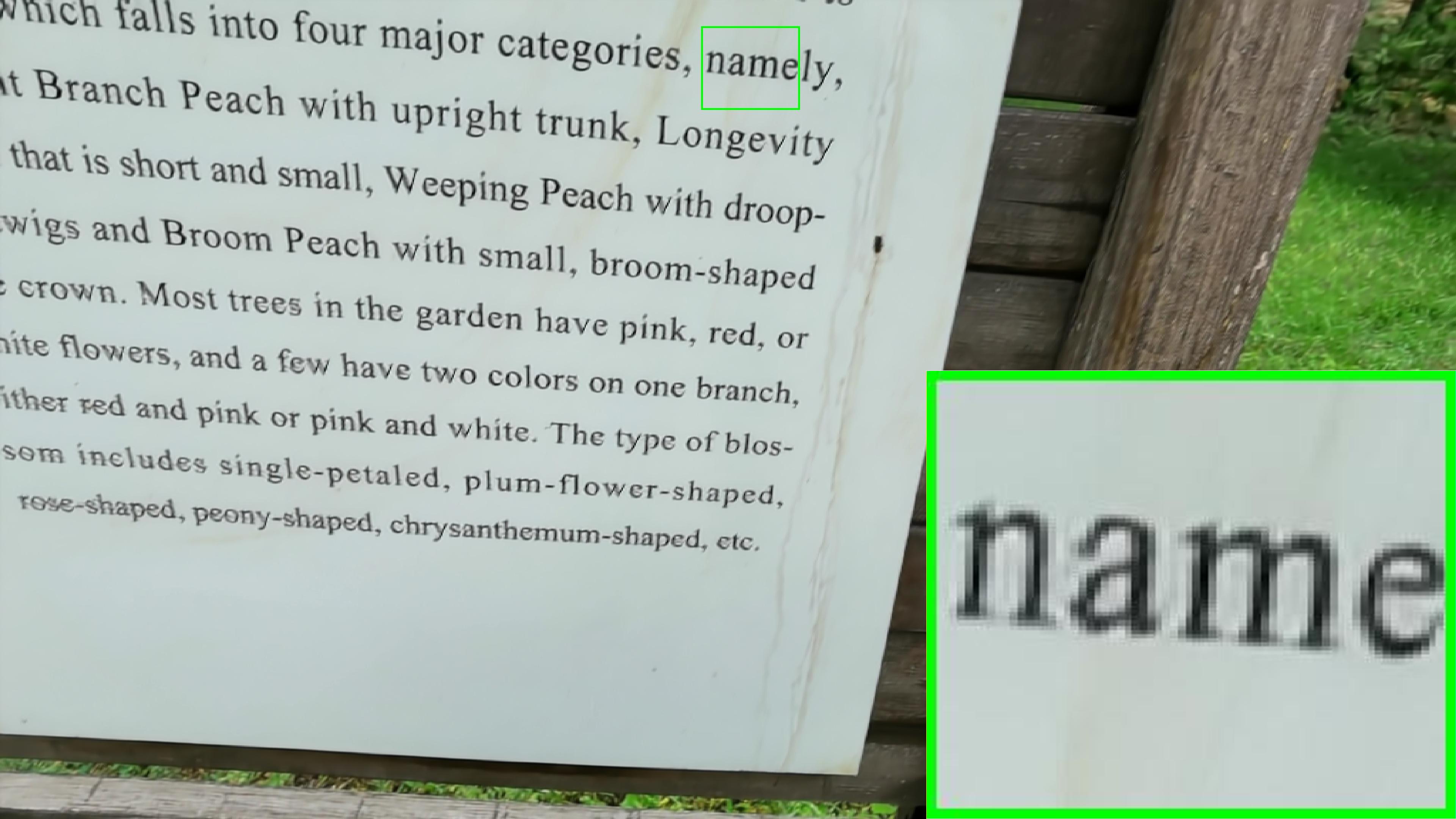} &\hspace{-4.75mm}
\includegraphics[width=0.2\linewidth]{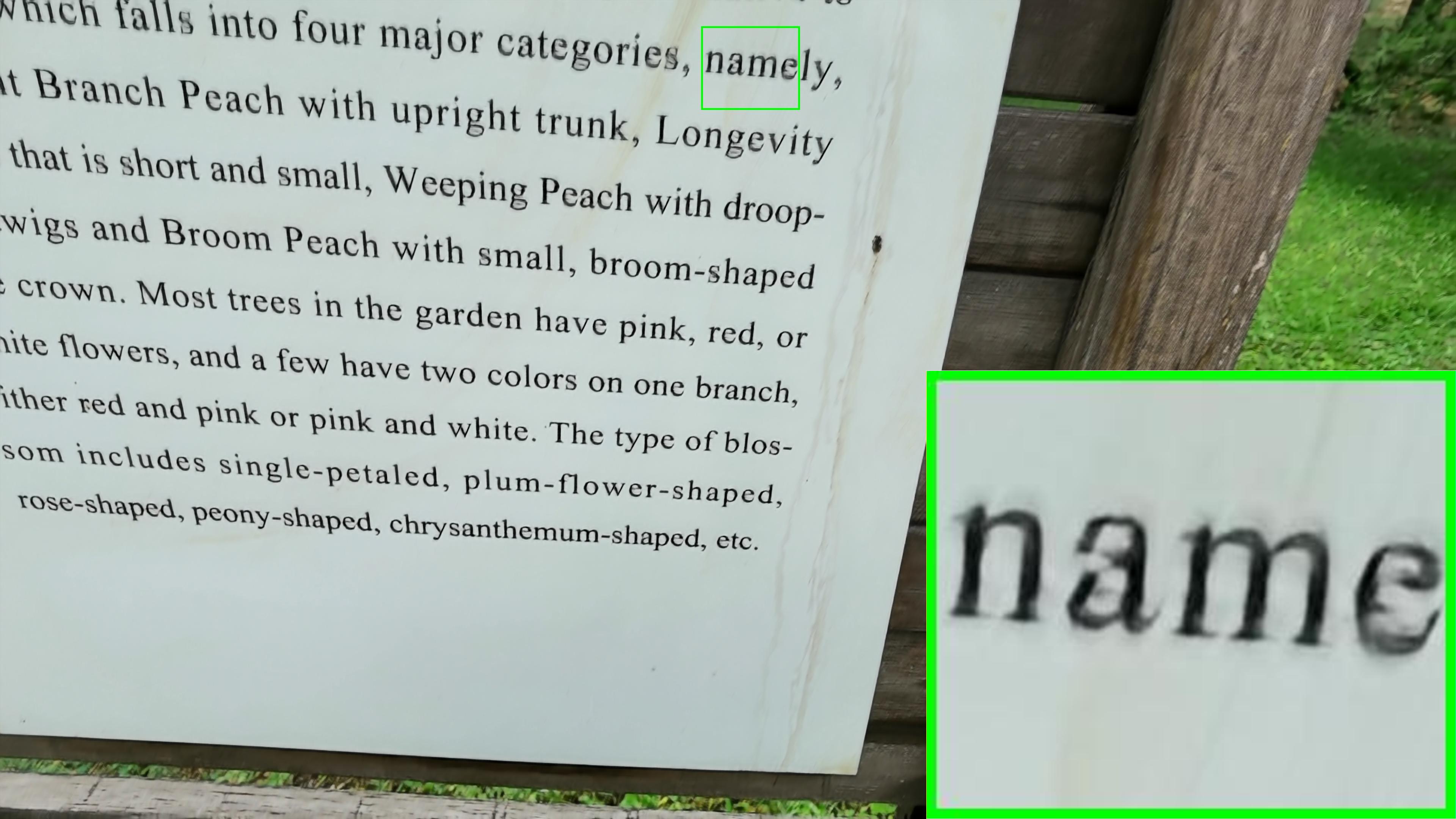}
&\hspace{-4.75mm}
\includegraphics[width=0.2\linewidth]{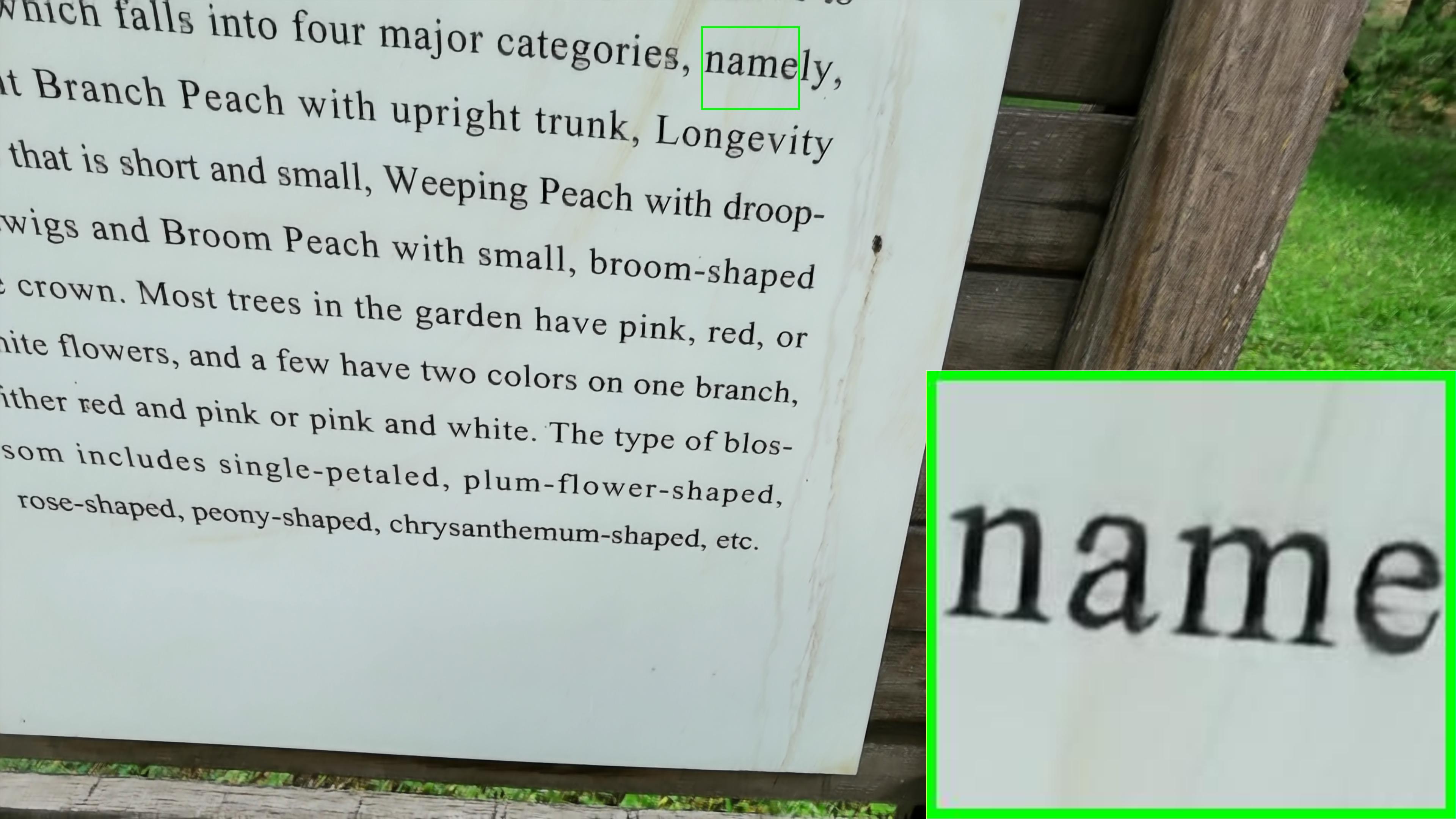}
\\
 (e) Stripformer~\cite{Tsai2022Stripformer}&\hspace{-4.75mm} (f) FFTformer~\cite{Kong_2023_CVPR_fftformer}&\hspace{-4.75mm} (g) UHDformer~\cite{aaai24wang_UHDformer}  &\hspace{-4.75mm} (h) \textbf{UHDformer++}  
\end{tabular}
\vspace{-2mm}
\caption{\textbf{Image deblurring on UHD-Blur \textit{trained on the UHD-Blur dataset}.}
% \textbf{UHDformer++} is able to generate much sharper results with fewer blur residuals.
}
\label{fig:Image deblurring on UHD-Blur.}
\end{center}
\vspace{-3mm}
\end{figure*}

%%%%%%%%%%%%%%%%%%%%%%%%%%%%%%%%%%

%%%%%%%%%%%%%%%%%%%%%%%%%%%%%%%%%%%%%%%%%%%%%%%%%%%%%%%%%%%%%%%%%%%%%%%%%%%%

\subsubsection{Image Dehazing Results} 
We evaluate image dehazing performance on the UHD-Haze benchmark under two training configurations: models trained on SOTS-ITS and models trained on UHD-Haze, with all methods assessed on the UHD-Haze test set. 
Tab.~\ref{tab:Image dehazing.} reports the quantitative results.
UHDformer++ consistently achieves superior performance over UHDformer~\cite{aaai24wang_UHDformer} across both PSNR, SSIM, PI, NIQE, and MIUSIQ metrics. 
Compared to DehazeFormer-B~\cite{DehazeFormer}, a recent method that requires resizing for UHD input, UHDformer++ not only reduces model size by 84\% but also achieves PSNR gains of $1.124$dB and $9.717$dB on the SOTS-ITS and UHD-Haze training sets, respectively. 
Furthermore, while GridNet~\cite{grid_dehaze_liu}, UHD~\cite{Zheng_uhd_CVPR21}, and MSBDN~\cite{msbdn_cvpr20_dong} are capable of processing full-resolution UHD images without resizing, they still exhibit limited dehazing effectiveness when trained on SOTS-ITS. 
In contrast, UHDformer++ maintains competitive performance under this cross-dataset setting, further corroborating the robustness and generalization ability of our approach.
Visual comparisons are presented in Fig.~\ref{fig:Image dehazing on UHD-Haze.}, where  UHDformer++ produces notably clearer outputs with substantially reduced haze residuals, whereas competing methods consistently leave considerable haze artifacts in their results.

\begin{table*}[!t]
\caption{\textbf{Image deraining on UHD-Rain dataset}. 
}
\vspace{-2mm}
\label{tab:Image deraining.} 
\tablestyle{1pt}{1}
\centering
\begin{tabular}{l|ccc|ccc|ccc|c}
\shline
 \multicolumn{1}{c|}{\multirow{3}{*}{\textbf{Method}}} & \multicolumn{6}{c|}{\textbf{Metrics}}&\multicolumn{3}{c|}{\multirow{2}{*}{\textbf{Computational Complexity}}}& \multirow{3}{*}{\textbf{Full Size}} 
 \\
  \cline{2-7}
&\multicolumn{3}{c|}{\textbf{Full Reference}}&\multicolumn{3}{c|}{\textbf{No-Reference}}&\multicolumn{3}{c|}{}
 \\
 \cline{2-10}
&\textbf{PSNR (dB)~$\uparrow$} 
&\textbf{SSIM~$\uparrow$}
&\textbf{LPIPS~$\downarrow$}
&\textbf{PI~$\downarrow$} 
&\textbf{NIQE~$\downarrow$} 
&\textbf{MIUSIQ~$\uparrow$}
&\textbf{Params. (M)~$\downarrow$} 
&\textbf{FLOPs (G)~$\downarrow$}
&\textbf{Time (s)~$\downarrow$}
\\
\shline
Restormer~\cite{Zamir2021Restormer}&19.408& 0.7105 &0.4775&8.8409&11.2731 &24.9277& 26.1 & 2255.85&1.27&\XSolidBrush\\
Uformer~\cite{wang2021uformer}&19.494& 0.7163 &0.4598&8.0579&9.9422 &25.0728& 20.6 & 657.45&0.43 &\XSolidBrush\\
SFNet~\cite{0001TBRGC0K23sfnet}&20.091 &0.7092& 0.4768 &7.1914&8.0167 &22.3553&13.23&1991.03&0.42&\XSolidBrush\\
UHDformer~\cite{aaai24wang_UHDformer}&\underline{37.348} &\underline{0.9748}& \underline{0.0554} &\underline{5.0464}&\underline{4.9256} &\underline{30.4390} &\textbf{0.3393} & \underline{51.63}&\underline{0.16}& \CheckmarkBold\\
\textbf{UHDformer++ (Ours)}&\textbf{38.839}&\textbf{0.9791}&\textbf{0.0394}&\textbf{5.0381}&\textbf{4.9024}&\textbf{30.5399}&\underline{0.3962}&\textbf{49.65}&\textbf{0.12}& \CheckmarkBold\\
\shline
\end{tabular}
% \vspace{-3mm}
\end{table*}
%%%%%%%%%%%%%%%%%%%%%%%%%%%%%%%%%%%%%%%
\begin{figure*}[!t]
\centering
\begin{center}
\begin{tabular}{cccccc}
\hspace{-1mm}\includegraphics[width=0.14\linewidth]{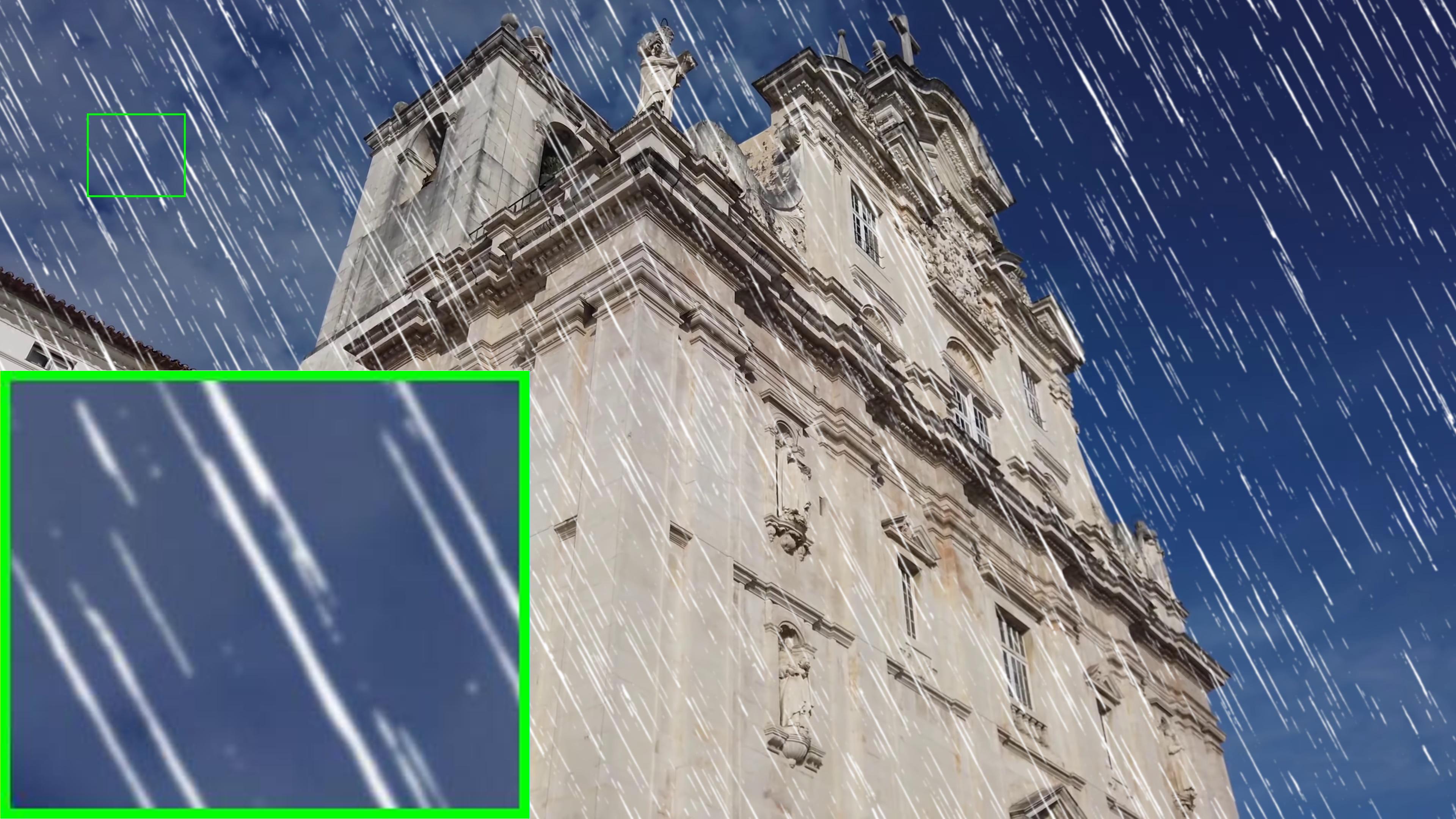} &\hspace{-4mm}
\includegraphics[width=0.14\linewidth]{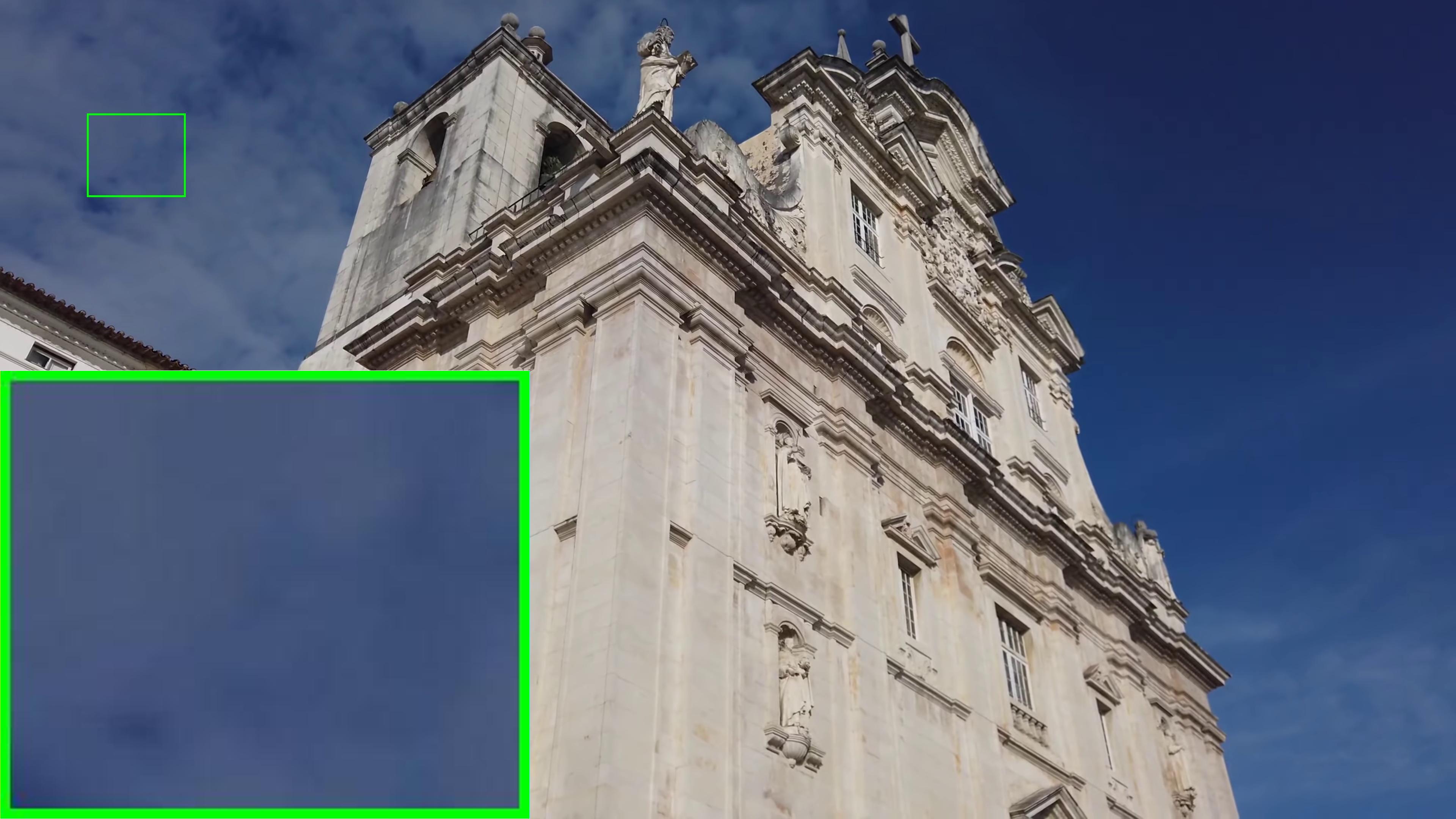} &\hspace{-4mm}
\includegraphics[width=0.14\linewidth]{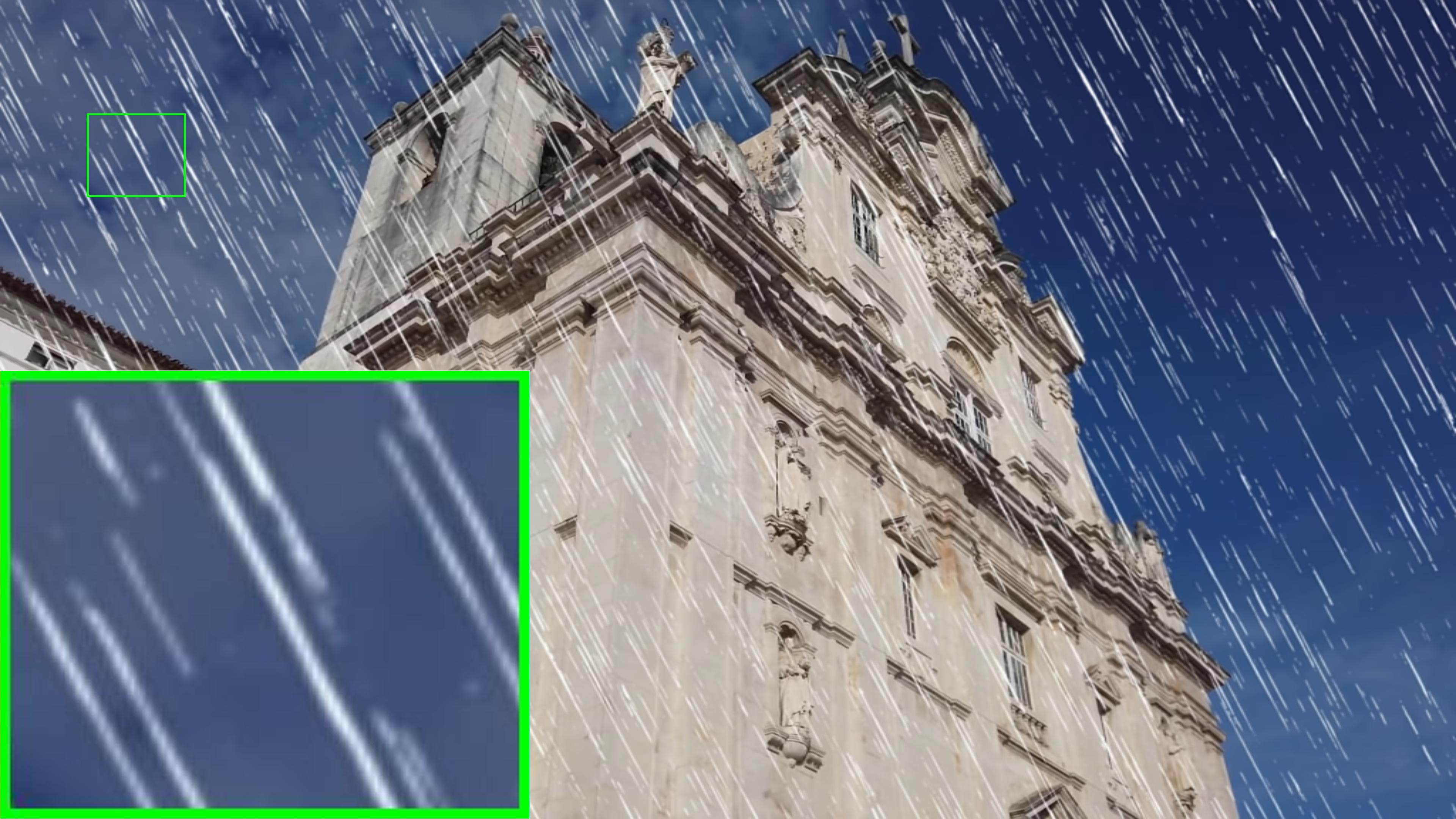} &\hspace{-4mm}
\includegraphics[width=0.14\linewidth]{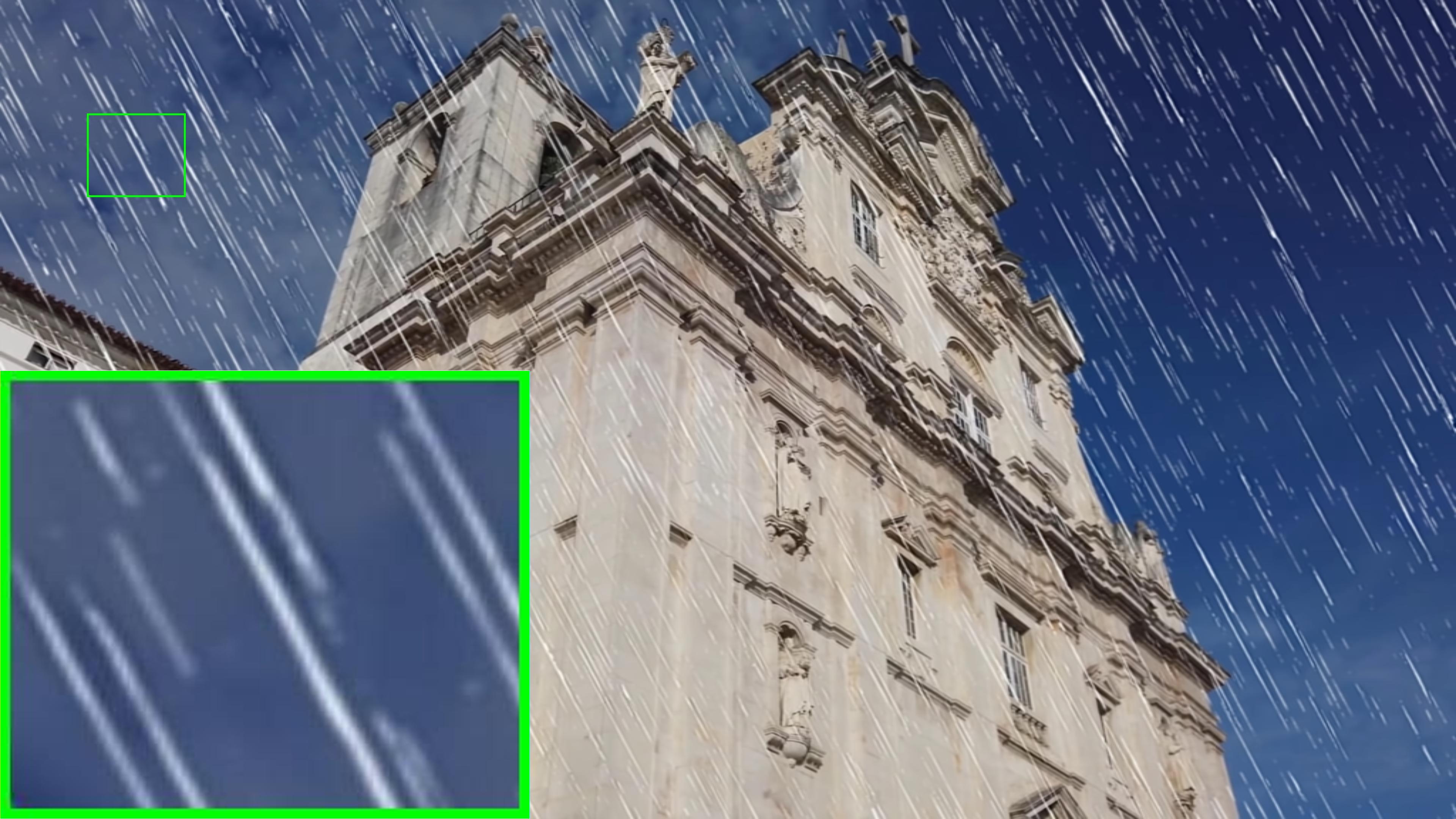} 
&\hspace{-4mm}
\includegraphics[width=0.14\linewidth]{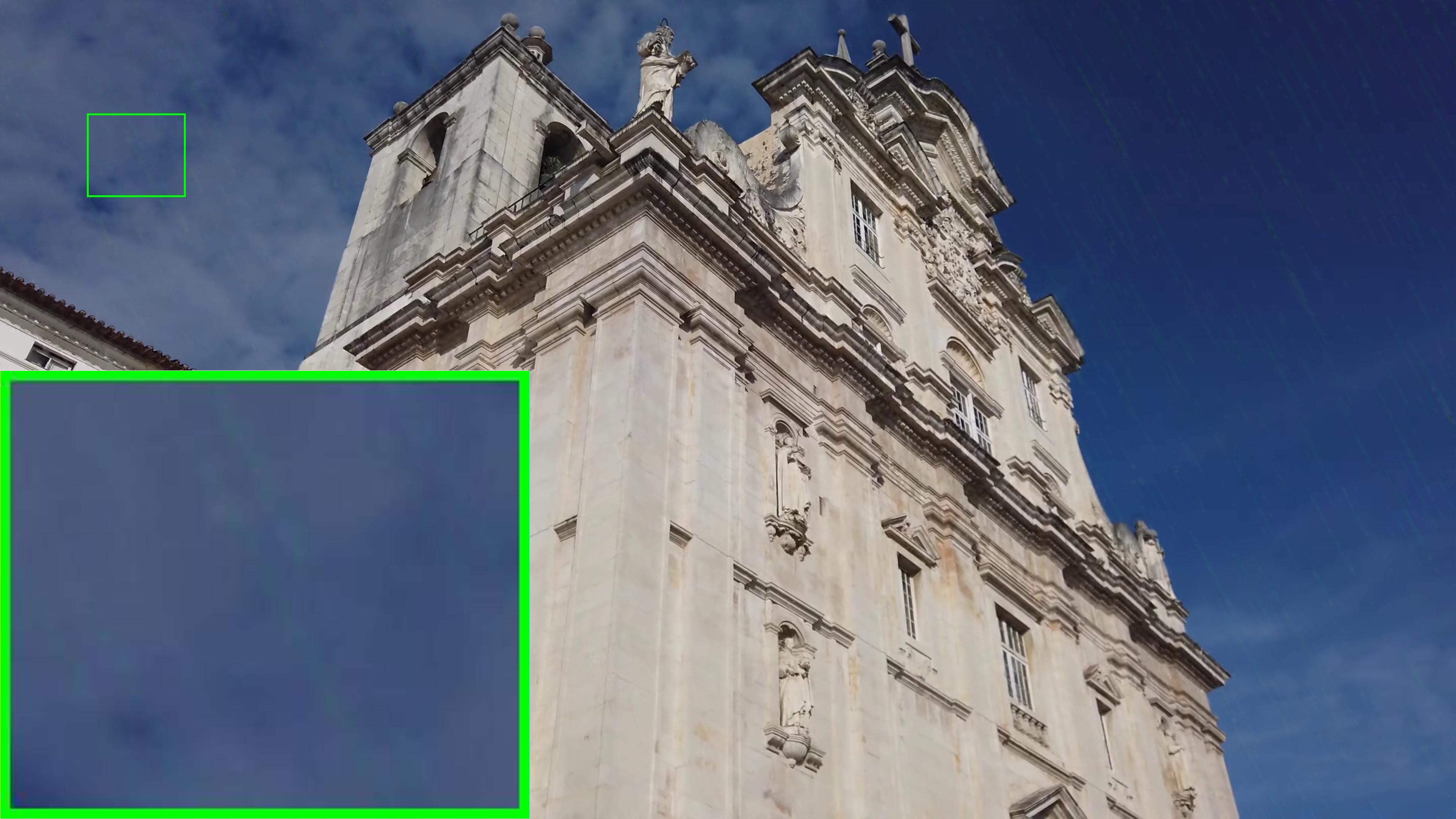} 
&\hspace{-4mm}
\includegraphics[width=0.14\linewidth]{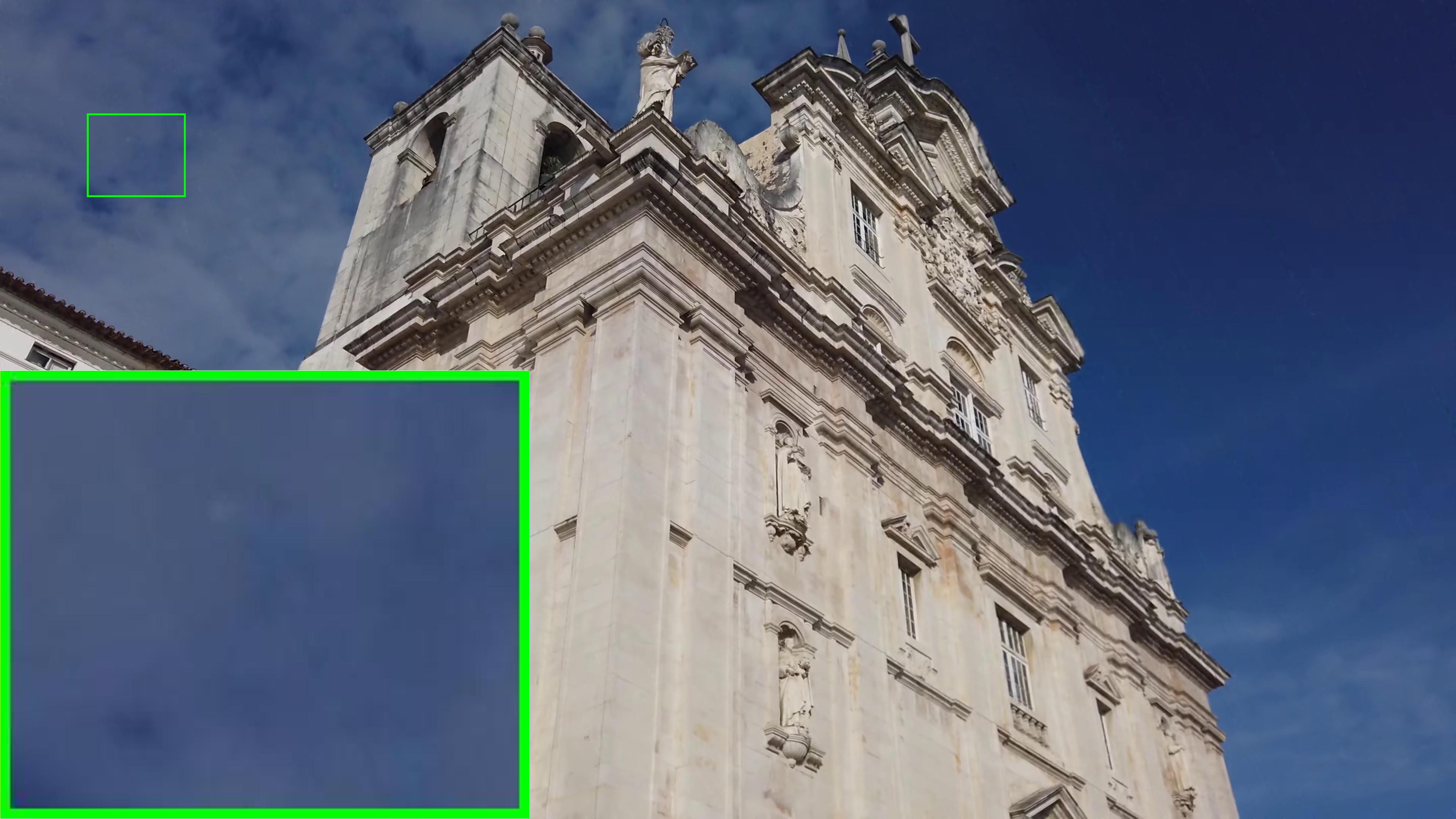} 
\\
\hspace{-1mm}\includegraphics[width=0.14\linewidth]{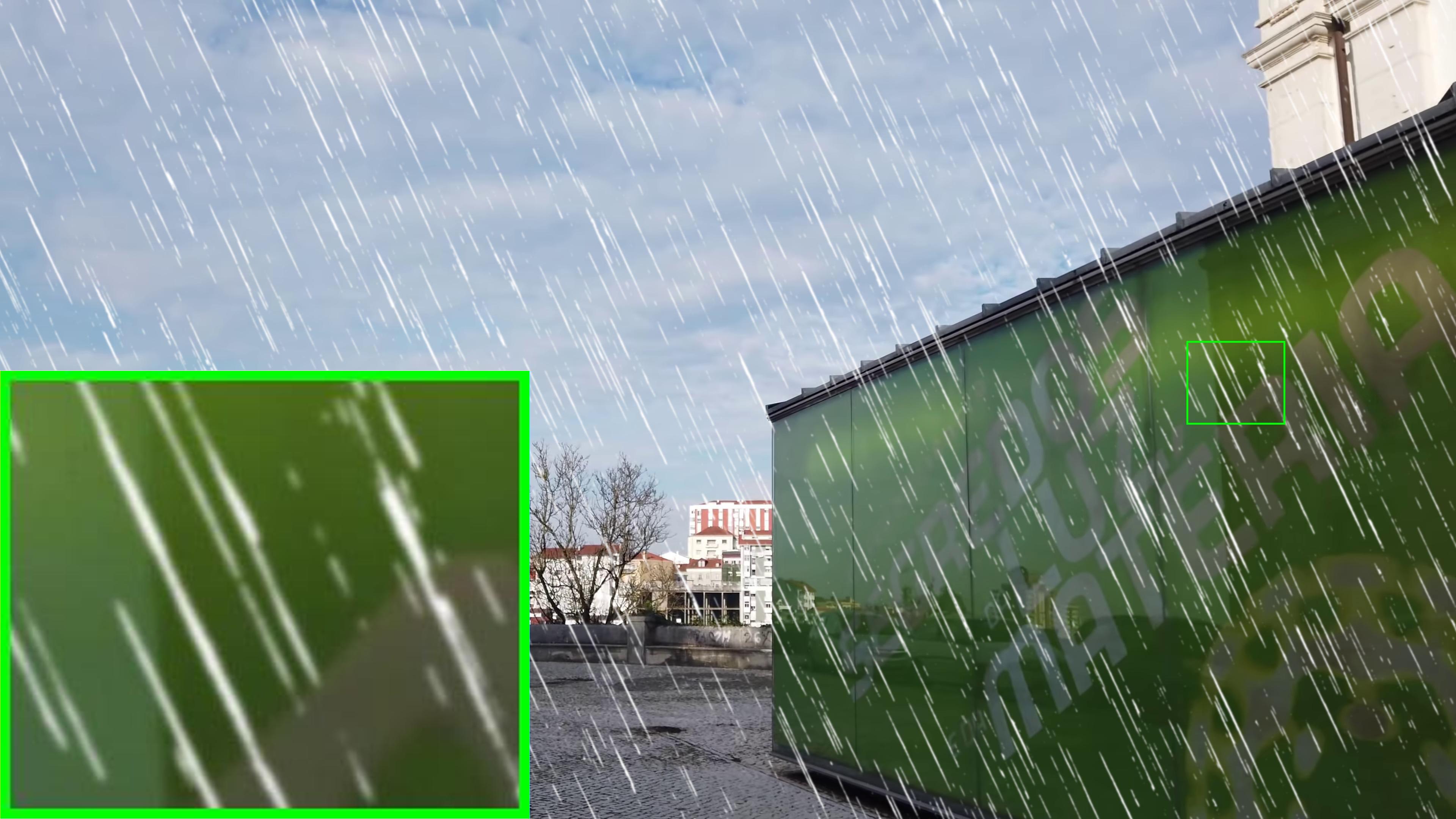} &\hspace{-4mm}
\includegraphics[width=0.14\linewidth]{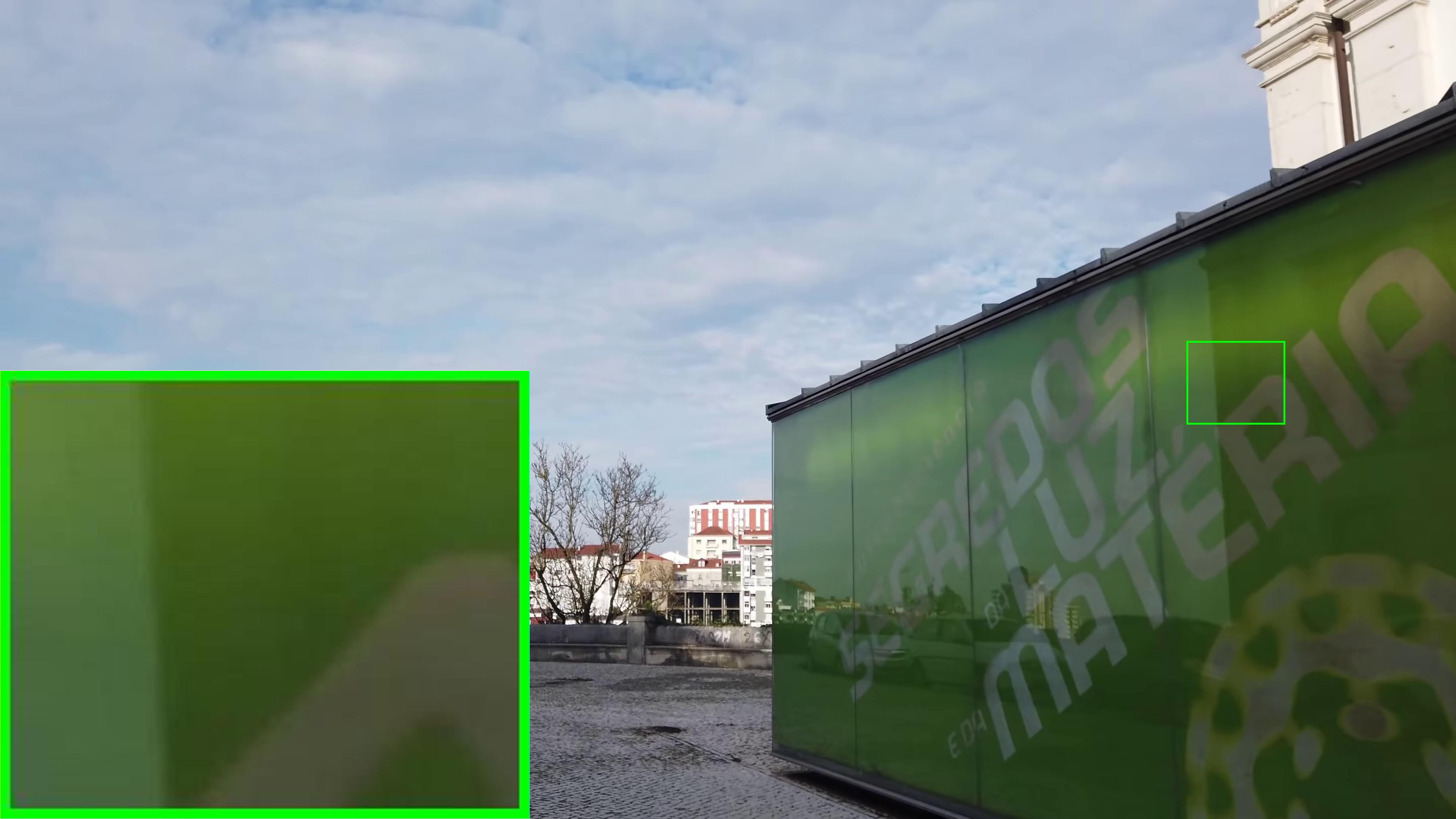} &\hspace{-4mm}
\includegraphics[width=0.14\linewidth]{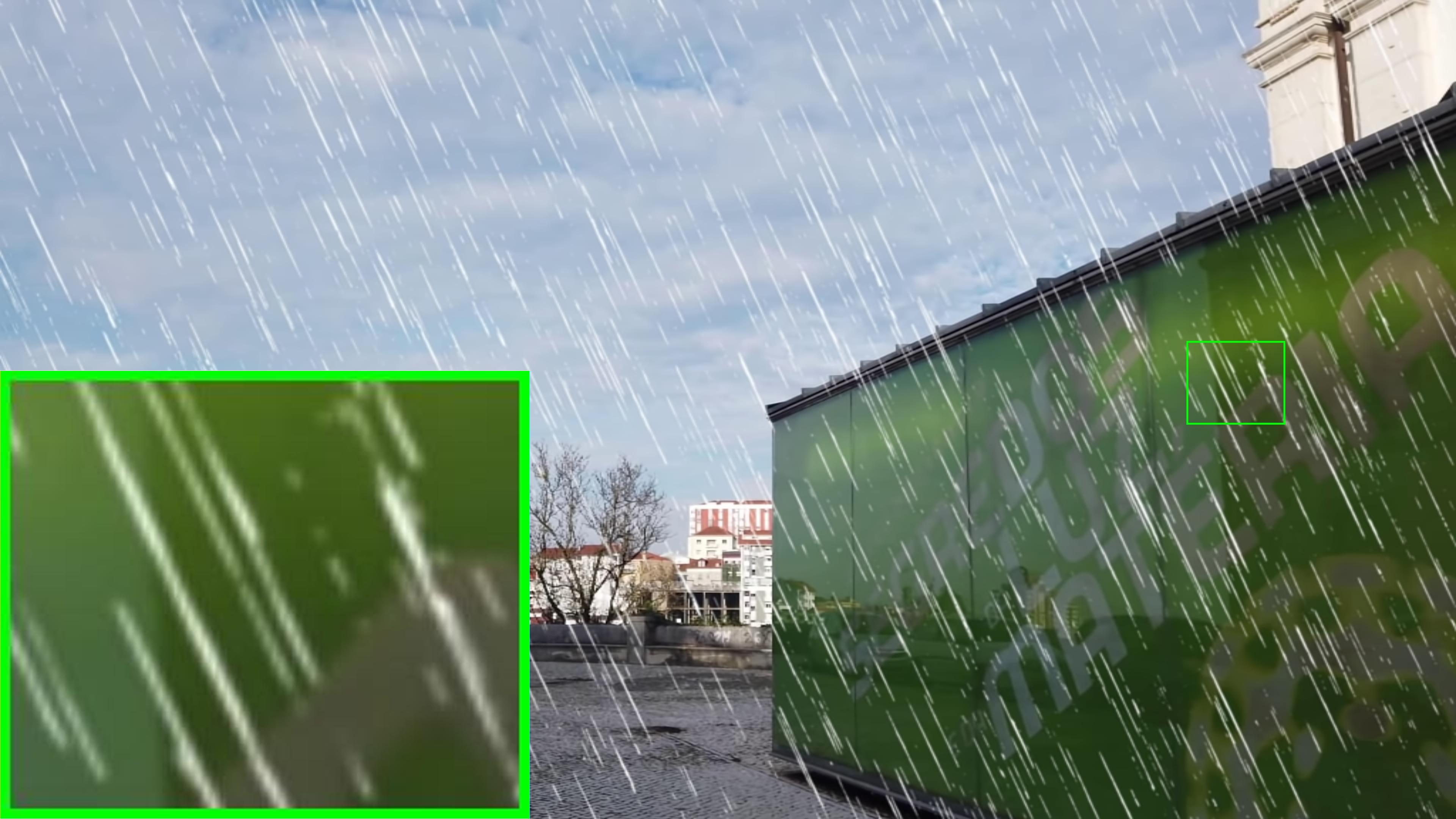} &\hspace{-4mm}
\includegraphics[width=0.14\linewidth]{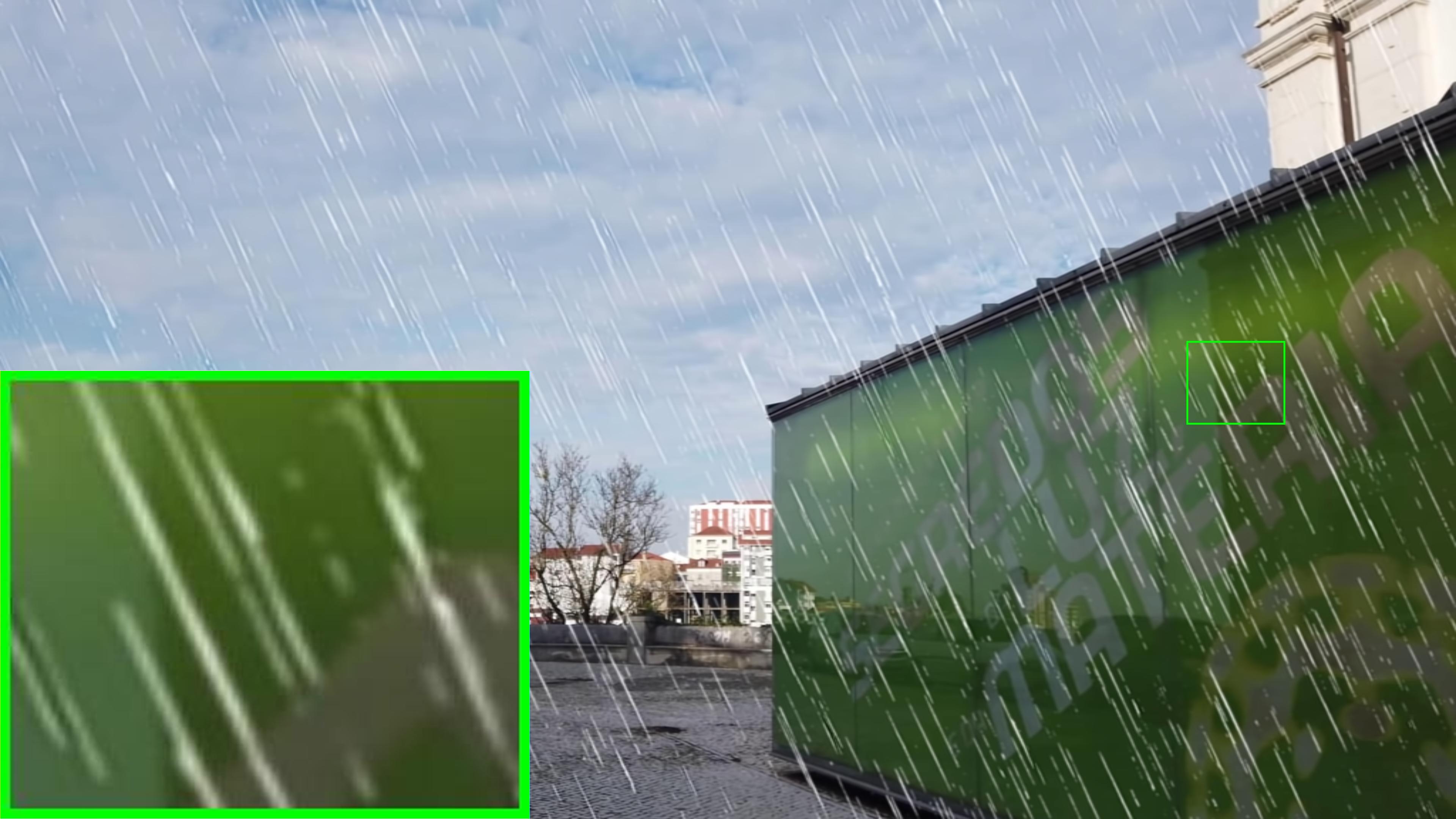} 
&\hspace{-4mm}
\includegraphics[width=0.14\linewidth]{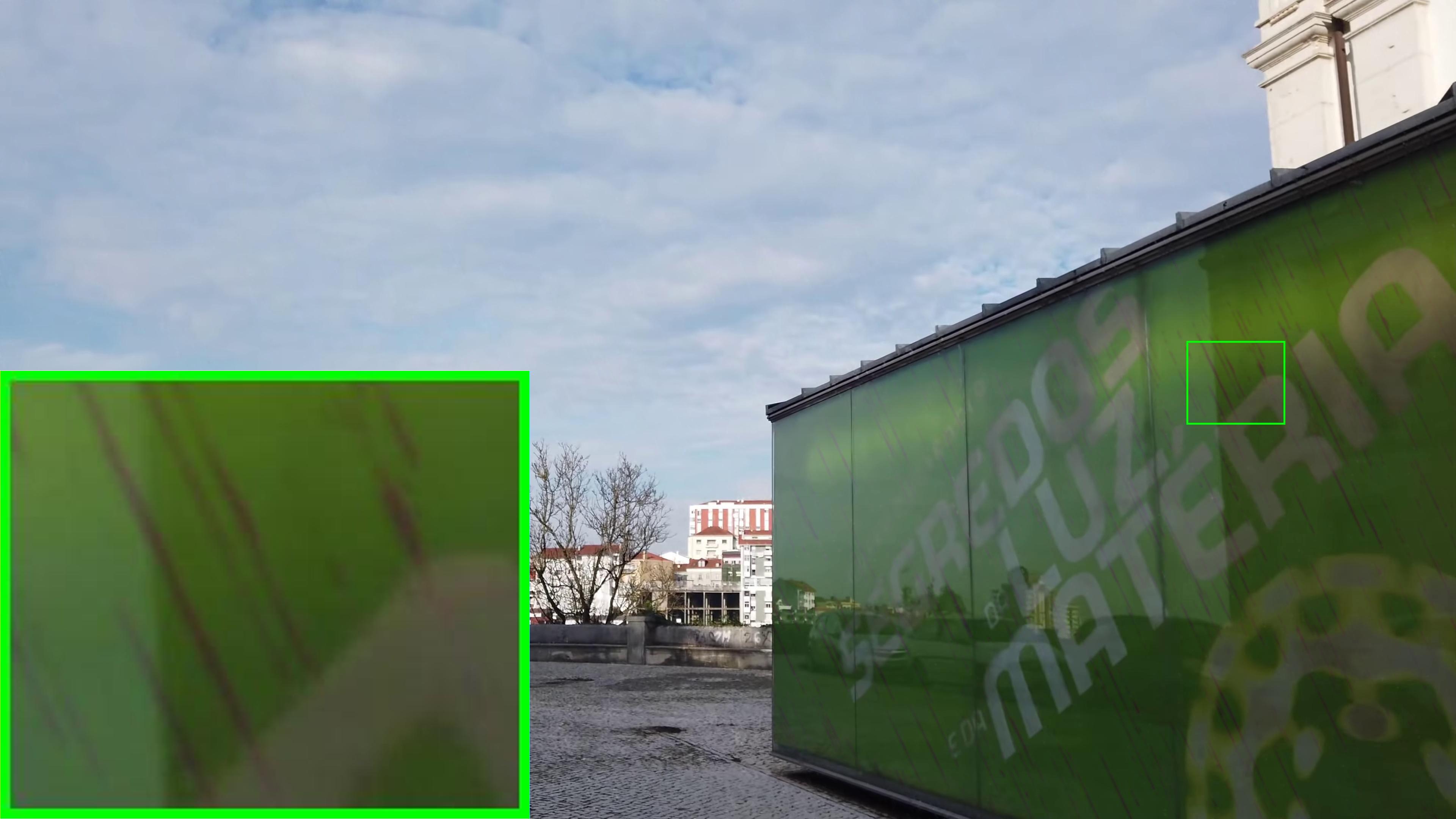} 
&\hspace{-4mm}
\includegraphics[width=0.14\linewidth]{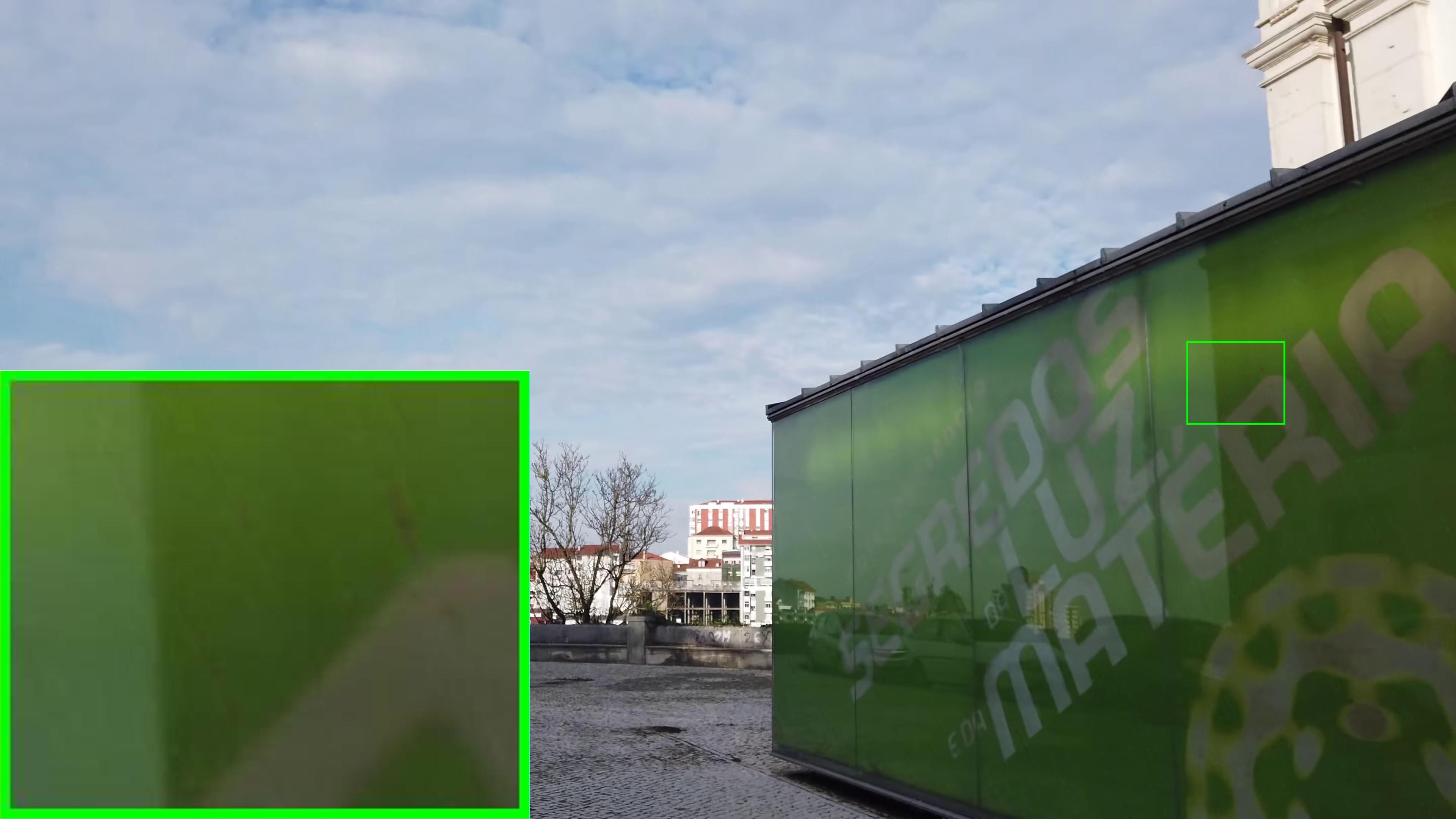} 
\\
 (a) Input&\hspace{-4mm} (b) GT&\hspace{-4mm}(c) Restormer~\cite{Zamir2021Restormer} &\hspace{-4mm} (d) SFNet~\cite{0001TBRGC0K23sfnet} &\hspace{-4mm}(e) UHDformer~\cite{aaai24wang_UHDformer} &\hspace{-4mm} (f) \textbf{UHDformer++}
\end{tabular} 
 \vspace{-2mm}
\caption{\textbf{Image deraining on UHD-Rain \textit{trained on the UHD-Rain dataset}.}
% \textbf{UHDformer++} is able to generate much sharper results with blur residuals.
}
\label{fig:Image deraining on UHD-Rain.}
\end{center}
\vspace{-3mm}
\end{figure*}
\begin{table*}[!t]
\caption{\textbf{Image desnowing on UHD-Snow dataset}. 
%
% \textbf{UHDformer++} with at least 98\% fewer parameters advances state-of-the-art methods.
}
\vspace{-2mm}
\label{tab:Image desnowing.} 
\tablestyle{1pt}{1}
\centering
\begin{tabular}{l|ccc|ccc|ccc|c}
\shline
 \multicolumn{1}{c|}{\multirow{3}{*}{\textbf{Method}}} & \multicolumn{6}{c|}{\textbf{Metrics}}&\multicolumn{3}{c|}{\multirow{2}{*}{\textbf{Computational Complexity}}}& \multirow{3}{*}{\textbf{Full Size}} 
 \\
  \cline{2-7}
&\multicolumn{3}{c|}{\textbf{Full Reference}}&\multicolumn{3}{c|}{\textbf{No-Reference}}&\multicolumn{3}{c|}{}
 \\
 \cline{2-10}
&\textbf{PSNR (dB)~$\uparrow$} 
&\textbf{SSIM~$\uparrow$}
&\textbf{LPIPS~$\downarrow$}
&\textbf{PI~$\downarrow$} 
&\textbf{NIQE~$\downarrow$} 
&\textbf{MIUSIQ~$\uparrow$}
&\textbf{Params. (M)~$\downarrow$} 
&\textbf{FLOPs (G)~$\downarrow$}
&\textbf{Time (s)~$\downarrow$}
\\
\shline
Restormer~\cite{Zamir2021Restormer}&24.142 &0.8691 &0.3190&8.0490& 9.4711& 26.0491& 26.1 & 2255.85&1.27&\XSolidBrush\\
Uformer~\cite{wang2021uformer}&23.717& 0.8711 &0.3095 &7.5036&8.5964 &26.1323&20.6 & 657.45&0.43 &\XSolidBrush\\
SFNet~\cite{0001TBRGC0K23sfnet}&23.638& 0.8456& 0.3528&7.1527& 7.3665& 22.5295&13.23&1991.03&0.42&\XSolidBrush\\
UHDformer~\cite{aaai24wang_UHDformer}&\underline{36.614}& \underline{0.9881} &\underline{0.0245}&\textbf{5.0046}& \textbf{4.7680} &\underline{31.1335} &\textbf{0.3393} & \underline{51.63}&\underline{0.16}& \CheckmarkBold\\
\textbf{UHDformer++ (Ours)}&\textbf{38.962}&\textbf{0.9898}&\textbf{0.0202}&\underline{5.0141}&\underline{4.7764}&\textbf{31.1452}&\underline{0.3962}&\textbf{49.65}&\textbf{0.12}& \CheckmarkBold\\
\shline
\end{tabular}
% \vspace{-3mm}
\end{table*}
%%%%%%%%%%%%%%%%%%%%%%%%%%%%%%%%%%%%%%%
\begin{figure*}[!t]
\centering
\begin{center}
\begin{tabular}{cccccc}
\hspace{-1mm}\includegraphics[width=0.14\linewidth]{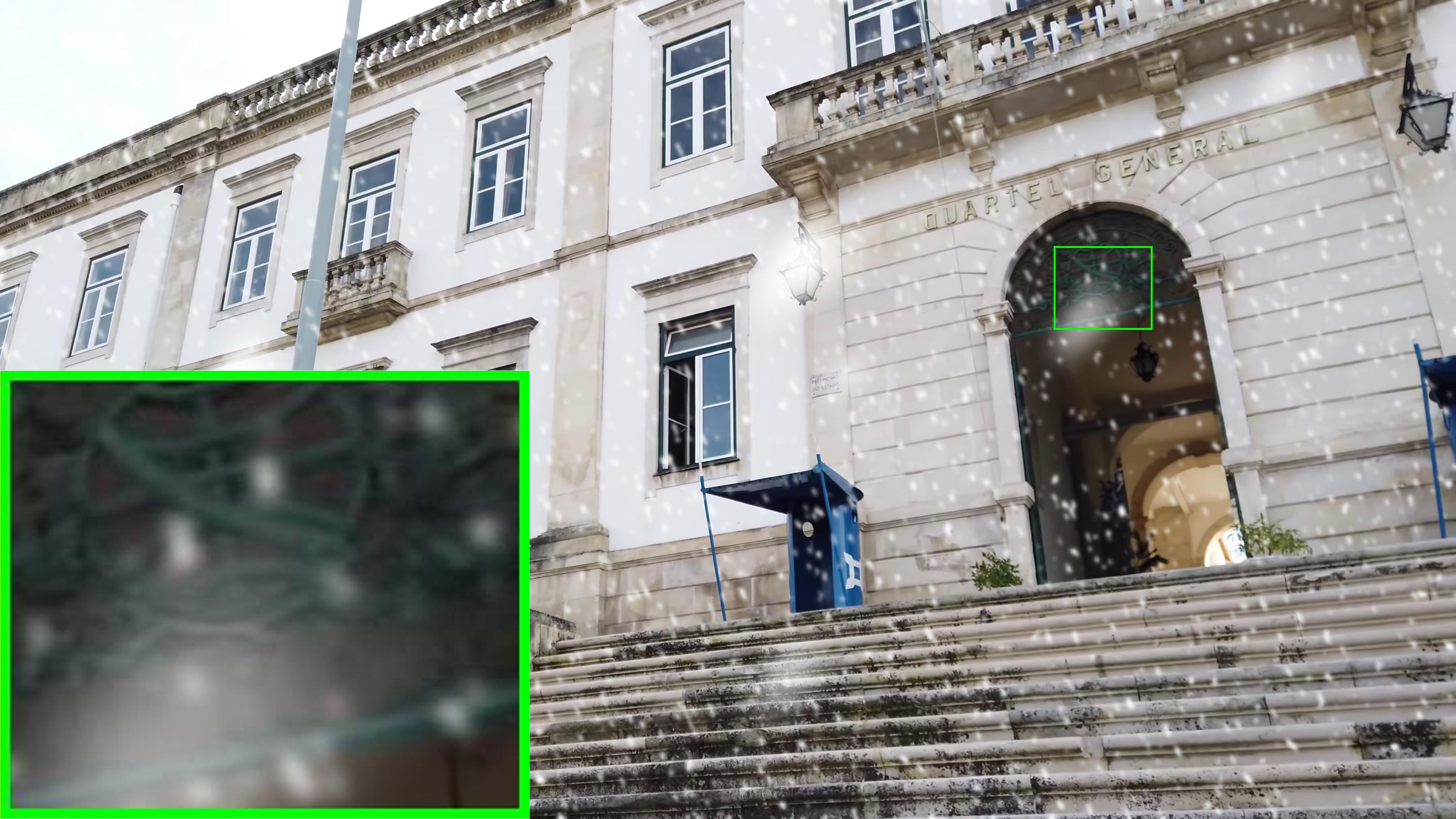} &\hspace{-4mm}
\includegraphics[width=0.14\linewidth]{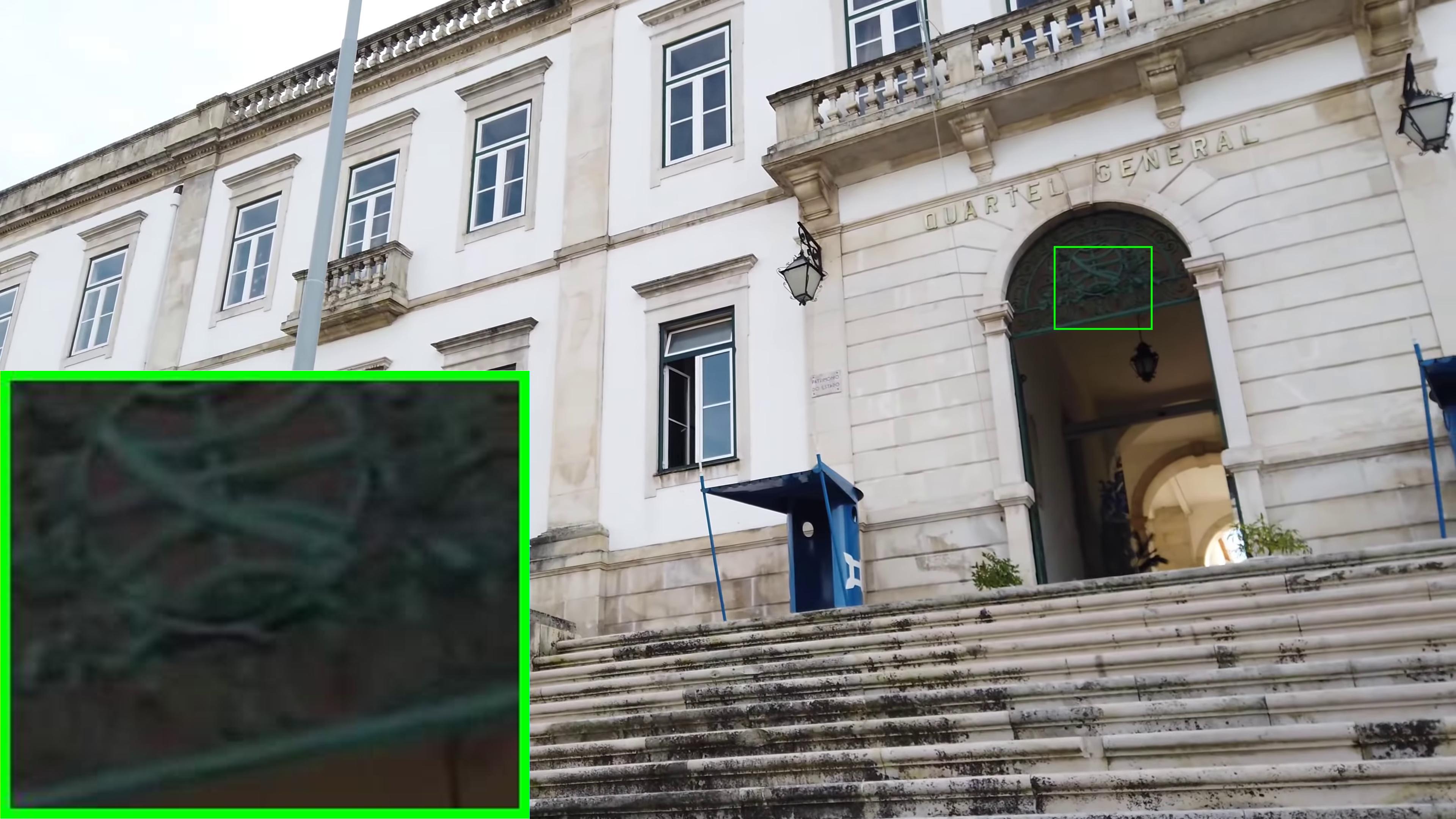} &\hspace{-4mm}
\includegraphics[width=0.14\linewidth]{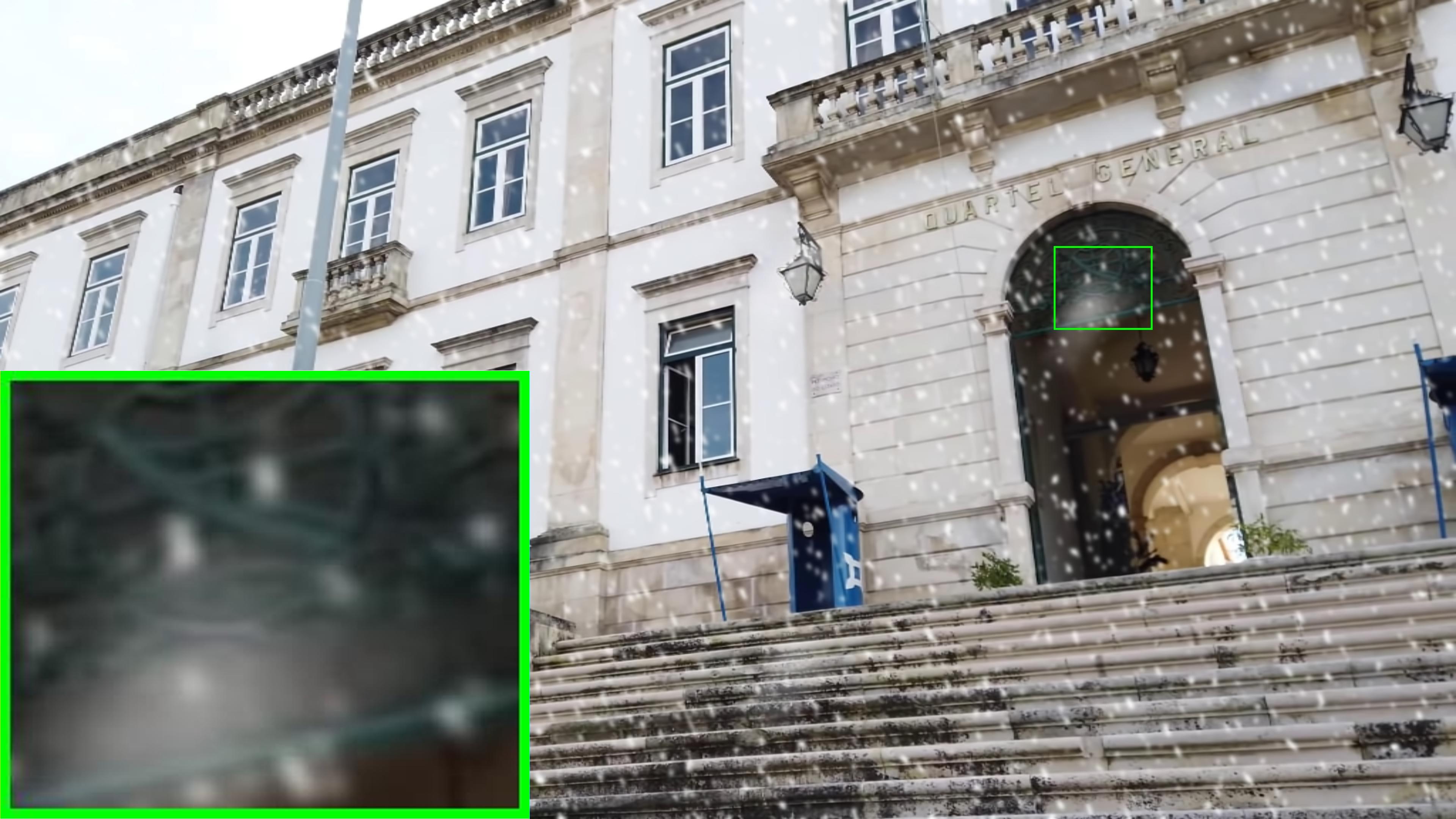} &\hspace{-4mm}
\includegraphics[width=0.14\linewidth]{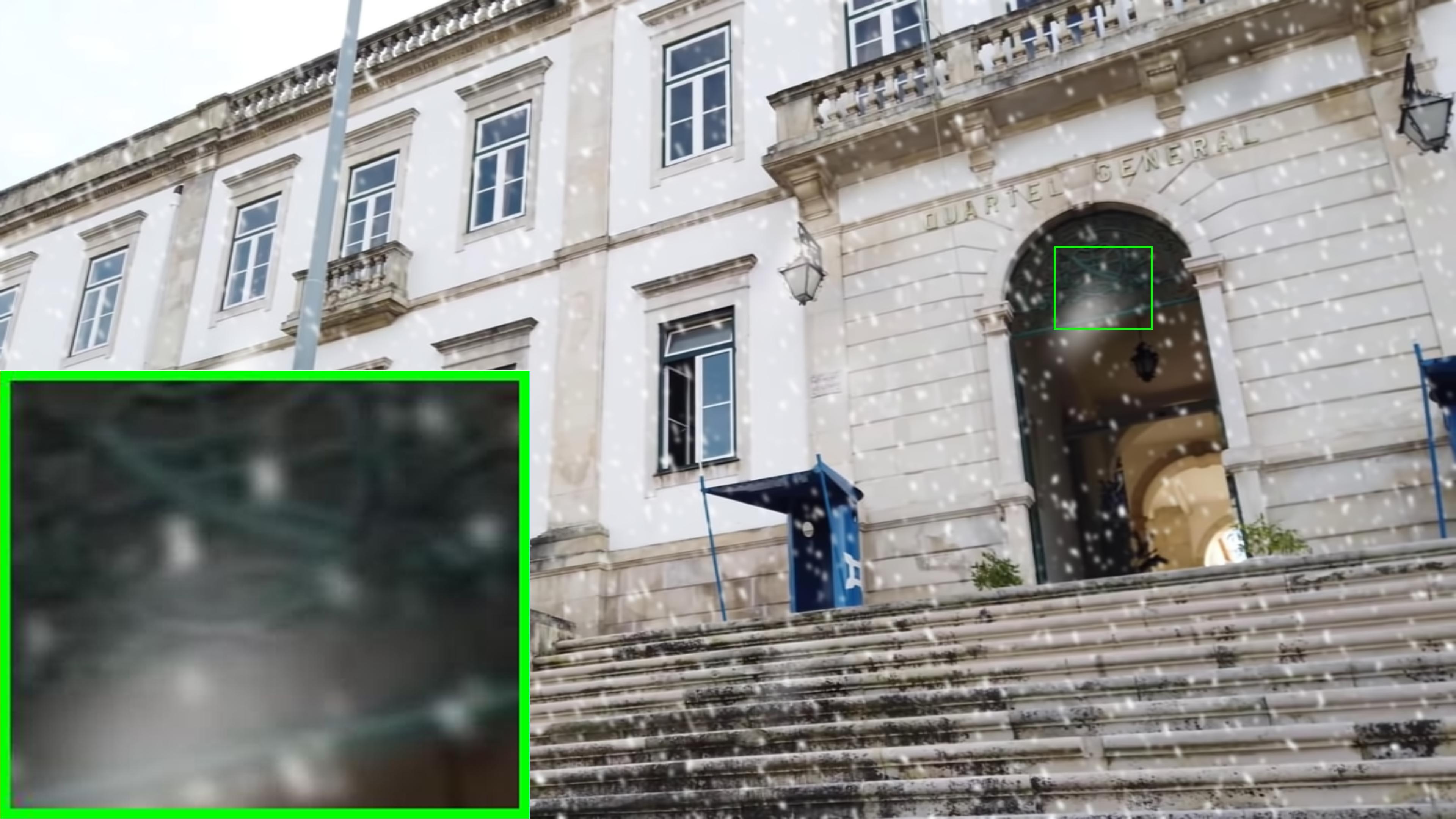} 
&\hspace{-4mm}
\includegraphics[width=0.14\linewidth]{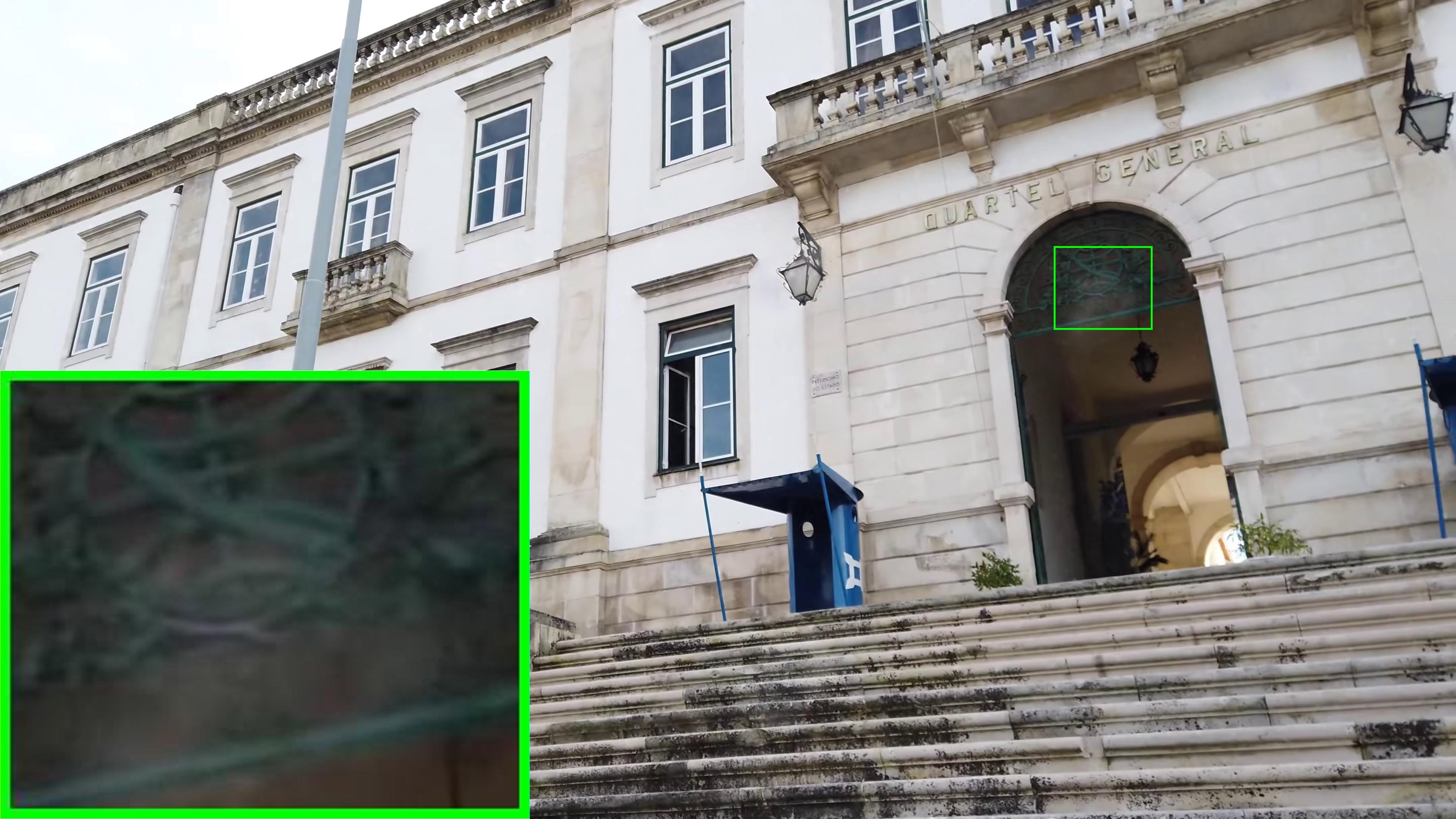} 
&\hspace{-4mm}
\includegraphics[width=0.14\linewidth]{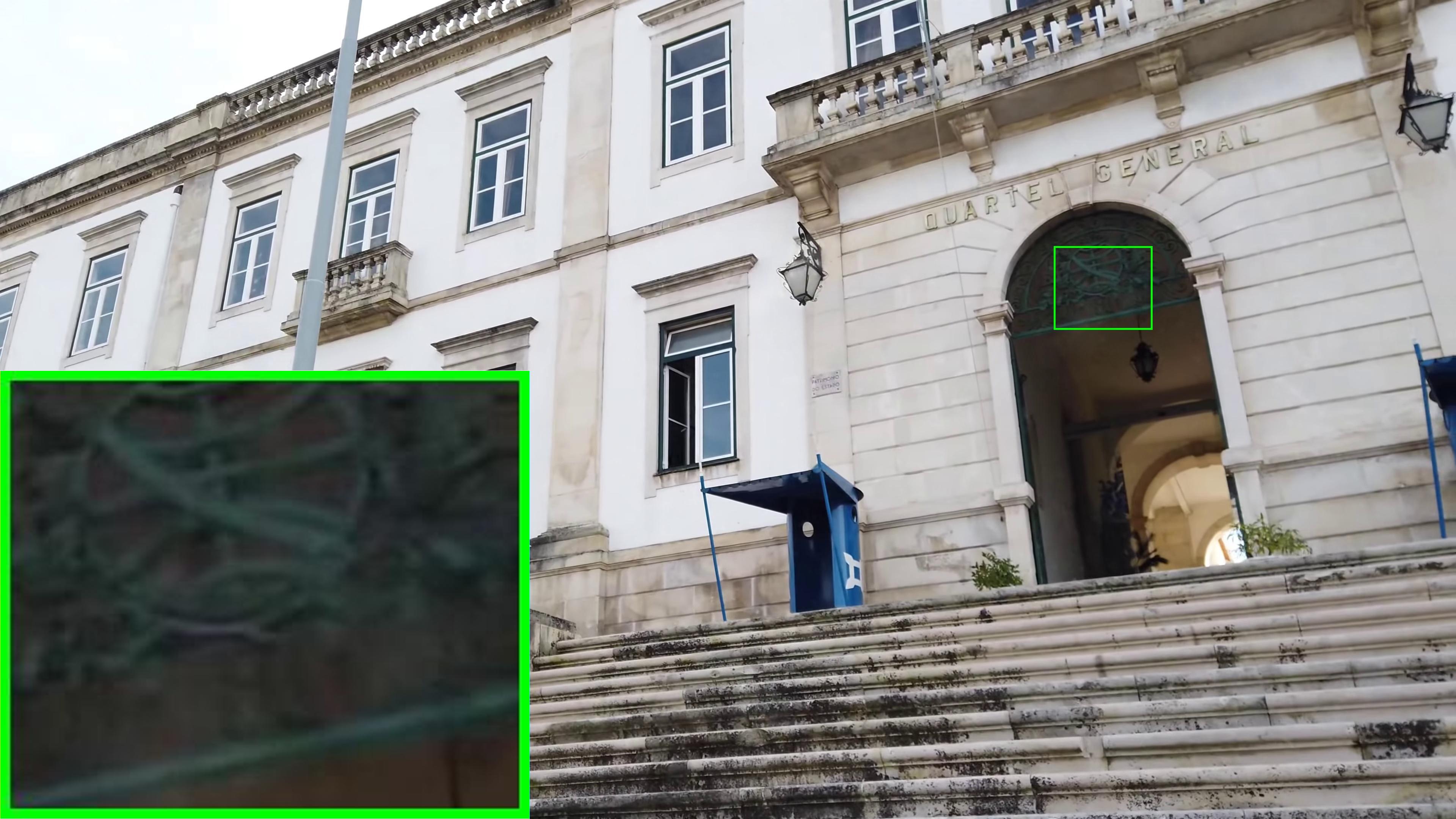} 
\\
\hspace{-1mm}\includegraphics[width=0.14\linewidth]{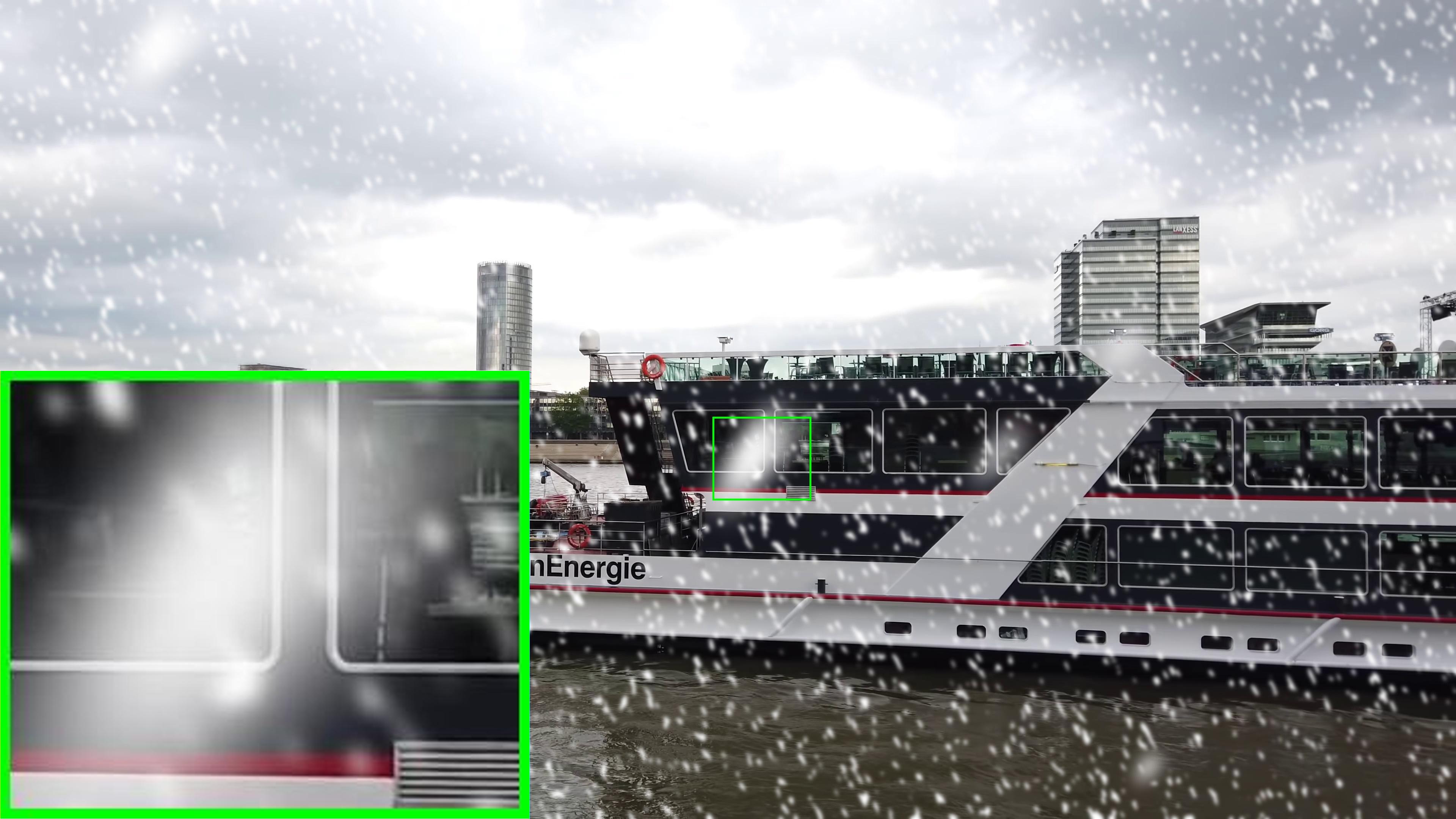} &\hspace{-4mm}
\includegraphics[width=0.14\linewidth]{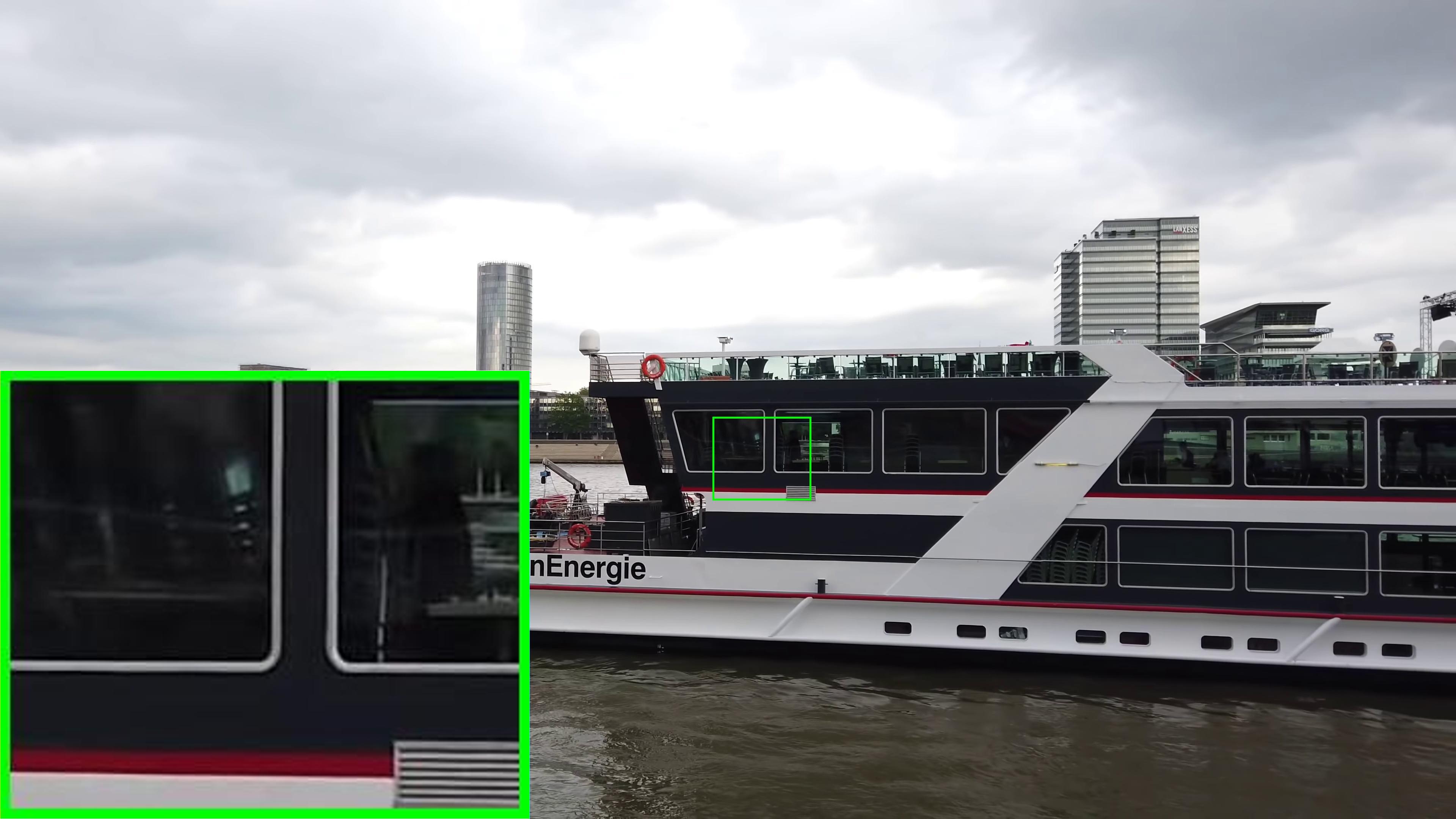} &\hspace{-4mm}
\includegraphics[width=0.14\linewidth]{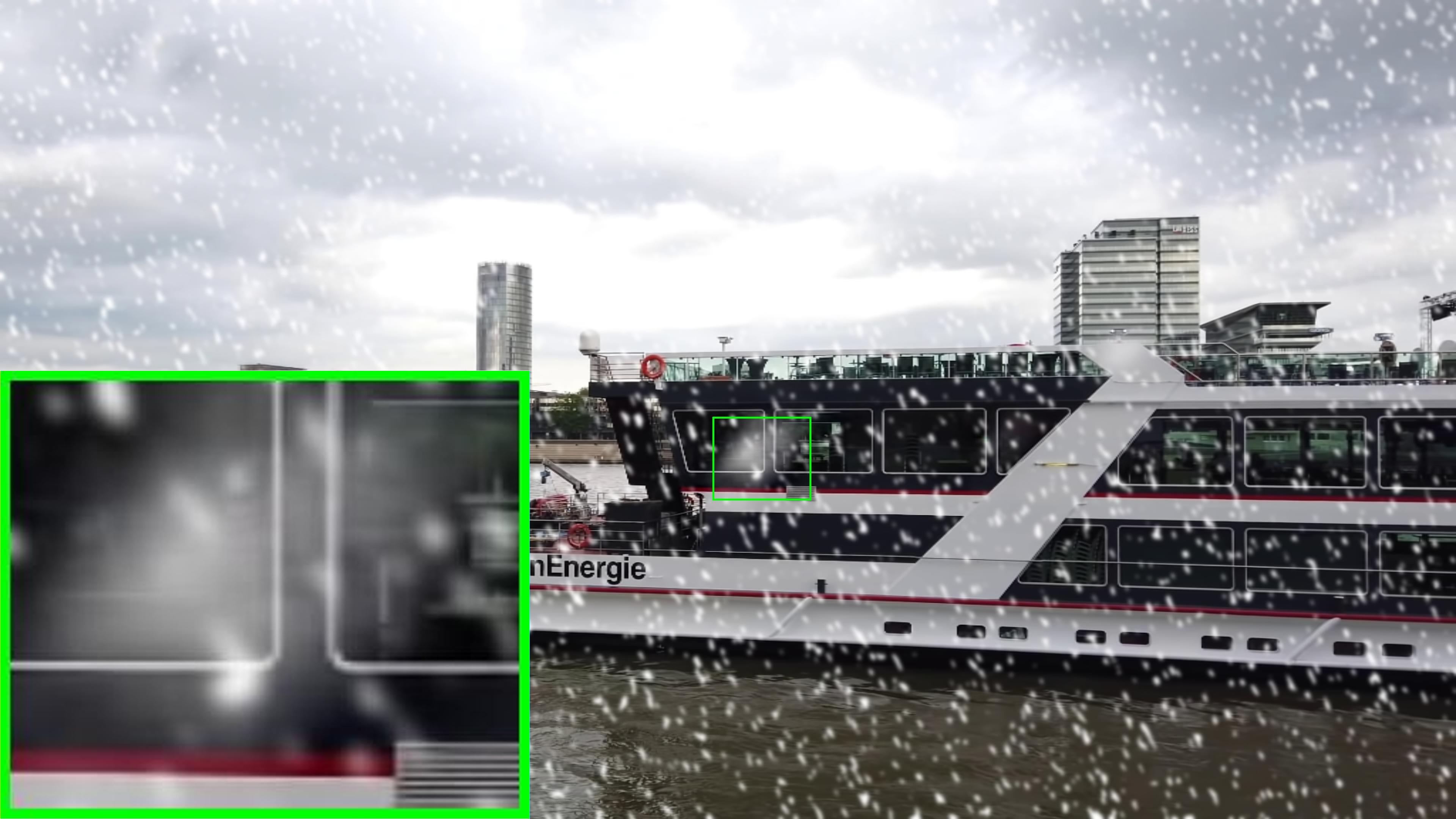} &\hspace{-4mm}
\includegraphics[width=0.14\linewidth]{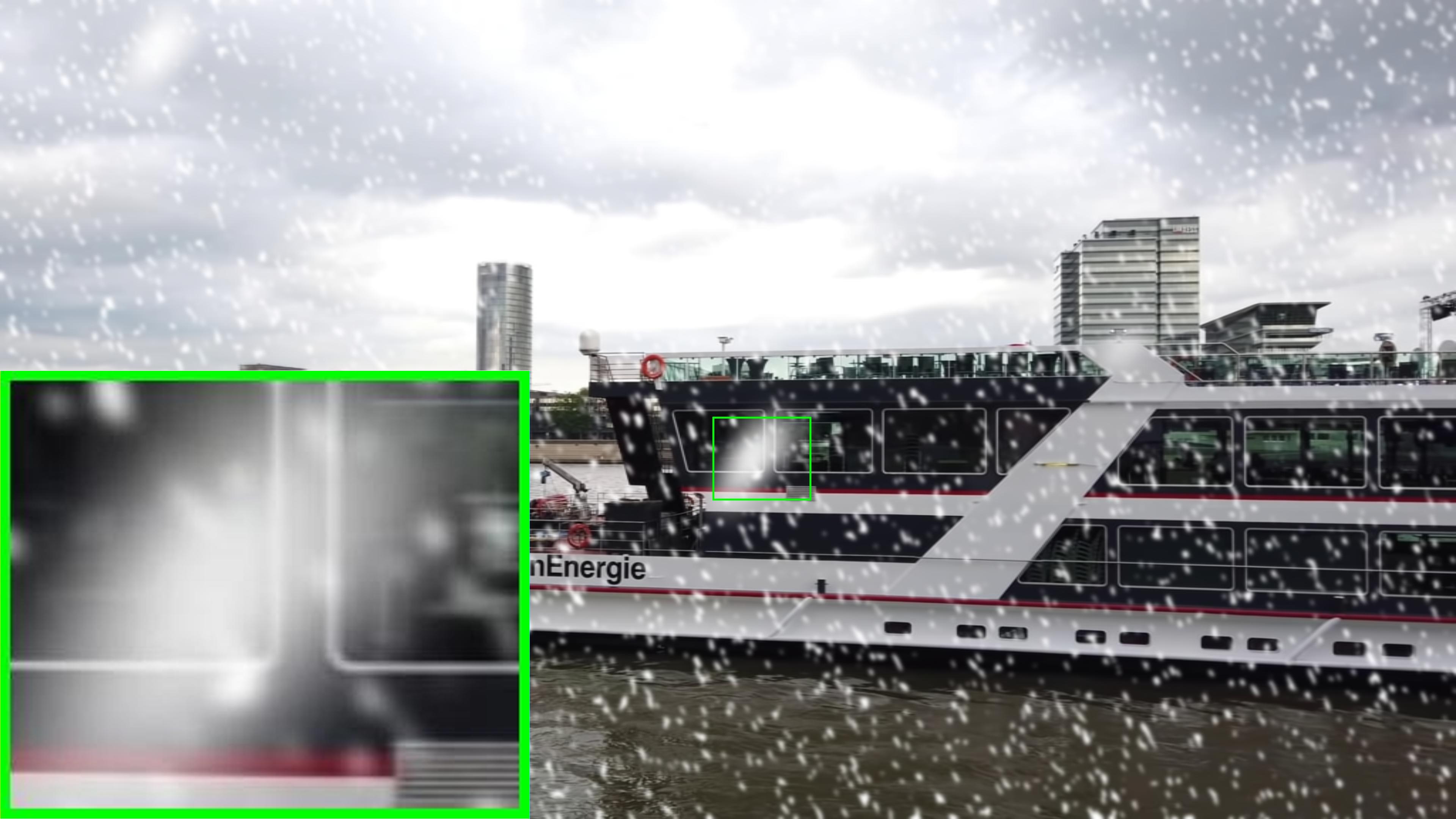} 
&\hspace{-4mm}
\includegraphics[width=0.14\linewidth]{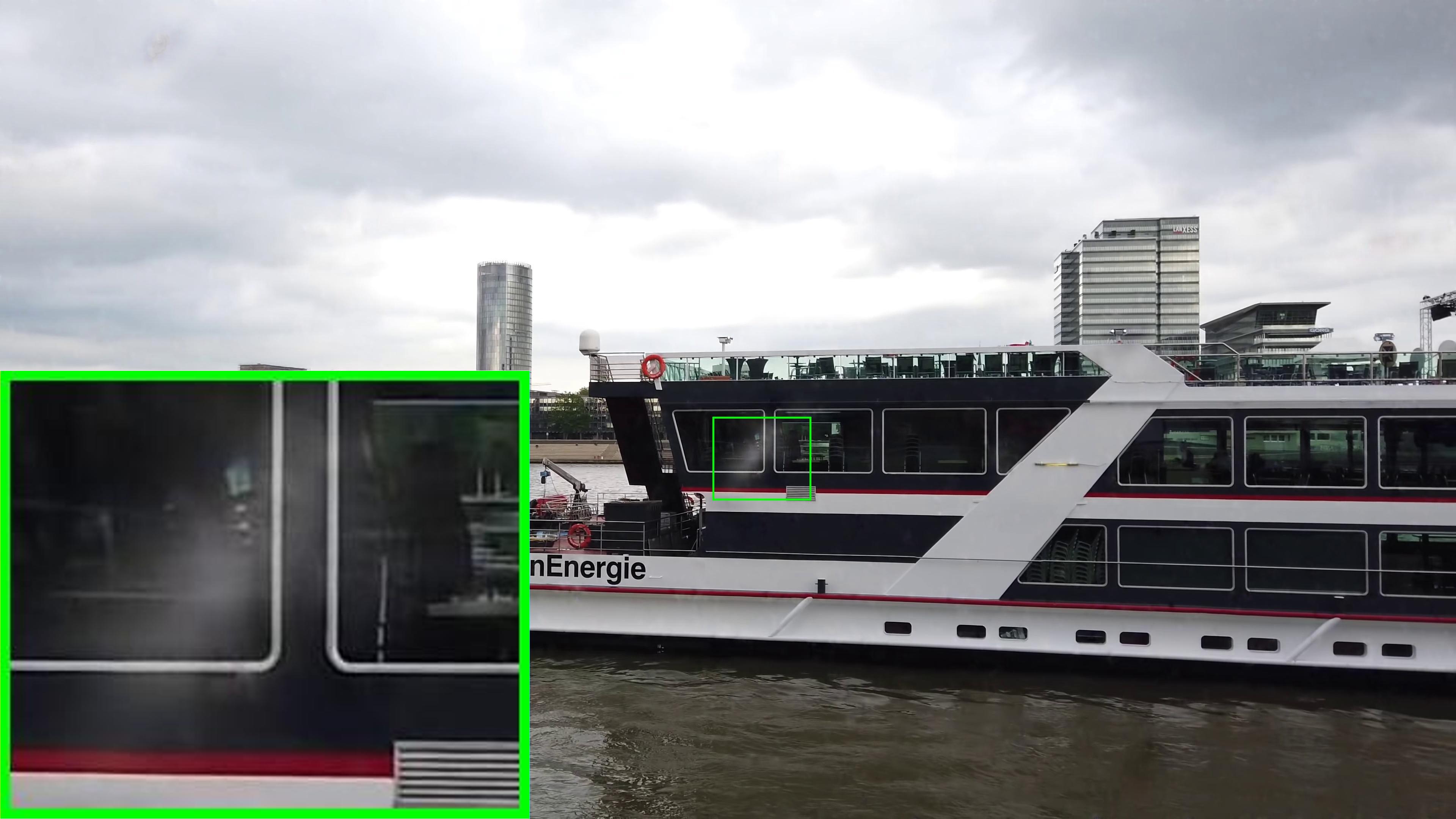} 
&\hspace{-4mm}
\includegraphics[width=0.14\linewidth]{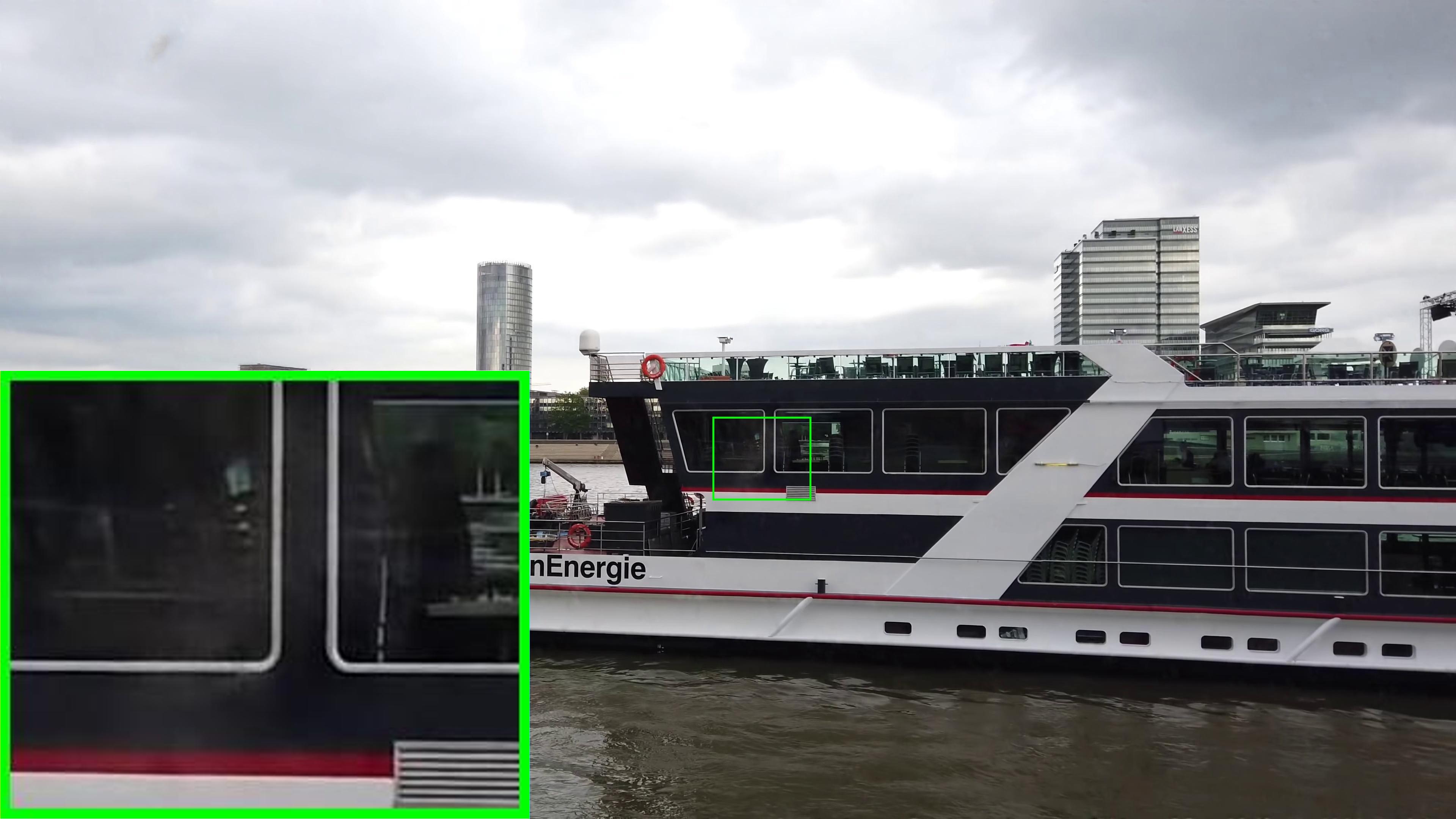} 
\\
 (a) Input&\hspace{-4mm} (b) GT&\hspace{-4mm}(c) Restormer~\cite{Zamir2021Restormer} &\hspace{-4mm} (d) SFNet~\cite{0001TBRGC0K23sfnet} &\hspace{-4mm}(e) UHDformer~\cite{aaai24wang_UHDformer} &\hspace{-4mm} (f) \textbf{UHDformer++}
\end{tabular} 
\vspace{-2mm}
\caption{\textbf{Image desnowing on UHD-Snow \textit{trained on the UHD-Snow dataset}.}
}
\label{fig:Image desnowing on UHD-Snow.}
\end{center}
\vspace{-3mm}
\end{figure*}
%%%%%%%%%%%%%%%%%%%%%%%%%%%%%%%%%%%%%%%%%%%%%%%%%%%%%%%%%%%%%%%%%%%%%%%%%%%%
%%%%%%%%%%%%%%%%%%%%%%%%%%%%%%%%%%%%%%%%%%%%%%%%%%%%%%%%%%%%%%%%%%%%%%%%%%%%

%
\subsubsection{Image Deblurring Results} 
We evaluate UHD image deblurring on the UHD-Blur benchmark under two training configurations: models trained on GoPro~\cite{gopro2017} and models trained on UHD-Blur, with all methods tested on the UHD-Blur test set. 
Tab.~\ref{tab:Image deblurring.} summarizes the quantitative results, where UHDformer++ advances the state-of-the-art approach on UHD-Blur training sets but slightly lower than UHDformer on the GoPro training set in terms of PSNR and SSIM.
Compared to the recent state-of-the-art deblurring method FFTformer~\cite{Kong_2023_CVPR_fftformer}, UHDformer++ achieves PSNR gains of $2.693$dB and $3.786$dB on the GoPro and UHD-Blur training sets, respectively. 
It is also worth noting that although DMPHN~\cite{dmphn2019}, MIMO-Unet++~\cite{cho2021rethinking_mimo}, and MPRNet~\cite{Zamir_2021_CVPR_mprnet} are capable of processing full-resolution UHD images without resizing, they require at least $97.5$\% more parameters than UHDformer++ when trained on GoPro, while still yielding a PSNR degradation of at least $0.828$dB.
Visual comparisons are provided in Fig.~\ref{fig:Image deblurring on UHD-Blur.}, where  UHDformer++ consistently produces sharper outputs with finer structural detail, while existing state-of-the-art methods struggle to reliably handle the challenges posed by full-resolution UHD images.
%%%%%%%%%%%%%%%%%%%%%%%%%%%%%%%%%%%%%%%%%%%%%%%%%%%%%%%%%%%%%%%%%%%%%%%%%%%%
%
%%%%%%%%%%%%%%%%%%%%%%%%%%%%%%%%%%

\subsubsection{Image Deraining Results} 
We evaluate UHD image deraining on the UHD-Rain benchmark~\cite{wang2025ultra}, with quantitative results reported in Tab.~\ref{tab:Image deraining.}. 
UHDformer++ achieves the best performance across all metrics in both full-reference and no-reference, demonstrating its effectiveness for this task.
Compared to SFNet~\cite{cui2023selective}, which is not capable of processing full-resolution UHD images, UHDformer++ achieves a substantial improvement of over $18$dB in PSNR. 
Furthermore, relative to UHDformer~\cite{aaai24wang_UHDformer}, UHDformer++ attains superior performance across all metrics while requiring fewer FLOPs and lower inference time.
Visual comparisons are presented in Fig.~\ref{fig:Image deraining on UHD-Rain.}, where UHDformer++ visibly restores cleaner and sharper outputs with significantly reduced rain artifacts compared to competing methods.

\subsubsection{Image Desnowing Results} 

We use UHD-Snow benchmark~\cite{wang2025ultra} to evaluate the UHD image desnowing task, and quantitative results are summarized in Tab.~\ref{tab:Image desnowing.}. 
UHDformer++ achieves excellent performance across the evaluated metrics, with particularly notable gains in PSNR. 
Specifically, UHDformer++ surpasses UHDformer~\cite{aaai24wang_UHDformer} by $2.348$dB in PSNR, and outperforms SFNet~\cite{cui2023selective} by over $15$dB, further corroborating the effectiveness of our approach.
Visual comparisons are presented in Fig.~\ref{fig:Image desnowing on UHD-Snow.}, where UHDformer++ visibly removes snow artifacts more thoroughly, producing outputs with substantially fewer snowflake residuals compared to recent methods.
%%%%%%%%%%%%%%%%%%%%%%%%%%%%%%%%%%%%%%%%%%%%%%%%%%%%%%%%%%%%%%%%%%%%%%%%%%%%

%
\subsection{Analysis and Discussion}
In this section, we conduct extensive analysis and discussion to demonstrate the effectiveness of our proposed UHDformer++.
We use the UHD-LL dataset~\cite{Li2023ICLR_uhdfour} to conduct the ablation experiments.
\begin{table*}[!t]
\caption{\textbf{Effect on the Normalization Layer (DyT), Feed-Forward Neural Network (CMTFN), and FR-CMT in the UHDformer++}.
Each component used in this paper is more effective than our previous version used in UHDformer~\cite{aaai24wang_UHDformer}.
}
 \vspace{-2mm}
\label{tab:Effect on Ablation study on LN}
% \tablestyle{3.2pt}{1.1}
\tablestyle{10pt}{1}
\centering
\begin{tabular}{l|l|ccc|cc}
\shline
\multicolumn{1}{c|}{\multirow{2}{*}{\textbf{ID}}}& \multicolumn{1}{c|}{\multirow{2}{*}{\textbf{Experiment}}}&\multicolumn{3}{c|}{\textbf{Metrics}}&\multicolumn{2}{c}{\textbf{Computational Complexity}}
\\
\cline{3-7}
&& \textbf{PSNR (dB)}~$\uparrow$ & \textbf{SSIM}~$\uparrow$ &\textbf{LPIPS}~$\downarrow$&\textbf{Params. (M)}~$\downarrow$ &\textbf{FLOPs (G)}~$\downarrow$ 
\\
\shline
\textbf{(a)}& w/o LN &26.812&	0.9279&0.2369&0.3955	&49.65\\
\textbf{(b)}& LN in UHDformer~\cite{aaai24wang_UHDformer} &27.119&	0.9277&	0.2493&	0.3962&	49.65\\
\textbf{(c)}& \textbf{DyT in UHDformer++} &\textbf{27.256}&	\textbf{0.9308}&	\textbf{0.2304}&	0.3962	&49.65 \\
\shline
\textbf{(d)}& GDFN in Restormer~\cite{Zamir2021Restormer}&26.499&	0.9270&	0.2366&	0.3064	& 38.06 \\
\textbf{(e)}& CMTN in UHDformer~\cite{aaai24wang_UHDformer} &27.090&	0.9290&	0.2394&	0.3379&	48.70\\
\textbf{(f)}& \textbf{CMTFN in UHDformer++}&\textbf{27.256}&	\textbf{0.9308}&	\textbf{0.2304}&	0.3962	&49.65 \\
\shline
\textbf{(g)}& DualCMT in UHDformer~\cite{aaai24wang_UHDformer} &\textbf{27.519}&	0.9286&	0.2422&	0.4602&	49.24\\
\textbf{(h)}& \textbf{FR-CMT in UHDformer++ }&27.256&	\textbf{0.9308}&	\textbf{0.2304}&	0.3962	&49.65 \\
\shline
\end{tabular}
\end{table*}
%%%%%%%%%%%%%%%%%%%%%%%%%%%%%%%%%%%%%%%%%%%%%%%%%%%%%%%%%

\begin{figure*}[!t]
\centering
\begin{center}
\begin{tabular}{ccccc}
\hspace{-1mm}\includegraphics[width=0.18\linewidth]{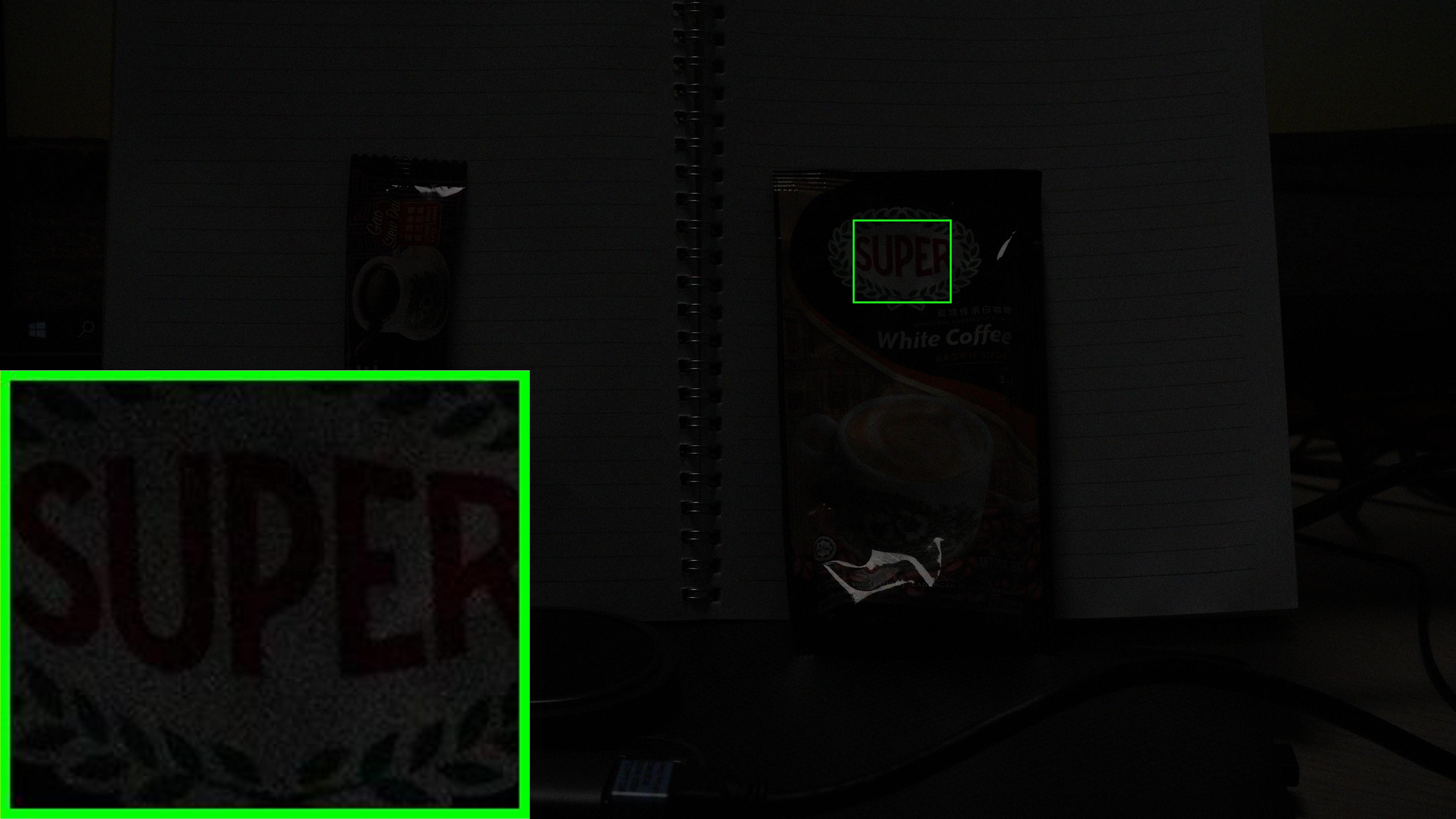} &\hspace{-4mm}
\includegraphics[width=0.18\linewidth]{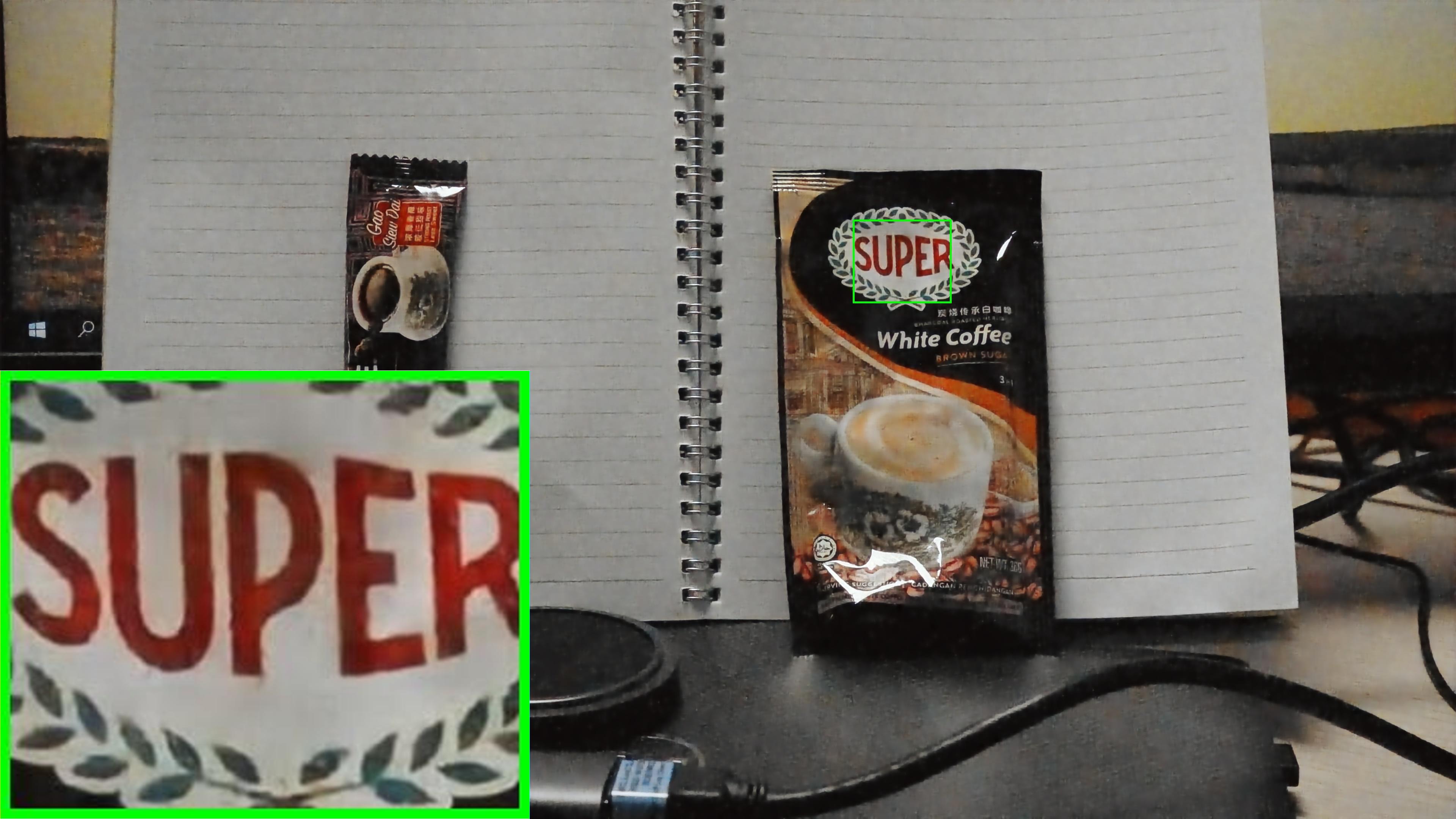} &\hspace{-4mm}
\includegraphics[width=0.18\linewidth]{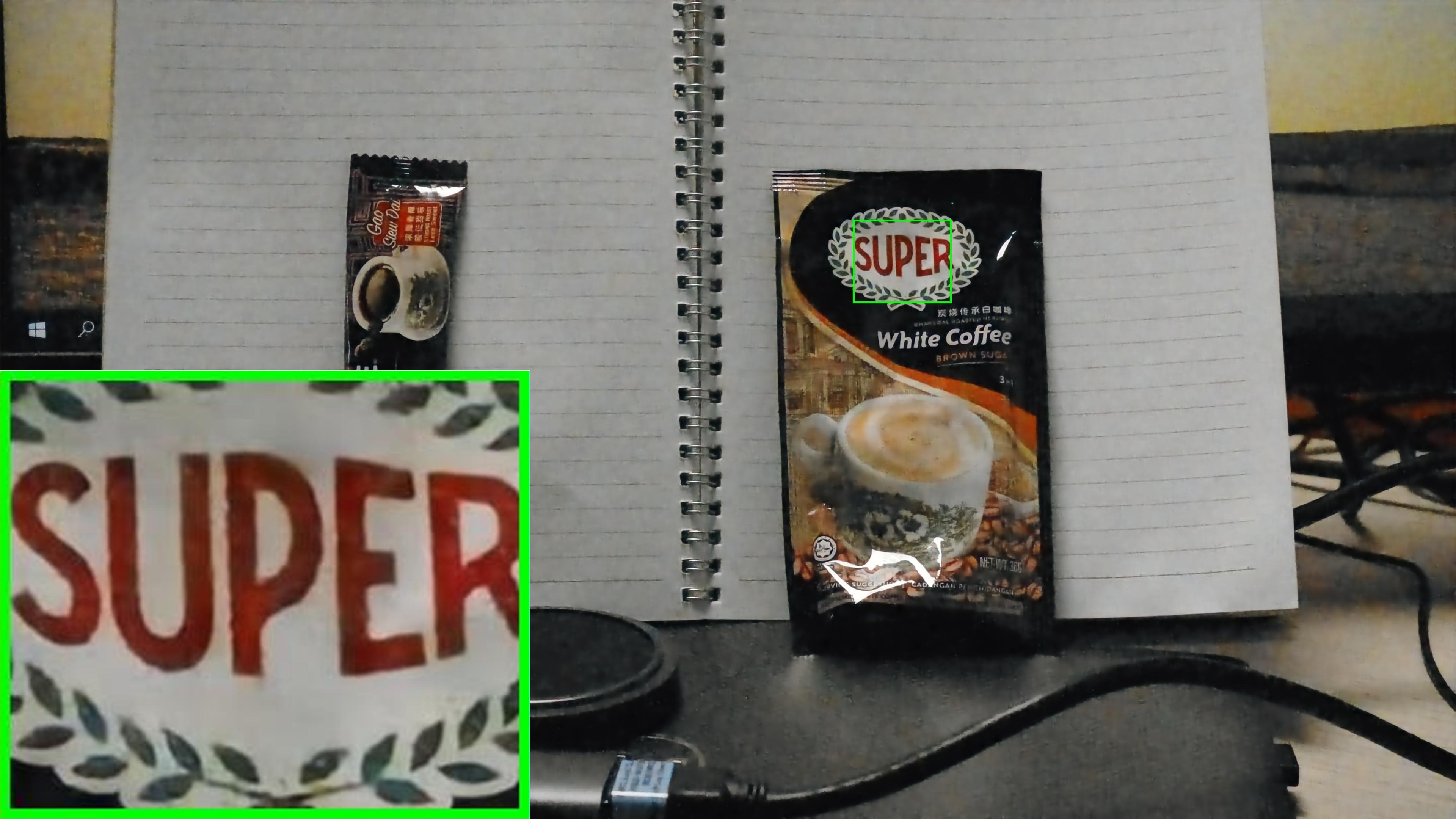} &\hspace{-4mm}
\includegraphics[width=0.18\linewidth]{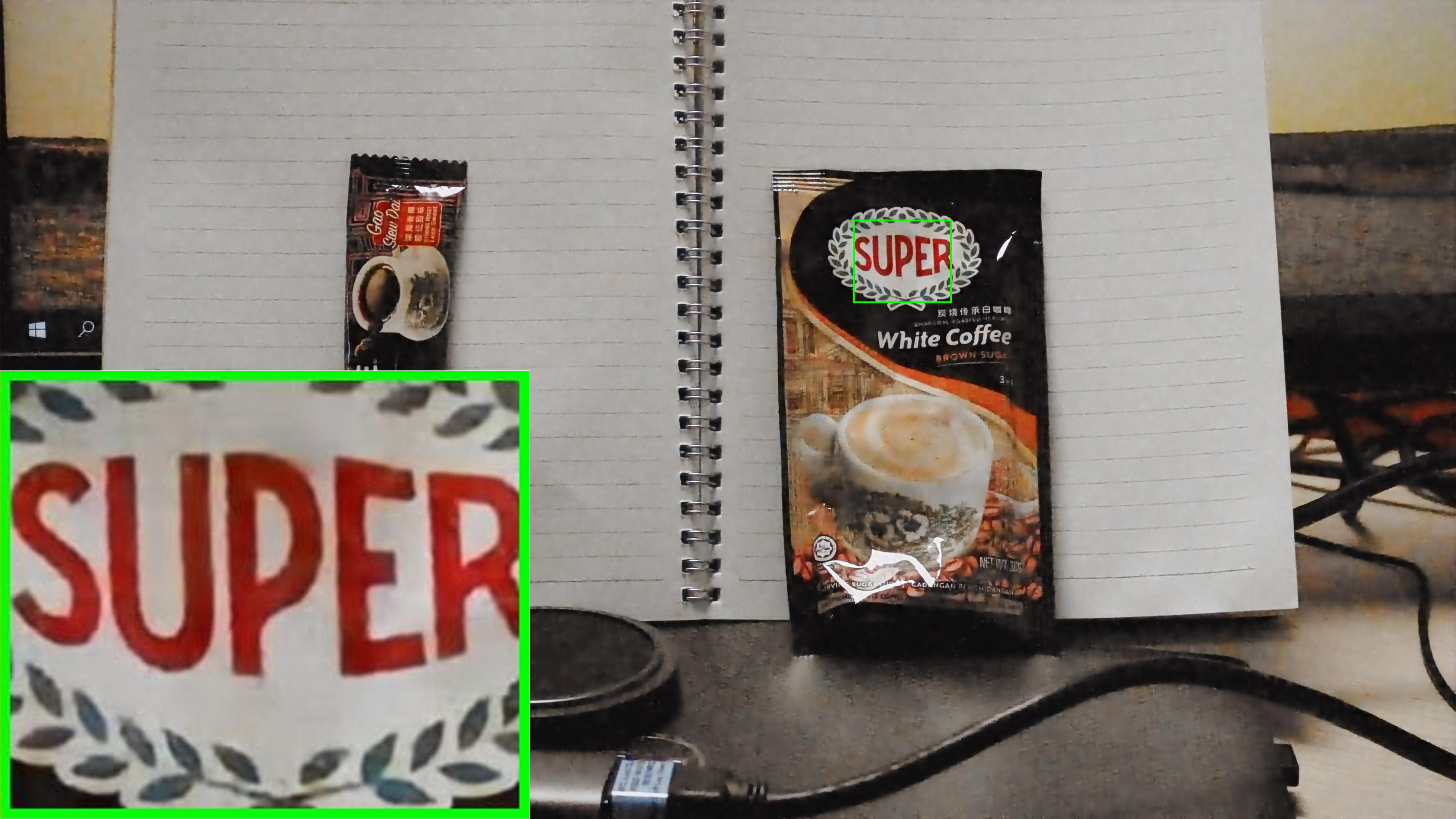} 
&\hspace{-4mm}
\includegraphics[width=0.18\linewidth]{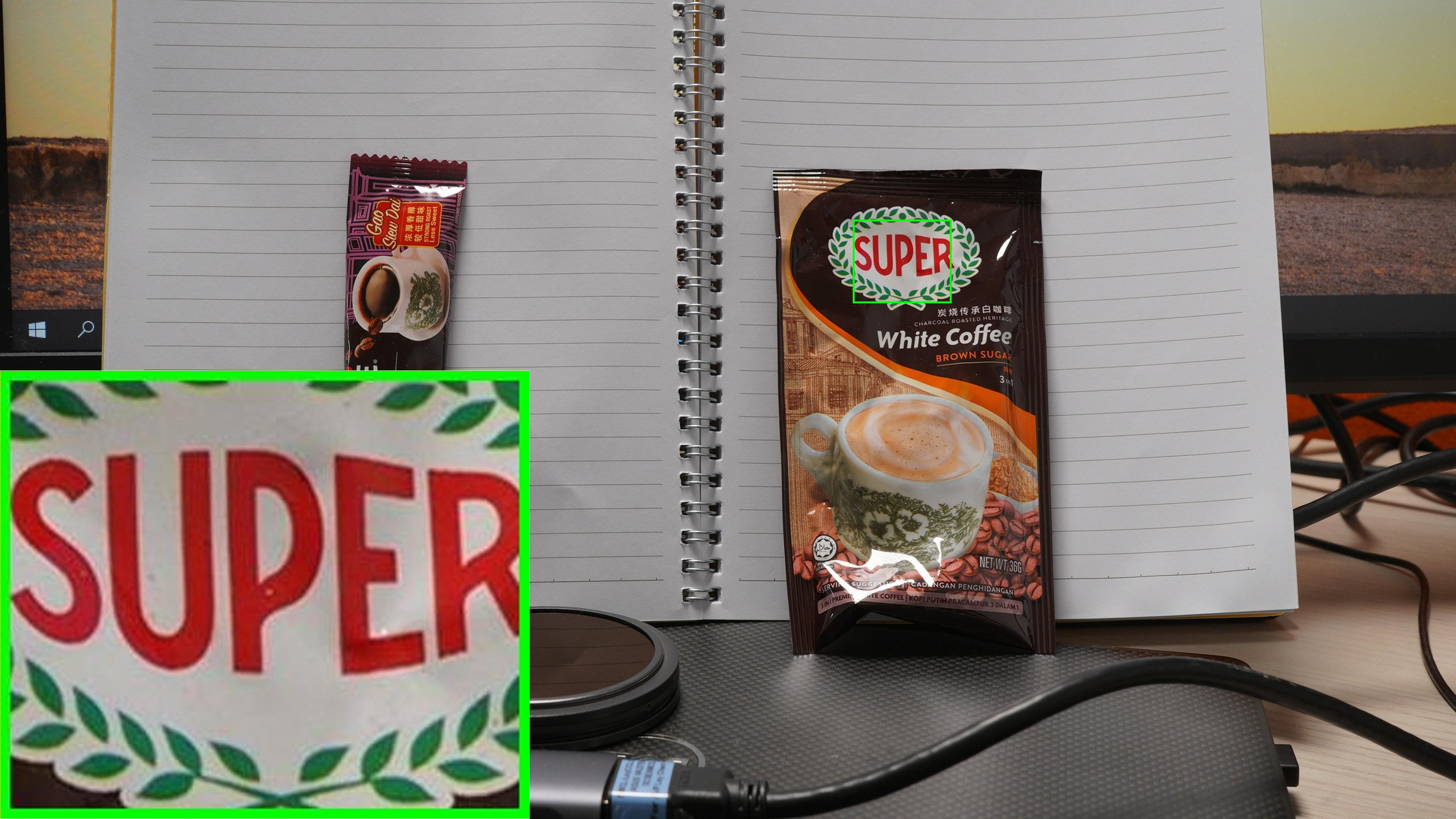} 
\\
 (a1) Input&\hspace{-4mm} (b1) w/o LN&\hspace{-4mm}(c1) LN in \cite{aaai24wang_UHDformer} &\hspace{-4mm} (d1) Ours &\hspace{-4mm}(e1) GT
 \\
 \hspace{-1mm}\includegraphics[width=0.18\linewidth]{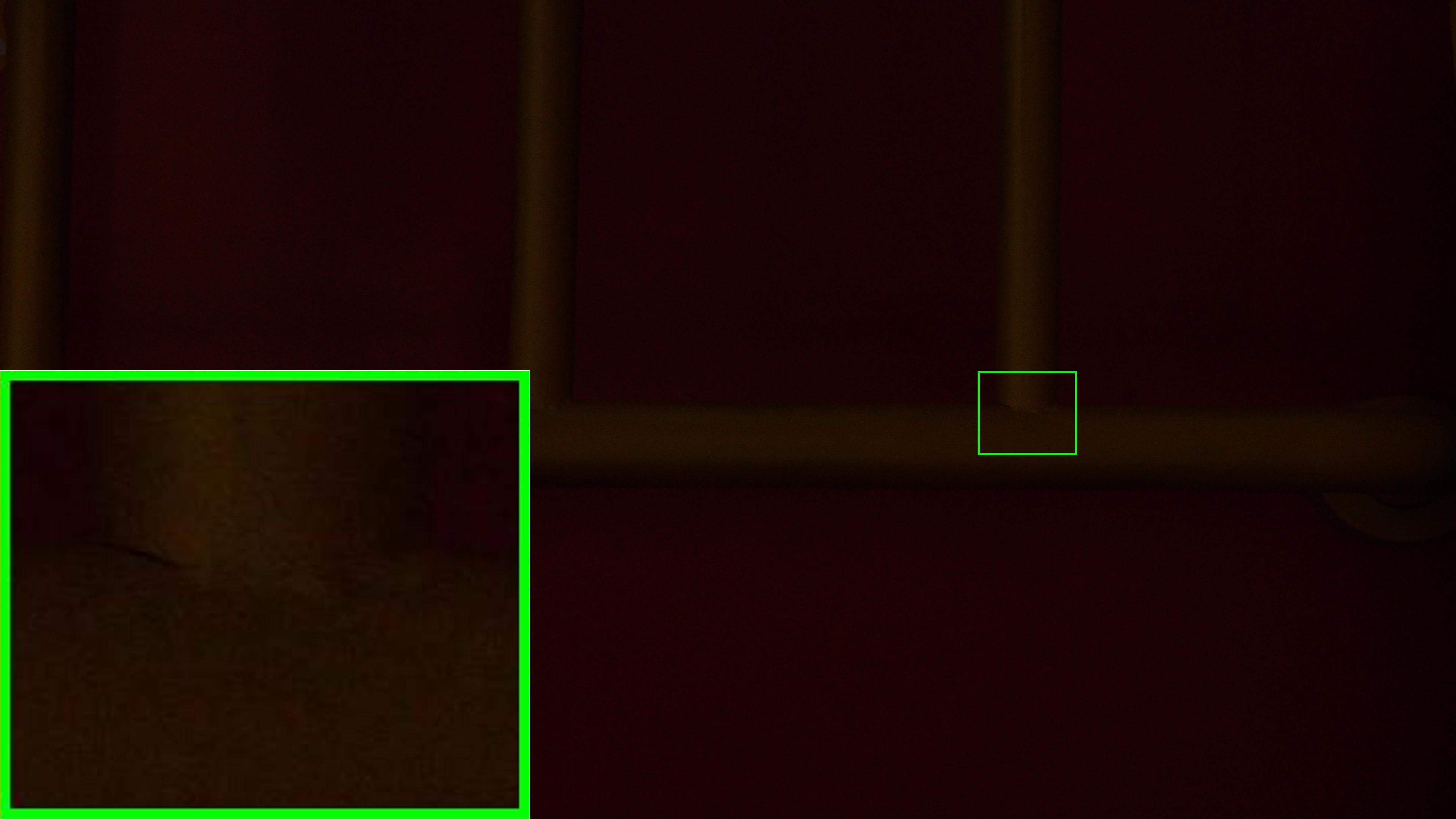} &\hspace{-4mm}
\includegraphics[width=0.18\linewidth]{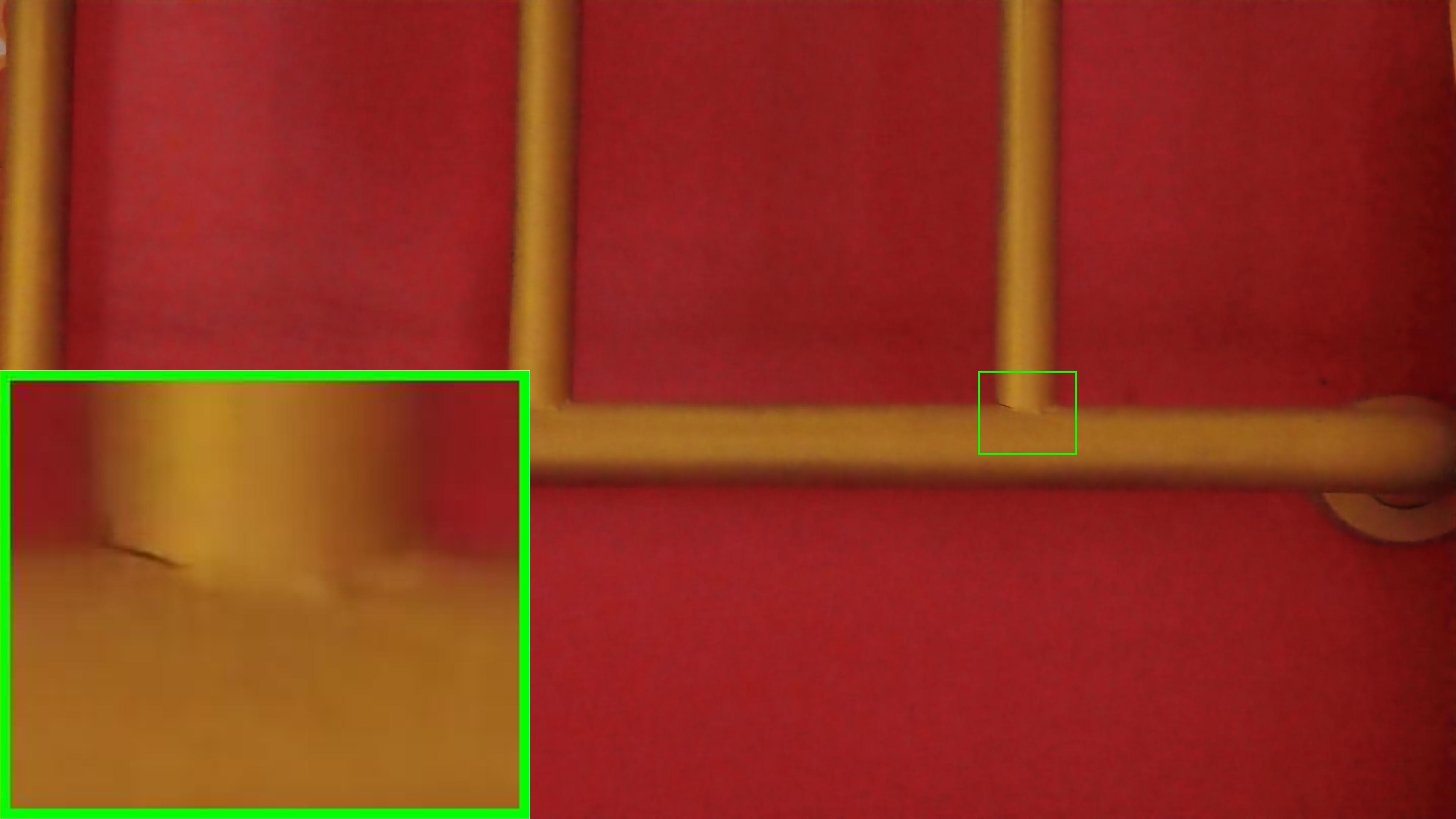} &\hspace{-4mm}
\includegraphics[width=0.18\linewidth]{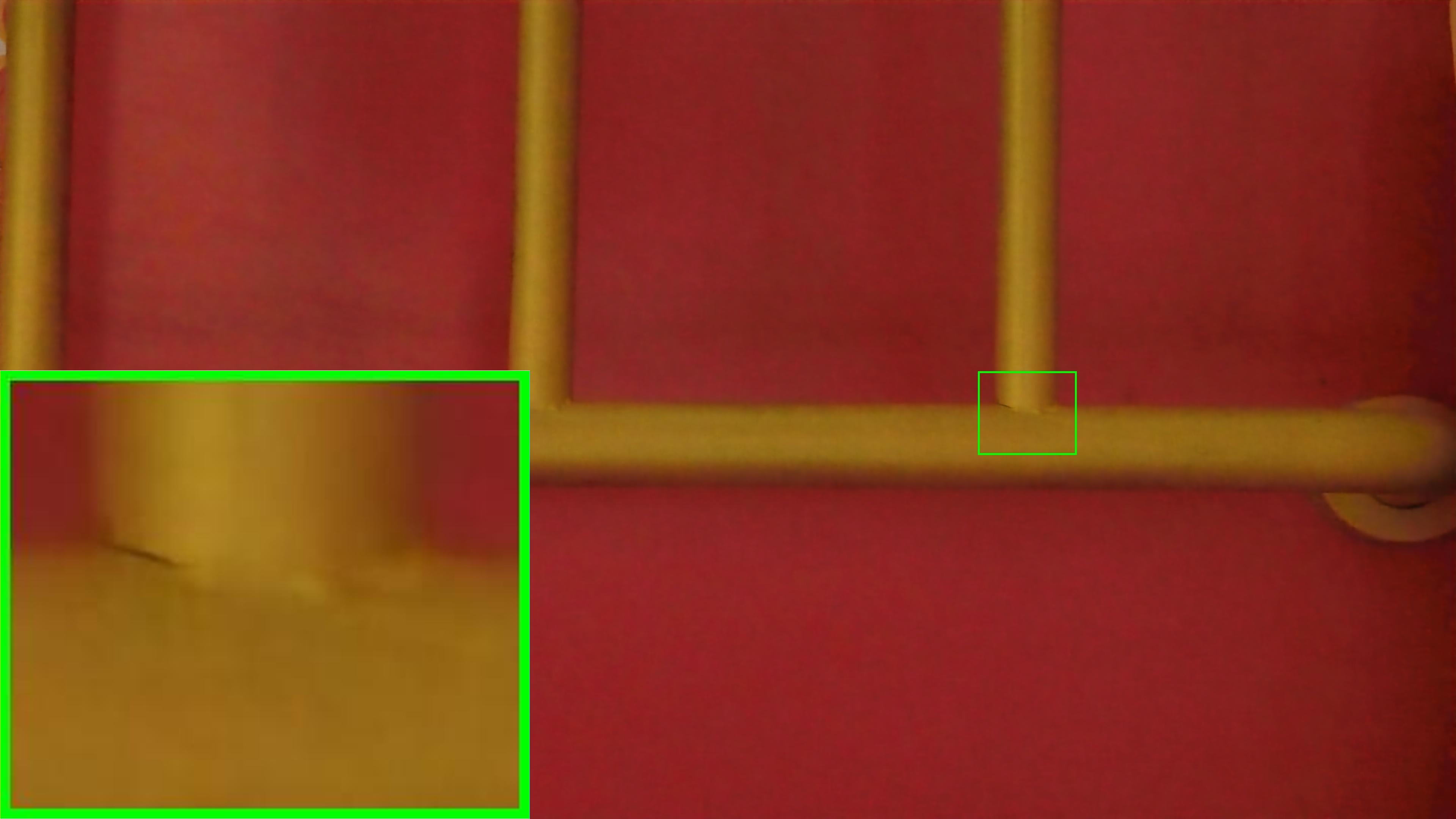} &\hspace{-4mm}
\includegraphics[width=0.18\linewidth]{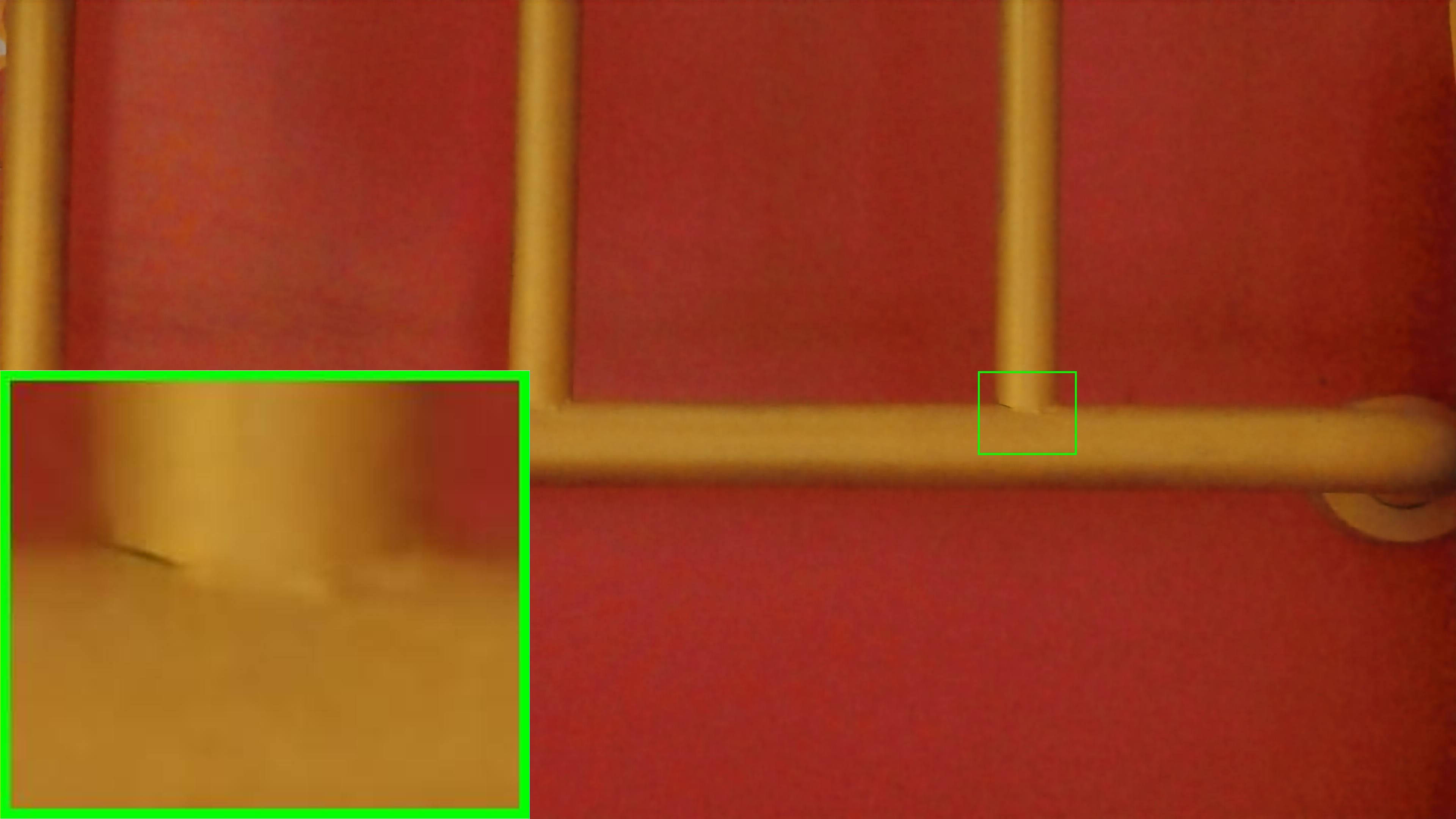} 
&\hspace{-4mm}
\includegraphics[width=0.18\linewidth]{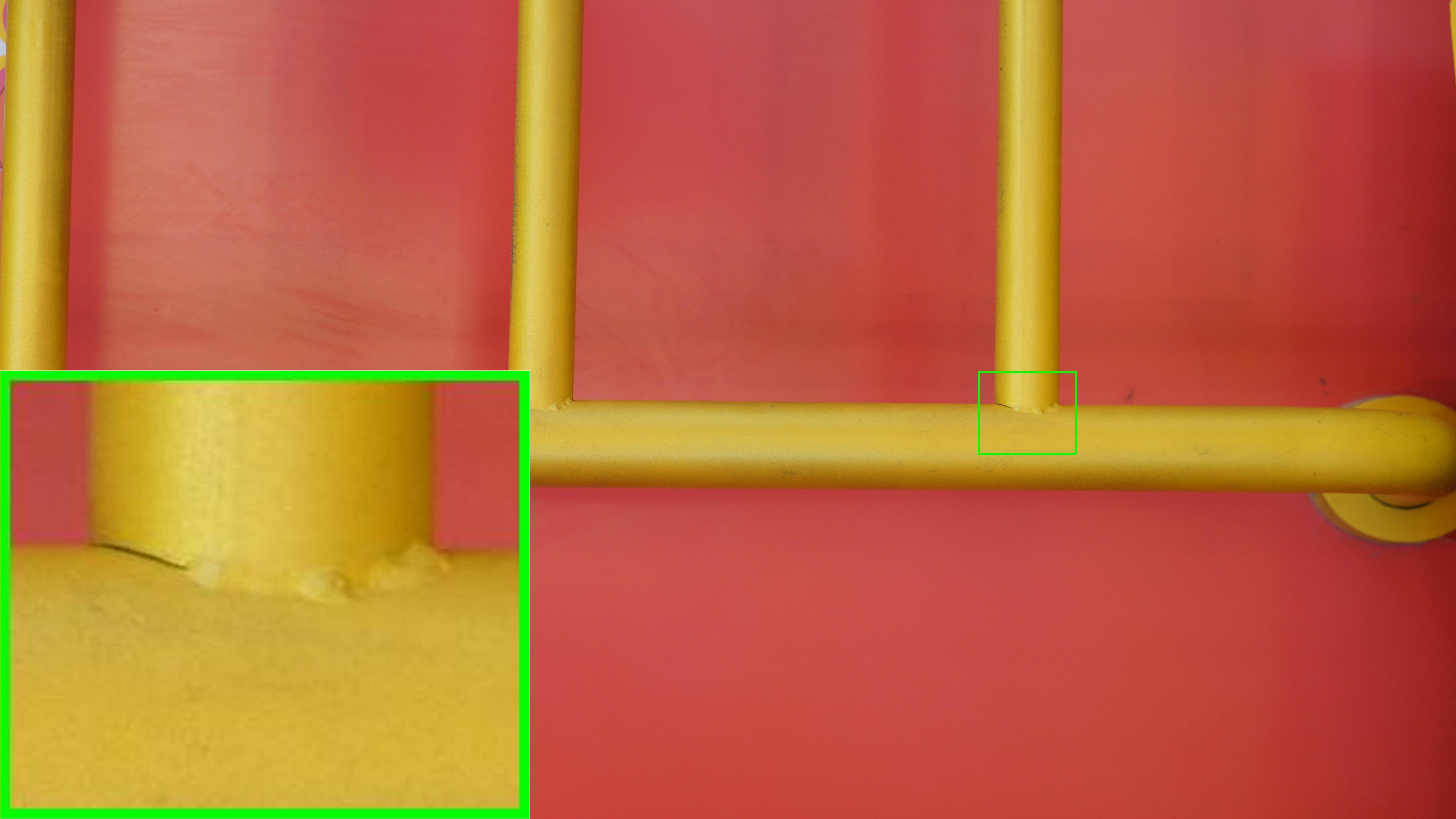} 
\\
 (a2) Input&\hspace{-4mm} (b2) GDFN in \cite{Zamir2021Restormer}&\hspace{-4mm}(c2) CMTN in \cite{aaai24wang_UHDformer} &\hspace{-4mm} (d2) Ours&\hspace{-4mm}(e2) GT
  \\
 \hspace{-1mm}\includegraphics[width=0.18\linewidth]{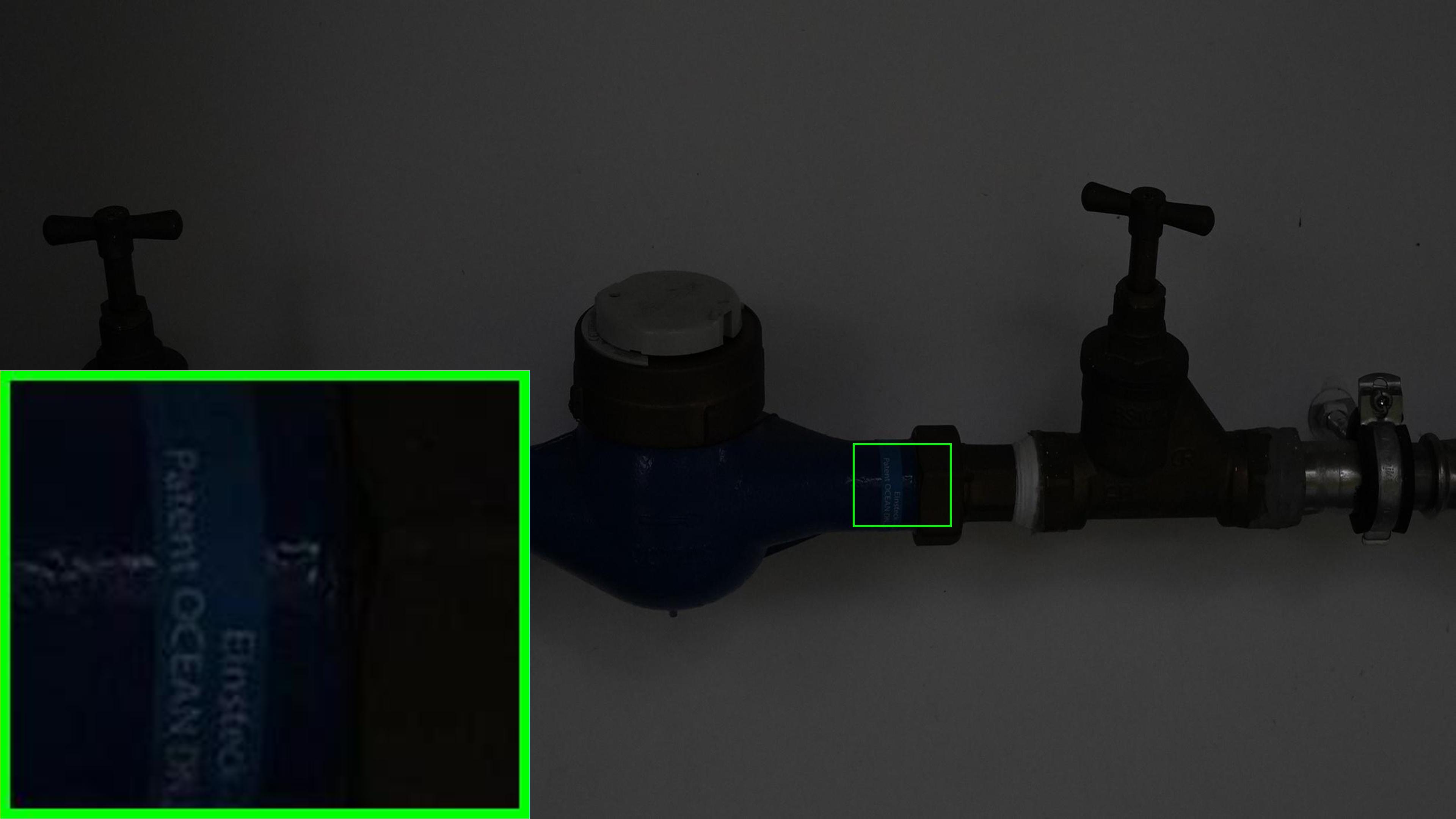} &\hspace{-4mm}
  \hspace{-1mm}\includegraphics[width=0.18\linewidth]{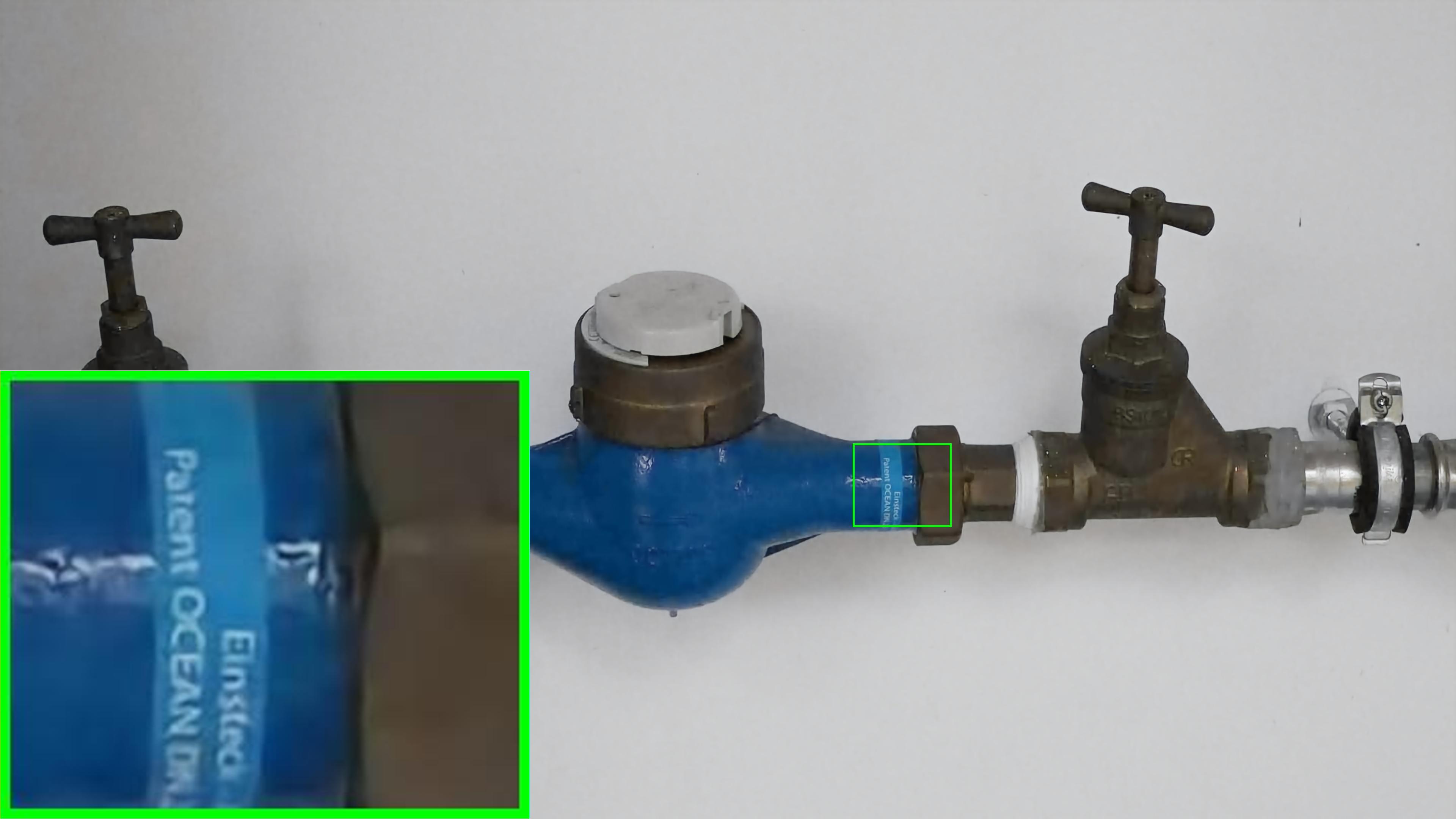} &\hspace{-4mm}
\includegraphics[width=0.18\linewidth]{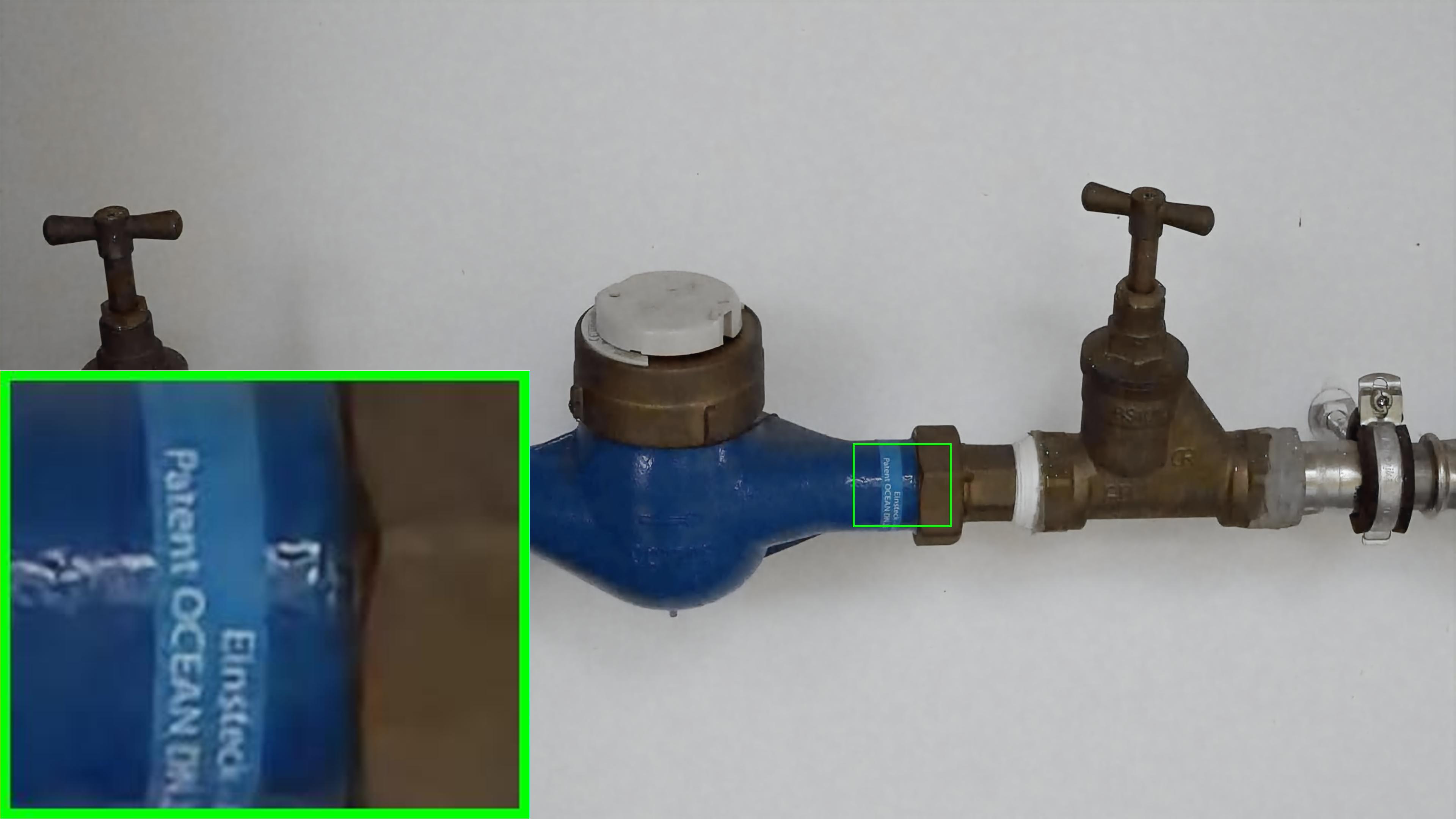} &\hspace{-4mm}
\includegraphics[width=0.18\linewidth]{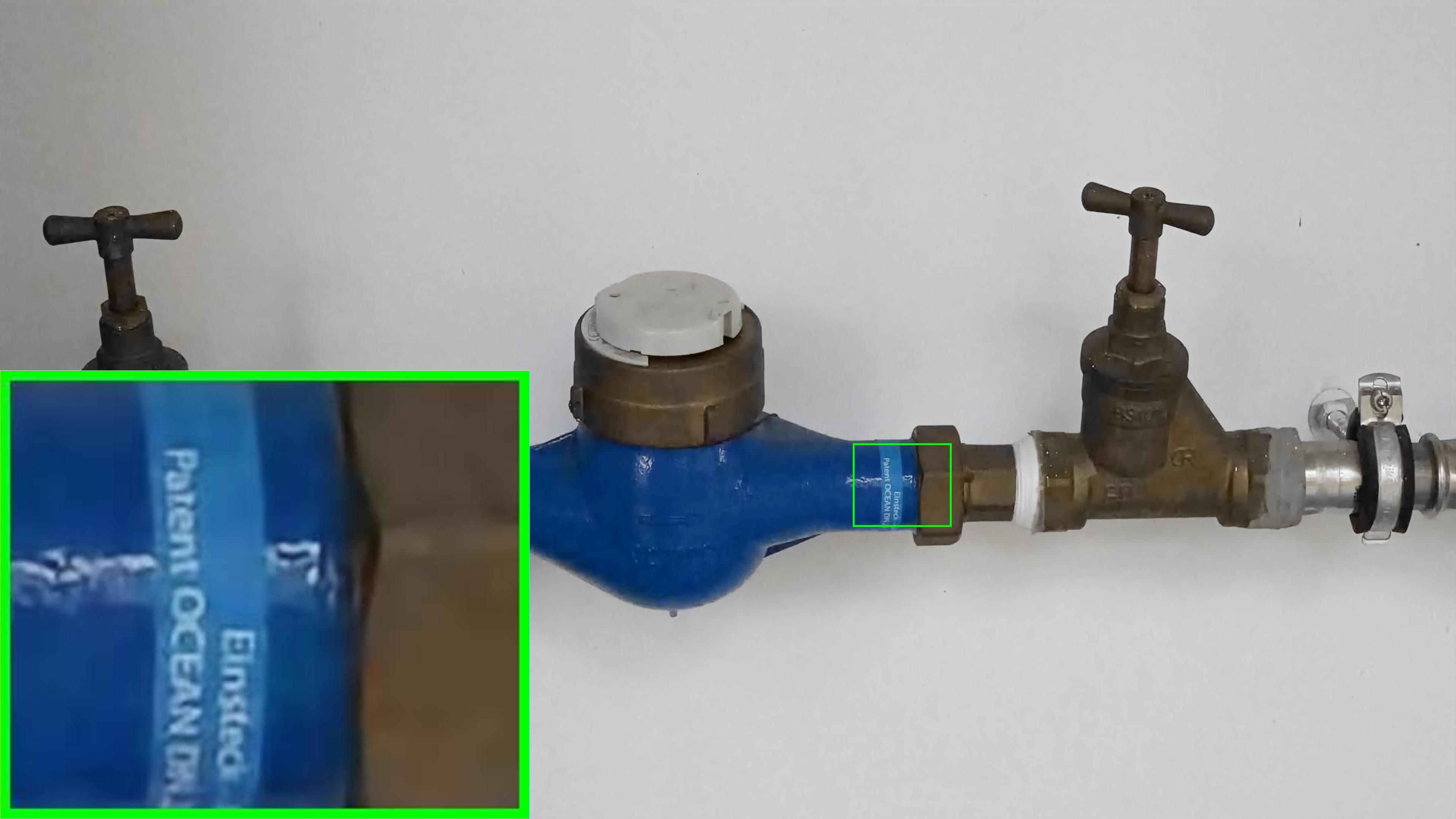} &\hspace{-4mm}
\includegraphics[width=0.18\linewidth]{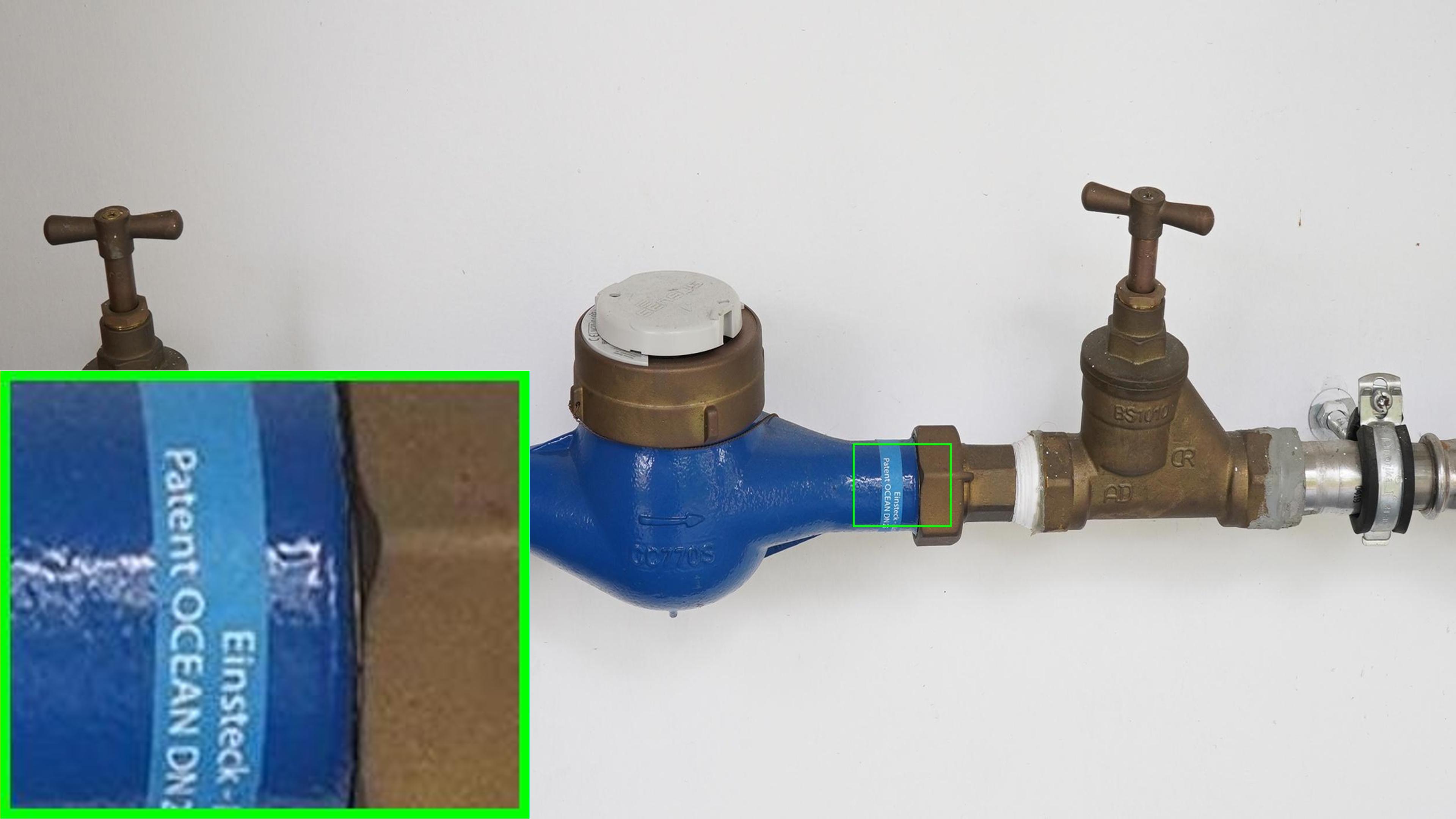} 
\\
 (a3) Input&\hspace{-4mm} (b3) w/o FR-CMT &\hspace{-4mm}(c3) DualCMT in \cite{aaai24wang_UHDformer} &\hspace{-4mm}(d3) Ours  &\hspace{-4mm}(e3) GT
\end{tabular} 
\vspace{-2mm}
\caption{\textbf{Effect on different basic components.} 
}
\label{fig:Compared to the normalization layer (LN).}
\end{center}
\vspace{-3mm}
\end{figure*}

%%%%%%%%%%%%%%%%%%%%%%%%%%%%%%%%%
\begin{table*}[!t]
\caption{\textbf{Ablation study on Feature-Refined Correlation Matching Transformation (FR-CMT)}.
}
\label{tab:Effect on Ablation study on MCM}
% \tablestyle{3.2pt}{1.1}
\vspace{-2mm}
\tablestyle{8pt}{1}
\centering
\begin{tabular}{l|l|ccc|cc}
\shline
\multicolumn{1}{c|}{\multirow{2}{*}{\textbf{ID}}}& \multicolumn{1}{c|}{\multirow{2}{*}{\textbf{Experiment}}}&\multicolumn{3}{c|}{\textbf{Metrics}}&\multicolumn{2}{c}{\textbf{Computational Complexity}}
\\
\cline{3-7}
&& \textbf{PSNR (dB)}~$\uparrow$ & \textbf{SSIM}~$\uparrow$ &\textbf{LPIPS}~$\downarrow$&\textbf{Params. (M)}~$\downarrow$ &\textbf{FLOPs (G)}~$\downarrow$ 
\\
\shline
\textbf{(a)}& w/o FR-CMT in CMTA \& CMTN& 26.933&	0.9279&	0.2433&	0.3060	&27.94 \\
\textbf{(b)}& w/o FR-CMT in CMTA&26.879&0.9278&	0.2406&	0.3511	&38.8\\
\textbf{(c)} &w/o FR-CMT in CMTFN&26.650&	0.9256&	0.2347& 0.3511	& 38.8 \\
\textbf{(d)}& w/o Using Max-Pooling in FR-CMT&27.079&	0.9270&	0.2406&	0.3962 	&49.45 \\
\textbf{(e)}& w/o Using Mean-Pooling in FR-CMT&26.911&	0.9281&	0.2469&	0.3962&	49.45\\
\shline
\textbf{(f)} &\textbf{Full Model (\textit{Ours})}&\textbf{27.256}&	\textbf{0.9308}&	\textbf{0.2304}&	0.3962	&49.65\\
\shline
\end{tabular}
\end{table*}
\begin{figure}[!t]
% \vspace{-2mm}
\centering
\begin{center}
\begin{tabular}{ccccccccc}
\hspace{-1mm}\includegraphics[width=0.4\linewidth]{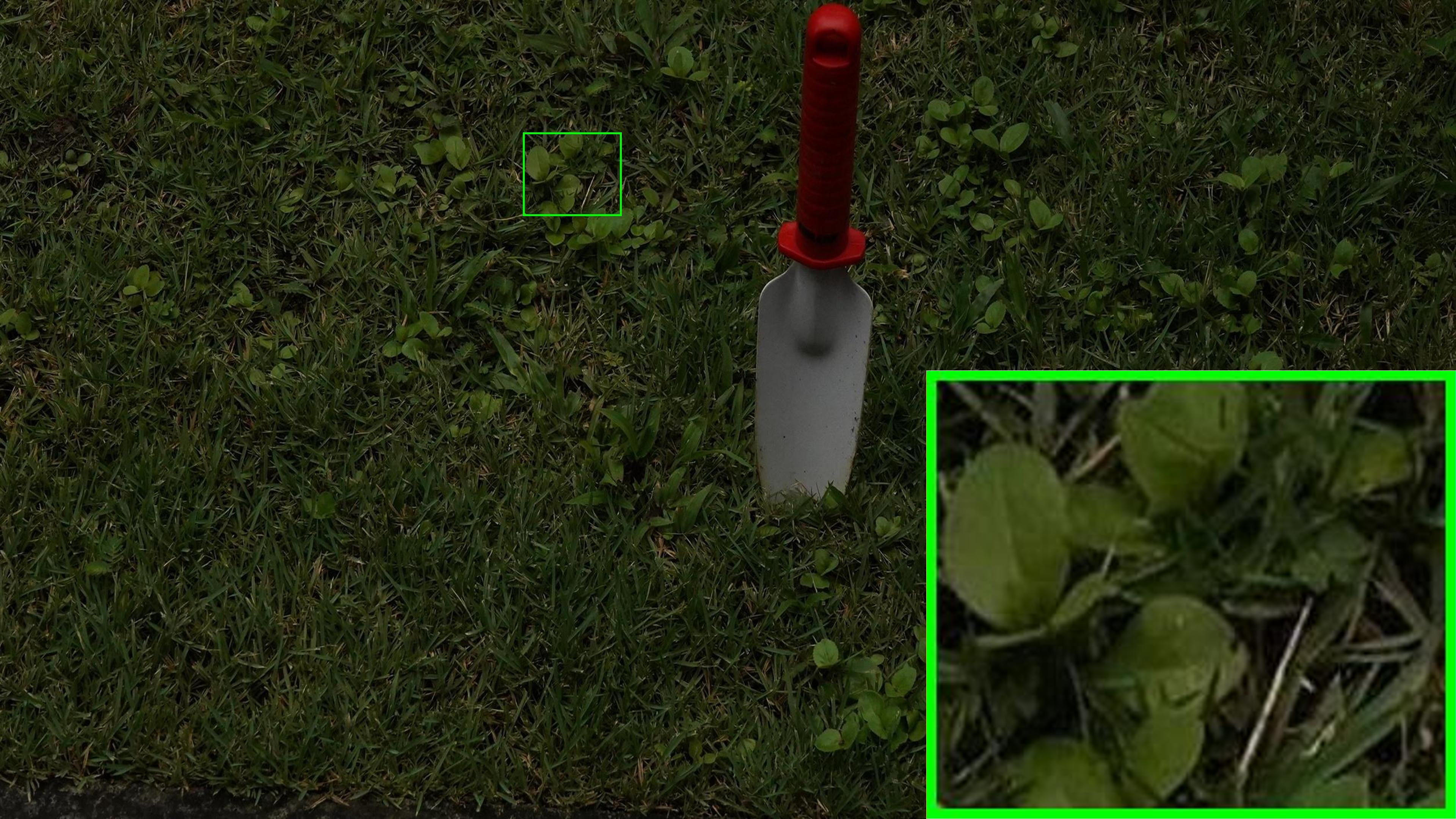} &\hspace{-4.75mm}
\includegraphics[width=0.4\linewidth]{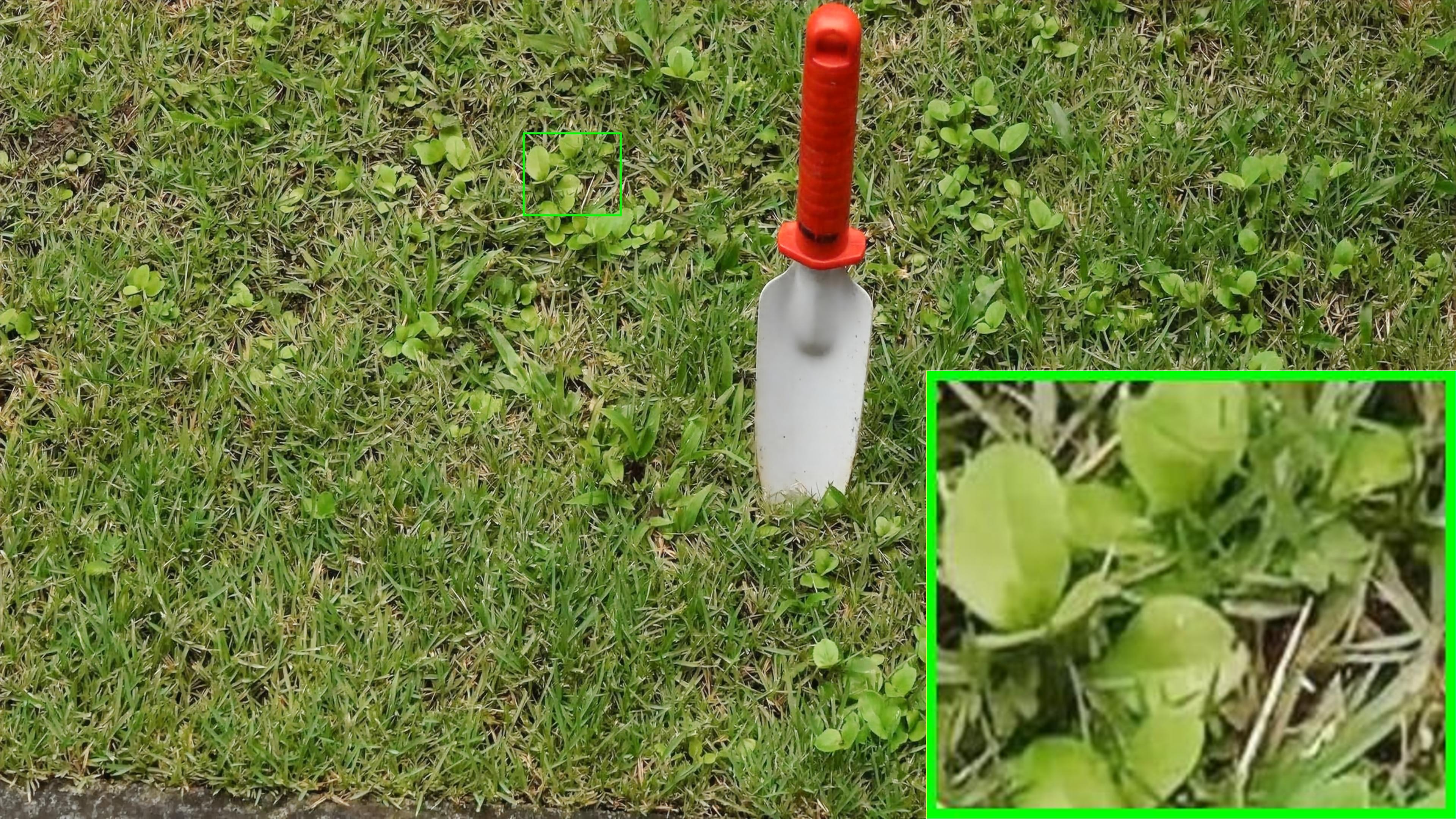} 
\\
\hspace{-1mm} (a) Input&\hspace{-4.75mm}(b) w/o FR-CMT
\\
\hspace{-1mm}\includegraphics[width=0.4\linewidth]{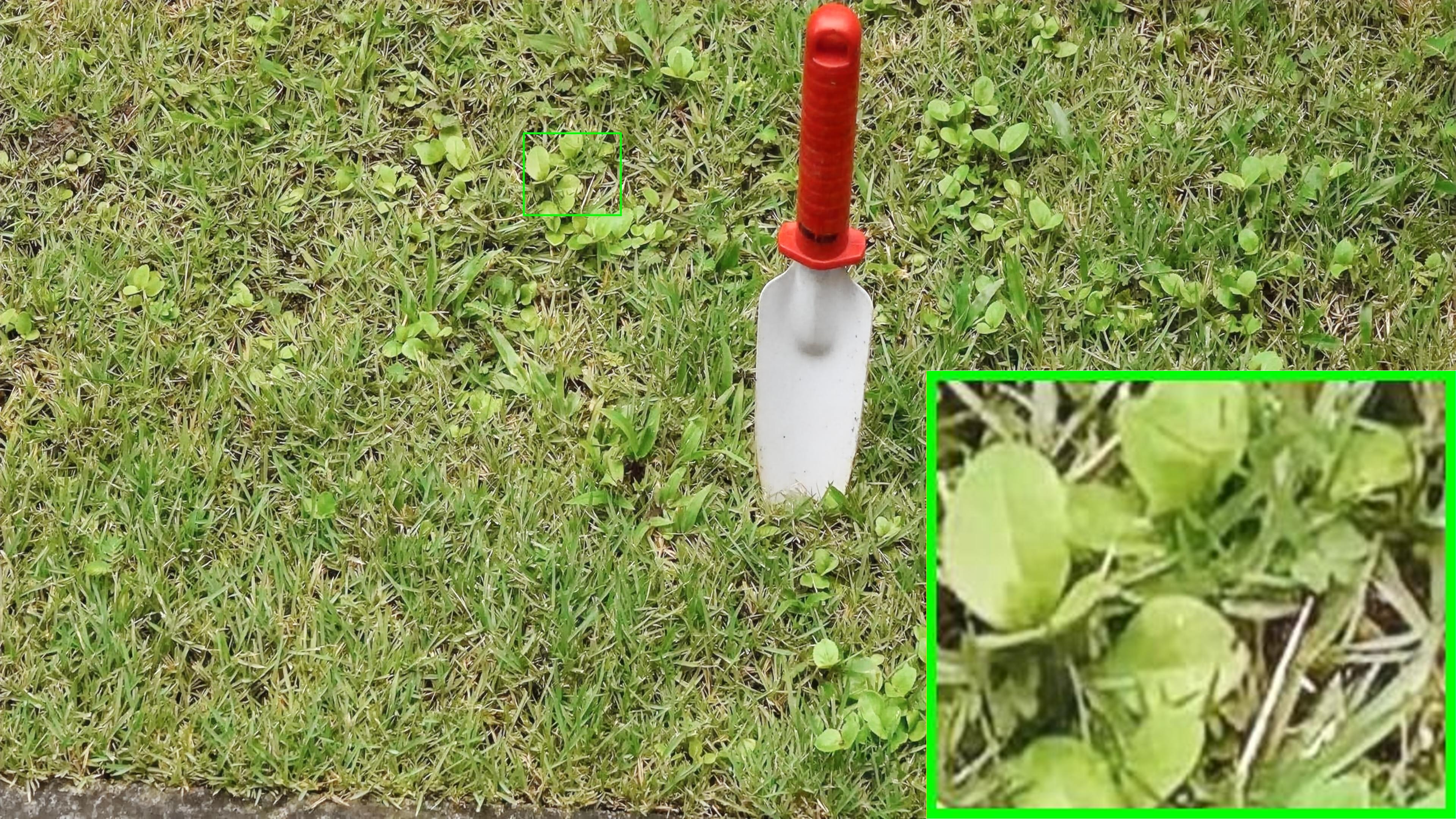} &\hspace{-4.75mm}
\includegraphics[width=0.4\linewidth]{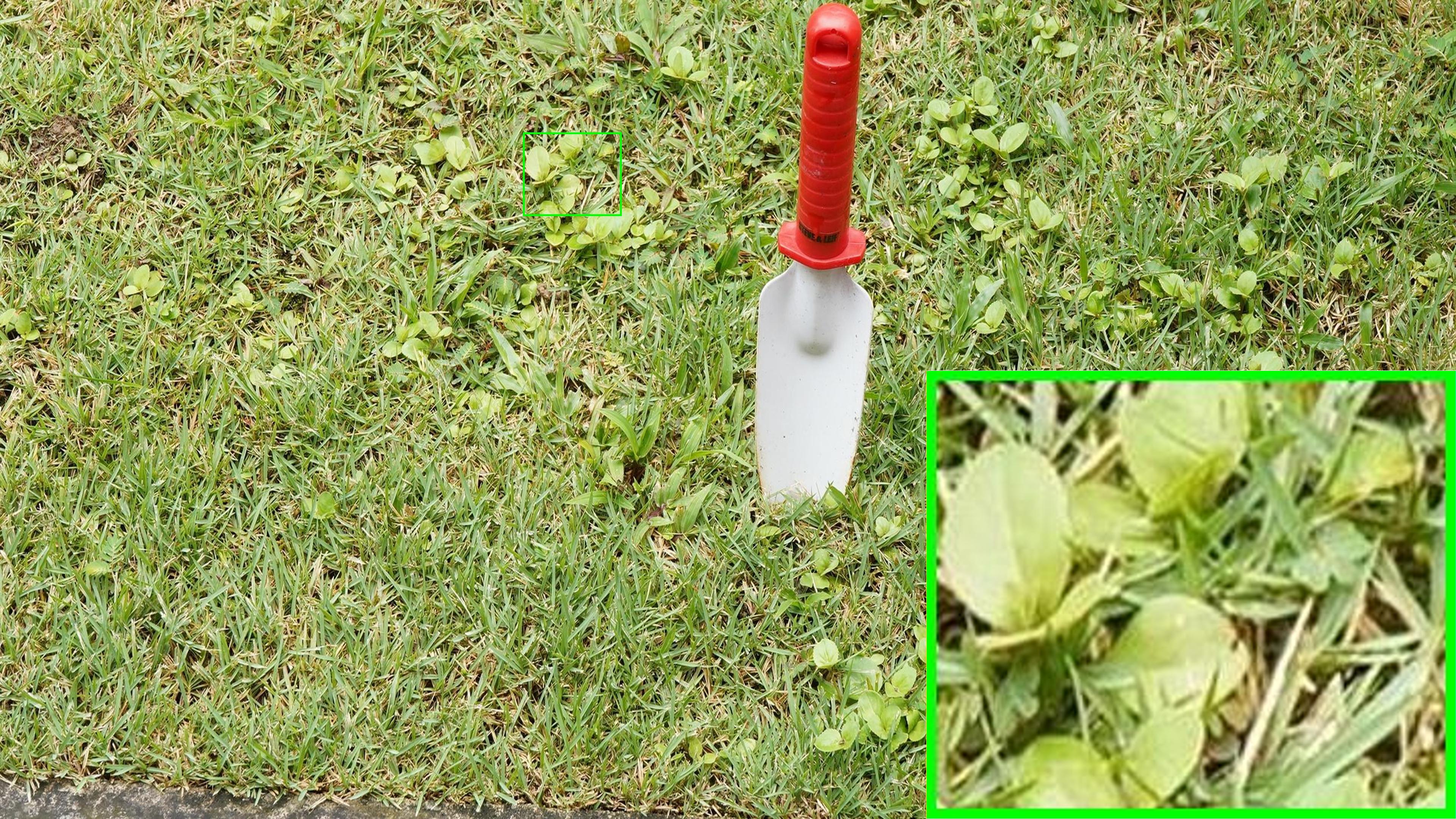} 
\\
\hspace{-1mm}(c) w/ FR-CMT (\textit{\textbf{Ours}}) &\hspace{-4.75mm}(d) GT
\end{tabular}
 \vspace{-2mm}
\caption{\textbf{Visual effect on Feature-Refined Correlation Matching Transformation} (FR-CMT).
% Using FR-CMT is able to help produce clearer visual results.
}
\label{fig:Visual effect on Maximal Correlation Matching Module}
\end{center}
 \vspace{-3mm}
\end{figure}

\subsubsection{Effect on Basic Components}\label{sec: Effect on Basic Component}
Compared with UHDformer~\cite{aaai24wang_UHDformer}, our UHDformer++ replaces the layer normalization, feed-forward network, and matching module with a newly designed module.
To verify the contribution of each modification, we conduct a systematic ablation study, with results reported in Tab.~\ref{tab:Effect on Ablation study on LN}.
The results show that DyT~\cite{zhu2025dyt} is more effective than standard layer normalization~\cite{ba2016layer}, and that both normalization strategies outperform the absence of any normalization. 
Furthermore, the newly proposed feed-forward network consistently surpasses both the CMTFN used in UHDformer~\cite{aaai24wang_UHDformer} and the GDFN used in Restormer~\cite{Zamir2021Restormer}. 
Finally, the proposed FR-CMT outperforms the DualCMT adopted in~\cite{aaai24wang_UHDformer}. 
Collectively, these ablation results confirm that each of the newly introduced components contributes meaningfully to overall restoration performance.

\subsubsection{Effect on Feature-Refined Correlation Matching Transformation}\label{sec: Effect on Dual-path Correlation Matching Transformation}
As FR-CMT is one of the core designs of UHDformer++, we conduct a detailed ablation study to investigate its impact on UHD restoration quality. 
Specifically, we assess variants in which FR-CMT is disabled within either the CMTA or CMTFN, or in which the Max-Pooling or Mean-Pooling operations within FR-CMT are individually removed, comparing each against the full model. 
The ablation results are presented in Tab.~\ref{tab:Effect on Ablation study on MCM}.
The full model with FR-CMT (Tab.~\ref{tab:Effect on Ablation study on MCM}(f)) achieves a PSNR gain of $0.323$dB over the variant without FR-CMT (Tab.~\ref{tab:Effect on Ablation study on MCM}(a)), confirming its contribution to restoration quality.
Moreover, disabling FR-CMT in either CMTA (Tab.~\ref{tab:Effect on Ablation study on MCM}(b)) or CMTFN (Tab.~\ref{tab:Effect on Ablation study on MCM}(c)) leads to a consistent PSNR degradation, and the concurrent use of both Max-Pooling and Mean-Pooling is shown to further benefit restoration performance. 
Visual comparisons in Fig.~\ref{fig:Visual effect on Maximal Correlation Matching Module} corroborate these findings, demonstrating that FR-CMT contributes to more natural color rendition and reduced visual artifacts.
We further investigate the effect of the squeezing factor $r$, which controls the compression level of feature maps following the matching operation in FR-CMT. 
As reported in Tab.~\ref{tab:Effect on Effect on Matching Factor.}, PSNR is maximized when $r=4$, suggesting that a moderate degree of feature compression, which may suppress redundant information while preserving discriminative features for subsequent learning, is preferable to either more aggressive or more conservative squeezing.

\begin{figure}[!t]
\begin{center}
\begin{tabular}{cccccccccc}
\includegraphics[width=0.4\linewidth]{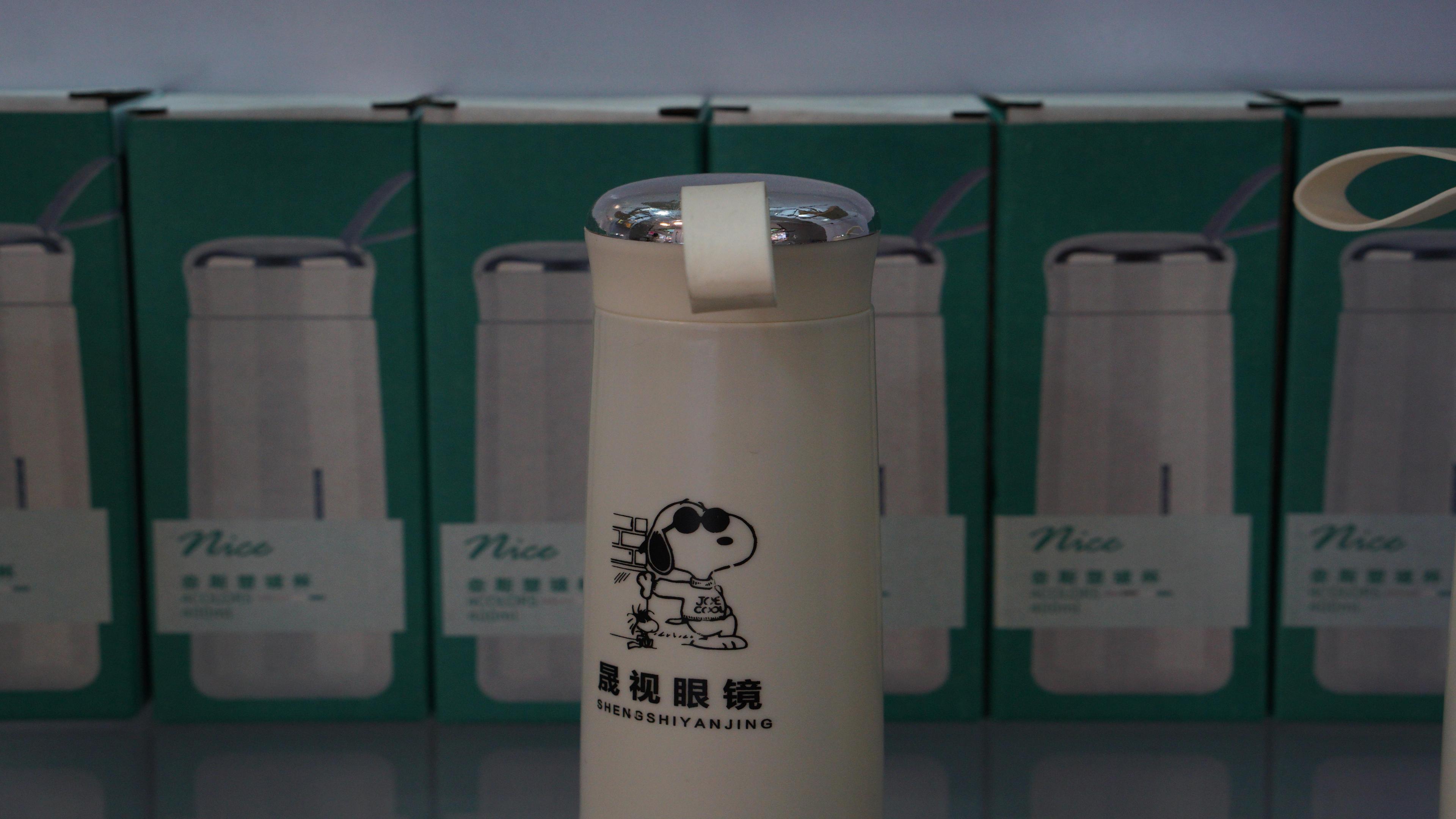}&\hspace{-4.5mm}
\includegraphics[width=0.4\linewidth]{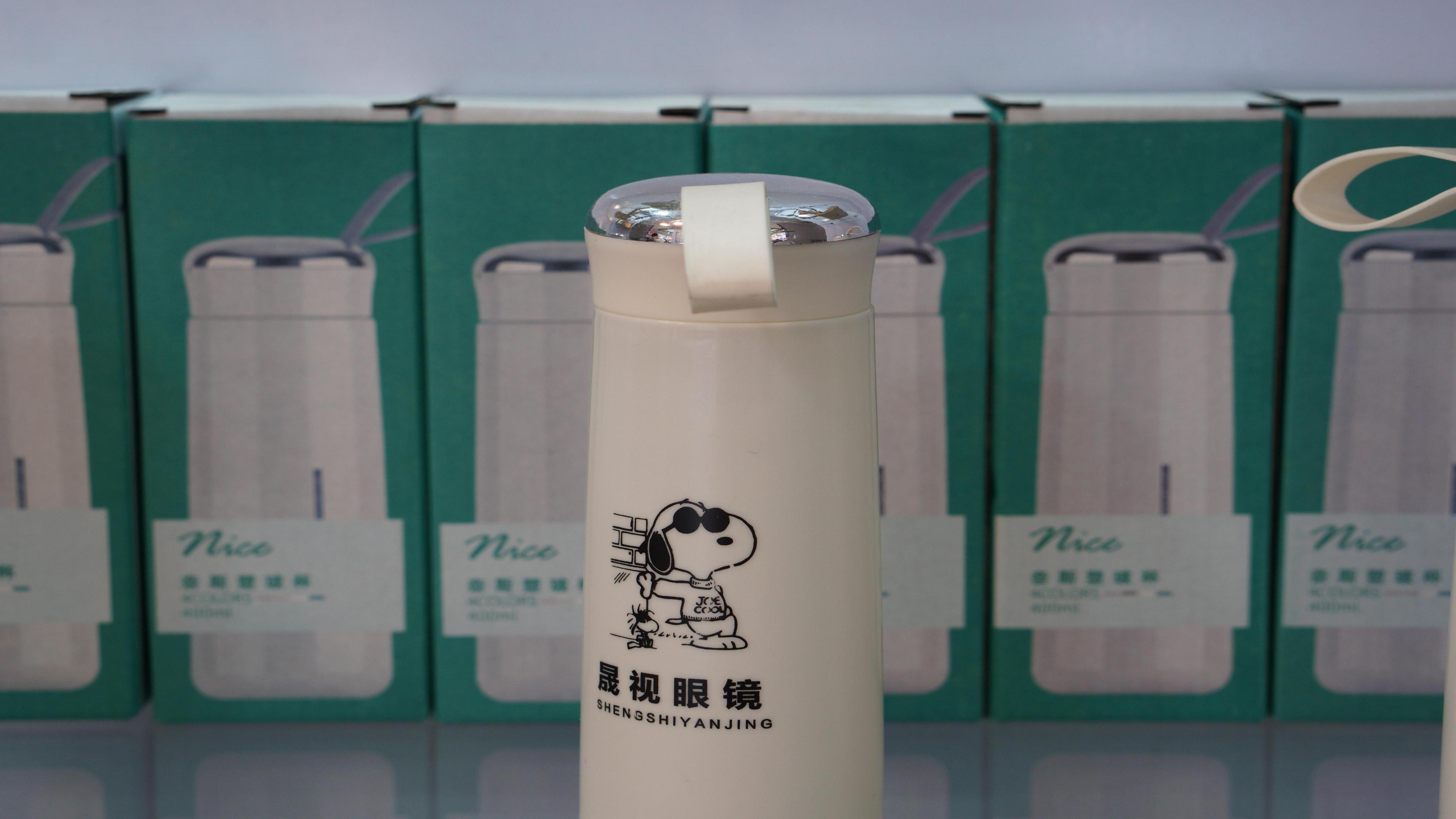}
\\
\footnotesize (a) Input &\hspace{-4.5mm}\footnotesize (b) GT
\\
\includegraphics[width=0.4\linewidth]{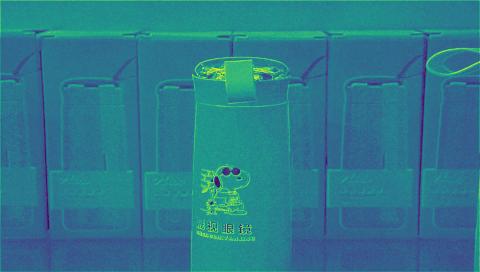}&\hspace{-4.5mm}
\includegraphics[width=0.4\linewidth]{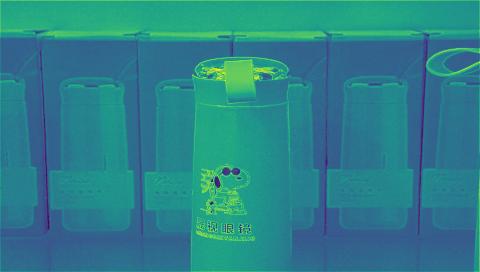}
\\
\footnotesize (c)  before CMT&\hspace{-4.5mm}\footnotesize (d) after CMT
\\
% &\hspace{3mm}\footnotesize (c) &\hspace{-4.5mm}\footnotesize (d) 
\end{tabular}
\vspace{-2mm}
\caption{\textbf{Feature visualization} on correlation matching transformation.
Please zoom in on a high-resolution display.
}
\label{fig: Feature visualization on correlation matching for forward network.}
\end{center}
 \vspace{-3mm}
\end{figure}

To better understand the functional role of FR-CMT, the core design of UHDformer++, we visualize the feature maps before and after the matching operation in both CMTA and CMTN, as presented in Fig.~\ref{fig: Feature visualization on correlation matching for forward network.}. 
As shown, the transformed features (Fig.~\ref{fig: Feature visualization on correlation matching for forward network.}(d)) exhibit notably clearer and more salient representations compared to the untransformed low-resolution features (Fig.~\ref{fig: Feature visualization on correlation matching for forward network.}(c)) in both modules. 
This demonstrates that the correlation matching transformation operation effectively enriches low-resolution feature representations with high-resolution contextual information, thereby facilitating more accurate and detailed UHD image restoration.

%%%%%%%%%%%%%%%%%%%%%%%%%%%%%%%%%%%%%%%%%%%%%%%%%%%
%

%%%%%%%%%%%%%%%%%%%%%%%%%%%%%%%%%%%%%%%%%%%%%%%%%%%%%%%%%%%%%%%%%%%%
\begin{table}[!t]
\caption{\textbf{Effect on Squeezing Factor $r$} in  Feature-Refined Correlation Matching Transformation.
}
% \vspace{-2mm}
\label{tab:Effect on Effect on Matching Factor.} 
\vspace{-2mm}
\tablestyle{7pt}{1}
\centering
\begin{tabular}{l|cccc}
\shline
 $r$ &1&2 &\textbf{4 (\textit{Default})} &8
\\
\shline
\textbf{PSNR (dB)}~$\uparrow$& 26.7669&	27.3854&	\textbf{27.2567}&	27.1235 \\
\textbf{SSIM}~$\uparrow$&0.9271&0.9288&\textbf{0.9308}&0.9278\\
\textbf{LPIPS}~$\downarrow$&0.2376 &0.2383&\textbf{0.2304}&0.2345  \\
 \shline
\textbf{Params. (M)}~$\downarrow$& 0.3962  &0.3962  & 0.3962 & 0.3962  \\
\textbf{FLOPs (G)}~$\downarrow$& 49.65   &  49.65 & 49.65  & 49.65   \\
\shline
\end{tabular}
\vspace{-3mm}
\end{table}
%%%%%%%%%%%%%%%%%%%%%%%%%%%%%%%%%%%%%%%%%%%%%%%%%%%%%%%%%%%%%%%%%%%%%%%%%%%%%%%%%%%%%%%%%%%%%%%%%
\subsubsection{Effect on Adaptive Channel Modulator}

\begin{figure}[!t]
% \vspace{-2mm}
\centering
\begin{center}
\begin{tabular}{ccccccccc}
\hspace{-1mm}\includegraphics[width=0.4\linewidth]{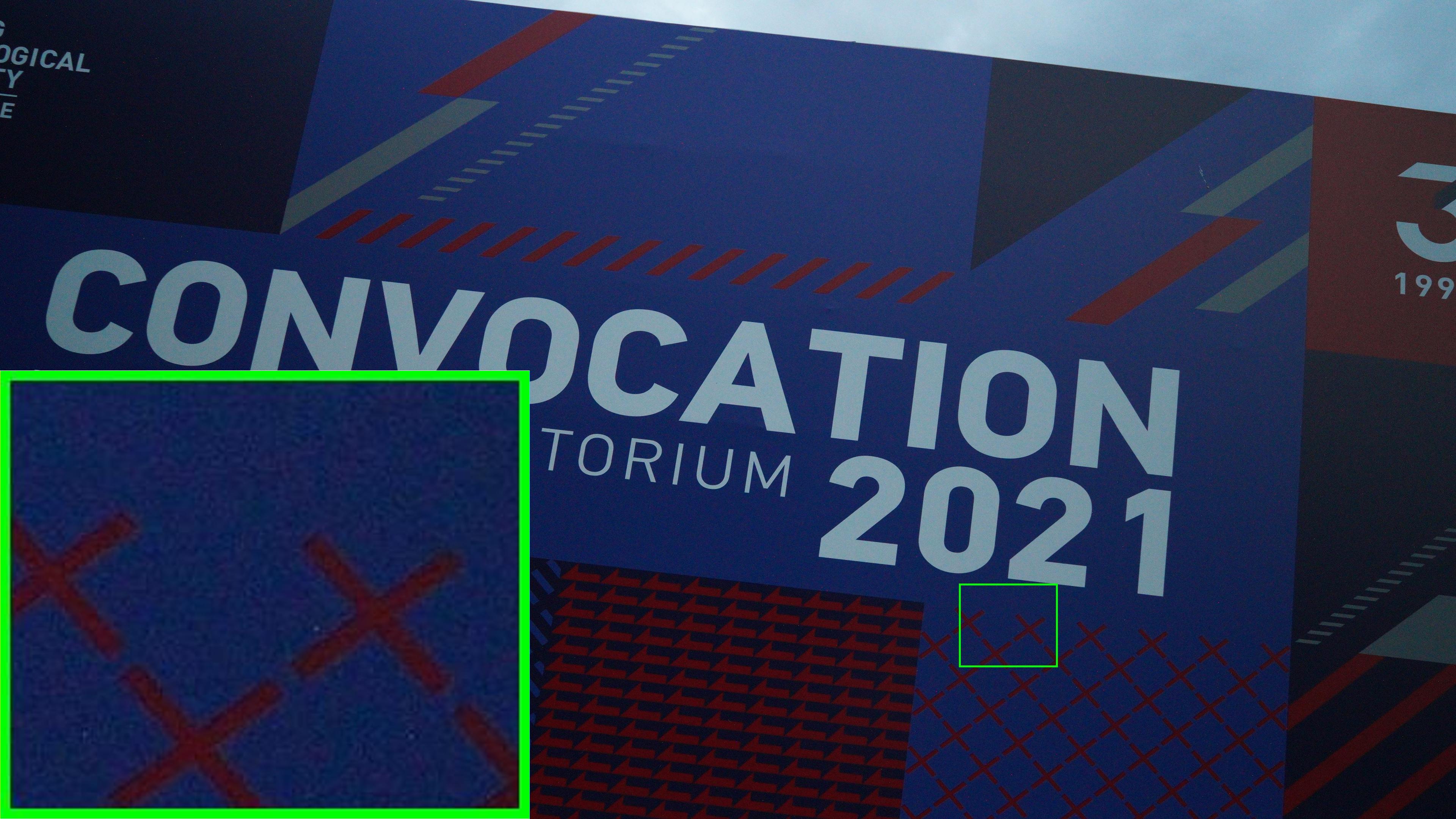} &\hspace{-4.75mm}
\includegraphics[width=0.4\linewidth]{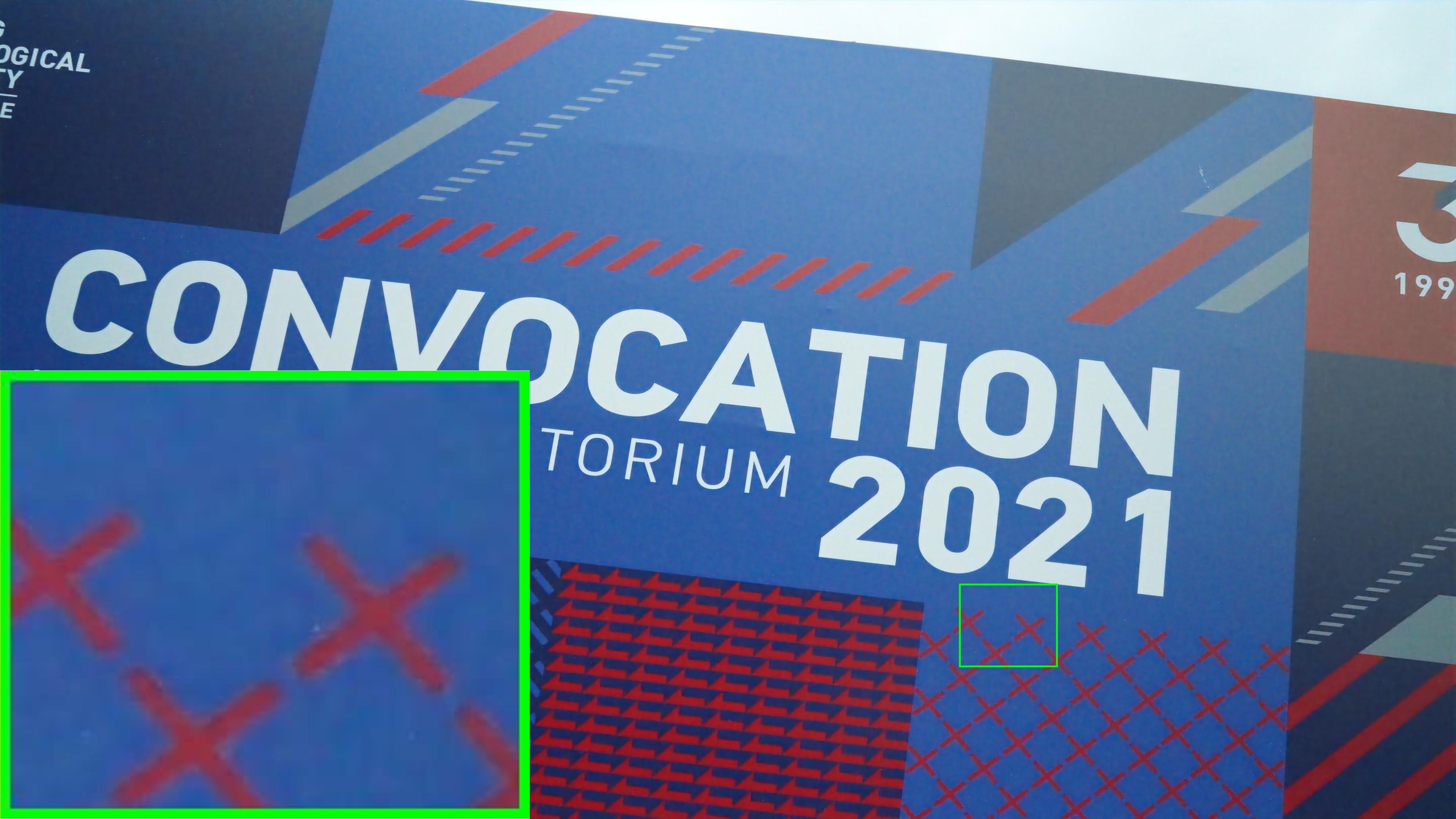} 
\\
\hspace{-1mm} (a) Input&\hspace{-4.75mm} (b) w/o ACM
\\
\hspace{-1mm}\includegraphics[width=0.4\linewidth]{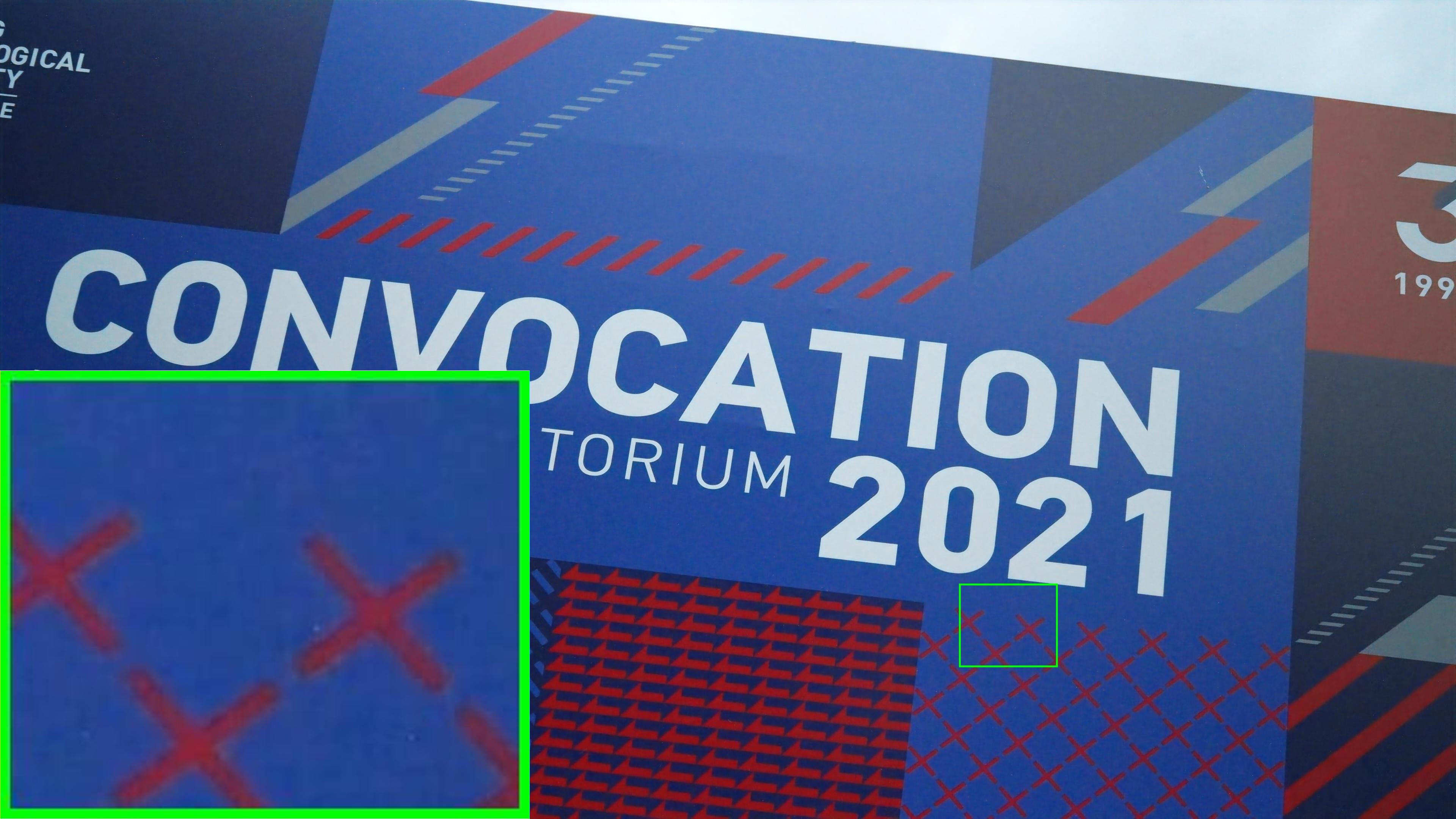} &\hspace{-4.75mm}
\includegraphics[width=0.4\linewidth]{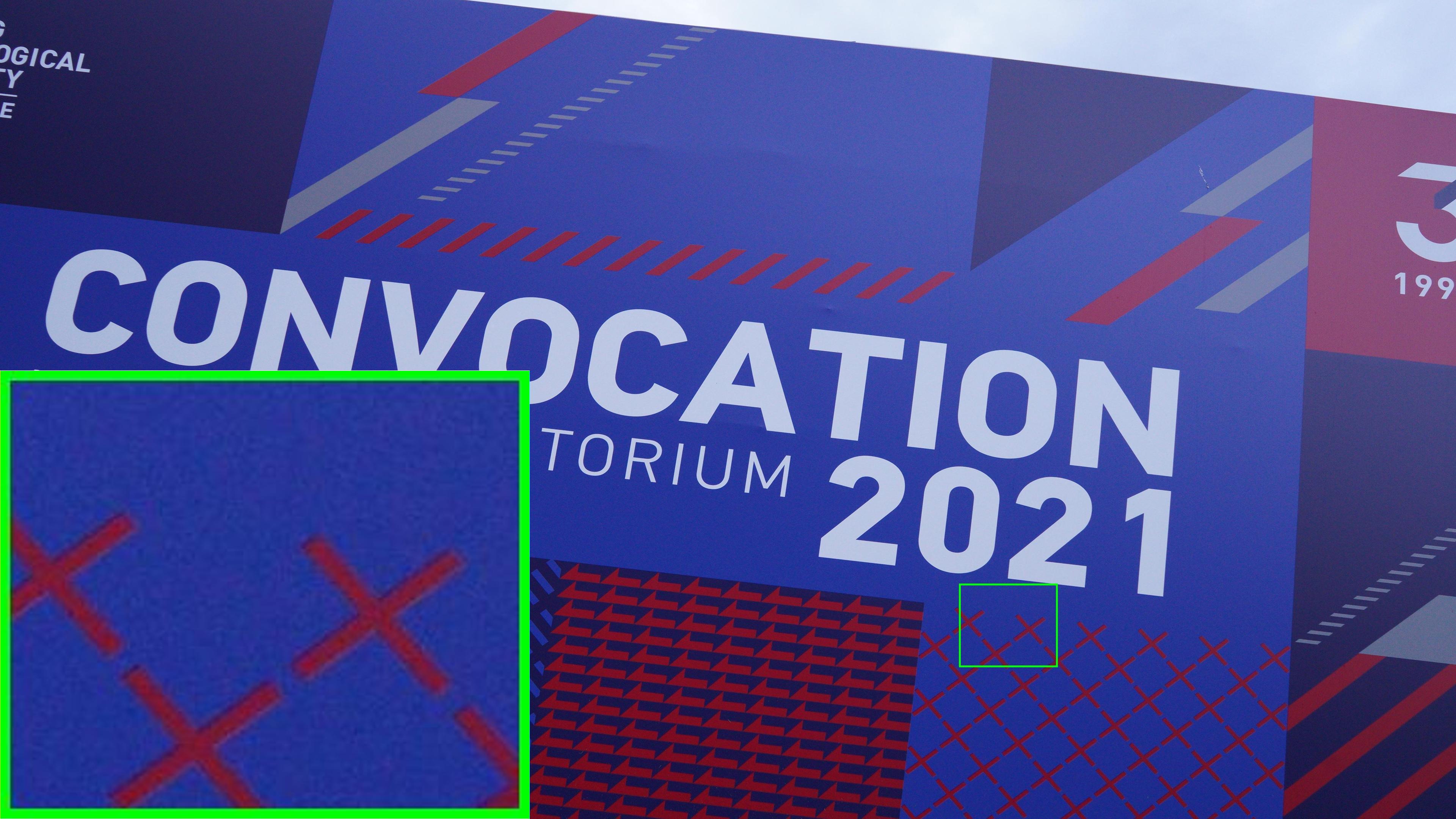} 
\\
\hspace{-1mm}  (c)  w/ ACM (\textit{\textbf{Ours}}) &\hspace{-4.75mm} (d) GT
\end{tabular}
\vspace{-2mm}
\caption{\textbf{Visual effect on Adaptive Channel Modulator} (ACM).
}
\label{fig:Visual effect on Adaptive Channel Modulator}
\end{center}
\vspace{-3mm}
\end{figure}

We further analyze the impact of the proposed Adaptive Channel Modulator (ACM), which adaptively modulates multi-level high-resolution features from a channel-wise perspective. 
To this end, we conduct a systematic ablation study comprising three configurations: 1) removing ACM entirely; 2) replacing it with Efficient Channel Attention (ECA)~\cite{wang2020eca}; and 3) replacing it with Squeeze-and-Excitation channel attention (SE)~\cite{se}. 
The results are summarized in Tab.~\ref{tab:Effect on Adaptive Channel Modulator}.
ACM consistently outperforms both the ECA~\cite{wang2020eca} and SE~\cite{se} attention modules. 
Notably, ACM achieves a PSNR gain of approximately $0.506$dB over ECA, despite both modules having nearly identical parameter counts, which strongly demonstrates the effectiveness of our design. 
Visual comparisons in Fig.~\ref{fig:Visual effect on Adaptive Channel Modulator} further support this observation that the model with ACM produces more natural and faithful restoration results (Fig.~\ref{fig:Visual effect on Adaptive Channel Modulator}(c)), whereas the model without ACM tends to yield under-restored outputs (Fig.~\ref{fig:Visual effect on Adaptive Channel Modulator}(b)).

\begin{table*}[!t]
\caption{\textbf{Ablation study on Adaptive Channel Modulator} (ACM).
}
\vspace{-2mm}
\label{tab:Effect on Adaptive Channel Modulator} 
%\tablestyle{1.2pt}{1.1}
\tablestyle{3pt}{1}
\centering
\begin{tabular}{l|l|ccc|cc}
\shline
\multicolumn{1}{c|}{\multirow{2}{*}{\textbf{ID}}}& \multicolumn{1}{c|}{\multirow{2}{*}{\textbf{Experiment}}}&\multicolumn{3}{c|}{\textbf{Metrics}}&\multicolumn{2}{c}{\textbf{Computational Complexity}}
\\
\cline{3-7}
&& \textbf{PSNR (dB)}~$\uparrow$ & \textbf{SSIM}~$\uparrow$ &\textbf{LPIPS}~$\downarrow$&\textbf{Params. (M)}~$\downarrow$ &\textbf{FLOPs (G)}~$\downarrow$ 
\\
\shline
(a) &w/o Adaptive Channel Modulator (ACM)&26.953&	0.9289&	0.2357&	0.3953	&48.65 \\
(b) &w/ Efficient Channel Attention (ECA)~\cite{wang2020eca}&26.750&	0.9278&	0.2407&	0.3953	&48.7 \\
(c) &w/ Squeeze-and-Excitation Attention (SE)~\cite{se}&27.008&0.9286&	0.2355&	0.3969	&48.7  \\
\shline
\textbf{(d)} &\textbf{w/ Adaptive Channel Modulator (ACM) (\textit{Ours})}&\textbf{27.256}&\textbf{0.9308}&	\textbf{0.2304}&	0.3962	&49.65\\
\shline
\end{tabular}
\vspace{-2mm}
\end{table*}
%%%%%%%%%%%%%%%%%%%%%%%%%%%%%%%%%%%%%%%%%%%%%%%%%%%%%%%%%%%%%%%%%%%%%%%%%%%%%%%%%%%%

We further investigate whether assigning an individual ACM to each Transformer block, rather than sharing a single ACM across all blocks, would yield improved performance. 
As shown in Tab.~\ref{tab:Ablation study on the learning manner of Adaptive Channel Modulator}, the shared ACM configuration consistently outperforms the individual ACM counterpart. 
This may be attributed to the fact that a shared ACM encourages the model to learn more generalizable and representative features from the high-resolution space, which in turn provides more consistent guidance to facilitate feature learning in the low-resolution space.
\begin{table*}[!t]
\caption{\textbf{Ablation study on the learning manner of Adaptive Channel Modulator}.
}
\vspace{-2mm}
\label{tab:Ablation study on the learning manner of Adaptive Channel Modulator} 
% \tablestyle{3pt}{1.1}
\tablestyle{4pt}{1}
\begin{tabular}{l|l|ccc|cc}
\shline
\multicolumn{1}{c|}{\multirow{2}{*}{\textbf{ID}}}& \multicolumn{1}{c|}{\multirow{2}{*}{\textbf{Experiment}}}&\multicolumn{3}{c|}{\textbf{Metrics}}&\multicolumn{2}{c}{\textbf{Computational Complexity}}
\\
\cline{3-7}
&& \textbf{PSNR (dB)}~$\uparrow$ & \textbf{SSIM}~$\uparrow$ &\textbf{LPIPS}~$\downarrow$&\textbf{Params. (M)}~$\downarrow$ &\textbf{FLOPs (G)}~$\downarrow$ 
\\
\shline
(a) &Individual Adaptive Channel Modulator&  26.981&	0.9281&	0.2404&	0.4062	&60.07\\
\shline
\textbf{(b)}& \textbf{Shared Adaptive Channel Modulator} (\textit{\textbf{Ours}})&\textbf{27.256}&	\textbf{0.9308}&	\textbf{0.2304}&	0.3962&	49.65\\
\shline
\end{tabular}
\vspace{-2mm}
\end{table*}
%%%%%%%%%%%%%%%%%%%%%%%%%%%%%%%%%
\begin{figure}[!t]
% \vspace{-2mm}
\centering
\begin{center}
\begin{tabular}{ccccccccc}
\hspace{-1mm}\includegraphics[width=0.4\linewidth]{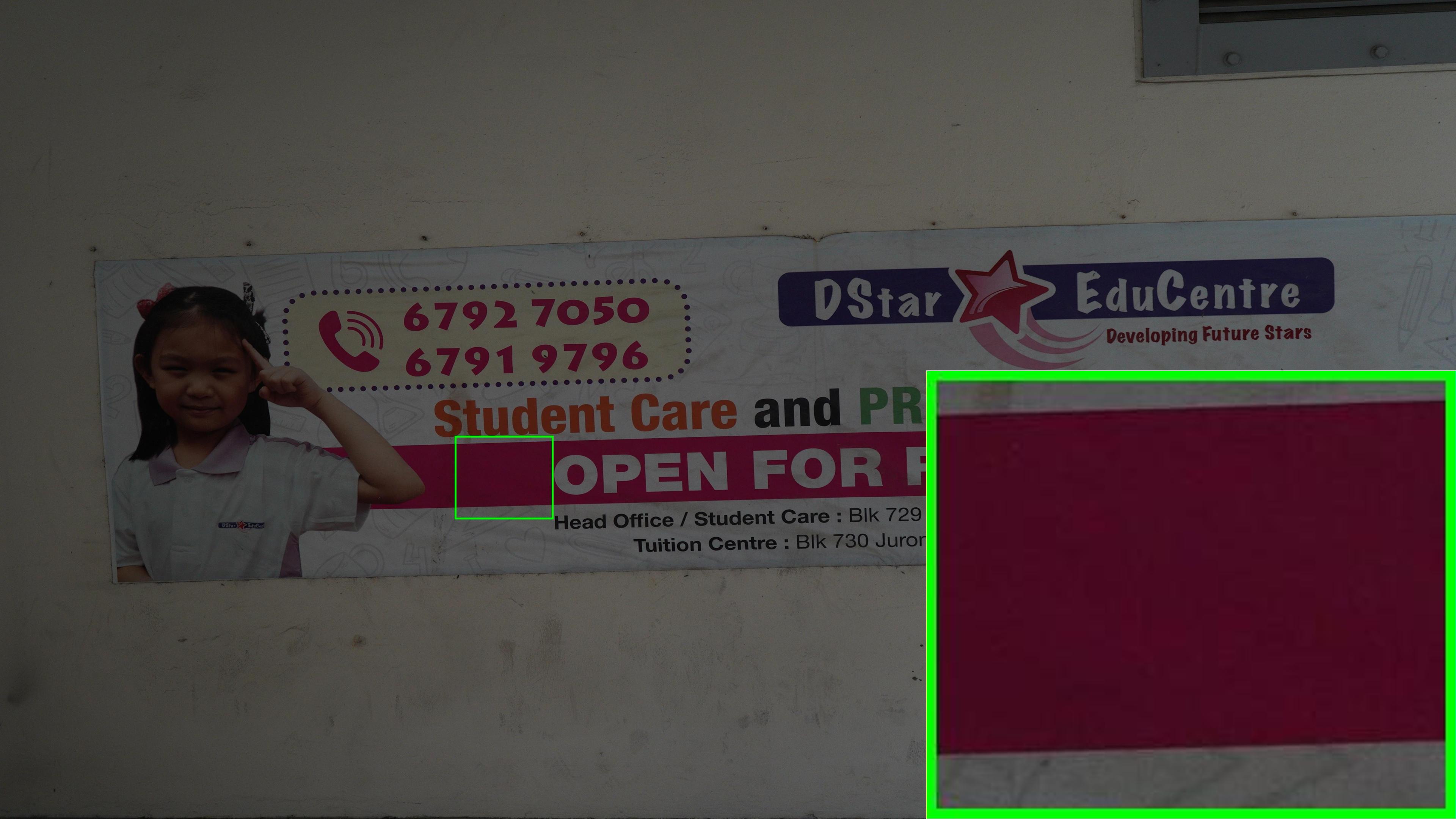} &\hspace{-4.75mm}
\includegraphics[width=0.4\linewidth]{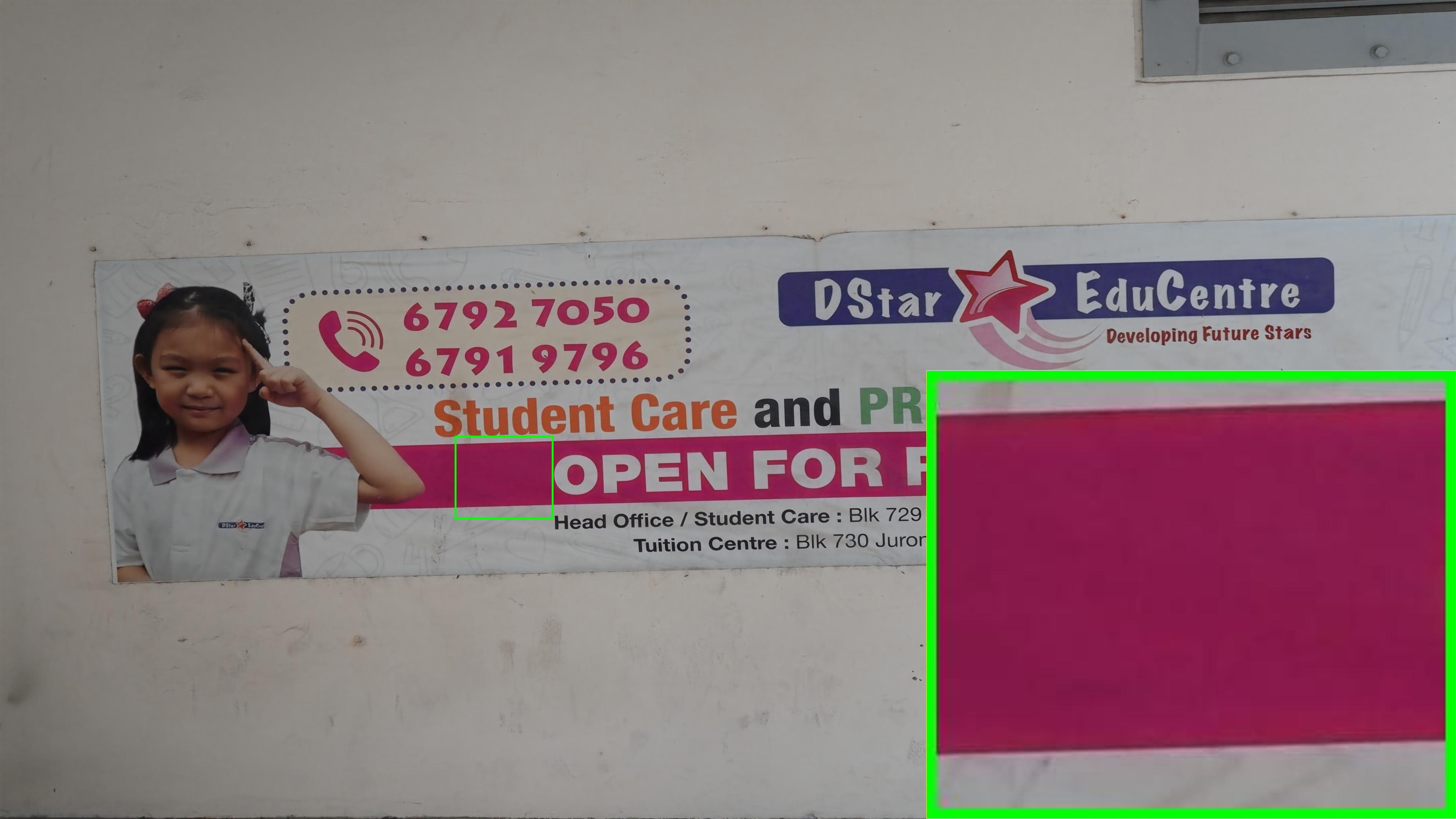} 
\\
\hspace{-1mm} (a) Input&\hspace{-4.75mm} (b) Individual ACM
\\
\hspace{-1mm}\includegraphics[width=0.4\linewidth]{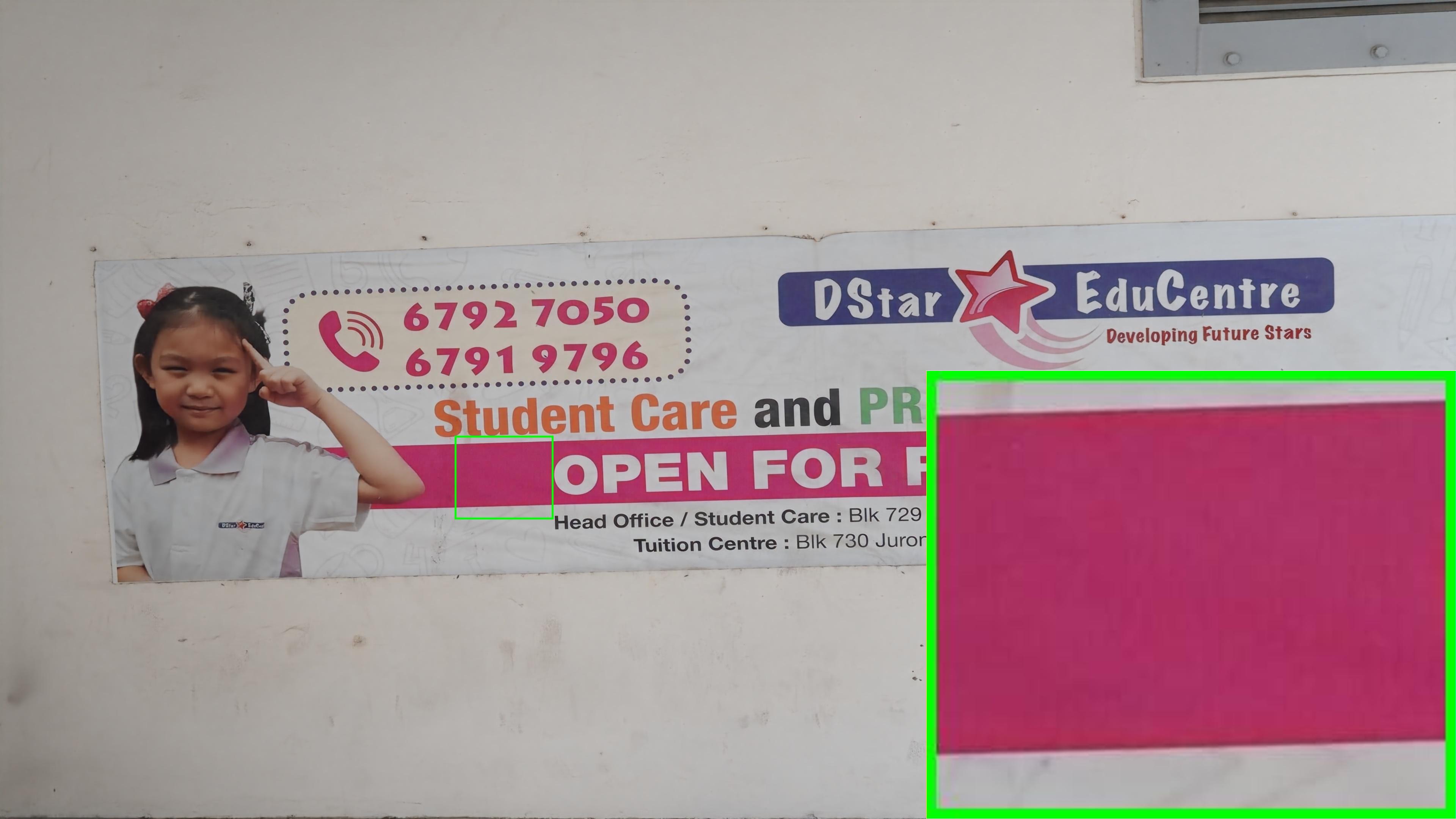} &\hspace{-4.75mm}
\includegraphics[width=0.4\linewidth]{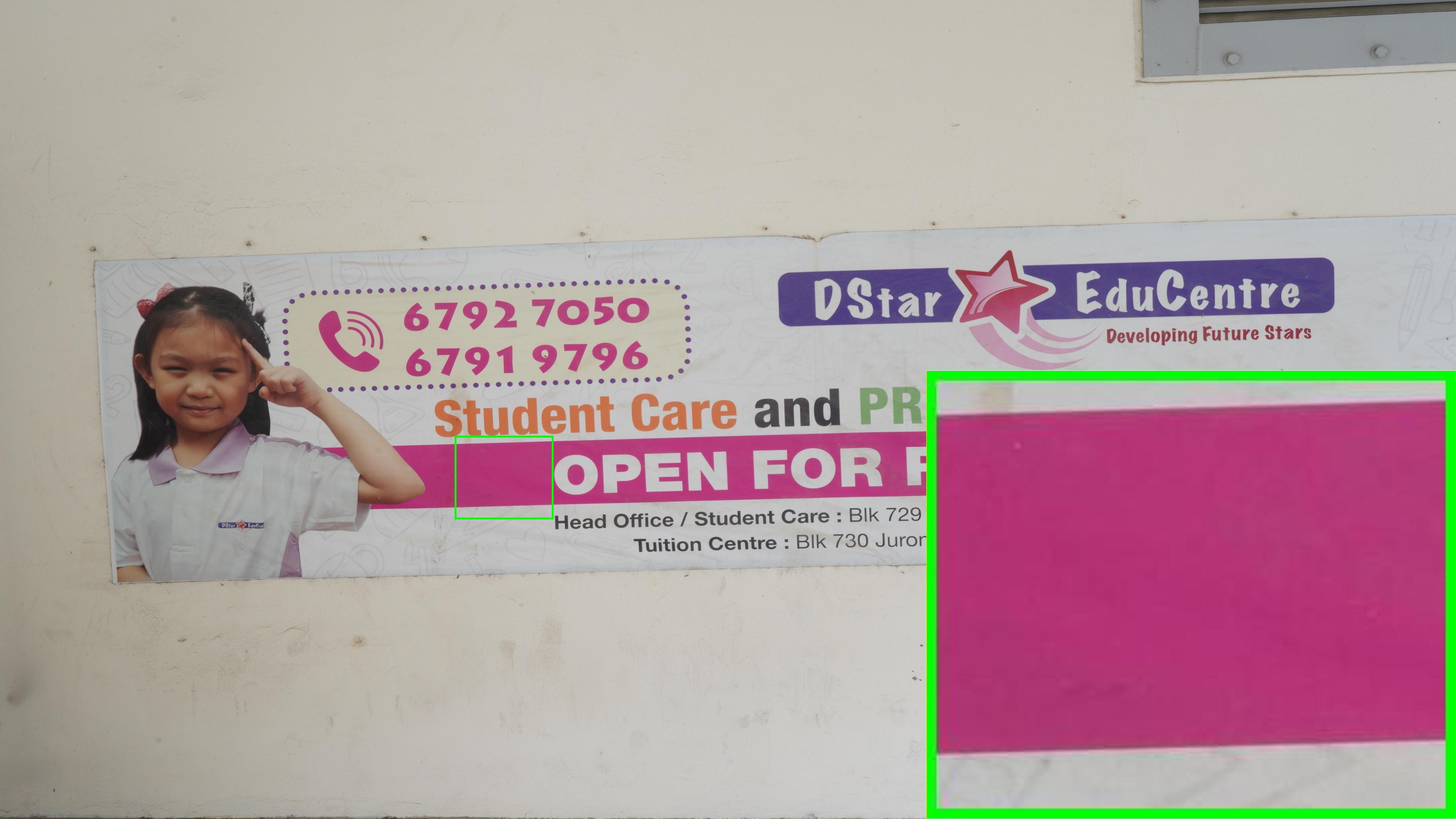} 
\\
\hspace{-1mm}  (c)  Shared ACM (\textit{\textbf{Ours}}) &\hspace{-4.75mm} (d) GT
\end{tabular}
\vspace{-2mm}
\caption{\textbf{Visual effect on learning manner of Adaptive Channel Modulator (ACM)}.
% The proposed ACM helps produce results with vivid color.
}
\label{fig: ablation study on the learning manner of Adaptive Channel Modulator}
\end{center}
\vspace{-3mm}
\end{figure}

\subsubsection{Effect on Gated Feature Refinement}
\begin{figure}[!t]
% \vspace{-2mm}
\centering
\begin{center}
\begin{tabular}{ccccccccc}
\hspace{-1mm}\includegraphics[width=0.4\linewidth]{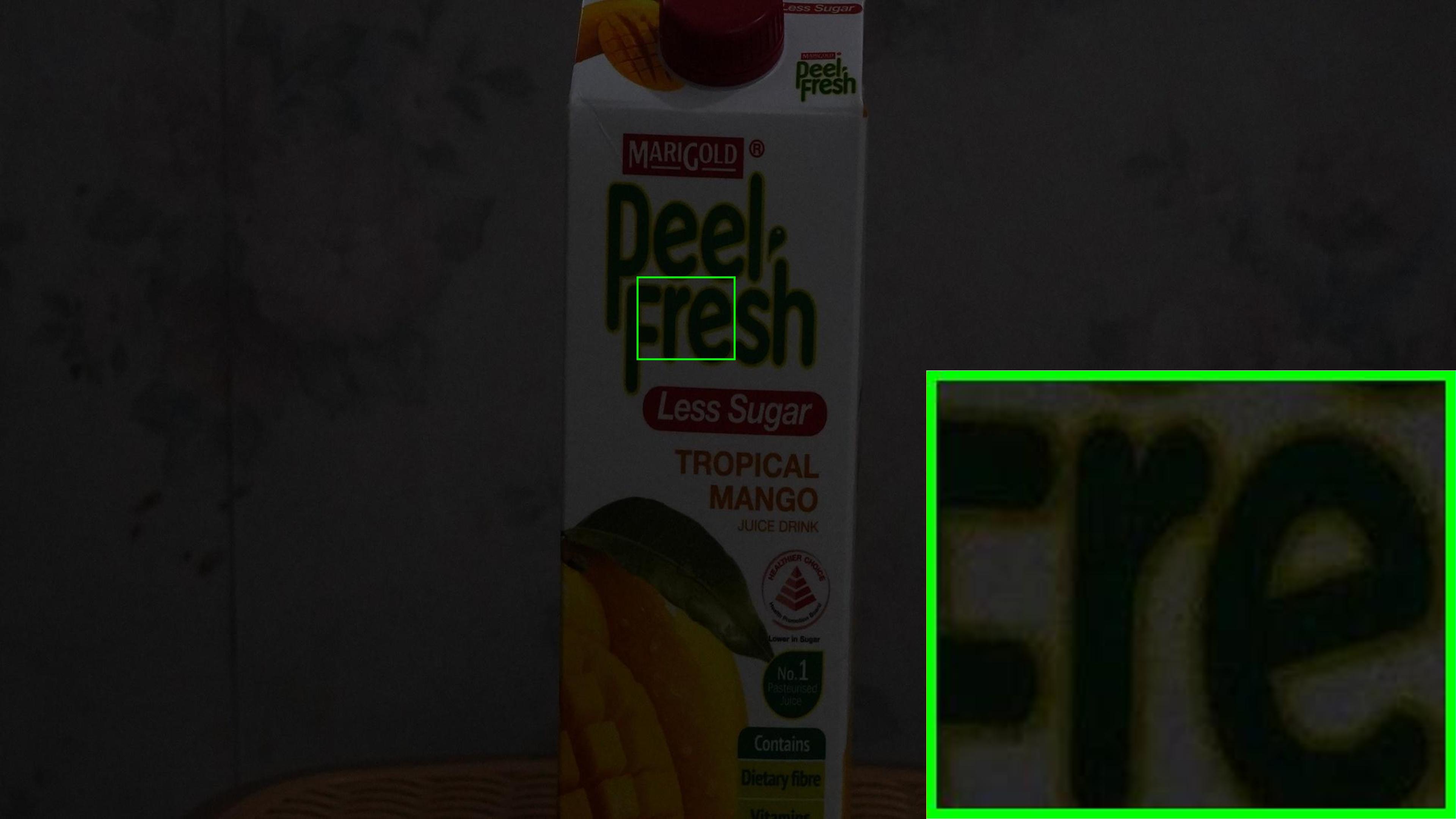} &\hspace{-4.75mm}
\includegraphics[width=0.4\linewidth]{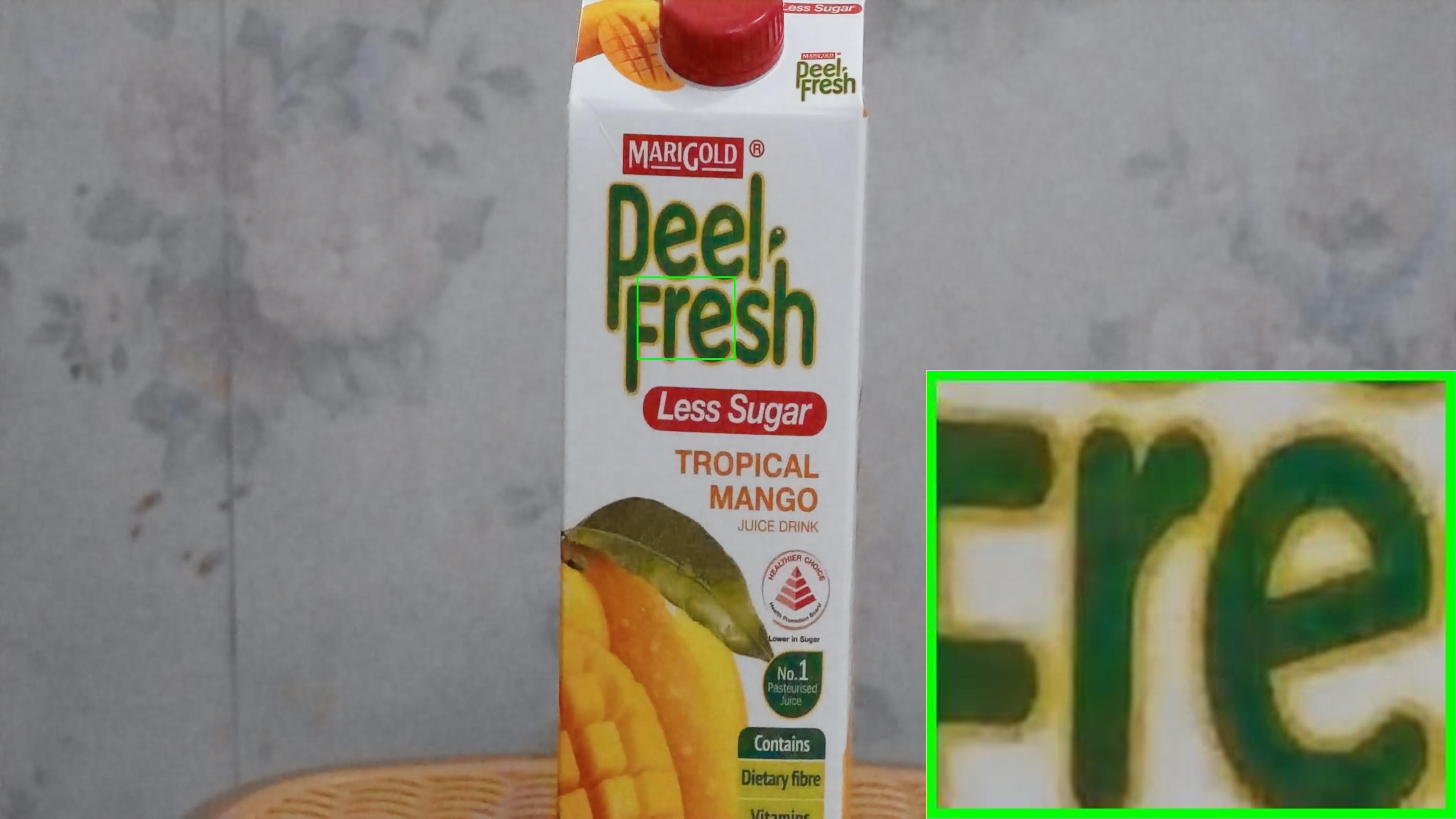} 
\\
\hspace{-1mm} (a) Input&\hspace{-4.75mm} (b) w/o GFR
\\
\hspace{-1mm}\includegraphics[width=0.4\linewidth]{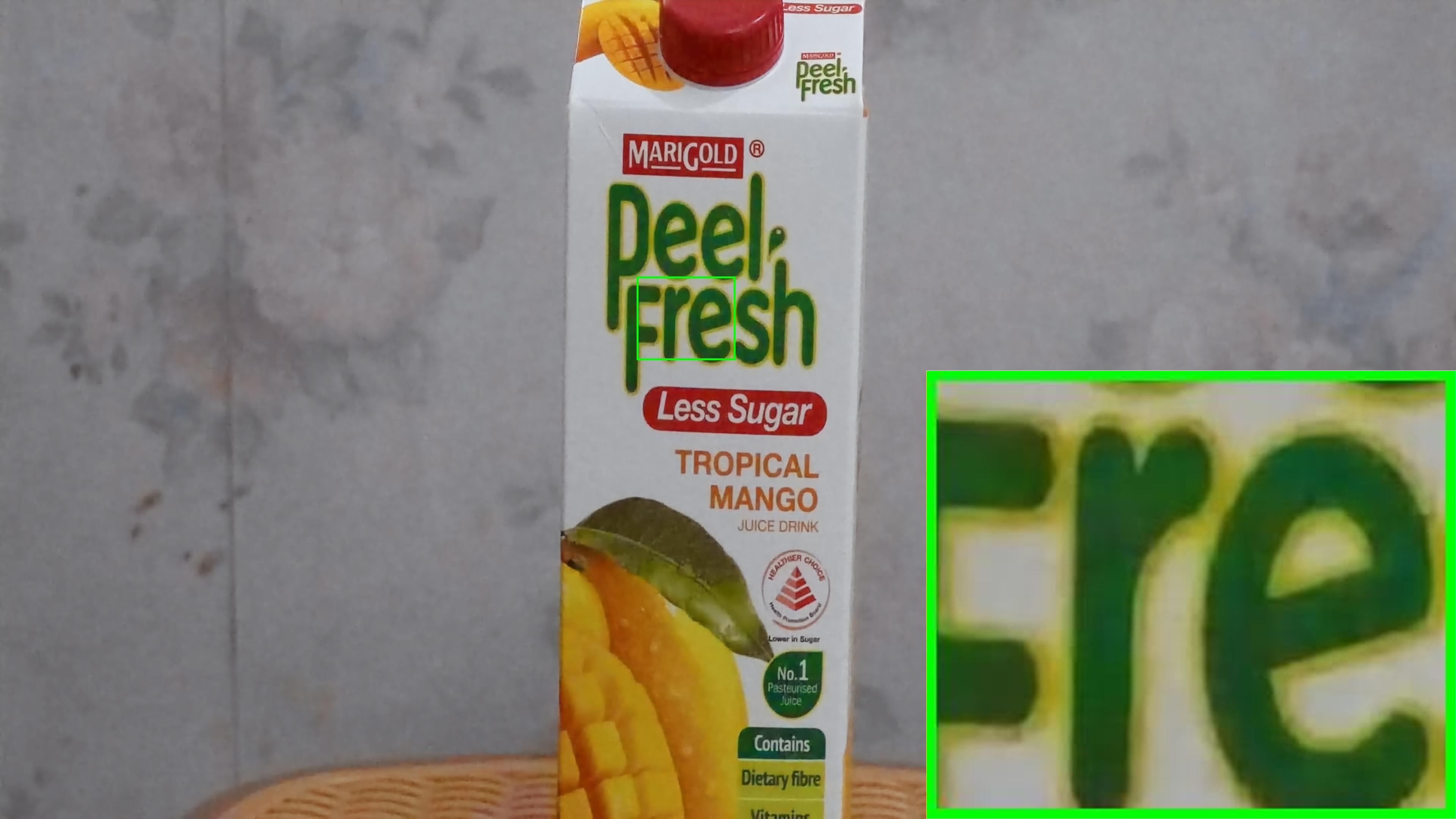} &\hspace{-4.75mm}
\includegraphics[width=0.4\linewidth]{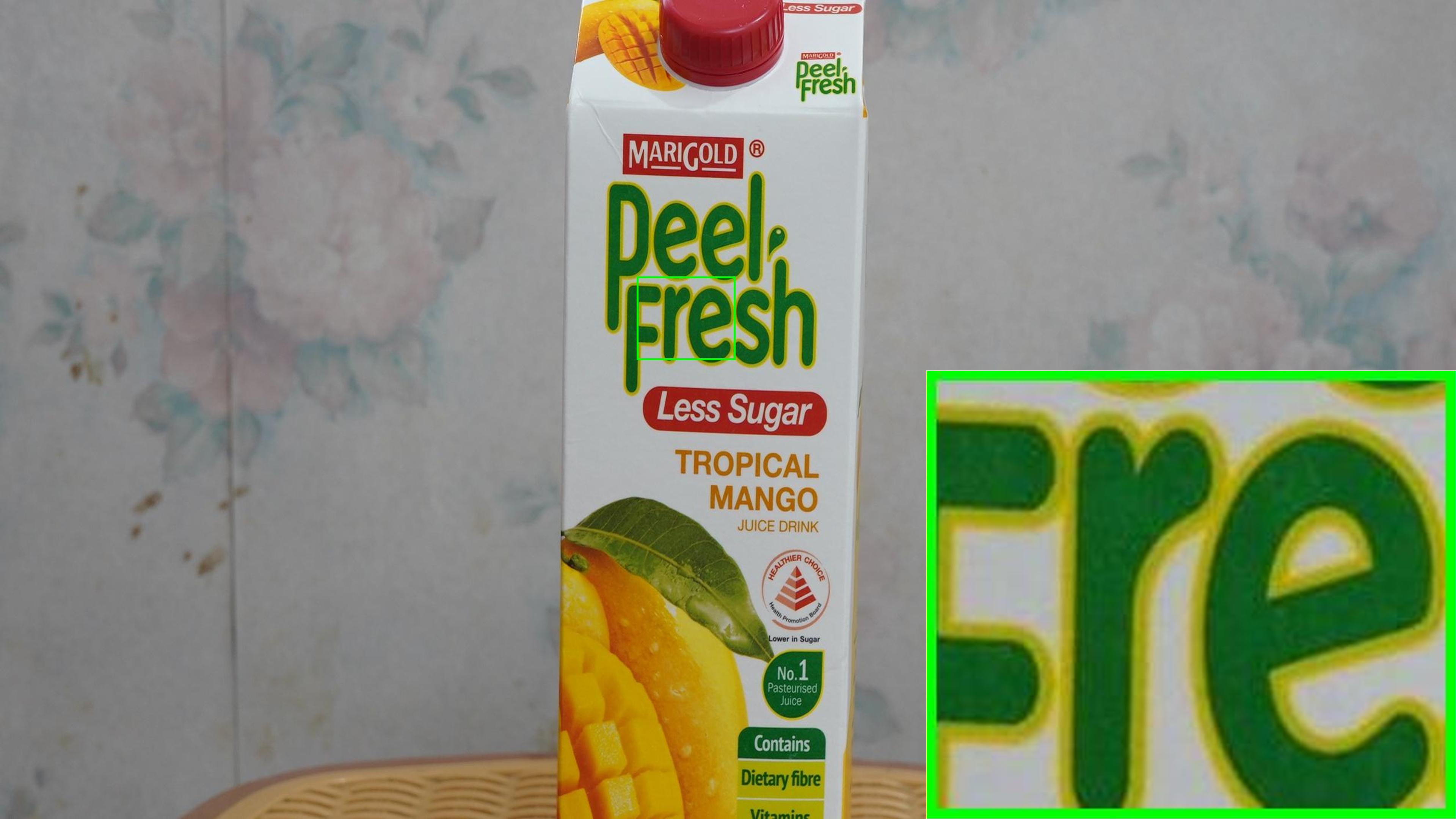} 
\\
\hspace{-1mm}  (c) w/ GFR (\textit{\textbf{Ours}}) &\hspace{-4.75mm} (d) GT
\end{tabular}
\vspace{-2mm}
\caption{\textbf{Visual effect on Gated Feature Refinement} (GFR).
%
% Using GFR helps produce visually pleasing results.
}
\label{fig:Visual effect on Gated Feature Refinement}
\end{center}
\vspace{-2mm}
\end{figure}

We further examine the effect of the Gated Feature Refinement (GFR) module, which is employed within FR-CMT to refine matched features. 
To isolate its contribution, we replace GFR with a $1$$\times$$1$ convolution as the baseline comparison. 
As reported in Tab.~\ref{tab:Effect on Ablation study on Gated Feature Refinement}, GFR improves restoration quality by $0.243$dB in PSNR over this baseline, confirming its effectiveness. 
Visual comparisons in Fig.~\ref{fig:Visual effect on Gated Feature Refinement} further demonstrate that GFR produces outputs with more consistent color contour with GT, whereas the model without GFR exhibits color excursion.
\begin{table*}[!t]
\caption{\textbf{Effect on Gated Feature Refinement (GFR)}.
}
\vspace{-2mm}
\label{tab:Effect on Ablation study on Gated Feature Refinement} 
\tablestyle{9.5pt}{1}
\centering
\begin{tabular}{l|l|ccc|cc}
\shline
\multicolumn{1}{c|}{\multirow{2}{*}{\textbf{ID}}}& \multicolumn{1}{c|}{\multirow{2}{*}{\textbf{Experiment}}}&\multicolumn{3}{c|}{\textbf{Metrics}}&\multicolumn{2}{c}{\textbf{Computational Complexity}}
\\
\cline{3-7}
&& \textbf{PSNR (dB)}~$\uparrow$ & \textbf{SSIM}~$\uparrow$ &\textbf{LPIPS}~$\downarrow$&\textbf{Params. (M)}~$\downarrow$ &\textbf{FLOPs (G)}~$\downarrow$ 
\\
\shline
(a) &w/o GFR& 27.013&	0.9270&	0.2442&	0.3705&	49.23 \\
% \toprule
% (b)& w/ Individual GFR &  26.553& 0.9267  \\
\shline
\textbf{(b)} &\textbf{w/ Shared GFR} (\textit{\textbf{Ours}})&\textbf{27.256}&	\textbf{0.9308}&	\textbf{0.2304}&	0.3962	&49.65\\
% (b)  GFR (\textit{\textbf{Ours}})&\textbf{27.113} & \textbf{0.9271}  \\
\shline
\end{tabular}
\vspace{-2mm}
\end{table*}
\begin{table*}[!t]
\caption{\textbf{Effect on the use of SR Space}.
}
% \vspace{-2mm}
\label{tab:Effect on SR branch} 
\tablestyle{10pt}{1.1}
\centering
\begin{tabular}{l|l|ccc|cc}
\shline
\multicolumn{1}{c|}{\multirow{2}{*}{\textbf{ID}}}& \multicolumn{1}{c|}{\multirow{2}{*}{\textbf{Experiment}}}&\multicolumn{3}{c|}{\textbf{Metrics}}&\multicolumn{2}{c}{\textbf{Computational Complexity}}
\\
\cline{3-7}
&& \textbf{PSNR (dB)}~$\uparrow$ & \textbf{SSIM}~$\uparrow$ &\textbf{LPIPS}~$\downarrow$&\textbf{Params. (M)}~$\downarrow$ &\textbf{FLOPs (G)}~$\downarrow$ 
\\
\shline
(a) &w/o SR Space& 26.449&	0.9274&	0.2355&	0.3929	&46.21\\
\shline
\textbf{(b)} &\textbf{w/ SR Space} (\textit{\textbf{Ours}})&\textbf{27.256}&	\textbf{0.9308}&	\textbf{0.2304}&	0.3962	&49.65\\
\shline
\end{tabular}
\end{table*}
%%%%%%%%%%%%%%%%%%%%%%%%%%%%%%%%%%%%%%%%%%%%%%%%%%%
\begin{figure}[!t]
% \vspace{-2mm}
\centering
\begin{center}
\begin{tabular}{cccccc}
\hspace{-1mm}\includegraphics[width=0.4\linewidth]{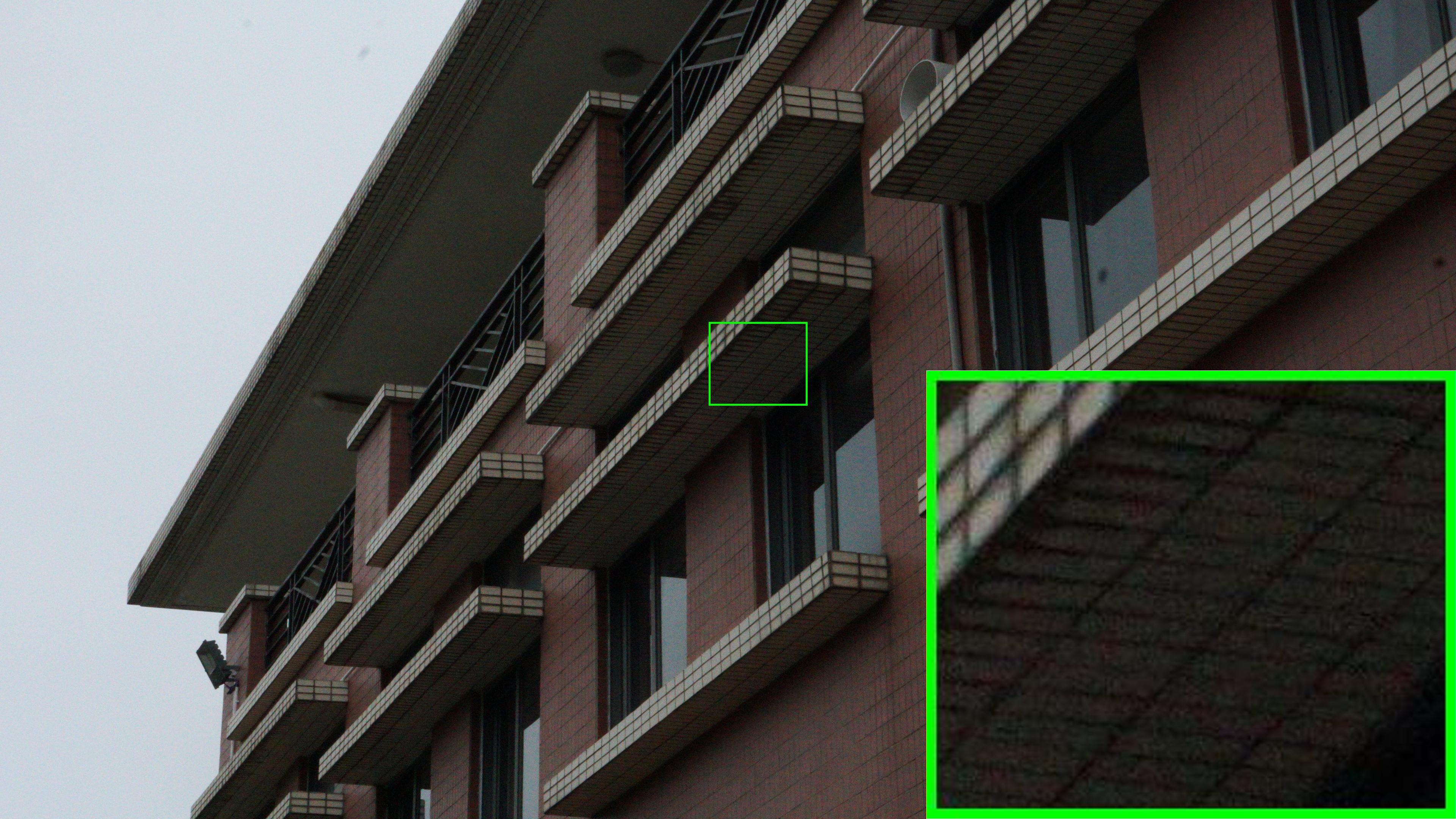} &\hspace{-4.75mm}
\includegraphics[width=0.4\linewidth]{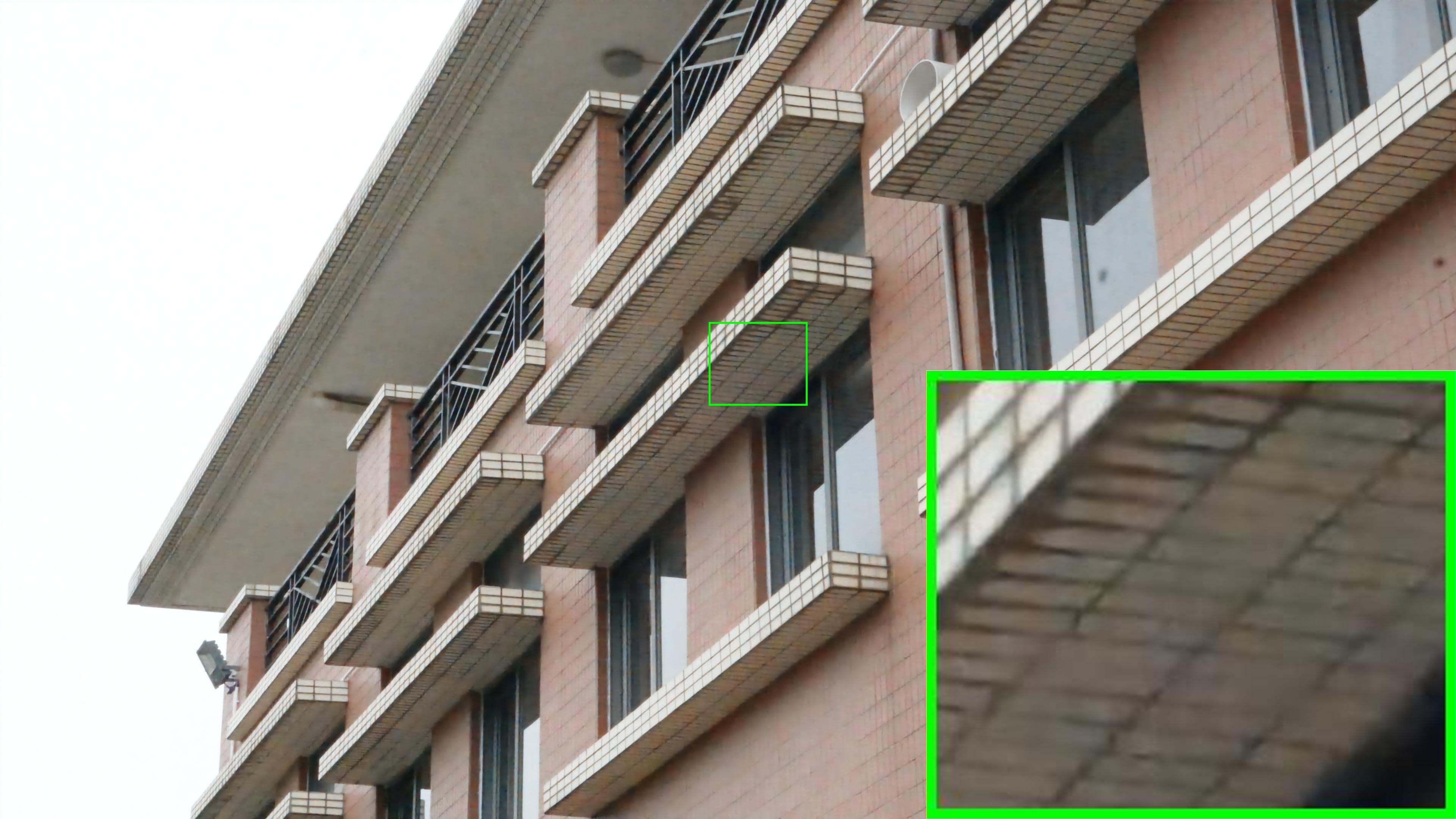} 
\\
\hspace{-1mm} (a) Input&\hspace{-4.75mm} (b)  w/o SR Branch 
\\
\hspace{-1mm}\includegraphics[width=0.4\linewidth]{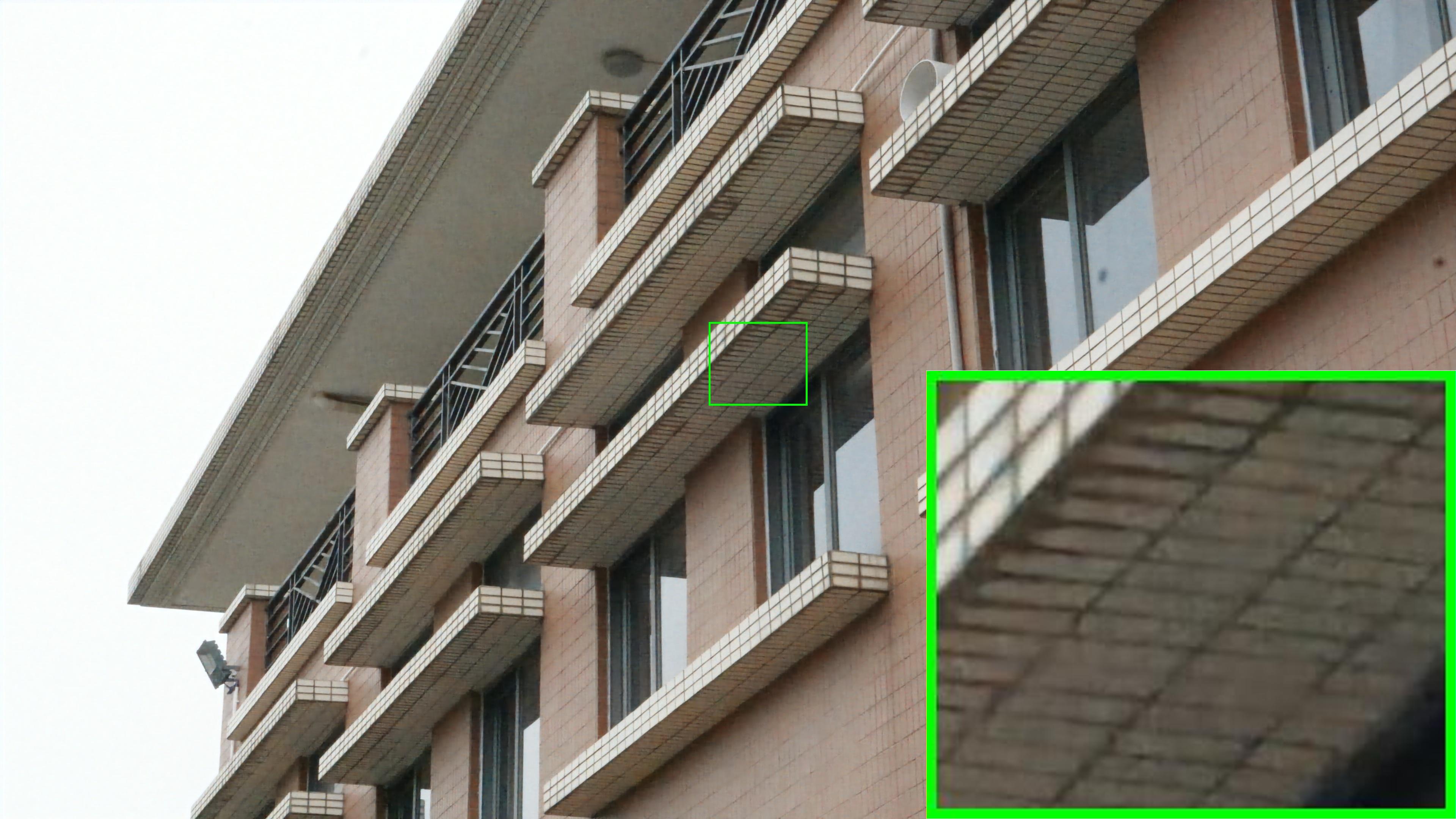} &\hspace{-4.75mm}
\includegraphics[width=0.4\linewidth]{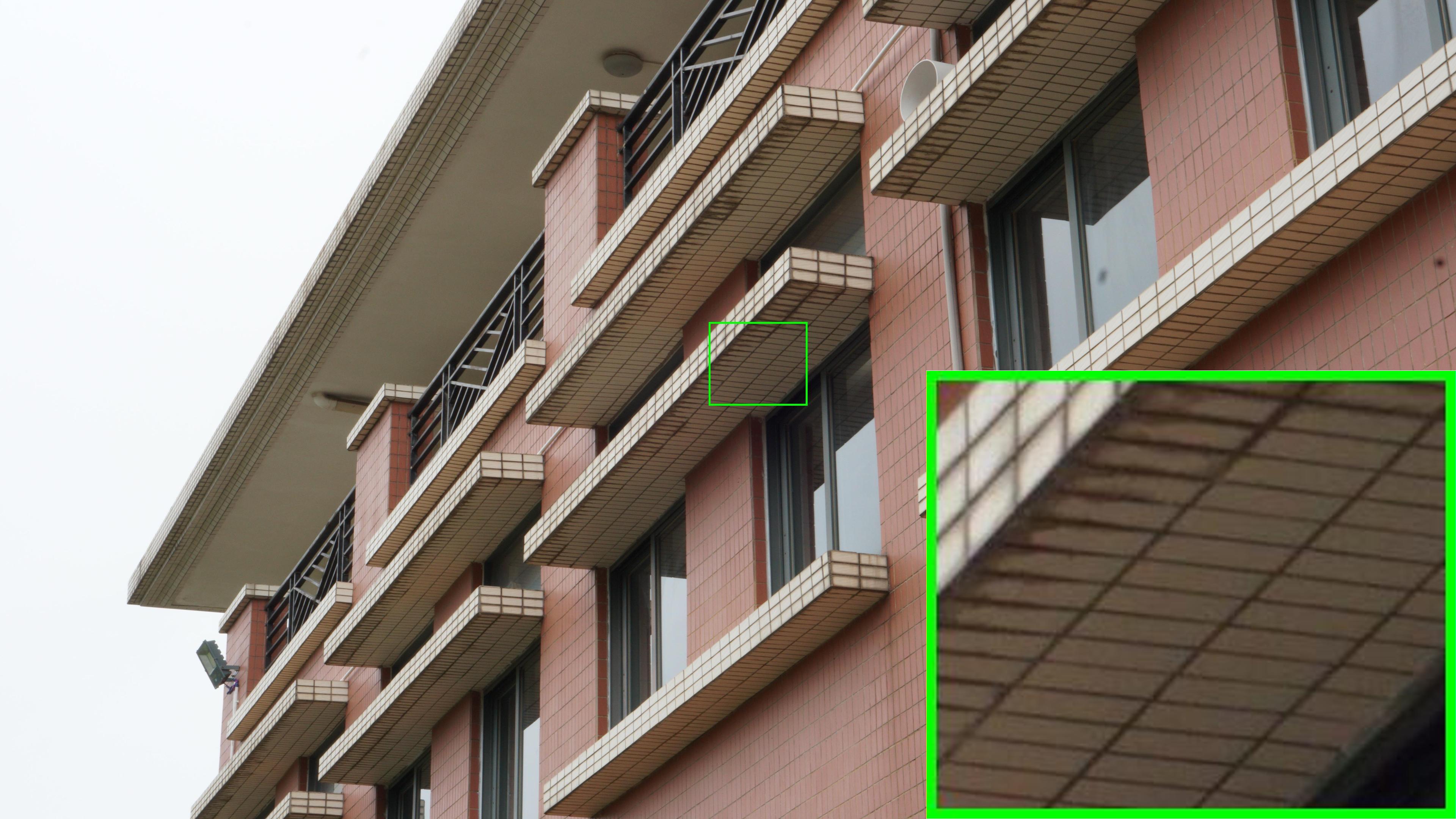} 
\\
\hspace{-1mm}  (c)  w/ SR Branch (\textit{\textbf{Ours}}) &\hspace{-4.75mm} (d) GT
\end{tabular}
\vspace{-2mm}
\caption{\textbf{Visual effect on SR space}.
% Using SR space to reconstruct UHD images is beneficial to the restoration quality.
}
\label{fig: Visual effect on SR  Branch}
\end{center}
\vspace{-3mm}
\end{figure}

\subsubsection{Effect on SR Space}
We further analyze the effect of the SR space on the final UHD image reconstruction, with results reported in Tab.~\ref{tab:Effect on SR branch}. 
To ensure a fair comparison, we augment the variant without SR space by appending an additional ConvNeXt block to the reconstruction stage, thereby maintaining a comparable model capacity. 
As shown, incorporating the SR space consistently improves restoration quality across all three reference metrics, namely PSNR, SSIM, and LPIPS. 
This is further corroborated by the visual example in Fig.~\ref{fig: Visual effect on SR Branch}, where the SR space enables the reconstruction of finer structural details in the restored output.

\begin{table}[!t]
\caption{\textbf{Effect on the number of channels ($C$)}.
}
\vspace{-2mm}
\label{tab: Effect on number of channel.} 
\tablestyle{4pt}{1}
\centering
\begin{tabular}{l|ccccc}
\shline
$C$ & 4 & 8 &\textbf{16 (\textit{Default})}&24&32 
\\
\shline
\textbf{PSNR (dB)}~$\uparrow$&22.789 & 24.773&\textbf{27.256} & 26.690& 27.158\\
\textbf{SSIM}~$\uparrow$&0.9086 & 0.9162 & 0.9308&0.9298& \textbf{0.9312}\\
\textbf{LPIPS}~$\downarrow$&0.306&0.2756&0.2304&0.2305&\textbf{0.2250}\\
\shline
\textbf{Params. (M)}~$\downarrow$&0.0399 &0.1193& 0.3962 & 0.8310M  &1.4237\\
\textbf{FLOPs (G)}~$\downarrow$&4.07 &13.69& 49.65 & 107.89  &188.39 \\
\shline
\end{tabular}

\end{table}

\subsubsection{Effect on Shuffled Down Factor}
We analyze the effect of the pixel-shuffle downsampling factor on restoration performance and computational cost, with results reported in Tab.~\ref{tab:Effect on Shuffle Down Factor}.
Although a shuffle-down factor of $2$ yields the best PSNR and SSIM, it incurs twice the FLOPs compared to our default setting of $8$$\times$$8$. 
Among configurations with larger shuffle-down factors, the $8$$\times$$8$ setting achieves the best performance across all three reference metrics, including PSNR, SSIM, and LPIPS, making it the most favorable trade-off between restoration quality and computational efficiency.

\subsubsection{Effect on Number of Channels}
We set the number of channels as $16$ in all the blocks in our UHDformer, except the last convolutional layer. 
We analyze the effect of the number of feature channels $C$ on restoration performance, with results reported in Tab.~\ref{tab: Effect on number of channel.}. 
As shown, UHDformer++ achieves the best PSNR when $C=16$, while increasing the channel count to 24 or 32 does not yield further performance improvement, suggesting that a compact channel configuration is sufficient for effective UHD image restoration.

\subsubsection{Effect on Number of Correlation Matching Transformation Transformer Blocks}
We analyze the effect of the number of Correlation Matching Transformation Transformer Blocks (CMT-TBs) on restoration performance, with results reported in Tab.~\ref{tab:Effect on Number of Correlation Matching Transformation Transformer Blocks}. 
As shown, PSNR gradually decreases when the number of CMT-TBs exceeds 12, while SSIM reaches its peak at 16 CMT-TBs. 
Balancing both metrics, we adopt $L=16$ as the default configuration in our model.

\begin{table}[!t]
\caption{\textbf{Effect on Number of Correlation Matching Transformation Transformer Blocks ($N$)}.
}
\vspace{-2mm}
\label{tab:Effect on Number of Correlation Matching Transformation Transformer Blocks} 
\tablestyle{6pt}{1}
\centering
\begin{tabular}{l|ccccc}
\shline
 \textbf{$N$} & 8 & 12 &\textbf{16 (\textit{Default})}& 20 & 24 
\\
\shline
 \textbf{PSNR (dB)}~$\uparrow$& 27.188 & \textbf{27.452}&27.256& 26.858&26.583 \\ 
\textbf{SSIM}~$\uparrow$&0.9284&0.9299&\textbf{0.9308}&0.9274&0.9275  \\
\textbf{LPIPS}~$\downarrow$&0.2438&0.2378&\textbf{0.2304}&0.2401&0.2350\\
\shline
\textbf{Params. (M)}~$\downarrow$&0.2913 &0.3438& 0.3962 & 0.4487  &0.5011\\
\textbf{FLOPs (G)}~$\downarrow$&37.83 &43.74& 49.65 & 55.57  &61.48 \\
\shline
\end{tabular}
\end{table}

\begin{table}[!t]
\caption{\textbf{Effect on Shuffle Down Factor ($S$)}.
}
\vspace{-2mm}
\label{tab:Effect on Shuffle Down Factor} 
\tablestyle{6pt}{1}
\centering
\begin{tabular}{l|ccccc}
\shline
 $S$&2  &4 & \textbf{8 (\textit{Default})} & 16 &  32
\\
\shline
   \textbf{PSNR (dB)}~$\uparrow$&  27.602& 27.109 & \textbf{27.256 }&27.051 &26.682\\ 
  \textbf{SSIM}~$\uparrow$&0.9315&0.9295&\textbf{0.9308}&0.9280&0.9253 \\ 
\textbf{LPIPS}~$\downarrow$&0.2335&0.2370&\textbf{0.2304}&0.2375& 0.2426\\
\shline
\textbf{Params. (M)}~$\downarrow$&0.2426&0.2733&0.3962&0.8878&2.8538 \\
\textbf{FLOPs (G)}~$\downarrow$&96.27&58.98&49.65&47.32&46.74 \\
\shline
\end{tabular}
\end{table}
%%%%%%%%%%%%%%%%%%%%%%%%%%%%%%%%%%%%%%%

%%%%%%%%%%%%%%%%%%%%%%%%%%%%%%%%%%%%%%%%%%%%%%%%%%%

%%%%%%%%%%%%%%%%%%%%%%%%%%%%%%%%%%%%%%%%%%%%%%%%%%%
\begin{table}[!t]
\caption{\textbf{Effect on weight of loss function}.
}
\vspace{-2mm}
\label{tab:Effect on weight of loss function} 
\tablestyle{0.5pt}{1}
\centering
\begin{tabular}{l|cccccc}
\shline
 $\lambda_{freq}$&0  &0.01 & \textbf{0.05 (\textit{Default})} & 0.1 &  0.2 &0.5
\\
\shline
   \textbf{PSNR (dB)}~$\uparrow$&26.828&\textbf{27.3703}&27.256&26.634&27.215&27.005\\ 
  \textbf{SSIM}~$\uparrow$&0.9118&0.9257&\textbf{0.9308}&0.9281&0.9291&0.9305\\ 
\textbf{LPIPS}~$\downarrow$&0.2545&0.2464&\textbf{0.2304}&0.2392&0.2360&0.2351\\
\shline 
$\lambda_{re\_sr}$&0  &0.1 & 0.25&\textbf{0.5 (\textit{Default})} & 0.75 & 1
\\
\shline \textbf{PSNR (dB)}~$\uparrow$&26.683&26.421&26.734&\textbf{27.256}&26.864&26.897\\ \textbf{SSIM}~$\uparrow$&0.9284&0.9283&0.9280&\textbf{0.9308}&0.9275&0.9290\\ 
\textbf{LPIPS}~$\downarrow$&0.2380&0.2379&0.2404&\textbf{0.2304}&0.2402&0.2341\\
\shline
$\lambda_{freq\_sr}$&0  &0.01 & 0.05  & \textbf{0.1 (\textit{Default})} &  0.2 &0.5
\\
\shline \textbf{PSNR (dB)}~$\uparrow$&26.662&26.580&27.075&\textbf{27.256}&26.549&26.740\\ \textbf{SSIM}~$\uparrow$&0.9277&0.9273&0.9285&\textbf{0.9308}&0.9272&0.9275\\ 
\textbf{LPIPS}~$\downarrow$&0.2401&0.2376&0.2374&\textbf{0.2304}&0.2402&0.2396\\
\shline
\end{tabular}
\end{table}
%%%%%%%%%%%%%%%%%%%%%%%%%%%%%%%%%%%%%%%%%%%%%%%%%%%

\subsubsection{Effect on Weight of Loss Function}
We analyze the contribution of each loss component, including the $L_1$ loss and frequency loss applied in both the reconstruction and SR spaces, with results reported in Tab.~\ref{tab:Effect on weight of loss function}. 
As shown, incorporating frequency loss in both spaces consistently benefits restoration quality, demonstrating the complementary role of frequency-domain supervision alongside pixel-wise reconstruction loss.

%%%%%%%%%%%%%%%%%%%%%%%%%%%%%%%%%%%%%%%%%%%%%%%%%%%

%%%%%%%%%%%%%%%%%%%%%%%%%%%%%%%%%%%%%%%%%%%%%%%%%%%
\begin{table}[!t]
\caption{\textbf{Comparisons on general low-light image enhancement} on the LOL~\cite{retinexnet_wei_bmvc18}. 
UHDFour-2/8 means that the UHDFour version uses the shuffle-down factor with 2/8.
}
\vspace{-2mm}
\label{tab:Low-light image enhancement-lol-lolv2.} 
\tablestyle{6pt}{1}
\centering
\begin{tabular}{l|cccccccc}
\shline
\textbf{Method}&\textbf{PSNR (dB)}~$\uparrow$ & \textbf{SSIM}~$\uparrow$&\textbf{Params.}~$\downarrow$
  
\\
\shline
Retinex-Net~\cite{retinexnet_wei_bmvc18}&16.77 &0.54  &0.45M \\
Zero-DCE~\cite{zerodce_lowlight_guo}&16.79&0.67  & 79.416K\\
AGLLNet~\cite{AGLLNet}&17.52& 0.77   &3.438K\\
Zhao et. al.~\cite{zhao_lie}&21.67 & 0.87  & 11.560M \\
RUAS~\cite{RUAS_liu_cvpr21}&16.44 & 0.70   &3K\\
SCI~\cite{ma2022toward}&14.78 & 0.62  &0.258K \\
URetinex-Net~\cite{Wu_2022_CVPR}  &19.84&0.87 &0.3196M\\
UHDFour-2~\cite{Li2023ICLR_uhdfour}& 23.09 &0.87   &28.518M\\
UHDFour-8~\cite{Li2023ICLR_uhdfour}& 25.23 & 0.87 & 17.537M \\
\shline
UHDformer~\cite{aaai24wang_UHDformer}&25.36 & 0.91 &0.3393M  \\
UHDformer++ (Ours)&\textbf{25.75}&\textbf{0.92}&0.3962M \\
\shline
\end{tabular}
\vspace{-2mm}
\end{table}
%%%%%%%%%%%%%%%%%%%%%%%%%%%%%%%%%%%%%%%%%%%%%%%%%%%

\begin{figure}[!t]\footnotesize
% \vspace{-1mm}
\begin{center}
\begin{tabular}{cccccccccc}
\includegraphics[width=0.32\linewidth]{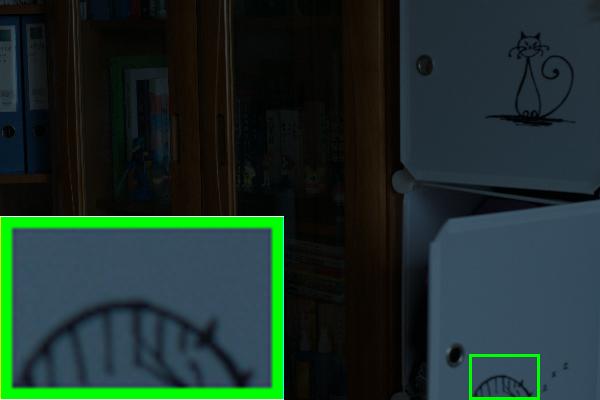}&\hspace{-4.5mm}
\includegraphics[width=0.32\linewidth]{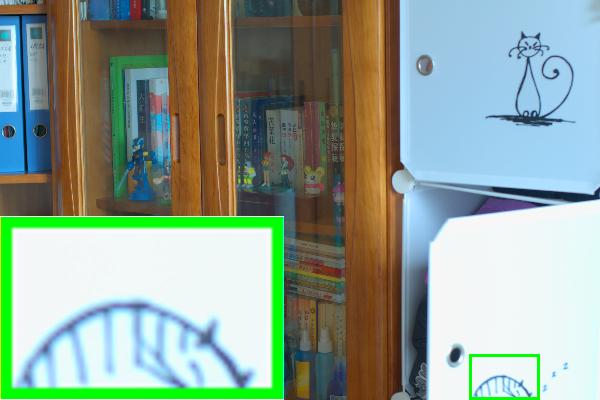}&\hspace{-4.5mm}
\includegraphics[width=0.32\linewidth]{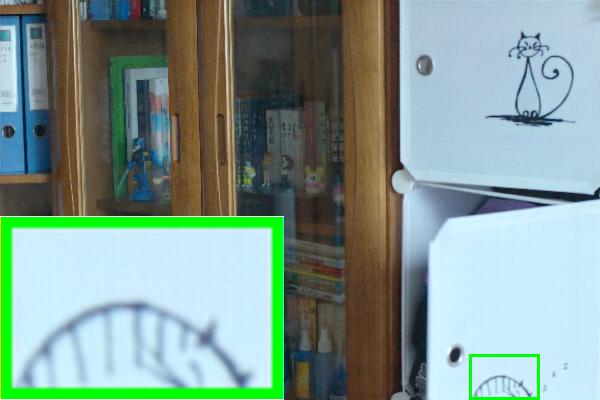}
\\
(a1) Input&\hspace{-4.5mm} (b1) GT &\hspace{-4.5mm}(c1) LLformer~\cite{LLformer}
\\
\includegraphics[width=0.32\linewidth]{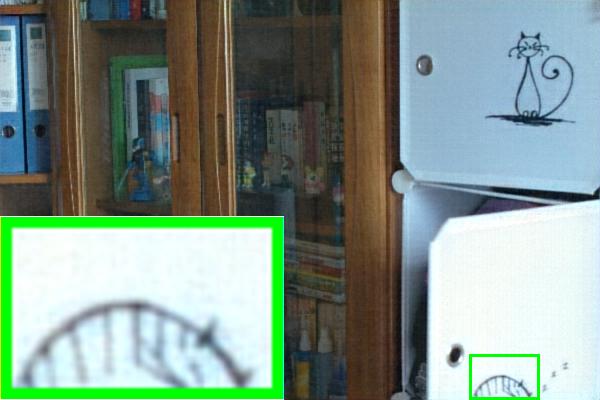}&\hspace{-4.5mm}
\includegraphics[width=0.32\linewidth]{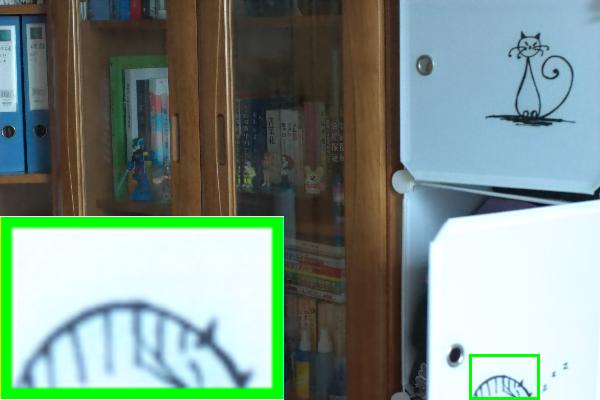}&\hspace{-4.5mm}
\includegraphics[width=0.32\linewidth]{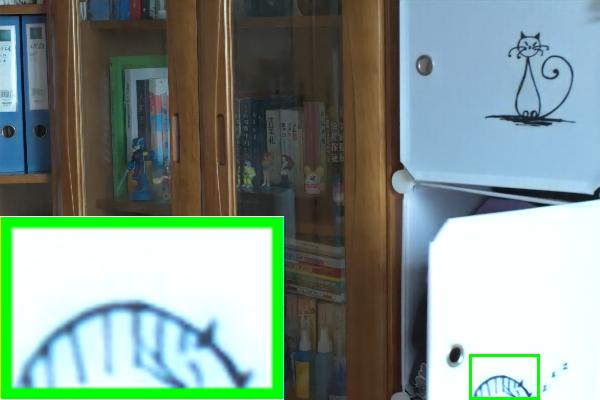}
\\
(d1) UHDFour~\cite{Li2023ICLR_uhdfour}  &\hspace{-4.5mm} (e1) UHDformer &\hspace{-4.5mm} (f1) \textbf{UHDformer++} 
\\
\includegraphics[width=0.32\linewidth]{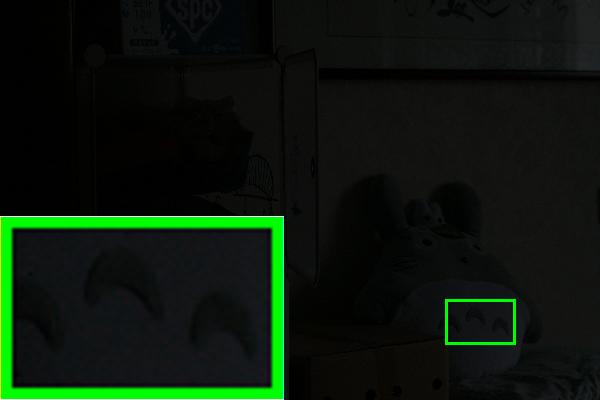}&\hspace{-4.5mm}
\includegraphics[width=0.32\linewidth]{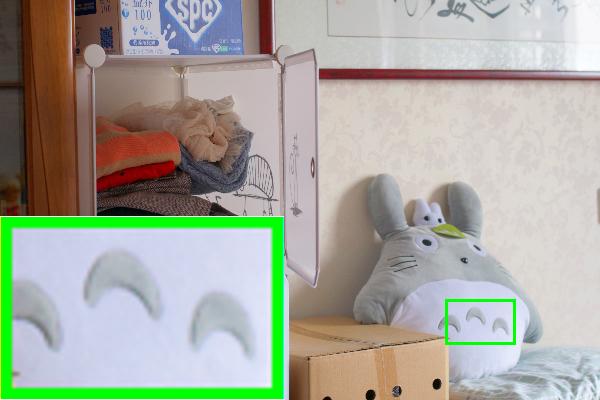}&\hspace{-4.5mm}
\includegraphics[width=0.32\linewidth]{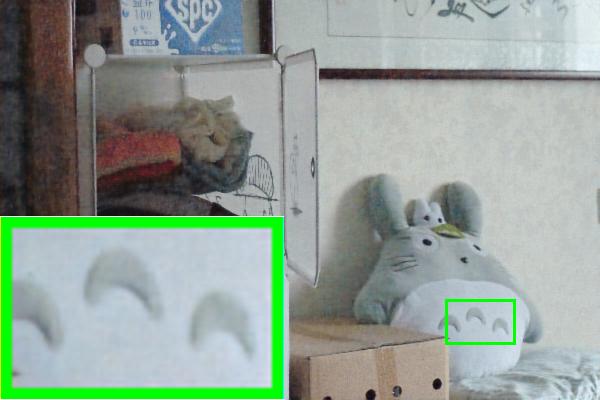}
\\
(a2) Input&\hspace{-4.5mm} (b2) GT &\hspace{-4.5mm}(c2) LLformer~\cite{LLformer} 
\\
\includegraphics[width=0.32\linewidth]{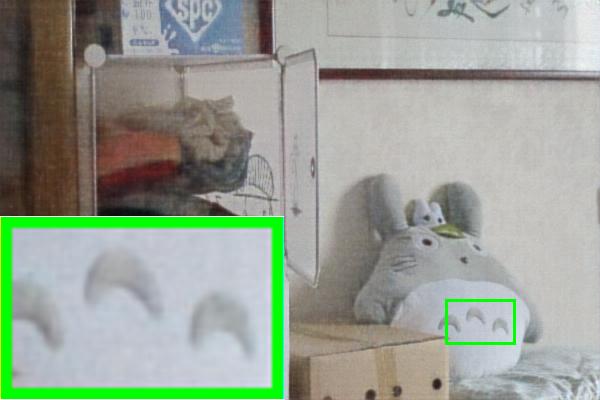}&\hspace{-4.5mm}
\includegraphics[width=0.32\linewidth]{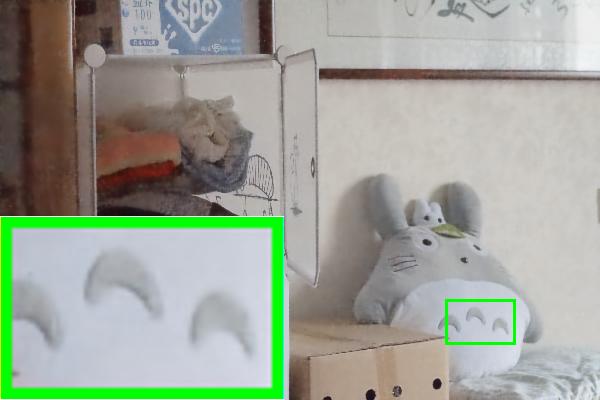}&\hspace{-4.5mm}
\includegraphics[width=0.32\linewidth]{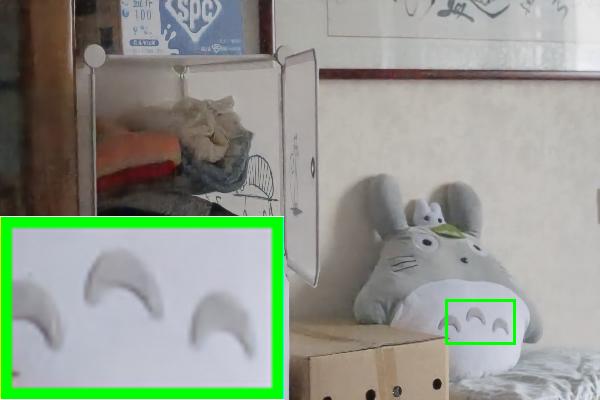}
\\
(d2) UHDFour~\cite{Li2023ICLR_uhdfour} &\hspace{-4.5mm} (e2) UHDformer &\hspace{-4.5mm} (f2) \textbf{UHDformer++} 
\end{tabular}
\vspace{-2mm}
\caption{\textbf{Low-light image enhancement} example on LOL~\cite{retinexnet_wei_bmvc18}.
Although ourUHDformer++ consumes only 0.3962M parameters, it achieves comparable or superior performance on the low-light image enhancement problem.
}
\label{fig: Low-light image enhancement example on LOL-start.}
\end{center}
\vspace{-3mm}
\end{figure}
%%%%%%%%%%%%%%%%%%%%%%%%%%%%%%%%%%%%%%%%%%%%%%%%%%%%%%%%%%

%%%%%%%%%%%%%%%%%%%%%%%%%%%%%%%%%%%%%%%%%%
\section{Comparisons with General Image Restoration}
Our UHDformer++ is designed for UHD images, where most operations are performed in a low-resolution space.
One may wonder if UHDformer++ can work on general image restoration or not.
Hence, we further compare our UHDformer++ with general image restoration methods on general image restoration benchmarks.
Following UHDFour~\cite{Li2023ICLR_uhdfour}, we use LOL~\cite{retinexnet_wei_bmvc18} as the general low-light image enhancement benchmark.
Following general image deblurring methods Stripformer~\cite{Tsai2022Stripformer} and FFTformer~\cite{Kong_2023_CVPR_fftformer}, we use the commonly-used GoPro dataset~\cite{gopro2017} as the general image deblurring benchmark.
%%%%%%%%%%%%%%%%%%%%%%%%%%%%%%%%%%%%%%%

We further benchmark UHDformer++ against state-of-the-art methods on general low-light image enhancement datasets, with quantitative results on LOL~\cite{retinexnet_wei_bmvc18} reported in Tab.~\ref{tab:Low-light image enhancement-lol-lolv2.}. 
Despite being specifically designed for UHD images, where the majority of operations are performed in the low-resolution space, UHDformer++ nonetheless achieves state-of-the-art performance on this general benchmark, demonstrating its strong generalization capability beyond the UHD setting. 
Visual comparisons on LOL are presented in Fig.~\ref{fig: Low-light image enhancement example on LOL-start.}, where UHDformer++ consistently produces clearer and more natural restoration results.

%%%%%%%%%%%%%%%%%%%%%%%%%%%%%%%%%%%%%%%%%%%%%%%

%%%%%%%%%%%%%%%%%%%%%%%%%%%%%%%%%%%%%%%%%%%%%%%%%%
\begin{table}[!t]
\caption{\textbf{Results on general image deblurring} on the GoPro~\cite{gopro2017} benchmarks as well as the limitations of our UHDformer++.
Note that the PSNR and SSIM may be different from existing papers as we use the commonly used IQA PyTorch Toolbox (https://github.com/chaofengc/IQA-PyTorch) to recompute the results.
}
\vspace{-2mm}
\label{tab:Comparisons on GoPro.} 
\tablestyle{6pt}{1}
\centering
\begin{tabular}{l|cccccccc}
\shline
\textbf{Method} &  \textbf{PSNR (dB)}~$\uparrow$ &  \textbf{SSIM}~$\uparrow$&  \textbf{Params.}~$\downarrow$
\\
\shline
% Nah et. al.~\cite{gopro2017}&CVPR'17&17.900 &0.7379   \\
SRN~\cite{tao2018scale}&31.751 & 0.9144 &6.8M  \\
DMPHN~\cite{dmphn2019}&31.734&0.9130&21.7M  \\
MIMO-Unet+~\cite{cho2021rethinking_mimo}&33.277 & 0.9348& 16.1M \\
MPRNet~\cite{Zamir_2021_CVPR_mprnet}&34.195 & 0.9454& 20.1M  \\
Restormer~\cite{Zamir2021Restormer}&34.483 & 0.9488 & 26.1M\\
Uformer~\cite{wang2021uformer}&34.668 & 0.9502& 20.6M \\
Stripformer~\cite{Tsai2022Stripformer}&34.678&0.9506&19.7M\\
FFTformer~\cite{Kong_2023_CVPR_fftformer}&35.798&0.9594&16.6M\\
\shline
UHDformer~\cite{aaai24wang_UHDformer}&29.407 & 0.8660 &0.3393M  \\
UHDformer++ (Ours)&29.713 & 0.8712 &0.3962M  \\
\shline
\end{tabular}
\vspace{-2mm}
\end{table}
%%%%%%%%%%%%%%%%%%%%%%%%%%%%%%%%%%%%%%%%%%%%%%%%%

%
\subsection{Comparisons on Image Deblurring as well as Limitations of Our UHDformer Family}
Despite its strong performance on UHD images, UHDformer++ exhibits limitations when applied to general image deblurring tasks, as evidenced by the results on the GoPro dataset~\cite{gopro2017} in Tab.~\ref{tab:Comparisons on GoPro.}. 
We attribute this to two factors. 
First, general image deblurring typically demands large model capacities to handle complex and spatially varying blur patterns. 
However, UHDformer++ employs fewer than $0.4$M parameters, which may be insufficient for modeling such complexity. 
Second, effective general deblurring often requires operating on high-resolution spatial features, whereas UHDformer++ is designed to perform the majority of its computations in an $8$$\times$$8$ downsampled feature space, a design choice optimized for UHD efficiency that inherently limits its capacity to capture fine-grained blur patterns at full resolution. 
Addressing these limitations remains an important direction for future work.

%%%%%%%%%%%%%%%%%%%%%%%%%%%%%%%%%%%%%%%%%%%%%%%%%%%%%%%%%

%%%%%%%%%%%%%%%%%%%%%
\section{Conclusion}
In this paper, we have proposed a general UHDformer++ framework to solve UHD image restoration tasks.
To generate more useful and representative features for low-resolution space to facilitate better restoration, we have proposed to build the transformation from the high-resolution space to the low-resolution one using the developed FR-CMT and ACM.
FR-CMT is used to explore the exact correlation matching between high-resolution features and low-resolution ones in attention and forward networks to select more useful features from high-resolution space to replace the low-resolution features to improve the feature representation for facilitating better learning in low-resolution space. 
The ACM is used to adaptively modulate multi-level high-resolution features, providing more representative content to the low-resolution space.
Extensive experiments have demonstrated that our UHDformer family significantly reduces model sizes while outperforming state-of-the-art approaches on $5$ UHD image restoration tasks, including low-light image enhancement, image dehazing, image deblurring, image deraining, and image desnowing.

% \subsection{Futher Works}
% xxxx
% \clearpage
% ---- Bibliography ----
%
% BibTeX users should specify bibliography style 'splncs04'.
% References will then be sorted and formatted in the correct style.
%
% \bibliographystyle{splncs04}
% \bibliography{egbib}
\bibliographystyle{IEEEtran}
% \bibliography{aaai23}
\bibliography{aaai23}

\vspace{-12mm}
\begin{IEEEbiography}[{\includegraphics[width=1in,height=1.25in,clip,keepaspectratio]{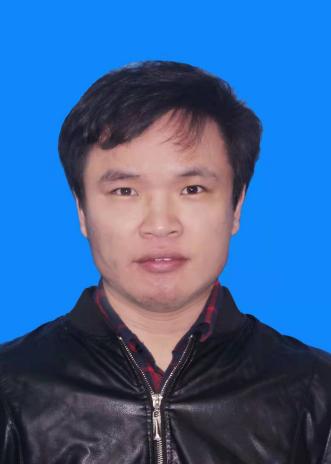}}]{Cong Wang} received the Ph.D. degree from the Department of Computing, The Hong Kong Polytechnic University, in 2024. He is currently a Postdoctoral Fellow at the Department of Radiology and Biomedical Imaging, University of California, San Francisco, CA, USA. His research interests include computer vision, deep learning, and AI for healthcare.
\end{IEEEbiography}
\vspace{-12mm}
\begin{IEEEbiography}[{\includegraphics[width=1in,height=1.25in,clip,keepaspectratio]{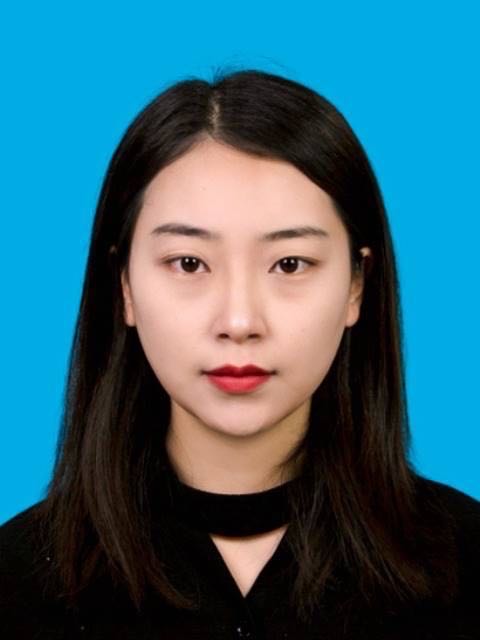}}]{Liyan Wang} is currently a Ph.D. student at the School of Mathematical Sciences, Dalian University of Technology. She received the Master’s degree and the Bachelor’s Degree in computer science and technology from the School of Computer and Information Technology, Liaoning Normal University, Dalian,  China. Her research interests include computer vision and deep learning.
\end{IEEEbiography}
\vspace{-12mm}
\begin{IEEEbiography}[{\includegraphics[width=1in,height=1.25in,clip,keepaspectratio]{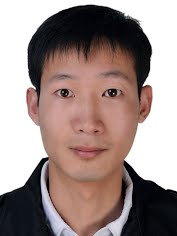}}]{Jinshan Pan} (Senior Member, IEEE) is a professor of the School of Computer Science and Engineering, Nanjing University of Science and Technology. He received the Ph.D. degree in computational mathematics from Dalian University of Technology, China, in 2017. He was a joint training Ph.D. student in Electrical Engineering and Computer Science at the University of California, Merced from 2014 to 2016. His research interests include image analysis and enhancement, and related vision problems.
\end{IEEEbiography}

\vspace{-12mm}
\begin{IEEEbiography}[{\includegraphics[width=1in,height=1.25in,clip,keepaspectratio]{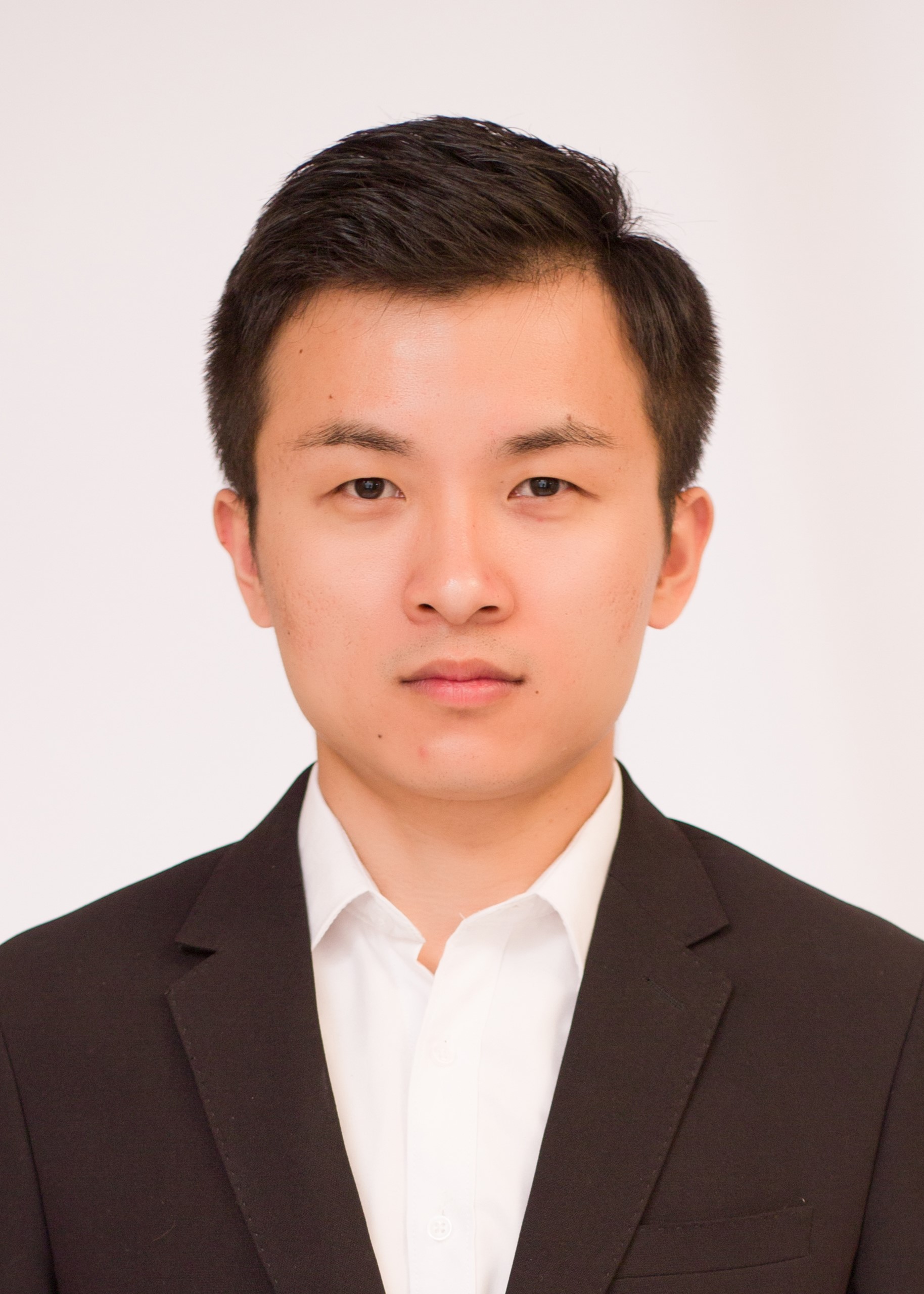}}]{Wei Wang} received the Ph.D. degree from the School of Software Technology, Dalian University of Technology, in 2022. He is currently an Associate Professor with the School of Cyber Science and Technology, Shenzhen Campus of Sun Yat-sen University. His research interests include deep learning, transfer learning, and image processing.
\end{IEEEbiography}

\vspace{-12mm}

\begin{IEEEbiography}[{\includegraphics[width=1in,height=1.25in,clip,keepaspectratio]{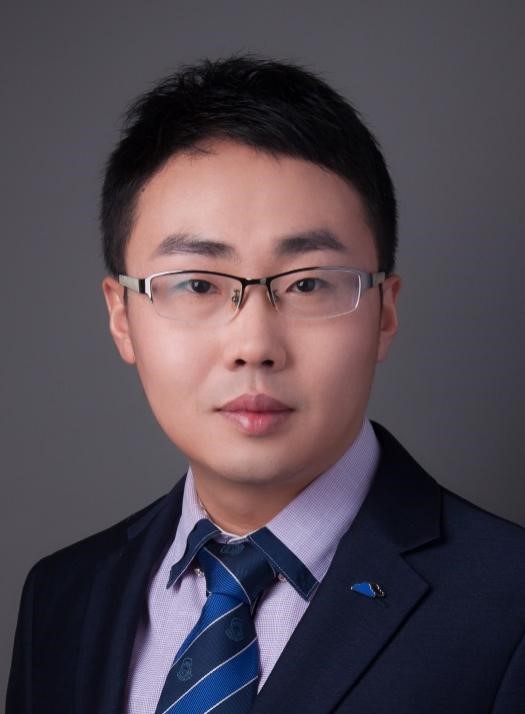}}]{Wenqi Ren} (Senior Member, IEEE) received the Ph.D. degree from Tianjin University, Tianjin, China, in 2017. From 2015 to 2016, he was supported by the China Scholarship Council and worked with Prof. Ming-Husan Yang as a joint-training Ph.D. Student with the Electrical Engineering and Computer Science Department, University of California at Merced. He is currently an Associate Professor with the School of Cyber Science and Technology, Shenzhen Campus of Sun Yat-sen University, Shenzhen, China. His research interests include image processing and related high-level vision problems.
\end{IEEEbiography}
\vspace{-12mm}
\begin{IEEEbiography}[{\includegraphics[width=1in,height=1.25in,clip,keepaspectratio]{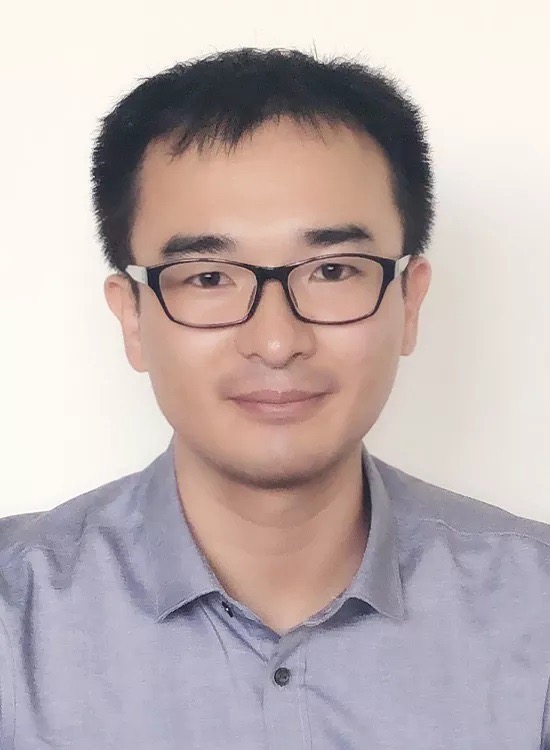}}]{Jun Liu} (Senior Member, IEEE) is a Professor at the School of Computing and Communications, Lancaster University, Lancaster, UK. He obtained his PhD from Nanyang Technological University in 2019, subsequently serving as faculty at Singapore University of Technology and Design from 2019 to 2024. He serves as Associate Editor-in-Chief for Pattern Recognition and Senior Area Editor for IEEE Transactions on Image Processing. He is the General Chair of BMVC 2026 and Program Chair of BMVC 2025. His research focuses on computer vision, machine learning, and digital health applications.
\end{IEEEbiography}

\vspace{-12mm}
\begin{IEEEbiography}[{\includegraphics[width=1in,height=1.25in,clip,keepaspectratio]{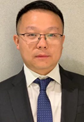}}]{Xiaochun Cao} (Senior Member, IEEE) is a Professor and Dean of the School of Cyber Science and Technology, Shenzhen Campus of Sun Yat-sen University. He received the B.E. and M.E. degrees both in computer science from Beihang University (BUAA), China, and the Ph.D. degree in computer science from the University of Central Florida, USA, with his dissertation nominated for the university-level Outstanding Dissertation Award. After graduation, he spent about three years at ObjectVideo Inc. as a Research Scientist. From 2008 to 2012, he was a professor at Tianjin University. Before joining SYSU, he was a professor at the Institute of Information Engineering, Chinese Academy of Sciences. He has authored and co-authored over 200 journal and conference papers. In 2004 and 2010, he was the recipient of the Piero Zamperoni Best Student Paper Award at the International Conference on Pattern Recognition. He is on the editorial boards of IEEE Transactions on Pattern Analysis and Machine Intelligence and IEEE Transactions on Image Processing, and was on the editorial boards of IEEE Transactions on Circuits and Systems for Video Technology and IEEE Transactions on Multimedia.   
\end{IEEEbiography}

% that's all folks
\end{document}